\documentclass[sn-mathphys-num]{sn-jnl}

\usepackage{graphicx}%
\usepackage{multirow}%
\usepackage{amsmath,amssymb,amsfonts}%
\usepackage{amsthm}%
\usepackage{mathrsfs}%
\usepackage[title]{appendix}%
\usepackage{xcolor}%
\usepackage{textcomp}%
\usepackage{manyfoot}%
\usepackage{booktabs}%
\usepackage{algorithm}%
\usepackage{algorithmicx}%
\usepackage{algpseudocode}%
\usepackage{framed}
\usepackage{ulem}

\usepackage{listings}%
\usepackage{soul, color, xcolor}
\usepackage[export]{adjustbox}
\usepackage{appendix}

\theoremstyle{thmstyleone}%
\theoremstyle{thmstyletwo}%

\theoremstyle{thmstylethree}%

\newtheorem{myDef}{Definition}
\newtheorem{myTheo}{Theorem}
\newtheorem{myProof}{Proof}

\begin{document}

\title[Article Title]{LayeredMAPF: a decomposition of MAPF instance to reduce solving costs}


\author[1]{\fnm{Zhuo} \sur{Yao}}

\author[1]{\fnm{Wang} \sur{Wei}}

\author*[1]{\fnm{Yueri} \sur{Cai}}\email{caiyueri@buaa.edu.cn}


\affil*[1]{\orgdiv{School of Mechanical Engineering and Automation}, \orgname{Beihang University}, \orgaddress{\street{Haidian District}, \postcode{100191}, \state{Beijing}, \country{China}}}




\abstract{Multi-agent pathfinding (MAPF) holds significant utility within autonomous systems, however, the calculation and memory space required for multi-agent path finding (MAPF) grows exponentially as the number of agents increases. This often results in some MAPF instances being unsolvable under limited computational resources and memory space, thereby limiting the application of MAPF in complex scenarios. Hence, we propose a decomposition approach for MAPF instances, which breaks down instances involving a large number of agents into multiple isolated subproblems involving fewer agents. Moreover, we present a framework to enable general MAPF algorithms to solve each subproblem independently and merge their solutions into one conflict-free final solution, and avoid loss of solvability as much as possible. Unlike existing works that propose isolated methods aimed at reducing the time cost of MAPF, our method is applicable to all MAPF methods. In our results, we apply decomposition to multiple state-of-the-art MAPF methods using a classic MAPF benchmark\footnote{https://movingai.com/benchmarks/mapf.html}. The decomposition of MAPF instances is completed on average within 1s, and its application to seven MAPF methods reduces the memory usage or time cost significantly, particularly for serial methods. Based on massive experiments, we speculate the possibilty about loss of solvability caused by our method is $<$ 1\%. To facilitate further research within the community, we have made the source code of the proposed algorithm publicly available\footnote{https://github.com/JoeYao-bit/LayeredMAPF/tree/minimize\_dependence}. }




\maketitle

\section{Introduction}

Multi-agent pathfinding (MAPF), as its name suggests, computes a set of collision-free paths for multiple agents from their respective starting locations to target locations. MAPF is widely utilized in autonomous systems, such as automated warehouses \cite{honig2019persistent} and UAV traffic management \cite{ho2020decentralized}.

Existing methods for MAPF are capable of determining optimal or bounded suboptimal solutions, but efficiency remains a key factor limiting its application. Researchers have proposed novel methods to address this issue, such as trading off solution quality to reduce runtime, reducing search branch factors, and solving agents independently based on their priorities. However, these techniques are often not applicable to all MAPF methods, as they are limited to specific types of MAPF problems.

Motivated by the phenomenon that the cost of solving MAPF instances grows exponentially as the number of agents increases \cite{li2022mapf}, we propose a novel approach to reduce the cost of MAPF methods by decomposing a MAPF instance into multiple smaller subproblems. These subproblems are solved independently, while considering the solutions of other subproblems as dynamic obstacles.

This idea bears resemblance to Priority-Based Search (PBS) \cite{ma2019searching}, which assigns a unique priority to each agent and solves agents separately in decreasing priority order. PBS can be viewed as decomposing a MAPF instance into subproblems, each involving only one agent. While PBS is efficient, it lacks a guarantee of completeness. Compring to PBS, our method try to reduce possibility of loss of solvability by make each subproblem contain more than one agent. 
We formulate the decomposition of MAPF instances as a progressive optimization problem. Initially, we decompose the MAPF instance into multiple subproblems without restricting the order of solving, and then further split them into smaller subproblems with a limited solving order. To minimize loss of solvability and minimize the size of each subproblem, we evaluate the solvability of each step of decomposition, allowing only decompositions that pass a solvability check. As a result, we demonstrate the performance of our method across various maps and illustrate its improvement on multiple cutting-edge MAPF methods.

\begin{figure}[t] \scriptsize
\centerline{\includegraphics[width=8.8cm]{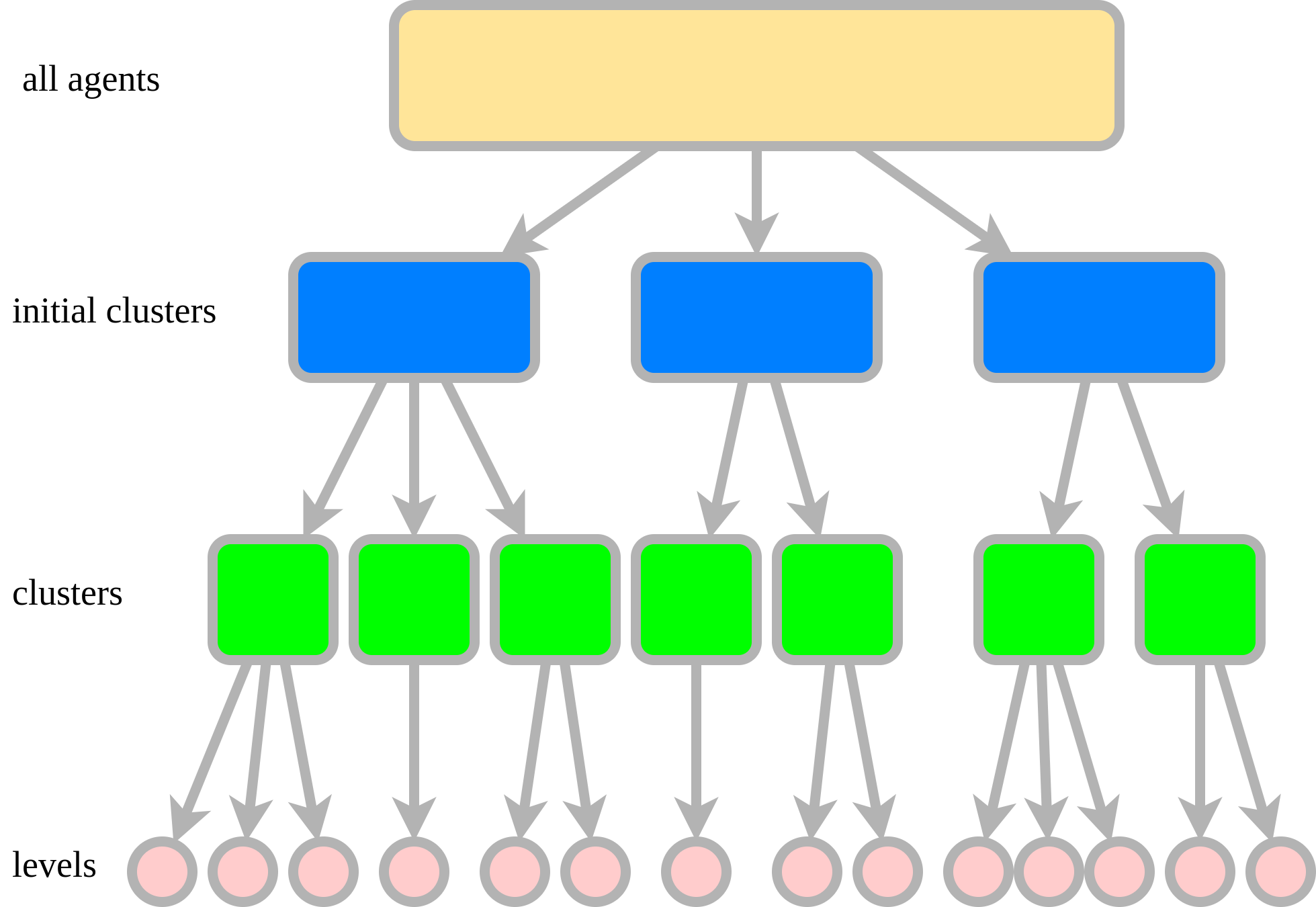}}
\caption{These figures illustrate the complete process of decomposing a MAPF instance, which includes decomposing agents into initial clusters (a type of subproblem), further decomposing initial clusters into smaller clusters, and decomposing clusters into levels (another type of subproblem). For more details, please refer to Section \ref{Methodology}.}
\label{overview}
\end{figure}

The contributions of this manuscript are listed as follows:

1. We propose a novel method that decomposes a MAPF instance into multiple subproblems with minimizing loss of solvability. An overview of our method is depicted in Fig. \ref{overview}.

2. We present a framework that enables general MAPF methods to solve subproblems independently and merge their solutions to obtain the solution of the original problem.

3. We conduct extensive testing to evaluate how the decomposition of MAPF instances influences various MAPF methods using classic MAPF datasets. We evaluate the impact in terms of time cost, memory usage, success rate, and solution quality.

The remainder of this article is organized as follows. In Section \ref{Relatedworks}, we review related problems and methods aimed at reducing the solving cost of MAPF instances. We then define basic terminology in Section \ref{Preliminaries} and introduce our method in Section \ref{Methodology}. Our test results regarding the performance of decomposition under various maps are presented in Section \ref{Resultsofdecomposition}, followed by an examination of how decomposition benefits multiple cutting-edge MAPF methods in Section \ref{application}. Finally, Section \ref{Conclusion} concludes this article.

\section{Related works}
\label{Relatedworks}

In this section, we provide an overview of existing works aimed at reducing the time cost of MAPF methods. MAPF methods typically involve searching for paths connecting the starting and target locations for multiple agents. Broadly, there are two approaches to determining the paths of agents:

1, plan one agent's path from its start to target while keeping other agents static and attempting to avoid conflicts with the paths of other agents. This approach is adopted by CBS-based methods, HCA*, LNS, and PBS.

2, plan for all agents simultaneously by avoiding conflicts with each other at each time step. This approach is employed by LaCAM, PIBT, and Push and Swap.

We refer to the first approach as serial MAPF methods and the second approach as parallel MAPF methods. This distinction is crucial in understanding how the decomposition of a MAPF instance applies to them. Further details about this difference can be found in Section \ref{solve_and_combing}.

\subsection{Conflict based search}

Confict-Based Search (CBS) \cite{sharon2015conflict} is a two-level complete and optimal multi-agent pathfinding algorithm. At the high level, CBS employs a constraint tree in which nodes represent conflicts between paths of two agents. By splitting a conflict, CBS obtains constraints for the related two agents, respectively. The low level utilizes heuristic single-agent searches to find new paths that satisfy all constraints imposed by the high-level constraint tree node. Therefore, CBS operates as a serial MAPF method. If CBS finds a new path, it checks for conflicts between the new path and previous paths. If no new conflicts are found, CBS exits with a conflict-free solution; otherwise, it inserts the new constraints generated from a  conflict as nodes into the conflict tree and repeats the process of splitting conflicts and searching for new paths. If no new conflicts are found, CBS exits with a conflict-free solution. If conflicts exist in the current high-level node, the algorithm first selects one conflict, then generates a pair of constraints to resolve the conflict. Finally, it generates two child nodes, each adding a new constraint (together with existing constraints inherited from the current node). Then in each child node, paths are then replanned to satisfy new constraints and conflicts are updated. 

While CBS can find optimal solutions, it can be time-consuming in certain cases. To accelerate CBS, various improvements have been proposed.

\noindent
$\bullet$ Trade off solution quality for efficency

Bounded CBS (BCBS) \cite{hatem2013bounded} incorporates focal search into the low-level search of CBS, enabling the low-level search to consider avoiding conflicts with other agents' paths and generate bounded suboptimal solutions. Enhanced CBS (ECBS) \cite{hatem2013bounded} utilizes the same low-level search as BCBS and applies focal search to the high-level search of CBS, aiming to minimize the number of nodes from the current constraint tree (CT) node to the CT node representing the conflict-free solution. Similar to BCBS, ECBS also produces bounded suboptimal solutions.

Explicit Estimation CBS (EECBS) \cite{li2021eecbs} employs online learning to estimate the cost of the solution beneath each high-level node. It utilizes Explicit Estimation Search (EES) \cite{thayer2011bounded} to select high-level nodes for expansion, mitigating two drawbacks of ECBS: (1) the cost of the solution beneath a CT node $N$ might exceed the cost of $N$ and thus could surpass the suboptimality bound; (2) the lower bound of ECBS seldom increases, leading to difficulties in finding a solution within a reasonable time if the optimal sum of costs is not within the initial suboptimality bound.

\noindent
$\bullet$ Bypassing conflicts

Bypassing Conflicts \cite{boyarski2015don} is a conflict resolution technique that alters the paths of agents involved in a chosen conflict instead of splitting a constraint tree (CT) node and searching for paths to avoid related constraints. When expanding a CT node $N$ and generating child CT nodes, if the cost of a child CT node equals the cost of $N$ and the number of conflicts of the paths in the child node is smaller than the number of conflicts of $N$, then the child node is used to replace $N$, and all generated child CT nodes are discarded. Otherwise, $N$ is split as before. It has been demonstrated that bypassing conflicts often results in smaller CTs and reduces the runtime of CBS.

\noindent
$\bullet$ Prioritizing conflicts

Prioritizing Conflicts \cite{boyarski2015don} is a conflict-selection technique. A conflict is considered cardinal if, when CBS uses the conflict to split a constraint tree (CT) node $N$, the costs of both child CT nodes are larger than the cost of $N$. It is semi-cardinal if the cost of one child CT node is larger than the cost of $N$, and the cost of the other node is not. It is non-cardinal if the costs of both child CT nodes are equal to the cost of $N$. CBS can significantly enhance its efficiency by resolving cardinal conflicts first, then semi-cardinal conflicts, and finally non-cardinal conflicts. This prioritization is effective because generating CT nodes with larger costs first typically improves the lower bound of the CT (i.e., the minimum cost of the CT nodes in the open list) faster, thereby producing a smaller constraint tree and accelerating CBS.

\noindent
$\bullet$ Symmetry reasoning

Symmetry Reasoning \cite{li2019symmetry, li2020new} is a technique aimed at avoiding the repeated resolution of conflicts between the same pair of agents due to symmetric paths and conflicts. It efficiently identifies each symmetry and resolves it through a single splitting action with specialized constraints. This process results in a smaller constraint tree and reduces the time cost of CBS.

\noindent
$\bullet$ Weighted Dependency Graph (WDG)

The Weighted Dependency Graph (WDG) heuristic \cite{li2019improved} is an admissible heuristic employed in the high-level search of CBS. It operates by constructing a weighted dependency graph for each constraint tree (CT) node $N$. In these graphs, vertices represent agents, and edges denote dependency between agents. This dependency is defined as follows: the minimum sum of costs of their conflict-free paths satisfying $N$'s constraints (computed via solving a 2-agent MAPF instance using CBS) is greater than the sum of costs of their paths in $N$'s paths (the shortest paths satisfying $N$'s constraints but not necessarily conflict-free). The edge weights signify the difference between the minimum sum of costs of conflict-free paths satisfying $N$'s constraints and the sum of costs of their paths in $N$'s paths.

The value of the edge-weighted minimum vertex cover of the graph serves as an admissible heuristic for the high-level search of CBS. Despite the runtime overhead associated with building the weighted dependency graphs and determining their edge-weighted minimum vertex cover, incorporating the WDG heuristic often leads to a reduction in constraint tree size and a decrease in the runtime of CBS.

\subsection{Large neighborhood search}

Large Neighborhood Search (LNS) is a classic algorithm for finding good solutions to challenging discrete optimization problems. Li et al.  \cite{li2021anytime} proposed the first (to our best knowledge) LNS approaches for MAPF, MAPF-LNS. 

MAPF-LNS starts with a given solution and deletes a portion of it, referred to as a neighborhood, while treating the remaining part as fixed. This simplifies the original problem, making it easier to solve. If the newly found solution is superior to the current one, it replaces the existing solution. Consequently, MAPF-LNS is capable of iteratively enhancing solution quality towards near-optimality. Its completeness depends on the algorithm used to generate initial paths. During each iteration, LNS selects a subset of agents and updates their paths without modifying those of other agents. Thus, MAPF-LNS operates as a serial MAPF method. MAPF-LNS can employ any desired approach to solve the simplified problem, provided it can account for the fixed information.

Building on this work, Li et al. introduced MAPF-LNS2 \cite{li2022mapf}, which efficiently finds a solution (rather than improving a given solution) for a MAPF instance. Initially, MAPF-LNS2 invokes a MAPF algorithm to solve the instance and acquire a (partial or complete) plan. For agents lacking a path, MAPF-LNS2 devises a path that minimizes collisions with existing paths.

\subsection{Priority based search}

Priority-Based Search (PBS) \cite{ma2019searching} is an incomplete, suboptimal algorithm designed for prioritized planning. It employs a depth-first search at the high level to dynamically construct a priority ordering, thereby forming a priority tree (PT). When confronted with a collision, PBS greedily determines which agent should be assigned a higher priority. It efficiently backtracks and explores alternative branches only if no solution is found in the current branch. Consequently, PBS incrementally constructs a single partial priority ordering until all collisions are resolved.

Once each agent is assigned a unique priority, PBS computes a minimum-cost path (in priority order) from its starting vertex to its target vertex without colliding with the paths of agents with higher priorities that have already been planned. Therefore, PBS operates as a serial MAPF method.

PBS ranks among the most efficient methods for solving MAPF. However, prioritized planning with an arbitrary priority ordering does not guarantee completeness or optimality in general \cite{ma2019searching}.

\noindent
$\bullet$ Greedy PBS

Given that PBS becomes less effective for MAPF instances with high densities of agents and obstacles, \cite{chan2023greedy} introduced Greedy PBS (GPBS). GPBS utilizes greedy strategies to enhance PBS by minimizing the number of collisions between agents. In essence, GPBS employs the numbers of conflicts and conflicting pairs as heuristics to guide the search on its low and high levels, respectively.

PBS can be regarded as a special decomposition of MAPF instances, where each subproblem involves only one agent. Unlike PBS, our decomposition does not necessitate that every subproblem contain only one agent, so our decomposition has less loss of solvability. Additionally, our decomposition serves as an auxiliary component rather than an isolated method, making it applicable to a wide range of MAPF algorithms.

\subsection{PIBT and LaCAM}

\subsubsection{Priority Inheritance with Backtracking (PIBT)}

PIBT \cite{okumura2019priority} is an incomplete and suboptimal MAPF method that assigns a unique priority to each agent at every timestep. This prioritization ensures that all movements are ordered, and all agents act within a single timestep according to their priority, making it a parallel MAPF method. Priority inheritance is utilized to effectively handle priority inversion in path adjustment within a small time window, and a backtracking protocol prevents agents from becoming stuck. 

Based on priorities and assuming only local interactions, PIBT is easily adaptable to decentralized contexts. Intuitively, PIBT is well-suited for large environments where intensive communication is prevalent, and path efficiency improves with a lower density of agents.

In PIBT, for each timestep, each agent must evaluate distances from surrounding nodes to its goal. While this operation could be implemented by calling A* on demand, it may also become a bottleneck. To address this issue, the authors of PIBT proposed PIBT+ \cite{okumura2022priority}, which saves computation time by preparing distance tables from each agent's goal.

\subsubsection{Lazy constraints addition for MAPF (LaCAM)}

LaCAM \cite{okumura2023lacam} is a complete and suboptimal two-level MAPF method. At the high level, it explores a sequence of configurations, where each search node corresponds to one configuration. At the low level, it searches for constraints specifying which agents go where in the next configuration, making it a parallel MAPF method.

For each high-level node, LaCAM performs a low-level search that generates constraints determining the agents' positions in the next configuration. Successors at the high level (i.e., configurations) are lazily generated while adhering to constraints from the low level, resulting in a significant reduction in search effort. The successor generator of LaCAM is implemented using PIBT.

It is noteworthy that when LaCAM encounters a known configuration, it reinserts the corresponding high-level node into the open set (which stores nodes for high-level search). This action prevents repeated appearance of configurations, thereby preventing LaCAM from avoiding conflicts with external paths (external path: path not belonging to current MAPF instance), as all agents may need to wait at times to avoid conflicts with external paths.

Okumura et al. proposed LaCAM2 \cite{okumura2023improving}, which introduces two enhancements to LaCAM. The first enhancement is its anytime version, called LaCAM*, which converges to optimality eventually, provided that solution costs accumulate transition costs. The second enhancement adds a swap action to the successor generator of LaCAM, enabling the quick generation of initial solutions.

\subsection{Independence detection}
The idea of splitting a MAPF problem into multiple smaller subproblems has been explored by researchers in recent years. Standley et al.\cite{Standley2010FindingOS, Standley2011CompleteAF} proposed that if the optimal paths of two agents have no conflicts, they can be solved independently. Standley\cite{Standley2011CompleteAF} introduced an independence detection (ID) algorithm to decompose a group of agents into the smallest possible groups. ID begins by assigning each agent to its own group and finds an initial path for each group independently. It then attempts to find alternative paths to avoid conflicts, via classic MAPF methods. If attempts to find conflict-free paths fail, ID merges the conflicting groups. The process continues until there are no conflicts between agents from different groups.

Sharon et al.\cite{Sharon2021MetaAgentCS} proposed a continuum between CBS and ID called Meta-Agent CBS (MA-CBS). MA-CBS introduces a predefined parameter \textit{B}, where conflicting agents are merged into a meta-agent and treated as a joint composite agent if the number of conflicts exceeds \textit{B}. The original CBS algorithm corresponds to the extreme case where \textit{B} = $\infty$ (never merge agents), while the ID framework\cite{Standley2010FindingOS, Standley2011CompleteAF} represents the other extreme, where \textit{B} = 0 (always merge agents when conflicts occur).

Compared to ID, our method considers the order of solving subproblems, thus our method have more possibility to decomposeMAPF instance into smaller subproblems. And our decomposition is decoupled from MAPF methods, while ID may need run multiple times of MAPF methods for an agent to find conflict-free paths. So in terms of time cost, our method has significant advantages. Since ID only considers making a group avoid conflict with one group at one time, a group may have conflict with a group it avoided before. For example, group A avoid conflicts wth group B, but A may have conflicts with B after A tries to avoid conflicts with group C. So it may be trapped into infinite loops. This phenomenon is not frequent when there are a few agents but more likely to happen if there are dense agents. In results, we compare our method with ID. And our method demonstrates significant advantages comparing to ID, in terms of max subproblem size, time cost and success rate to find conflict-free solution.

\subsection{SEQ(Sequence) and DSP(Delayed Shortest Path)}

Online Multi-Agent Path Finding (MAPF)\cite{Ji2019Online} is the online version of MAPF where new agents appear over time. One kind of Online MAPF assumes when an agent reaches its goal it stays there, this results in a setting similar to Lifelong MAPF\cite{2017Lifelong}. Another kind of Online MAPF assumes an agent disappears when reaching its goal.

For the second kind of online MAPF, non-conflicting paths can be generated by considering the agents sequentially according to a given order, taking their shortest paths from start to goal location and adding delays at the beginning of the paths to avoid conflicts. The algorithm SEQ (SEQUENCE)\cite{ma2021competitive} does that but, for each agent, naively chooses a relative delay that is equal to the length of the shortest path of the agent that precedes it in the order. Hence, in SEQ, one agent at a time follows its shortest path to destination while the other agents wait. Delayed Shortest Path (DSP)\cite{DSPAtzmon} improve SEQ by calculating safe delays, i.e. relative delays that are possibly shorter than the shortest path of the previous agent in the order but are long enough to safely avoid conflicts. 

DSP and SEQ use a given priority order for the agents. There are several types of priority order: 
\begin{itemize}
\item SH (Shortest path first) gives
a higher priority to agents with shorter paths. 

\item LH (Longer path first) is the opposite of SH and gives a higher priority to agents with longer paths. RND prioritizes the agents randomly. 

\item LD (Lowest delays first) is a greedy method that prioritizes agents according to the lowest safe delay. 

\end{itemize}

The similarity between our method and SEQ and DSP is that we also decompose a MAPF instance to subproblems with a priority order and introduce delays to get conflict-free solution. However, our method focuses on classic MAPF (i.e., offline MAPF), each agent stays at its start position before move and stay at its target position after finish its path. The biggest difference between offline MAPF and online MAPF is, agents staying in start or target positions may block other agents' path. So essentially, our method focus on different problem comparing to SEQ and DSP.

\section{Preliminaries}
\label{Preliminaries}
\subsection{Basic definitions}

In this section, we present fundamental definitions of multi-agent pathfinding. Since our method is dimension-independent, the provided definitions apply to any-dimensional cell space $\mathcal{C}_{\mathcal{N}}$.

\noindent
$\bullet$ Grid space

Let $\mathcal{C}_{\mathcal{N}}$ ($\mathcal{N} \geq 2$) denote a finite $\mathcal{N}$-dimensional integer Euclidean space, where the size of the space is defined as $\mathcal{D}$, with $\mathcal{D}=\{d_{1}, d_{2},\ldots,d_{i},\ldots,d_\mathcal{N}\}$ and $d_{i} \in \mathbb{N}$. The coordinates of an element $\textit{g}$ in this space are defined as a vector $(x_1,x_2,\ldots,x_i,\ldots,x_{\mathcal{N}})$, where $x_i \in ([0,d_i) \cap \mathbb{N})$.

\noindent
$\bullet$ Cell states

There are only two possible states for a cell/element in $\mathcal{C}_{\mathcal{N}}$: passable or unpassable. The set of all passable cells in $\mathcal{C}_{\mathcal{N}}$ is denoted as $\mathcal{F}$, while the set of all unpassable cells is denoted as $\mathcal{O}$. Therefore, $\mathcal{F} \cup \mathcal{O} = \mathcal{C}_{\mathcal{N}}$, $\mathcal{F} \cap \mathcal{O} = \emptyset$. 

\noindent
$\bullet$ Agents

Assuming there are $k$ agents $A=\{a_{1}, a_{2},\ldots,a_{k}\}$ in the grid space $\mathcal{C}_{\mathcal{N}}$, where each agent always occupies a passable cell. For convenience, we denote starts or target state of an agent $a_i$ as $S[a_i]$ or $T[a_i]$. 

Each agent has a unique starting cell and a targeting cell. Specifically, for all $i \in \{1,2,\ldots,k\}$ and $j \in \{1,2,\ldots,k\}$ with $i \neq j$, we have $S[a_i] \neq S[a_j]$, $T[a_i] \neq T[a_j]$, $S[a_i] \neq T[a_j]$, and $T[a_i] \neq S[a_j]$.

\noindent
$\bullet$ Timestep

Time is discretized into timesteps. At each timestep, every agent can either move to an adjacent cell or wait at its current cell.

\noindent
$\bullet$ Path

A path $p_{i}$ for agent $a_{i}$ is a sequence of cells that every pair of subsequent cells is adjacent or identical (indicating a wait action), starting at the start cell $s_{i}$ and ending at the target cell $t_{i}$. The cost of a path is the number of time steps it takes, i.e., the size of the path sequence $|p_{i}|$. We assume that the agents remain at their targets forever after completing their paths.

\noindent
$\bullet$ Conflict

We say that two agents have a conflict if they are at the same cell or exchange their positions at the same timestep (e.g.,  $p_i[t] == p_j[t+1]$ and $p_i[t+1] == p_j[t]$, where $p_i$ and $p_j$ means path of agent $a_i$ and $a_j$, $p_i[t]$ means $a_i$'s location at timestep $t$). 

\noindent
$\bullet$ Solution

A solution of a MAPF instance is a set of conflict-free paths $\{p_{1}, p_{2},\ldots, p_{k}\}$, one for each agent. Generally, an optimal solution is one with minimal \textit{sum of costs (SOC)} $\sum_{i=1}^{k} |p_{i}|$ or \textit{makespan} $\mathop{\mathrm{argmax}}\limits_{i\in \{1,2,\ldots,k\}}|p_{i}|$. In this manuscript, we consider the effects of decomposition on both \textit{SOC} and \textit{makespan}.

\noindent
$\bullet$ Solvability

If there exists a solution for a MAPF instance, it is solvable. In the manuscript, we only consider the decomposition of solvable MAPF instances.

A MAPF method is considered complete if it can always find a solution for a solvable problem; otherwise, it is incomplete. For example, EECBS is complete while PBS is incomplete.

A quick method to determine whether a MAPF instance is solvable is whether every agent can find a collision-free path (i.e., pass no unpassable cells) from the grid space. It is noteworthy that passing through other agents' starting or target cells is acceptable.

Here, we define a search path function:

\begin{myDef}
$search\_path(a_i, \mathcal{C}_{\mathcal{N}}, avoid\_node\_set)$: a complete method to search a path in $\mathcal{C}_{\mathcal{N}}$ that connects the start state and target state of agent $a_i$. The parameter $avoid\_node\_set$ refers to the set of nodes that cannot be part of the path. If a solution exists, we denote this as $search\_path(a_i, \mathcal{C}_{\mathcal{N}}, avoid\_node\_set) \neq \emptyset$.
\end{myDef}

$search\_path$ could be implemented using Best-First Search or Breadth-First Search. If a LA-MAPF instance is solvable, then for all agents $a_i \in A$, it holds that $search\_path(a_i, \mathcal{C}_{\mathcal{N}}, \emptyset) \neq \emptyset$.

\begin{figure}[t] \scriptsize
\begin{minipage}{.24\linewidth}
  \centerline{\includegraphics[width=2.2cm]{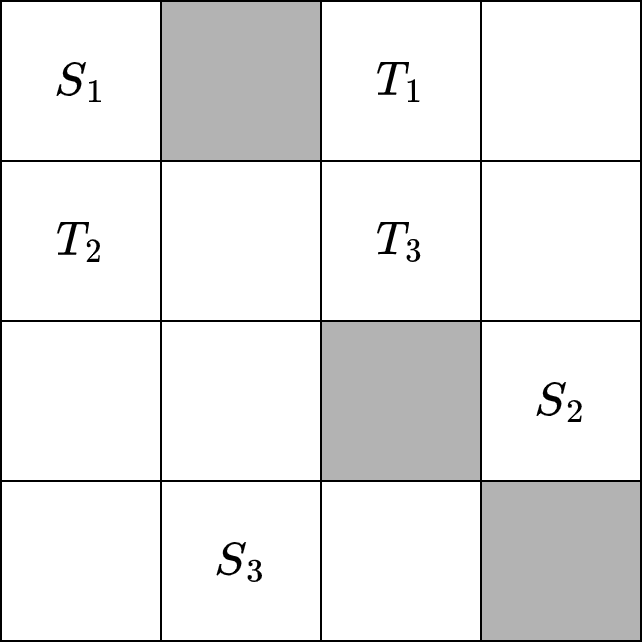}}
  \centerline{A}
\end{minipage}
\hfill
\begin{minipage}{.24\linewidth}
  \centerline{\includegraphics[width=2.2cm]{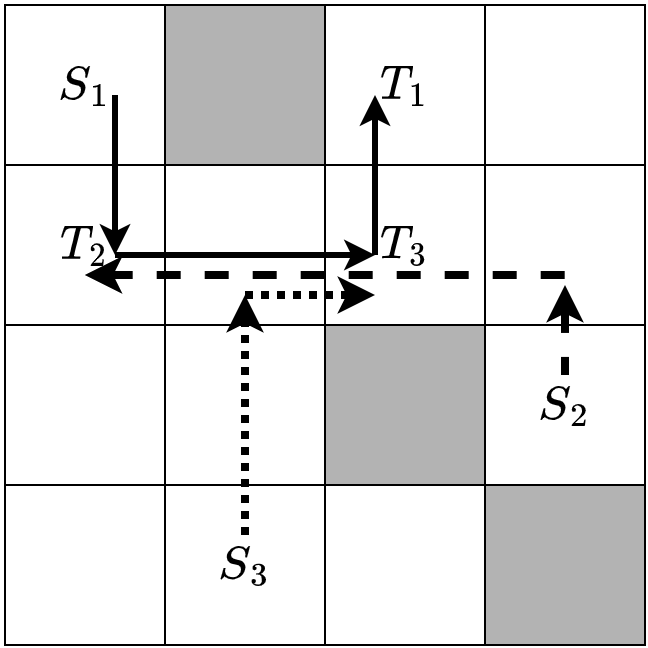}}
  \centerline{B}
\end{minipage}
\hfill
\begin{minipage}{.24\linewidth}
  \centerline{\includegraphics[width=2.2cm]{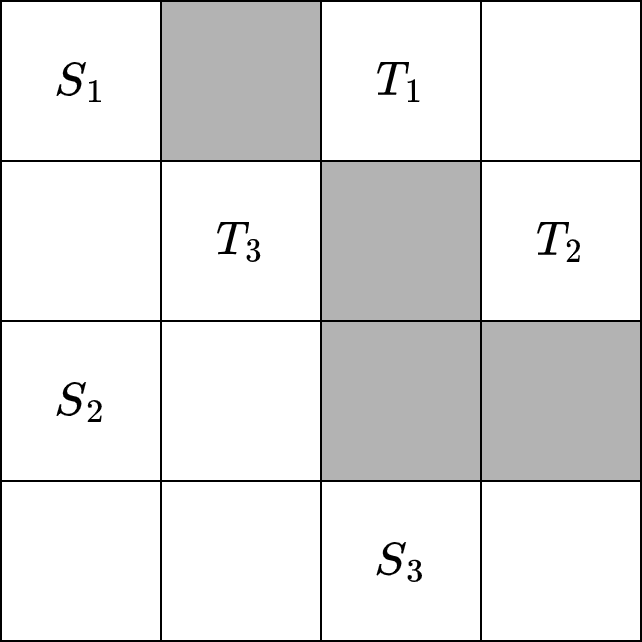}}
  \centerline{C}
\end{minipage}
\hfill
\begin{minipage}{.24\linewidth}
  \centerline{\includegraphics[width=2.2cm]{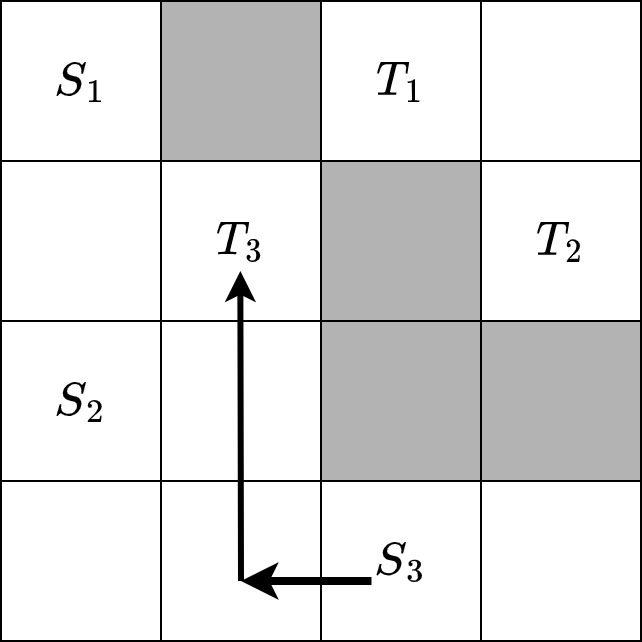}}
  \centerline{D}
\end{minipage}
\vfill
\caption{These figures show a solvable MAPF instance (Fig. A) and an unsolvable MAPF instance (Fig. C). $S_1, S_2, S_3$ and $T_1, T_2, T_3$ represent the start and target cells of the three agents in the instance, as follows. The paths of each agent in the instance are shown in Figures B and D, respectively. The instance in Fig. A passes the solvability check, as all its agents have a path to their target. However, the instance in Fig. C does not pass the solvability check and is unsolvable because only the 3rd agent has a path to its target, while the 1st and 2nd agents have no path to their target. Grey cells represent unpassable cells, while white cells represent passable cells, as follows.  } 
\label{solvable}
\end{figure}

\begin{figure}[h] 
\centerline{\includegraphics[width=3.0cm]{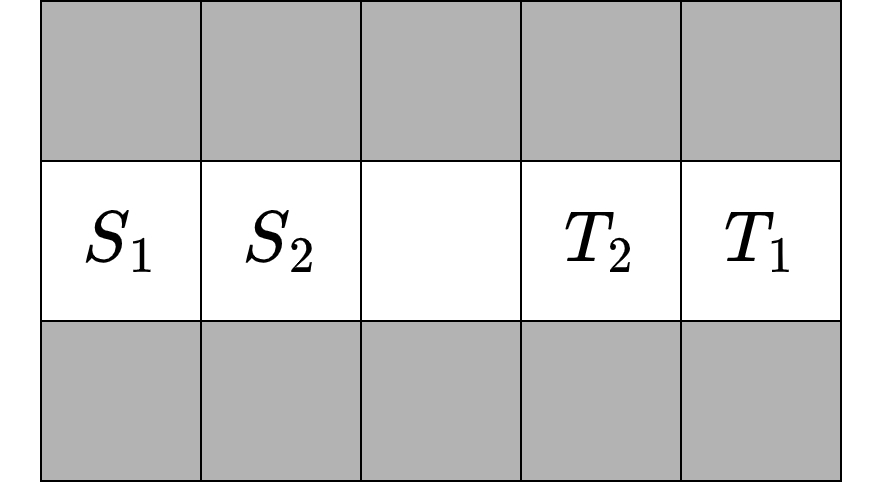}}
\caption{
This figure shows a MAPF instance that passes the mentioned solvability check but is unsolvable. $a_1$'s start $S_1$ have path to its target $T_1$ and $a_2$'s start $S_2$ have path to its target $T_2$, if another agent's start and target are passable. However, due to $a_2$ always block the path from $S_1$ to $T_1$, this instance is unsolvable.
}
\label{solva_failed}
\end{figure}

Here is an example about how to check a MAPF instance is solvable, as shown in Fig. \ref{solvable}. We notice if a MAPF instance failed to meet this condition, it must be unsolvable; however, if a instance meet this condition, it might be unsolvable, an example is shown in Fig. \ref{solva_failed}. So our decomposition minimizing every subproblem's possibility of unsolvable by making every subproblem meet this condition. 

\subsection{Decomposition of MAPF instance}

Given a MAPF instance's graph have $N$ nodes and $k$ agents, and the maximum length of path are $T$, the total number of solutions (including solution that have conflicts) is $N^{Tk}$, so the state space grows exponentially with agent size increases. In the worst cases, a complete MAPF algorithm need to travelsal all solutions to get a conflict-free solution. So the time cost of solve MAPF instance grows nearly exponentially as the number of agents increases\cite{li2022mapf}.

To reduce the cost of solving MAPF instance, we proposed a new technique: splitting a MAPF instance into multiple subproblems, each with fewer agents but under the same map, and solving these subproblems in a specific order. Here, we define the number of agents in each subproblem as the size of the subproblem. When solving a subproblem, it is crucial to ensure that the current paths do not conflict with the paths from previous subproblems. If every subproblem is solvable, and every subproblem's solution doesn't pass previous subproblem's target and later subproblem's start, we can combining all subproblems' results (by adding wait action at the begining of each subproblem's solution, related proof and examples can be found in Section \ref{Methodology}) yields the solution of the original MAPF instance. If we can't split the instance into subproblems that pass the mentioned solvability check, we don't split it. 

As the time cost of solving a MAPF instance grows nearly exponentially with the number of agents, we need to minimize the size of each subproblem to reduce the total time cost of solving all subproblems. So the criteria for judging whether one decomposition is better than another involves sorting the subproblems' sizes in decreasing order and comparing the sizes of the subproblems from the largest to the smallest. We define a decomposition that not worse than any other decomposition of this instance as a optimal decomposition. There might more than one optimal decomposition for a MAPF instace, when two decompositions have the same size of subproblems but subproblem contain different agents. The first smaller subproblem encountered indicates the better decomposition.

For example, suppose there are two decompositions of a MAPF instance with 100 agents. The sizes of the subproblems for the first decomposition are 40, 20, 15, 14, and 11, while for the second decomposition, they are 40, 20, 19, 13, and 8. The first decomposition is considered better than the second one because the third subproblem's size (15) in the first decomposition is smaller than the second decomposition's third subproblem size (19).

Considering all agent plans together can get better solution via coordination all agent's path while solving them separately can't coordinate agent of different subproblems, solving all agents together tends to finding a shorter solution compared to solving them separately. Therefore, MAPF methods with decomposition of MAPF instances tend to produce worse solutions than raw methods. However, we provide no theoretical analysis regarding the extent of this degradation in this manuscript; hence, we conduct empirical comparisons through extensive testing in Section \ref{application}. 

Here we define the decomposition of a MAPF instance:

\begin{myDef}
\textbf{Decomposition of a MAPF instance:} Splitting a set of $k$ agents, $A = \{a_1, a_2, \dots, a_k\}$, where $k \geq 1$, into $m$ subsets of agents, $A_1, A_2, \dots, A_m$, where $m \geq 1$ and $A_1 \cup A_2 \cup \dots \cup A_m = A$. Each subset (i.e., subproblem) should be solvable independently without updating the solutions of the other subsets. A decomposition of a MAPF instance is considered legal if each subproblem passes the solvability check.
\end{myDef}

Before discussing how to decompose a MAPF instance, we first introduce the criteria for determining whether a decomposition is legal.

\begin{figure}[h] 
\centerline{\includegraphics[width=10.0cm]{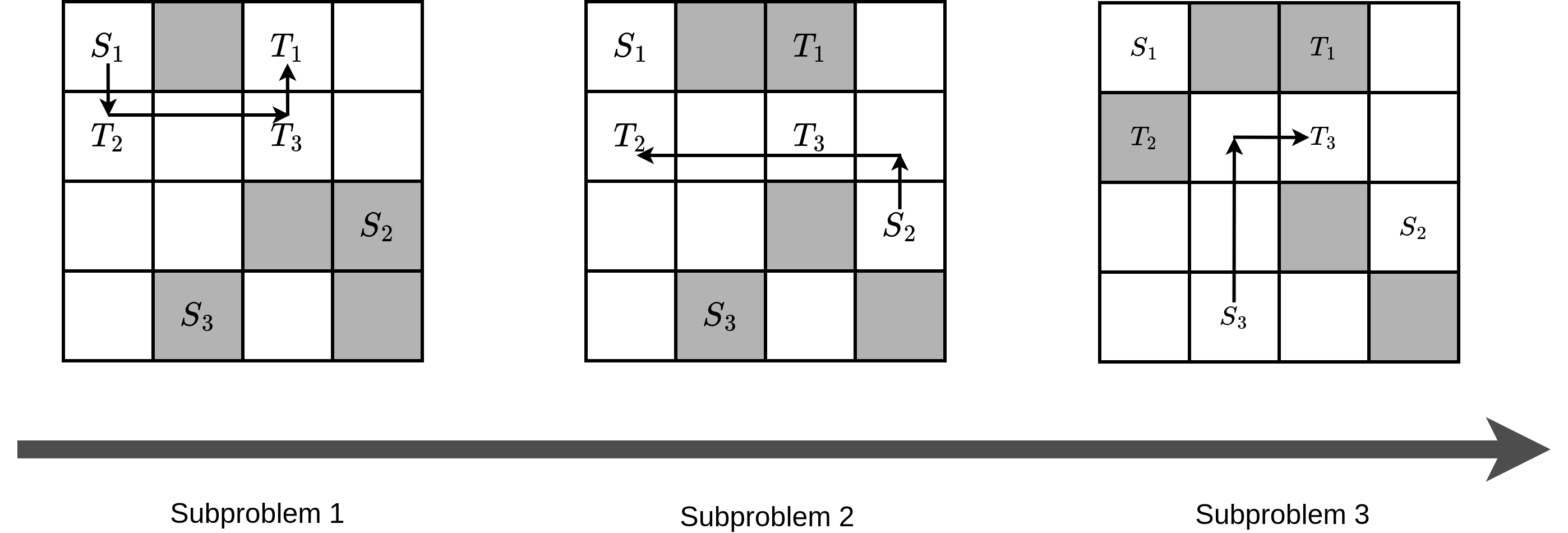}}
\caption{
This figure illustrates three subproblems ($1, \{a_1\}; 2, \{a_2\}; 3, \{a_3\}$) of the MAPF instance shown in Fig. \ref{solvable} A and checks whether these three subproblems pass the solvability check under the simplified scenario. As shown in the figure, the agent in each subproblem has a solution that avoids the target state of the previous subproblem and the start state of the next subproblem. Therefore, since every subproblem passes the solvability check, the instance can be decomposed into these three subproblems.
}
\label{layered_mapf_example}
\end{figure}

We simplify this by allowing the agents in each subproblem to start moving only after all agents in the previous subproblems have reached their target states, while avoiding the start states of subproblems that have not yet been solved. In other words, in this \textit{simplified scenario}, each subproblem treats the agents from other subproblems as static obstacles (remaining at their start state if solved later than the current subproblem, or staying at their target state if solved earlier). An example is shown in Fig. \ref{layered_mapf_example}.

If all subproblems of a decomposition pass the mentioned solvability check under this simplified scenario, we say that we have found a legal decomposition of the MAPF instance.

\begin{figure}[h] 
\centerline{\includegraphics[width=10.0cm]{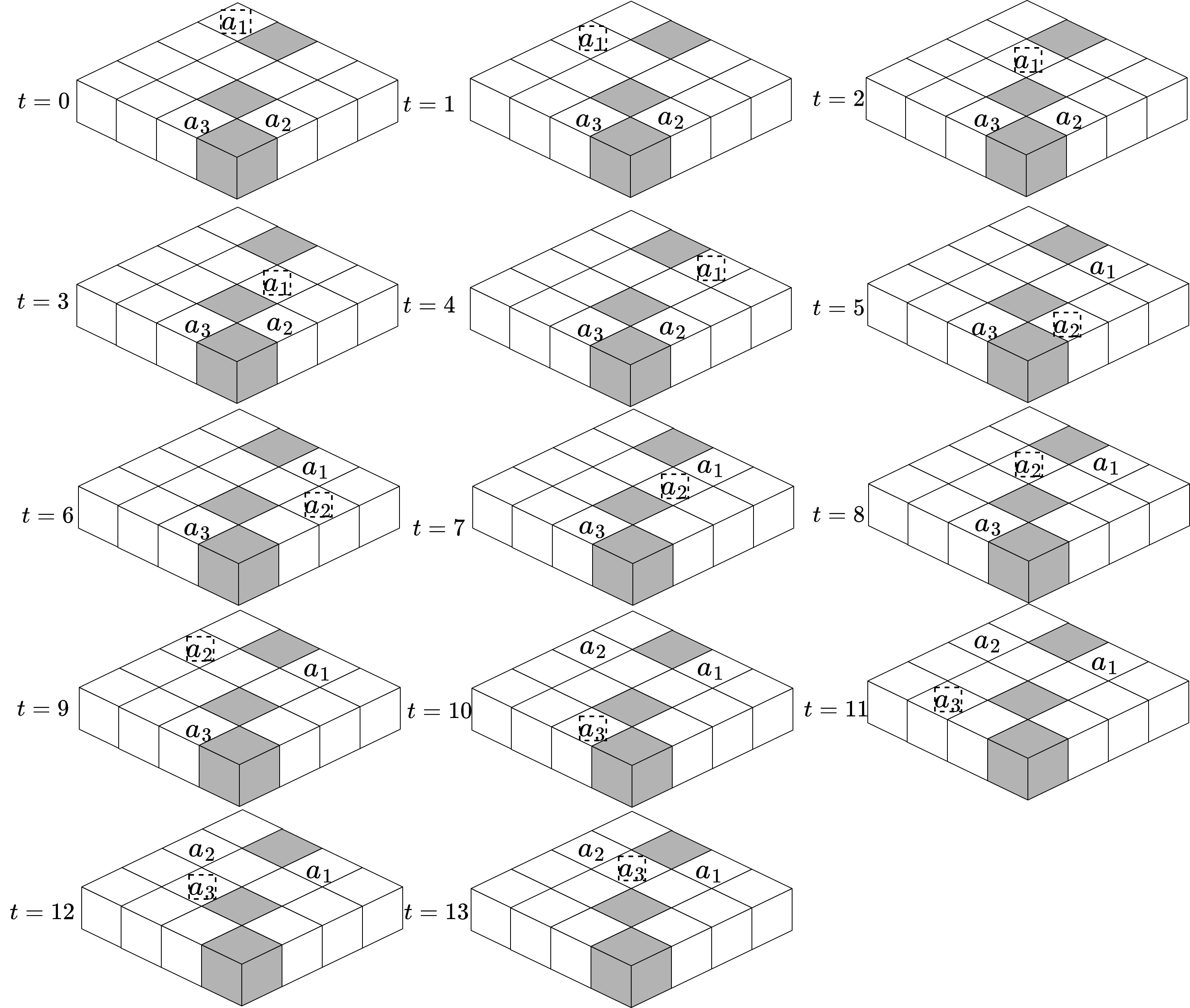}}
\caption{
This figure shows the total solution generated by adding wait actions at the start for the MAPF instance in Fig. \ref{layered_mapf_example}. In the total solution, only one agent from each subproblem (denoted as $a_i$) is moving (marked with a dashed-line box), while the agents from the other subproblems stay at their start or target states. Specifically, $a_1$ (from subproblem 1) moves when $0 \leq t \leq 4$; $a_2$ (from subproblem 2) moves when $5 \leq t \leq 9$; and $a_3$ (from subproblem 3) moves when $10 \leq t \leq 13$.
}
\label{merged_path}
\end{figure} 

If every subproblem passes the solvability check under this simplified scenario, and every subproblem is solved, we can obtain a conflict-free total solution by combining the solutions of each subproblem. Specifically, making an agent from a subproblem wait at the start state until the agents from previous subproblems have reached their target states in the spatial-time map ensures a conflict-free total solution. An example is shown in Fig. \ref{merged_path}.

\begin{figure}[h] 
\centerline{\includegraphics[width=3.5cm]{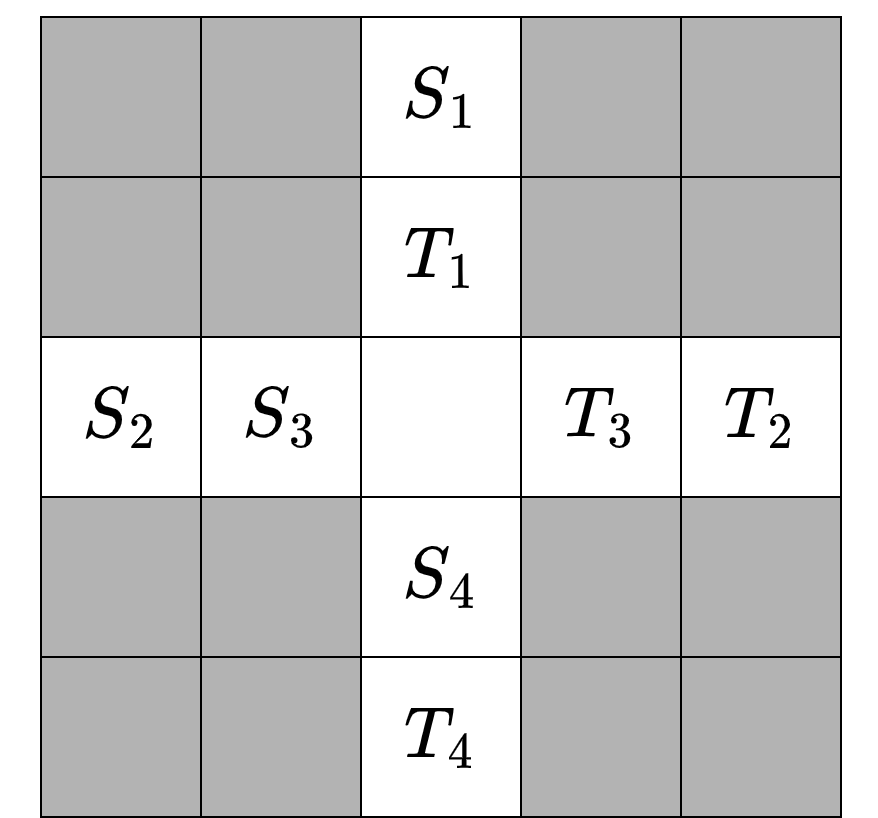}}
\caption{
This figure shows an instance where our decomposition may cause a loss of solvability. Our decomposition may result in the following three subproblems: 1, \{$a_1$\}; 2, \{$a_2$, $a_3$\}; 3, \{$a_4$\}. However, the second subproblem is unsolvable in the simplified scenario, as $T_1$ and $S_4$ are occupied. 
}
\label{imcomplete}
\end{figure} 

However, it is noteworthy that the mentioned solvability check is not always accurate; it may misjudge unsolvable subproblems as solvable, resulting in a loss of solvability. An example is shown in Fig. \ref{imcomplete}. Based on empirical observations, this phenomenon becomes more frequent in instances with dense agents. However, the decomposition of instances also becomes less effective in cases with many dense agents (which means the size of the largest subproblem will be close to the total number of agents, or the instance may not be split). In such cases, the side effect of loss of solvability is sometimes offset.

\subsubsection{Cluster and level}
\begin{myDef}
Here, we define two types of subproblems: \textbf{cluster}, a type of subproblem with no restrictions on the order of solving; and \textbf{level}, a type of subproblem that must be solved in a specific order.
\end{myDef}

For convenience, we denote all start or target states of a cluster $c$ and a level $l$ as $S[c]$ or $T[c]$, and $S[l]$ or $T[l]$, respectively.

\begin{figure}[t] \scriptsize
\begin{minipage}{.45\linewidth}
\centerline{\includegraphics[width=2.0cm]{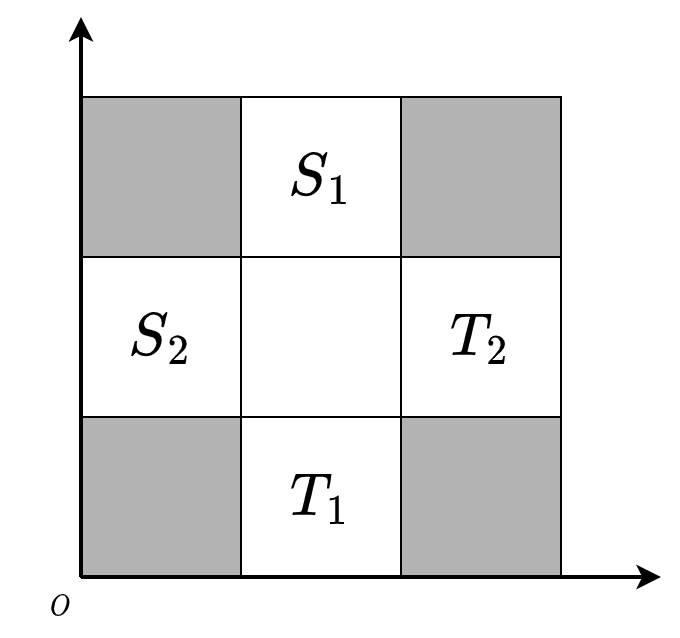}}
  \centerline{A: a MAPF instance about two clusters}
\end{minipage}
\hfill
\begin{minipage}{.45\linewidth}
  \centerline{\includegraphics[width=2.cm]{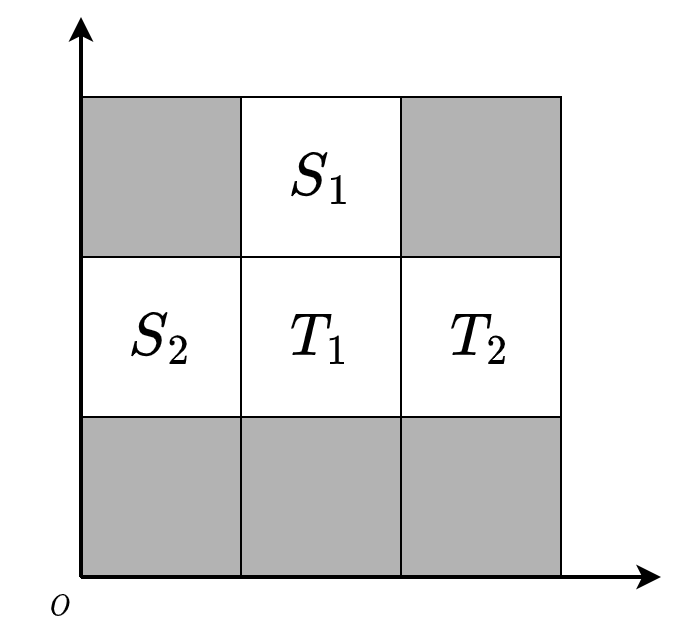}}
  \centerline{B: a MAPF instance about two levels}
\end{minipage}
\vfill

\caption{
Figure A and B show an example about two kinds of subproblem (cluster and level) in MAPF that can be solved independently.  \\
For MAPF instance in Figure A, we can solve $a_1$ and $a_2$ separately and have no limitation in order of solve. If we solve $a_1$ first, a conflict-free solution might be $a_1$ = $\{(1,2)\rightarrow(1,1)\rightarrow(1,0)\}$, and $a_2$ = $\{(0,1)\rightarrow(0,1)\rightarrow(1,1)\rightarrow(2,1)\}$. If we solve $a_2$ first, a conflict-free solution might be $a_1$ = $\{(1,2)\rightarrow(1,2)\rightarrow(1,1)\rightarrow(1,0)\}$, and $a_2$ = $\{(0,1)\rightarrow(1,1)\rightarrow(2,1)\}$. So we say MAPF instance in Figure A can be decompose into two clusters, cluster 1 = $\{a_1\}$, cluster 2 = $\{a_2\}$. \\
For MAPF instance in Figure B, we can solve $a_1$ and $a_2$ separately but there is a limitation in order of solve. If we solve $a_1$ first, the solution of $a_1$ = $\{(1,2)\rightarrow(1,1)\}$, then $a_2$ find no conflict free solution while considering $a_1$ as dynamic obstacles. If we solve $a_2$ first, a conflict-free solution might be $a_1$ = $\{(1,2)\rightarrow(1,2)\rightarrow(1,1)\}$, and $a_2$ = $\{(0,1)\rightarrow(1,1)\rightarrow(2,1)\}$. So we say MAPF instance in Figure B can only be decompose into two levels, level 1 = $\{a_2\}$, level 2 = $\{a_1\}$. Agents in level 1 must be solved before level 2 to pass the solvability check condition and ensure the solvability of subproblems. 
}
\label{cluster_and_level}
\end{figure}

A brief example of clusters and levels is shown in Fig. \ref{cluster_and_level}. A MAPF instance may be decomposed into multiple clusters, $C = \{c_1, c_2, \dots, c_m\}$, or multiple levels, $L = \{l_1, l_2, \dots, l_n\}$, where $1 \leq m \leq k$ and $1 \leq n \leq k$, with $k$ representing the total number of agents in the current instance.

To ensure that the decomposition of a MAPF instance into clusters or levels is legal, both the clusters and levels must satisfy certain conditions.

\begin{myTheo}
\textbf{Legal condition for decomposition into clusters:}

If a MAPF instance is decomposed into multiple clusters $C = \{c_1, c_2, \dots, c_m\}$, then $\forall c_i \in C$, the $avoid\_node\_set$ is defined as 
\[
avoid\_node\_set = \{s \mid s \in \{\bigcup_{j=0}^{m} S[c_j] \cup \bigcup_{j=0}^{m} T[c_j]\}, j \neq i \}. 
\]
If $\forall a \in c_i, search\_path(a, \mathcal{C}_{\mathcal{N}}, avoid\_node\_set) \neq \emptyset$, then the decomposition of the MAPF instance into clusters is considered legal. 

In other words, if the agents in each cluster have paths that do not pass through the start or target states of agents in other clusters, this subproblem passes the mentioned solvability check, and the decomposition of the MAPF instance into clusters is considered legal.
\end{myTheo}

\begin{myProof}
If the agents in each cluster have paths that do not pass through the start or target states of agents in other clusters, then the decomposition is legal under the mentioned simplified scenario, and the decomposition of the MAPF instance into clusters is considered legal.

\end{myProof}

\begin{myTheo}
\textbf{Legal condition for decomposition into levels: }

If a MAPF instance is decomposed into multiple levels $L = \{l_1, l_2, \dots, l_n\}$, then for each $l_i \in L$, the $avoid\_node\_set$ is defined as

\[
avoid\_node\_set = \{s \mid s \in \{\bigcup_{j=0}^{i-1} T[l_j] \cup \bigcup_{j=i+1}^{m} S[l_j]\}\}. 
\]

If $\forall a \in l_i, search_path(a, \mathcal{C}_{\mathcal{N}}, avoid_node_set) \neq \emptyset$, then the decomposition of the MAPF instance into levels is considered legal.

In other words, if the agents in each level have paths that do not pass through the target states of agents in previous levels or the start states of agents in subsequent levels, then the decomposition into levels is considered legal.

\end{myTheo}

\begin{myProof}
If the agents in each level have paths that do not pass through the target states of agents in previous levels or the start states of agents in subsequent levels, then the decomposition is legal under the mentioned simplified scenario, and the decomposition of the MAPF instance into levels is considered legal.
\end{myProof}

It is noteworthy that the legal requirements for decomposing into clusters and levels do not guarantee the solvability of subproblems. The legal requirements are used to avoid unsolvable subproblems that cannot pass the mentioned solvability check.

\subsubsection{Connectivity graph}
As our main focus is on whether an agent's path passes through the start or target states of other agents, and to reduce the time cost of searching for paths, we divide the passable cells into two types: one being the agent's start or target state, and the other being the remaining cells, which we refer to as free cells.

We then use connected component detection to split all free cells into subsets (which we call free cell groups), where every cell in a group has a path to another cell in the same group without passing through any agent's start or target state, or cells from other groups.

We then propose a connectivity graph, $\mathcal{G}_c$, whose nodes consist of free cell groups and the start and target states of agents, with edges representing the connectivity between these nodes. Since $\mathcal{G}c$ has a smaller size than $\mathcal{C}\mathcal{N}$, the time cost of searching for paths can be significantly reduced, especially for instances with sparse agents.

In the implementation, since free cell groups do not influence whether an agent's dependent path passes through other agents' states, we simplify $\mathcal{G}_c$ by directly connecting the nodes representing agent states if they can be connected via free cell groups. After simplification, $\mathcal{G}c$ has $2k$ nodes, where $k$ is the number of agents. Compared to $\mathcal{C}\mathcal{N}$, the simplified $\mathcal{G}_c$ has a significantly smaller size, which accelerates the process of determining whether two agents can be in different subproblems. 

\begin{figure}[h] \scriptsize
\begin{minipage}{.32\linewidth}
  \centerline{\includegraphics[width=2.2cm]{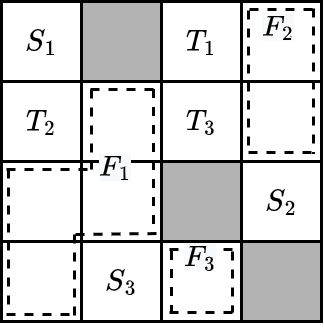}}
  \centerline{A}
\end{minipage}
\hfill
\begin{minipage}{.32\linewidth}
  \centerline{\includegraphics[width=3.2cm]{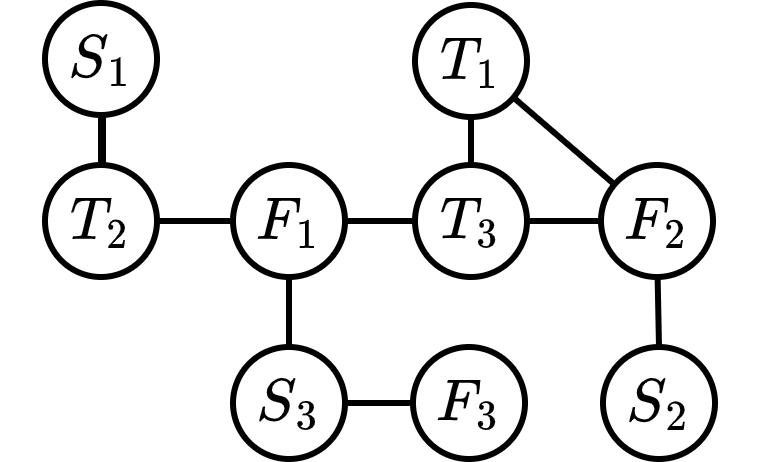}}
  \centerline{B}
\end{minipage}
\hfill
\begin{minipage}{.32\linewidth}
  \centerline{\includegraphics[width=3.2cm]{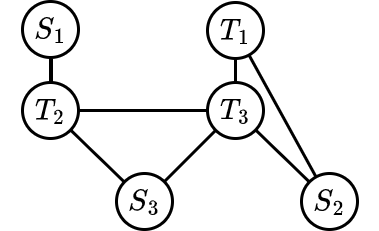}}
  \centerline{C}
\end{minipage}
\vfill
\caption{These figures show the free cell groups (marked as $F_1, F_2, \dots$, in Figure A), the connectivity graph $\mathcal{G}_c$ (in Figure B), and the simplified connectivity graph (in Figure C) of the instance in Fig. \ref{solvable} A. }
\label{connec_graph}
\end{figure}

An example of the connectivity graph is shown in Fig. \ref{connec_graph}.

\subsubsection{Variants of search path}
To reduce the time cost of searching for paths, we search for paths in $\mathcal{G}c$ rather than in $\mathcal{C}\mathcal{N}$. We denote the paths in $\mathcal{G}_c$ as dependence paths.

\begin{myDef}
The dependence path of an agent $a_i$ is defined as the path from $s_i$ to $t_i$, consisting of nodes in $\mathcal{G}_c$.
\end{myDef}

As an example, the dependence path for agent 1 in the instance shown in Fig. \ref{connec_graph} is ``$S_1 \rightarrow T_2 \rightarrow F_1 \rightarrow T_3 \rightarrow T_1$"; for agent 2, its dependence path is ``$S_2 \rightarrow F_2 \rightarrow T_3 \rightarrow F_1 \rightarrow T_2$"; and for agent 3, the dependence path is ``$S_3 \rightarrow F_1 \rightarrow T_3$".

The dependence path of an agent reflects its relationship with other agents. For a valid decomposition, each agent's path must not intersect with the targets of previous subproblems or the starts of later subproblems. The dependence paths of agents determine whether two agents must be solved in the same subproblem or the order in which they are solved across different subproblems. Specifically, there are four cases:

    1, if the dependence path of agent $a_i$ contains only the start of agent $a_j$, it indicates that $a_i$ should be solved after $a_j$ to enable they could be in different subproblems;
    
    2, if the dependence path of agent $a_i$ contains only the target of agent $a_j$, it implies that $a_i$ should be solved before $a_j$ to enable they could be in different subproblems;
    
    3, if the dependence path of agent $a_i$ contains both the start and target of agent $a_j$, it signifies that $a_i$ and $a_j$ must be solved in the same subproblem;
    
    4, if the dependence path of agent $a_i$ contains neither the start nor the target of agent $a_j$, it suggests that $a_i$ and $a_j$ can be solved in different subproblems, and the order of solving them does not matter.

It is worth noting that an agent's dependence path is not unique. By updating their dependence paths, we can alter the relationship between two agents (in implementation, we check for the existence of such paths rather than explicitly maintaining all agents' dependence paths). By changing the relationship between two agents, we can update the subproblems of a MAPF instance's decomposition. Considering that making an agent's dependence path avoid another agent might cause the path to pass through a third agent, and thus fail to generate smaller subproblems, we have no direct method to ensure that agents' dependence paths correspond to smaller subproblems, we have proposed several approaches to encourage the generation of smaller subproblems; their details can be found in Section \ref{Methodology}.

Here, we define two variants of the search path function used for decomposing the LA-MAPF instance into clusters:

\begin{myDef}
$search\_path\_agent(a_i, \mathcal{G}_c, available\_agents, avoid\_agents)$, where $available\_agents$ represents agents whose start or target states the dependence path of $a_i$ can pass through, and $avoid\_agents$ represents agents whose start or target states the dependence path of $a_i$ cannot pass through. If $available\_agents = \emptyset$, it means the path can pass through all nodes in $\mathcal{G}_c$, except those that belong to the start or target states of agents in $avoid\_agents$.

$search\_path\_agent$ performs a complete graph search on $\mathcal{G}_c$ and returns the agents whose dependence paths involve $a_i$. If no such dependence path exists, $search\_path\_agent$ returns $\emptyset$.

\end{myDef}

Similarly, we define a variant of $search\_path$ used in decomposing the instance into levels: 

\begin{myDef}
$search\_path\_SAT(a_i, \mathcal{G}_c, available\_SATs, avoid\_SATs)$, where $available\_SATs$ represents the start or target states that the dependence path of $a_i$ can pass through, and $avoid\_SATs$ represents the start or target states that the dependence path of $a_i$ cannot pass through. If $available\_SATs = \emptyset$, it means the path can pass through all nodes in $\mathcal{G}_c$, except those that belong to the start or target states of agents in $avoid_SATs$.

$search\_path\_SAT$ also performs a complete graph search on $\mathcal{G}_c$ and returns the start or target states of agents whose dependence paths are involved with $a_i$. If no such dependence path exists, $search\_path\_SAT$ returns $\emptyset$.
\end{myDef}

Here are some examples of the variants of the search path for the instance in Fig. \ref{connec_graph}: \ \centerline{$search\_path\_agent(a_1, \mathcal{G}_c, \{a_1, a_2, a_3\}, \emptyset) = \{a_1, a_2, a_3\}$;} \ \centerline{$search\_path\_agent(a_1, \mathcal{G}_c, \emptyset, \{a_2\}) = \emptyset$ (since $a_1$'s dependence path must pass through $T[a_2]$);} \ \centerline{$search\_path\_agent(a_1, \mathcal{G}_c, \{a_1, a_3\}, \emptyset) = \emptyset$ (since $a_1$'s dependence path must pass through $T[a_2]$);} \ \centerline{$search\_path\_SAT(a_1, \mathcal{G}_c, \emptyset, \emptyset) = \{S[a_1], T[a_1], T[a_2], T[a_3]\}$;} \ \centerline{$search\_path\_SAT(a_2, \mathcal{G}_c, \emptyset, \emptyset) = \{S[a_2], T[a_2], T[a_3]\}$;} \ \centerline{$search\_path\_SAT(a_2, \mathcal{G}_c, \emptyset, \{T[a_3]\}) = \emptyset$ (since $a_2$'s dependence path must pass through $T[a_3]$);} \ \centerline{$search\_path\_SAT(a_2, \mathcal{G}_c, \{S[a_2], T[a_2], T[a_3]\}, \emptyset) = \{S[a_2], T[a_2], T[a_3]\}$;} \\

There are two main advantages of $search\_path\_agent$ and $search\_path\_SAT$ compared to $search\_path$:

\begin{enumerate}
    \item $\mathcal{G}_c$ has a smaller size compared to $\mathcal{C}_\mathcal{N}$ (e.g., $\mathcal{C}_\mathcal{N}$ in Fig. \ref{connec_graph} has 4*4=16 nodes, while the related simplified $\mathcal{G}_c$ has only 2*3=6 nodes), so $search\_path\_agent$ and $search\_path\_SAT$ are faster compared to $search\_path$. 
    \item $search\_path\_agent$ and $search\_path\_SAT$'s return values provide direct information about the current agent's relationship with other agents, which is very useful in determining whether two agents must be in the same subproblem. 
\end{enumerate}

\section{Methodology}
\label{Methodology}

As mentioned previously, we have not identified a direct method to decompose a MAPF instance into multiple subproblems. Thus, we initiate the process with an initial decomposition, progressively refining it to generate smaller subproblems while ensuring that each step pass the mentioned solvability check. We define two types of subproblems encountered in various stages: clusters, which have no restrictions on the order of solving, and levels, which impose limitations on the order of solving.

The decomposition of a MAPF instance involves three steps:

1, identifying all agents' dependence path and establishing initial clusters based on them;
   
2, refining agent dependencies to bipartition clusters until they cannot be further divided into smaller clusters;
   
3, further refining agent dependencies to decompose clusters into levels and sorting.

Step 1 and 2 are described in Section \ref{Decomposition_to_clusters}, and step 3 is described in Section \ref{level_and_sorting}.

An overview of the decomposition process of a MAPF instance is depicted in Fig. \ref{overview}.

Following decomposition, we explore methods to solve subproblems independently and then combine their results to obtain a conflict-free solution for the original MAPF instance, as detailed in Section \ref{solve_and_combing}.

\subsection{Decomposetion to clusters}
\label{Decomposition_to_clusters}

The first step is to determine the connectivity graph $\mathcal{G}_c$ from a grid space $\mathcal{C}_{\mathcal{N}}$ and the related agents $\mathcal{A}$. We then determine initial clusters based on the dependence paths derived from the connectivity graph.

\begin{myDef}
Relevance of two agents: If an agent $a_i$'s dependence path contains another agent $a_j$'s start or target, we consider those two agents as relevant, regardless of whether $a_j$'s dependence path contains $a_i$'s start or target.
\end{myDef}

It is noteworthy that the relevance of two agents is determined by their dependence paths, so whether two agents are relevant changes as their dependence paths change.

\begin{myDef}
Graph of agents' relevance $\mathcal{G}_a$: an undirected graph where nodes represent agents and edges represent whether two agents are relevant.
\end{myDef}

Example about $\mathcal{G}_a$ can be found in Fig. \ref{bi_cluster}.

It is important to note that the relevance of two agents is determined by $avail\_agents$, $avoid\_agents$, and $\mathcal{G}_c$. While the subgraph $\mathcal{G}_c$ is constant, $avail\_agents$ and $avoid\_agents$ are variables, meaning that the relevance between two agents may change as these sets are modified.

Essentially, based on the definition of a cluster, a cluster is a maximal connected component in $\mathcal{G}_a$. Intuitively, a cluster is a set of agents from the raw MAPF instance, where any agents within it are only relevant to agents within the same set (based on their dependence paths). Additionally, we refer to an agent set that constitutes a cluster as an \textbf{independent} agent set.

\begin{figure}[h] \scriptsize
\centerline{\includegraphics[width=10.cm]{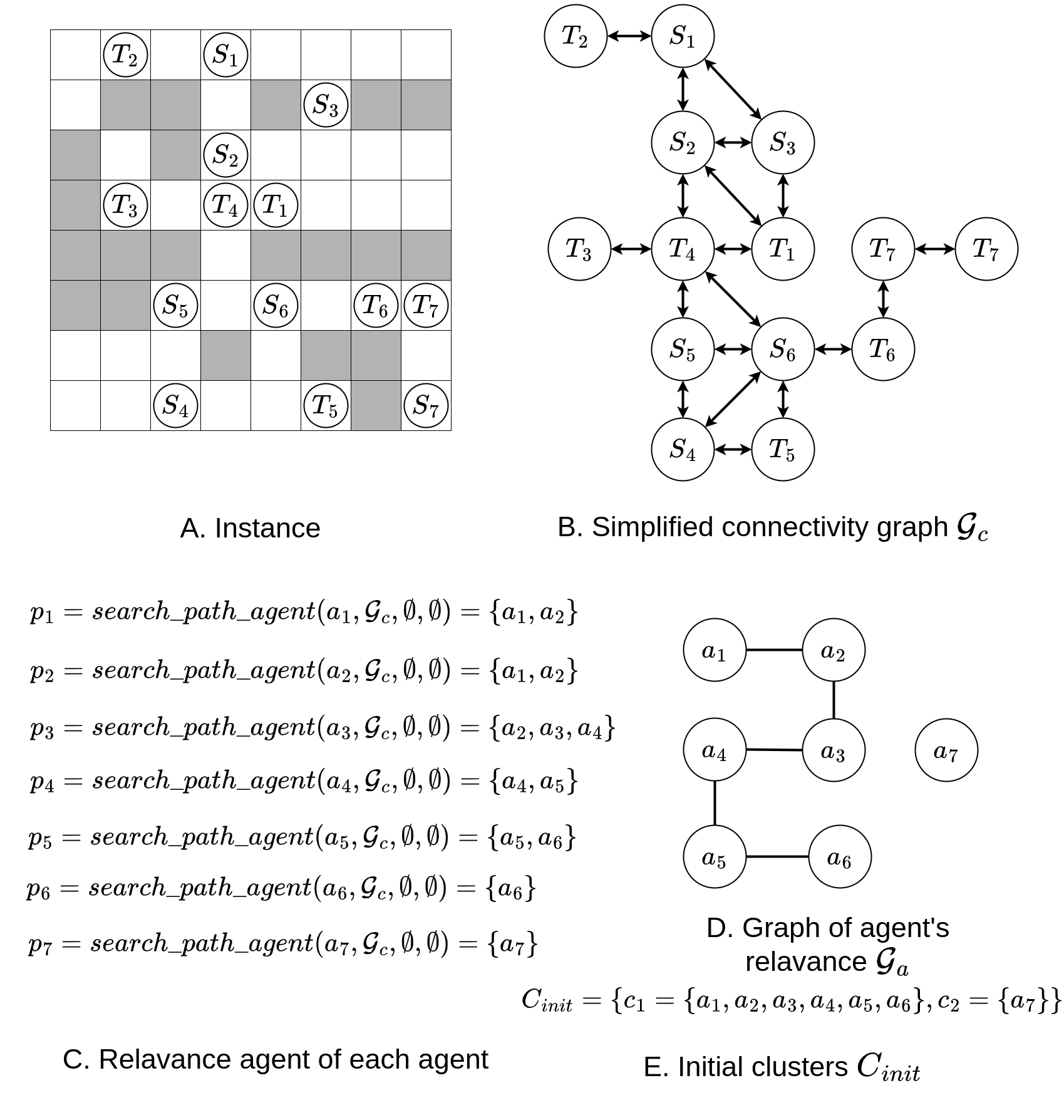}}
\caption{These figures show the initialization of clusters in a simple instance. Figure A illustrates the instance, where white cells represent passable cells and grey cells represent unpassable cells. The notations ``$S_i$'' and ``$T_i$'' denote the start and target states of agent $a_i$. \\
Figure B displays the simplified connectivity graph of agents. \\
Figure C shows the relevant agents for each agent, and Figure D presents the graph of agent relevance ($\mathcal{G}_a$). \\
Figure E shows the initial clusters determined by the graph of agent relevance, i.e., all connected components in $\mathcal{G}_a$. \\
} 
\label{init_cluster}
\end{figure}

\begin{figure}[h] \scriptsize
\centerline{\includegraphics[width=12.cm]{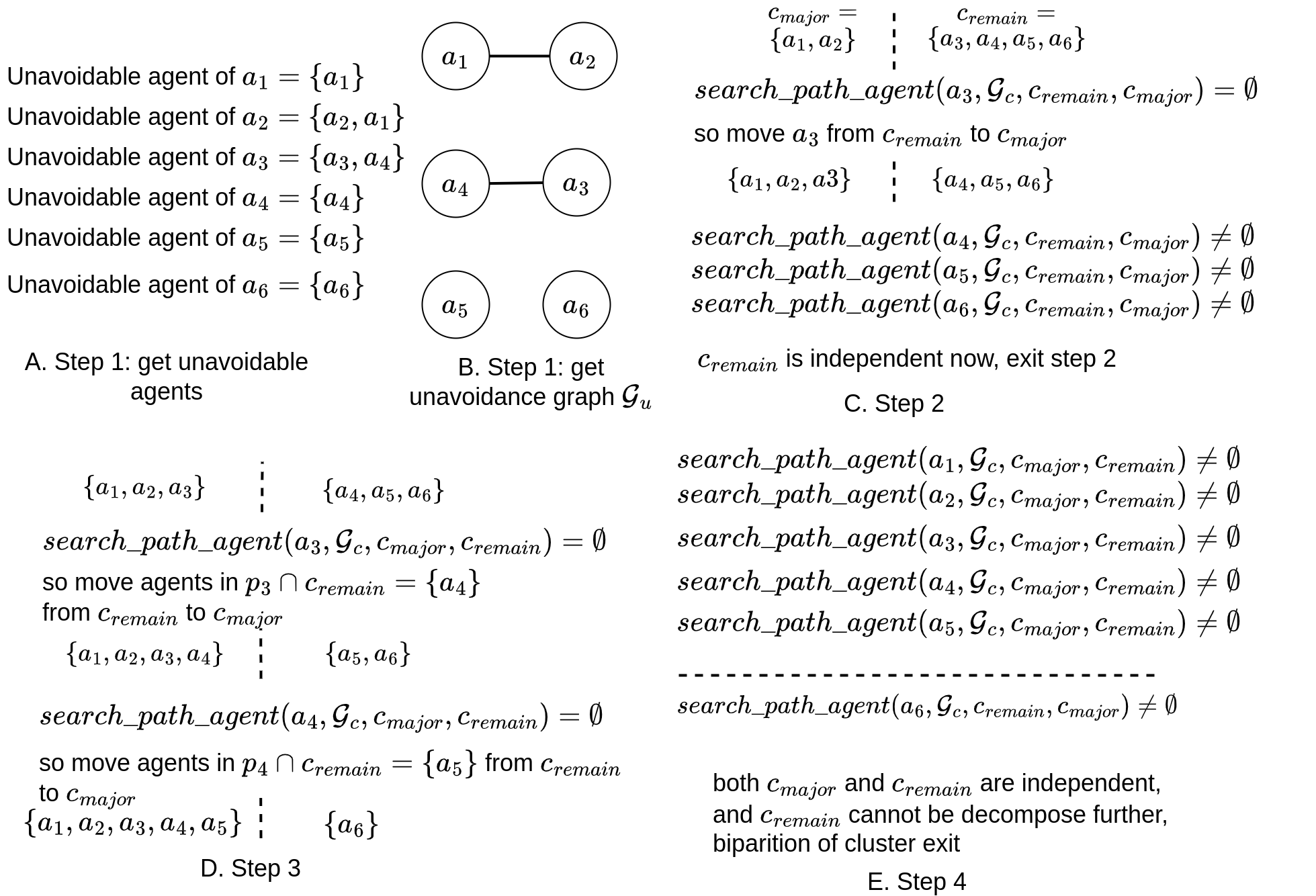}}
\caption{These figures demonstrate the process of bipartitioning cluster $C_1$, as mentioned in Fig. \ref{init_cluster}. \\
Figure A illustrates the unavoidable agents for each agent.
Figure B shows the related unavoidance graph $\mathcal{G}_u$.
Figures C, D, and E depict steps 2, 3, and 4 in the bipartition process of the cluster. }
\label{bi_cluster}
\end{figure}

An intuitive way to obtain clusters is by generating clusters from the graph of agent's relavance determined by the dependence paths that visit the fewest number of agents for each agent. We refer to such clusters as initial clusters(Line 1 $\sim$ 8 in Algorithm \ref{a1}). An example about initialize of clusters are shown in Fig. \ref{init_cluster}. 

However, it is evident that selecting dependence paths containing the fewest agents may not always result in the smallest possible clusters, as it does not consider how to guide dependence paths to avoid agents from other clusters. 

There is room for decomposing these initial clusters into smaller ones by updating the dependence paths of agents. Thus, we propose a method to iteratively bipartition initial clusters until further subdivision is not possible, aiming to minimize the size of subproblems. More details about bipartitioning can be found in the following section. This process ensures the legality of decomposition, and the more bipartitioning steps, the better the results obtained, although it does not guarantee the discovery of the optimal decomposition. An overview of the process of decomposing an instance into clusters is outlined in Algorithm \ref{a1}.

\begin{figure}[h] \scriptsize
\centerline{\includegraphics[width=10.cm]{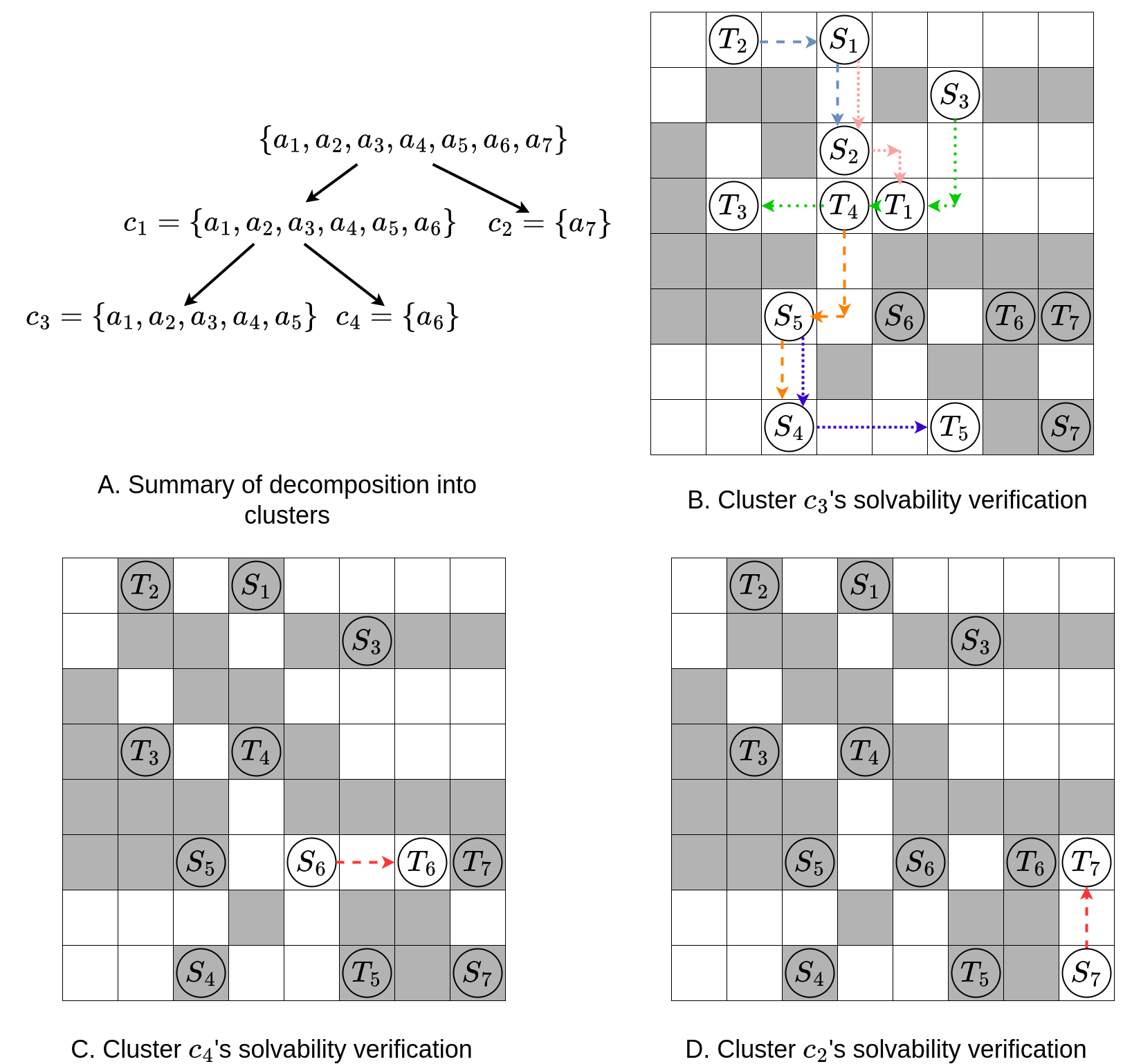}}
\caption{These figures show the verification of three clusters from the instance in Fig. \ref{bi_cluster}. Figure A illustrates the process of decomposing the instance into three clusters, similar to Fig. \ref{overview}.
Figures B, C, and D show the verification of the legality of the three clusters, i.e., checking whether there is a solution for each subproblem's agents while assuming the other clusters' agents occupy both their start and target states. Since we ensure that each cluster's agents are only related to the agents in the current cluster, all clusters are legal. The solutions for each agent are shown as dotted arrowed lines in different colors.
}
\label{veri_cluster}
\end{figure}


\begin{algorithm}
\caption{Decomposition instance to clusters}
\label{a1}
\begin{algorithmic}[1]
\Require $\mathcal{C}_{\mathcal{N}}, \mathcal{G}_c$
\Ensure $\mathcal{R}$
\State $//$ determine initial clusters
\State $P = \{p_1, p_2,...,p_k\}$; $//$ dependence path of all agents 
  \For {i = 1, 2,...,$k$} 
     \State {$p_i = search\_path\_agent(a_i, \mathcal{G}_c, \emptyset, \emptyset)$; }
  \EndFor  
  \State get $\mathcal{G}_a$ from $P$; 
  \State $//$ get initial clusters 
  \State $\mathcal{R}_{init}$ = all connected components of $\mathcal{G}_a$; 
  \State $\mathcal{R} = \emptyset$; $//$ final clusters 
  \For{$r \in \mathcal{R}_{init}$ } 
  	\State $temp\_cluster = r$; 
    \While{$temp\_cluster \neq \emptyset$} 
  		\State$\{r_{major}, r_{remain}\}$ = Bipartition of cluster ($temp\_cluster, \mathcal{G}_c$); 
  		\State add $r_{major}$ to $\mathcal{R}$; 
  		\State $temp\_cluster = r_{remain}$; 
  	\EndWhile
  \EndFor
  \State \textbf{return} $\mathcal{R}$;
\end{algorithmic}
\end{algorithm}

\subsection{Bipartition of clusters}
This section is dedicated to decomposing a cluster into two smaller clusters. However, before delving into the decomposition process, it's essential to introduce some necessary concepts.

\begin{algorithm}
\caption{Bipartition of cluster}
\label{a2}
\begin{algorithmic}[1]
\Require $r, \mathcal{G}_c$
\Ensure $\{r_{major}, r_{remain}\}$
   \State $//$ step 1
   \State $\mathcal{G}_u = \emptyset$; 
   \For{$a_i \in r$} 
     \For{$a_j \in r$} 
     \If{$a_i \neq a_j$} 
         \If{$search\_path\_agent(a_i, \mathcal{G}_c, r, \{a_j\}) = \emptyset$ } 
         \State add $(a_i \rightarrow a_j)$ as an edge to $\mathcal{G}_u$; 
         \EndIf  		
     \EndIf    
     \EndFor
   \EndFor
   \State $r_{major}$ = largest connected component of $\mathcal{G}_u$; 
   \State $r_{remain} = r - r_{major}$; 
   \While{$True$} 
     \State $//$ step 2 
     \State $new\_agents = \emptyset$; 
     \While{$True$} 
     \For{$a_i \in r_{remain}$} 
     \If{$search\_path\_agent(a_i, \mathcal{G}_c, r_{remain}$, $r_{major}$ $) = \emptyset$} 
         \State move $a_i$ from $r_{remain}$ to $r_{major}$; 
         \State add $a_i$ to $new\_agents$; 
     \EndIf
     \EndFor
     \If{$r_{remain}$ is independent $or$ $r_{remain}$ is empty} 
     \State break; 
     \EndIf
     \EndWhile
     \State $//$ step 3 
     \While{$True$} 
     \For{$a_j \in new\_agents$} 
     \If{$search\_path\_agent(a_i, \mathcal{G}_c, r_{major}, r_{remain}) = \emptyset$} 
         \State $//$ consider $a_i$ belong to cluster $r$, following search path always success 
         \State $p_i = search\_path\_agent(a_i, \mathcal{G}_c, r, \emptyset$); 
         \State move agents in ($p_i \cap r_{remain}$) from $r_{remain}$ to $r_{major}$; 
     \EndIf
     \EndFor
     \If{$r_{major}$ is independent}  
     \State break;
     \EndIf 
     \EndWhile
     \State $//$  step 4 
     \If{both $r_{major}$ and $r_{remain}$ are independent} 
     \State break;  	
     \EndIf  
   \EndWhile
   \State \textbf{return} $\{r_{major}, r_{remain}\}$;
\end{algorithmic}
\end{algorithm}

\begin{myDef}
Unavoidable agents of an agent $a$ in a cluster $r$ refer to those agents within $r$ that must be traversed by $a$'s dependence path. In other words, these agents must belong to the same cluster and cannot be further divided.

\end{myDef}

For example, for the instance in Fig. \ref{connec_graph}, $a_2$ and $a_3$ are unavoidable agents of $a_1$, as $a_1$'s dependence path must pass through $T[a_2]$ and $T[a_3]$.

\begin{myDef}
Unavoidance graph $\mathcal{G}_u$ of a cluster: an undirected representation of whether one agent is unavoidable to another agent within the cluster. Assume there are two agent $a_i$ and $a_j$, $a_i$ is unavoidable to $a_j$ if $a_j$'s dependence path must pass $a_i$'s start or target, meanwhile all dependence paths cannot pass a agent's start or target if the agent is not in current cluster. It is important to note that for a given cluster $r$, its unavoidance graph is unique.

\end{myDef}

An example of $\mathcal{G}_u$ can be found in Fig. \ref{bi_cluster}.

Both the unavoidance graph and the relevance graph depict relationships between agents within the cluster. 

\begin{myDef}
Maximum unavoidable agents of cluster: the largest connected component of the unavoidance graph $\mathcal{G}_u$ associated with the cluster. 

Intuitively, these maximum unavoidable agents represent the largest undividable subset within the current cluster, are referred to as ``major set" during the cluster bipartition process. Agents within the cluster, excluding those belonging to the maximum unavoidable agents, are referred to as ``remaining set" during the cluster bipartition process. An example of the maximum unavoidable agents of a cluster can be found in Fig. \ref{bi_cluster}.

It is noteworthy that during the bipartition process, agents in the remaining set will be moved to the major set, but agents in the remaining set will not move into the major set until both the major set and the remaining set are independent. 

\end{myDef}

\begin{algorithm}
\caption{Decomposing to levels and sorting}
\label{a3}
\begin{algorithmic}[1]
\Require $r, \mathcal{G}_c$
\Ensure $\mathcal{L}$
   \State $//$ determine initial dependence paths 
   \State $P = \emptyset$; $//$ $P[a]$ means agent a's denpendence path 
   \For{agent $a$ in $r$} 
   \State $p_i = search\_path\_SAT(a_i, \mathcal{G}_c, r, \emptyset)$; 
   \State add $p_i$ to $P$; 
   \EndFor
   \State get $\mathcal{G}_s$ from $P$; 
   \State determine all loop in $\mathcal{G}_s$ via strong component detect; 
   \State $//$ sort levels; 
   \State construct graph $\mathcal{G}_l$ via $\mathcal{G}_s$; 
   \State $root\_levels$ = level in $\mathcal{G}_l$ that not later than other level; 
   \State set order of all level to 0; 
   \State $levels$ = $root\_levels$; 
   \While{levels is not empty} 
   \State$next\_levels$ = $\emptyset$; 
   \For{level in levels} 
   \For{another\_level that later than level in $\mathcal{G}_l$} 
   \If{$another\_level$'s order $\leq$ level's order } 
    \State $another\_level$'s order = level's order + 1; 
    \State add $another\_level$ to $next\_levels$; 
   \EndIf
   \EndFor
   \EndFor
   \State $levels$ = $next\_levels$;
   \EndWhile
   \State \textbf{return} all levels in the order of last visit to them; 
\end{algorithmic}
\end{algorithm}
The bipartition of a cluster comprises four steps:

1, identify the maximum unavoidable agents of the cluster, referred to as the major set (Line 1 $\sim$ 12 in Algorithm \ref{a2});
   
2, examine each agent in the remaining set to determine if its dependence path must pass through agents of the major set. Move agents meeting this criterion to the major set, until the remain set is independent or empty(Line 16 $\sim$ 27 in Algorithm \ref{a2});

3, check any newly added agents to the major set to ascertain if their dependence paths must pass through agents in the remaining set. Transfer such agents from agents the remaining set to the major set, until the major set is independent(Line 29 $\sim$ 40 in Algorithm \ref{a2});

4, verify whether the major set and the remaining set both meet the legality requirements of the cluster (for every agent in a cluster, its dependence path does not pass through the start or target states of agents in other clusters). If they do (or if the remaining set is empty), exit and return the major set and the remaining set as the result of the bipartition. (Line 42 $\sim$ 44 in Algorithm \ref{a2}) Otherwise, proceed to step 2;

The bipartition process of a cluster concludes when both sets satisfy the cluster's requirements, or when there are no remaining agents, indicating that the cluster cannot be further decomposed into smaller clusters.

The pseudocode for bipartitioning a cluster is outlined in Algorithm \ref{a2}. An example of the bipartition of a cluster is shown in Fig. \ref{bi_cluster}. The verification of the legality of these clusters is shown in Fig. \ref{veri_cluster}.

\begin{algorithm}
\caption{Merge results}
\label{a4}
\begin{algorithmic}[1]
\Require $\{P_1,...,P_m\}$ $//$ assume there are $m$ subproblems and their solution is $P_i$
\Ensure $\{P_1,...,P_m\}$
   \State set last occupied time of all cells to 0; 
   \For{each solution $P$ in $\{P_1,...,P_m\}$} 
   \While{True} 
   \State $t$=1; $//$ current time index 
   \State $//$ check whether insert wait action to $P$ 
   \State $need\_wait$ = False; 
   \State $all\_finished$ = True; 
   \For{path $p$ in $P$} 
       \If{t $>$ size of $p - 1$}
       \State $//$ don't check path that have reach target   
       \State continue; 
       \Else 
       \State $all\_finished$ = False;      
       \EndIf
       \State $//$ $p[t]$ means path $p$'s location at time $t$ 
       \If{last occupied time of $p[t] \geq t$} 
       \State $need\_wait$ = True; 
       \State break; 
       \EndIf
   \EndFor
   \If{all\_finished} 
   \State break;
   \EndIf
   \State $//$ insert wait action to all path in $P$ 
   \If{need\_wait} 
   \For{path $p$ in $P$} 
       \If{t $>$ size of $p - 1$} 
         \State continue;
       \EndIf
       \State repeat $p[t]$ in $p$ at time index $t$; 
   \EndFor
   \Else
   \State $//$ update last occupied time of cells
   \For{path $p$ in $P$} 
       \If{t $>$ size of $p - 1$} 
           \State continue; 
       \EndIf
       \State set last occupied time of $p[t]$ to $t$; 
   \EndFor
   \EndIf
   \State $t$ = $t$ + 1; 
   \EndWhile
   \EndFor
   \State \textbf{return} $\{P_1,...,P_m\}$; 
\end{algorithmic}
\end{algorithm}

\begin{algorithm}
\caption{Layered MAPF}
\label{a5}
\begin{algorithmic}[1]
\Require $\mathcal{C}_{\mathcal{N}}, A$, MAPF
\Ensure $P$
   \State $//$ 1, decomposition of MAPF instance 
   \State construct $\mathcal{G}_c$ from $\mathcal{C}_{\mathcal{N}}, A$; 
    \State $R$ = Decomposing to cluster($\mathcal{G}_c$); 
    \State $L = \emptyset$; 
    \For{cluster $r$ in $R$} 
    \State $L'$ = Decomposing to leve and sorting($r$, $\mathcal{G}_c$);  
    \State merge $L'$ to $L$; 	   
    \EndFor
    \State $all\_subproblem\_paths$ = $\emptyset$; 
    \State $//$ 2, solve subproblems 
    \For{$l$ in $L$} 
    \If{MAPF is serial} 
    \State set later subproblem's start in $\mathcal{C}_{\mathcal{N}}$ to unpassable; 
    \State $current\_solution$ = MAPF($\mathcal{C}_{\mathcal{N}}, l, all\_subproblem\_paths$);  
    \State add $current\_solution$ to $all\_subproblem\_paths$; 
    \State reset $\mathcal{C}_{\mathcal{N}}$; 
    \Else
      \State set previous subproblem's target and later subproblem's start in $\mathcal{C}_{\mathcal{N}}$ to unpassable; 
      \State $current\_solution$ = MAPF($\mathcal{C}_{\mathcal{N}}, l$); 
      \State add $current\_solution$ to $all\_subproblem\_paths$; 
      \State reset $\mathcal{C}_{\mathcal{N}}$; 
    \EndIf
    \EndFor
    \State $//$ 3, merge solutions of subproblems 
    \If{MAPF is serial}
    \State $P$ = $all\_subproblem\_paths$;  
    \Else
    \State $P$ = Merge results($all\_subproblem\_paths$);    
    \EndIf
    \State \textbf{return} $P$; 
\end{algorithmic}
\end{algorithm}
\subsection{Decomposing to levels and sorting}
\label{level_and_sorting}

Clusters can be solved irrespective of the order of solving, thus providing an opportunity to decompose MAPF into smaller problems by considering the order of solving. By considering the order of solving, we can divide two agents into different subproblems even if one agent's dependence path contains the start or target of another agent. We refer to these smaller problems decomposed from clusters as levels. In this section, we focus on how to decompose clusters into levels and determine the order of solving. To facilitate this discussion, we introduce some new concepts.

\begin{myDef}
Order of solving agents: If an agent $a_i$ must be solved before another agent $a_j$, we denote it as $a_i > a_j$; if $a_i$ must be solved later than $a_j$, we denote it as $a_i < a_j$.
\end{myDef}

It is important to note that both $a_i > a_j$ and $a_i < a_j$ can coexist (indicating that both agents must be solved in the same problem) or neither can exist (indicating no order limitation between solving $a_i$ and $a_j$).

The order of solving agents is determined by whether an agent's dependence path contains another agent's start or target. If $a_i$'s dependence path contains $a_j$'s start, it implies that $a_i$ must be solved later than $a_j$ to ensure that $a_j$'s start is not occupied (to enable $a_i$ and $a_j$ could be in different subproblem). Conversely, if $a_i$'s dependence path contains $a_j$'s target, $a_i$ must be solved before $a_j$ to ensure that $a_j$'s target is not occupied. Furthermore, the order of solving levels is determined by the order of solving agents within them.

\begin{myDef}
Solving order graph $\mathcal{G}_s$ of a cluster: a directed graph representing whether an agent must be solved before another agent. Its nodes are agents, and edges indicate the order of solving between agents. The structure of the solving order graph is determined by each agent's dependence path. Specifically, if agent $a_i$ must be solved before agent $a_j$, the corresponding edge in $\mathcal{G}_s$ is denoted as $a_i \rightarrow a_j$.
\end{myDef}

\begin{figure}[h] \scriptsize
\centerline{\includegraphics[width=12.cm]{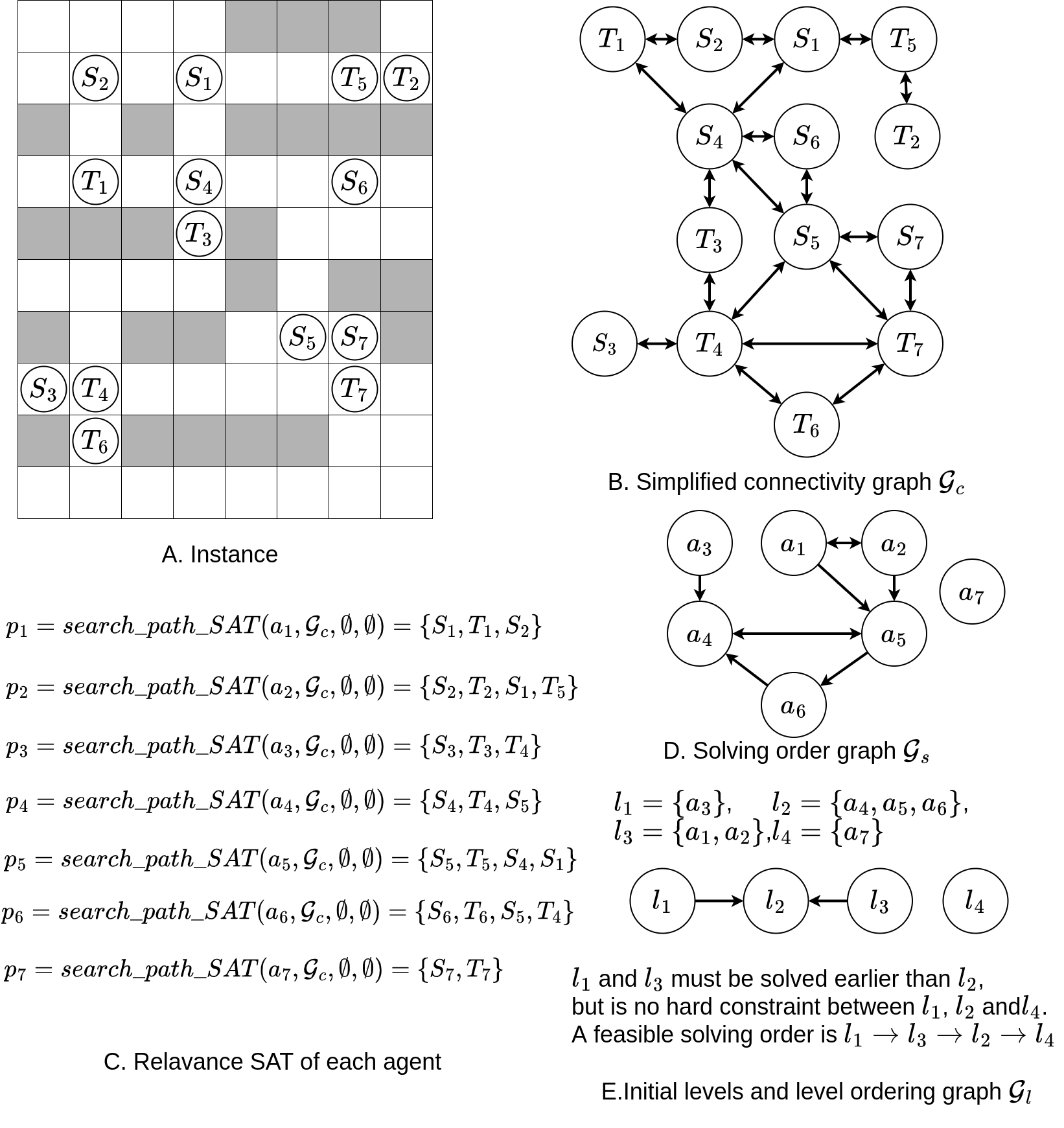}}
\caption{These figures illustrate the decomposition of a MAPF instance into levels in a simple scenario. Figure A shows the instance, where ``$S_i$'' and ``$T_i$'' represent the start state and target state of agent $a_i$. \\ Figure B presents the simplified connectivity graph of agents. \\ Figure C depicts the relevant SATs for each agent, while Figure D shows the solving order graph ($\mathcal{G}_s$). \\ Figure E displays the initial levels determined by the solving order graph, i.e., all strongly connected components in $\mathcal{G}_s$, along with the related level ordering graph $\mathcal{G}_l$.
}
\label{init_level}
\end{figure}

An example of the solving order graph can be found in Fig. \ref{init_level}.

\begin{figure}[h] \scriptsize
\centerline{\includegraphics[width=10.cm]{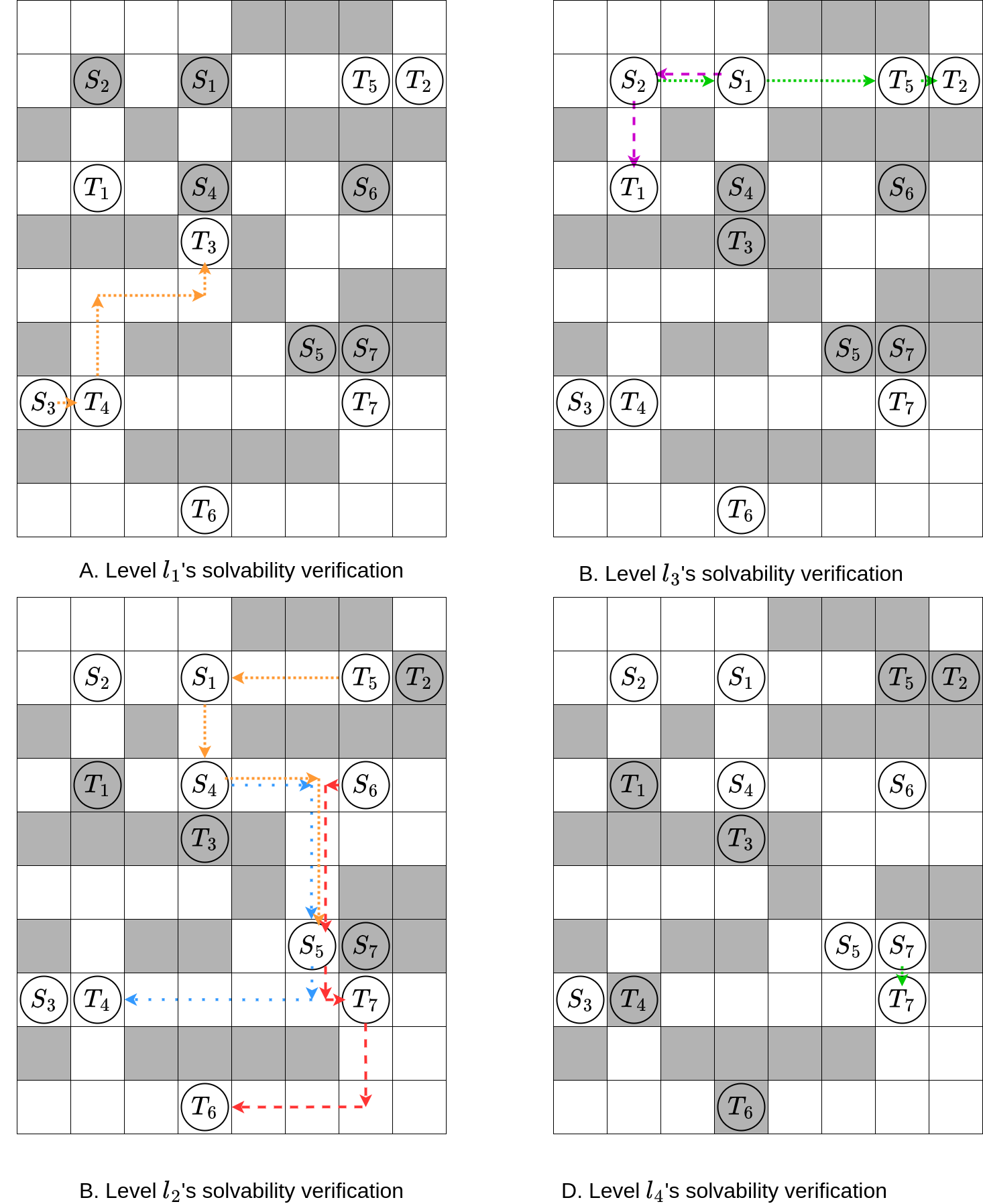}}
\caption{These figures show the verification of four levels from the instance in Fig. \ref{init_cluster}. We check whether there is a solution for each level's agents while assuming that the previous level's target states and the next level's start states are occupied. Since we ensure that each level's agents are not related to the previous level's target state and the next level's start state, all levels are considered legal. The corresponding solutions for each agent are shown as dotted arrowed lines in different colors.
}
\label{veri_level}
\end{figure}

Essentially, a level is a strongly connected component of $\mathcal{G}_s$. Intuitively, a level is a group of agents that form a loop in $\mathcal{G}_s$, meaning that the agents within the loop must be solved simultaneously. For example, if three agents, $A$, $B$, and $C$, are in the same loop such that $A \rightarrow B \rightarrow C \rightarrow A$, it implies that $A$ must be solved before $B$, $B$ before $C$, and $C$ before $A$. Therefore, $A$, $B$, and $C$ must be solved together in parallel. Unlike clusters, where the order of solving is arbitrary, the order of solving levels is strictly determined.

The order of solve levels is determine by the edge in $\mathcal{G}_s$ that connect them.

The order of solving levels is determined by the edges in $\mathcal{G}_s$ that connect them.

\begin{myDef}
Order of solving levels: if a level $l_i$ must be solved before another level $l_j$, we denote it as $l_i > l_j$; if $l_i$ must be solved later than $l_j$, we denote it as $l_i < l_j$.
\end{myDef}

\begin{myDef}
Level ordering graph $\mathcal{G}_l$ is a directed graph whose nodes are levels, and edges indicate whether a level must be solved earlier than another level. If there is no edge connecting two levels, this implies that there is no explicit order to solve them, although there may be an implicit order. For example, if level $A$ has no edge connecting to level $C$, but level $A$ is connected to level $B$ (A $>$ B) and level $B$ is connected to level $C$ (B $>$ C), it implies that level $A$ must be solved earlier than level $C$.

Essentially, $\mathcal{G}_l$ serves as a condensed version of $\mathcal{G}_s$.
\end{myDef}

There are five steps involved in decomposing a cluster into multiple levels and determining the order of solving:

1, determine each agent's dependence path, and available agents constrained to the agents of the current cluster (Line 1 $\sim$ 6 in Algorithm \ref{a3});

2, obtain the solving order graph $\mathcal{G}_s$ from the dependence paths of the current cluster(Line 7 in Algorithm \ref{a3});

3, identify all strong components of $\mathcal{G}_s$ as levels (Line 8 in Algorithm \ref{a3});

4, determine the relationships between levels by examining the edges that connect them. Construct the level ordering graph $\mathcal{G}_l$, which represents the order of solving between level (Line 9 $\sim$ 10 in Algorithm \ref{a3});

5, set levels that are not later than any other levels as root levels. Then, using Breadth First Search, traverse all levels in $\mathcal{G}_l$, where the order of the last visit to each level represents the order of solving them (Line 11 $\sim$ 25 in Algorithm \ref{a3}).

In implementation, Tarjan's algorithm \cite{tarjan1972depth} is utilized to determine strongly connected components from a directed graph.

It is important to note that after completing all steps of decomposition, we ensure that the final subproblems do not compromise legality. The entire process of decomposing into levels and sorting them is depicted in Algorithm \ref{a3}. An example of decomposing a cluster into levels can be found in Fig. \ref{init_level}.

\subsection{Solving and Combining}
\label{solve_and_combing}

After completing the decomposition of a MAPF instance, the next step involves considering how to solve the subproblems separately and combine their results without conflicts. There are two main types of MAPF methods: serial MAPF methods and parallel MAPF methods. Serial MAPF methods, such as CBS-based methods, can use external paths as constraints to avoid conflicts, as they search for an agent's path while the paths of other agents remain static. On the other hand, parallel MAPF methods, such as LaCAM and PIBT, cannot always treat external paths as dynamic obstacles to avoid, as they may avoid the same state (all agents' locations) occur multiple times. For simplicity in this article, we do not set external paths as dynamic obstacles to avoid for all parallel MAPF methods.

In serial MAPF methods, we only need to merge separate results into a single set. However, for the second type of MAPF method, we need to add wait actions in the resulting paths to avoid conflicts between paths from different subproblems, as shown in Algorithm \ref{a4}. 

In detail, this algorithm maintains a table that records the last visited time step for each cell, in order to determine whether a subproblem's solution collides with the solution of a previous subproblem. Starting from $t=1$, it checks each agent's solution to see if it conflicts with the solution of a previous subproblem (Lines 1 to 20 in Algorithm \ref{a4}). If a solution collides with another solution at time $t$, a wait action is added at $t$ to resolve the conflict (Lines 24 to 32 in Algorithm \ref{a4}). Otherwise, the table is updated, and the algorithm checks whether there is a conflict with other solutions at time $t+1$ (Lines 33 to 39 in Algorithm \ref{a4}), continuing until the last state is reached (i.e., the process is finished). Once all solutions are completed, the algorithm exits (Lines 21 to 23 in Algorithm \ref{a4}).

In the following, we define serial MAPF as \textit{MAPF}$(\mathcal{C}_\mathcal{N}, \mathcal{A}, P)$, and parallel MAPF methods as \textit{MAPF}$(\mathcal{C}_\mathcal{N}, \mathcal{A})$, where $\mathcal{A}$ represents agents and $P$ represents external paths that need to be avoided.

The overall process of decomposing a MAPF instance into multiple subproblems (Line 1 $\sim$ 9 in Algorithm \ref{a5}), solving the subproblems (Line 10 $\sim$ 23 in Algorithm \ref{a5}), and merging their results (Line 24 $\sim$ 29 in Algorithm \ref{a5}) is illustrated in Algorithm \ref{a5}.

As mentioned earlier, our algorithm aims to minimize the loss of solvability in subproblems by introducing a legality check. Although we have not provided a theoretical analysis regarding the percentage of solvability loss, we offer empirical analysis based on extensive testing. This is detailed in the comparison between Layered LaCAM2 and raw LaCAM2 in Section \ref{application}.

\subsection{Time complexity analysis}
\noindent
$\bullet$ Construction of connectivity graph $\mathcal{G}_c$

As mentioned in ``Data Structures and Network Algorithms''\cite{tarjan1983data}, since Tarjan's algorithm visits each node and edge only once, the time complexity of detecting connected components in a graph $G(V,E)$ is $\mathcal{O}(|V|+|E|)$. 

Assuming a MAPF instance ($\mathcal{N}$ = 2), the map's width and height are $w$ and $h$, respectively. It contains at most $w \times h$ cells and $4 \times w \times h$ edges (when all cells are passable). Therefore, the time complexity to determine the $\mathcal{G}_c$ of an instance is $\mathcal{O}(w \times h)$. By extension, for a MAPF instance with $\mathcal{N} > 2$, the time complexity to determine $\mathcal{G}_c$ is $\mathcal{O}(n)$, where $n$ represents the total number of states (e.g., for $\mathcal{N}$ = 2, $n$ is the area of the map; for $\mathcal{N}$ = 3, $n$ is the volume of the map).

\noindent
$\bullet$ Variants of path search

As mentioned earlier, the number of nodes in $\mathcal{G}_c$ is $2k$, where $k$ is the number of agents. In the worst case, the number of edges in $\mathcal{G}_c$ is $\frac{(2k)(2k-1)}{2}$. According to Introduction to Algorithms \cite{cormen2022introduction}, the time complexity of BFS is $\mathcal{O}(|V| + |E|)$ (where $|V|$ and $|E|$ represent the sizes of the nodes and edges, respectively). Therefore, the time complexity of $search_path_agent$ and $search_path_SAT$ is $\mathcal{O}(k^2)$.

\noindent 
$\bullet$ Decomposing to clusters

In initialization of clusters, every agents need using $search\_path\_agent$ one time. So the time complexity of $search\_path\_agent$ are $k*\mathcal{O}(k^2)=\mathcal{O}(k^3)$. 

Considering there are $k$ nodes and at most $\frac{k(k-1)}{2}$ edges in the graph of an agent's relevance, $\mathcal{G}_a$, and that the time complexity of detecting the connected components of a graph $G(V, E)$ is $\mathcal{O}(k^2)$.

In the best case, there are $k$ clusters, and each cluster has only one agent after detecting the connected components of $\mathcal{G}_a$, so no further decomposition is needed. Therefore, the total time complexity of the decomposition in the best case is $\mathcal{O}(k^3)$ (this usually happens in maps with sparse agents, such as the ``warehouse-20-40-10-2-2" map and ``Boston\_0\_256" in Section \ref{Resultsofdecomposition}). However, in the worst case, further decomposition is needed, such as the bipartition of clusters and decomposition into levels (like the ``random-32-32-20" map and ``room-32-32-4" in Section \ref{Resultsofdecomposition}).

In the bipartition of clusters, we first determine the complexity of decomposing a cluster into two clusters. Assuming there are $m$ agents in a cluster, in the worst case, it takes $m-1$ calls to $search_path_agent$ to determine that one agent needs to be moved from the remaining set to the major set, and $m-1$ agents need to be moved from the remaining set to the major set. Therefore, the time complexity of splitting one cluster into two clusters is $(m-1)^2 \times \mathcal{O}(k^2) = \mathcal{O}(m^2 k^2)$.

\begin{figure}[h] 
\centerline{\includegraphics[width=5.0cm]{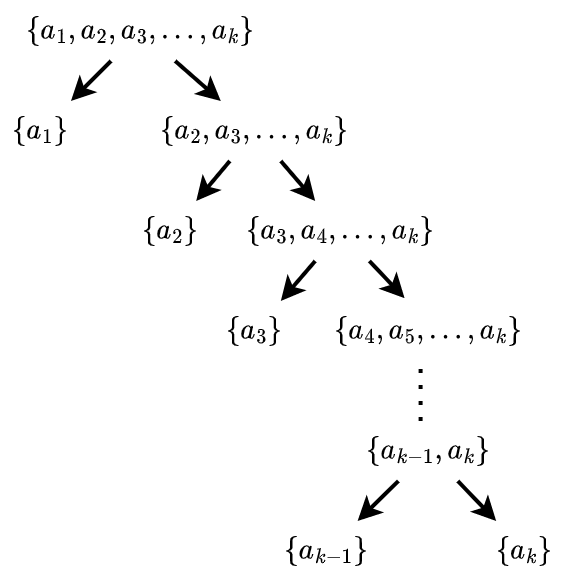}}
\caption{
This figure illustrates the case that maximize the time cost of biparition of cluster. 
}
\label{worst_case}
\end{figure}

Considering that the larger the cluster, the greater the time cost of splitting it, the worst case in the bipartition of clusters occurs when the major set always contains only one agent (i.e., maximizing the number of agents in the remaining set) until the remaining set is empty (as illustrated in Fig. \ref{worst_case}). This requires $m-1$ splits of a cluster into two clusters, where the number of agents in the clusters is $1, 2, \dots, k$. Therefore, the time complexity of the bipartition of a cluster in the worst case is: $\sum^{k}_{m=1} \mathcal{O}(m^2 k^2) = \mathcal{O}(\frac{k(k+1)(2k+1)}{6}(k^2)) = \mathcal{O}(k^5)$.

\noindent
$\bullet$ Decomposition to level and sorting

In the initialization of levels, each agent needs to use $search_path_SAT$ once. For a cluster with $m$ agents, the time complexity of $search_path_SAT$ is $m \times \mathcal{O}(k^2) = \mathcal{O}(mk^2)$.

Considering there are $m$ nodes and at most $m(m-1)$ edges in $\mathcal{G}_s$, and that in the construction of $\mathcal{G}_l$, we need to run Tarjan's algorithm on $\mathcal{G}_s$, whose time complexity is $\mathcal{O}(|V| + |E|)$. Therefore, the time complexity for constructing $\mathcal{G}_l$ is $\mathcal{O}(m^2)$.

The time complexity of level sorting is determined by a BFS search, so its complexity depends on the size of the level-ordering graph. In the worst case, each level contains only one agent, and there are $m$ levels in $\mathcal{G}_l$, resulting in at most $m(m-1)$ edges in $\mathcal{G}_l$. Since the time complexity of both DFS and BFS is $\mathcal{O}(|V| + |E|)$, the time complexity of level sorting in the worst case is $\mathcal{O}(m^2)$.

Considering that $m \leq k$, the time complexity for decomposing a cluster with $m$ agents into levels is $\mathcal{O}(mk^2) + \mathcal{O}(m^2) + \mathcal{O}(m^2) = \mathcal{O}(mk^2)$.

So, in summary, considering the total number of agents in all clusters is $k$, the worst-case time complexity of decomposing all clusters into levels is $\mathcal{O}(k^3)$.

\noindent
$\bullet$ Combining results

As mentioned before, serial MAPF methods can take extra paths as constraints to avoid (e.g., EECBS), so no extra actions are needed to merge subproblem solutions. However, parallel MAPF methods (e.g., LaCAM) need to add wait actions to avoid conflicts between the solutions of different subproblems.

For a MAPF instance with $k$ agents, assuming there are $m$ subproblems, let the number of agents in the $i$-th subproblem be denoted as $c_i$, and the makespan of their solution as $s_i$ for $0 < i < m$. In the worst case, every subproblem needs to add wait actions equal to the sum of the previous subproblem's makespan to ensure collision-free execution (i.e., waiting until all agents from previous subproblems have arrived at their targets, except for the first subproblem, as described in the \textit{simplified scenario}). Thus, the total number of added wait actions in the worst case is $\sum_{i=1}^{m-1} c_i \cdot s_{i-1}$. Assuming there is an upper bound on the makespan, $T$, we have: $\sum_{i=1}^{m-1} c_i*s_{i-1} <= \sum_{i=1}^{m-1} c_i*T < \sum_{i=1}^{m-1} k*T = (m-1)k*T < k^2T$. Therefore, the time complexity in the worst case is $\mathcal{O}(k^2T)$.

It is noteworthy that the makespan of solutions varies across different MAPF methods, so the time cost of merging results also differs depending on the method used.
\section{Results of decomposition}
\label{Resultsofdecomposition}

In theory, the decomposition of a MAPF instance reduces the cost of solving the instance, but it also incurs its own time and memory costs. Therefore, before analyzing how the decomposition of a MAPF instance contributes to solving the instance, it is essential to examine the time and memory usage of the decomposition process.

Specifically, we measure the peak memory usage as the memory usage, and record the memory usage of the program at every millisecond during implementation. However, due to the resolution of memory recording being 1KB, memory usage less than 1 KB is recorded as 0 MB. This phenomenon may occur under certain maps.

Furthermore, we analyze how each step of the decomposition process influences the overall decomposition by evaluating changes in the decomposition rate (maximum subproblem size / total number of agents) and the number of subproblems after each step. For simplicity, the initialization of clusters is denoted as ``IC", the bipartition of clusters as ``BC", and decomposing to levels and sorting as ``LS".

We employ a classic MAPF dataset \cite{stern2019mapf} as MAPF instances for our analysis, comprising 24 typical maps from the dataset. These maps feature an increasing number of agents, and each number of agents is randomly selected 100 times, resulting in a total of 22,300 MAPF instances. The same instances are utilized to evaluate how decomposition influences typical MAPF methods.

Experiments were conducted on a laptop running Ubuntu 20.04, equipped with a Ryzen 7 5800h (3.2GHz) CPU and 16GB of memory. All code is implemented in C++.

\begin{figure}[h] \tiny

\begin{minipage}{.04\linewidth}
\centerline{ }
\end{minipage}
\hfill
\begin{minipage}{.11\linewidth}
\leftline{\scalebox{0.8}{1.empty-16-16}}
\leftline{\scalebox{0.8}{16*16 (256)}}
\end{minipage}
\hfill
\begin{minipage}{.04\linewidth}
\leftline{\includegraphics[width=.5cm]{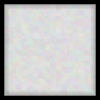}}
\end{minipage}
\hfill
\begin{minipage}{.11\linewidth}
\leftline{\scalebox{0.8}{2.empty-32-32}}
\leftline{\scalebox{0.8}{32*32 (1,024)}}
\end{minipage}
\hfill
\begin{minipage}{.04\linewidth}
\rightline{\includegraphics[width=.5cm]{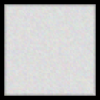}}
\end{minipage}
\hfill
\begin{minipage}{.11\linewidth}
\leftline{\scalebox{0.8}{3.maze-32-32-2}}
\leftline{\scalebox{0.8}{32*32 (666)}}
\end{minipage}
\hfill
\begin{minipage}{.04\linewidth}
\leftline{\includegraphics[width=.5cm]{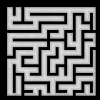}}
\end{minipage}
\hfill
\begin{minipage}{.11\linewidth}
\leftline{\scalebox{0.8}{4.maze-32-32-4}}
\leftline{\scalebox{0.8}{32*32 (790)}}
\end{minipage}
\hfill
\begin{minipage}{.04\linewidth}
\leftline{\includegraphics[width=.5cm]{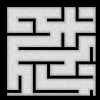}}
\end{minipage}
\hfill
\begin{minipage}{.11\linewidth}
\leftline{\scalebox{0.8}{5.maze-128-128-2}}
\leftline{\scalebox{0.8}{128x128 (10,858)}}
\end{minipage}
\hfill
\begin{minipage}{.04\linewidth}
\leftline{\includegraphics[width=.5cm]{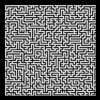}}
\end{minipage}
\hfill
\begin{minipage}{.11\linewidth}
\leftline{\scalebox{0.8}{6.maze-128-128}}
\leftline{\scalebox{0.8}{-10}}
\leftline{\scalebox{0.8}{128x128 (14,818)}}
\end{minipage}
\hfill
\begin{minipage}{.04\linewidth}
\leftline{\includegraphics[width=.5cm]{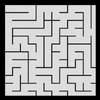}}
\end{minipage}
\vfill

\begin{minipage}{.04\linewidth}
  \rotatebox{90}{\scalebox{0.8}{decomposition rate}}
\end{minipage}
\hfill
\begin{minipage}{.15\linewidth}
  \centerline{\includegraphics[width=2.1cm]{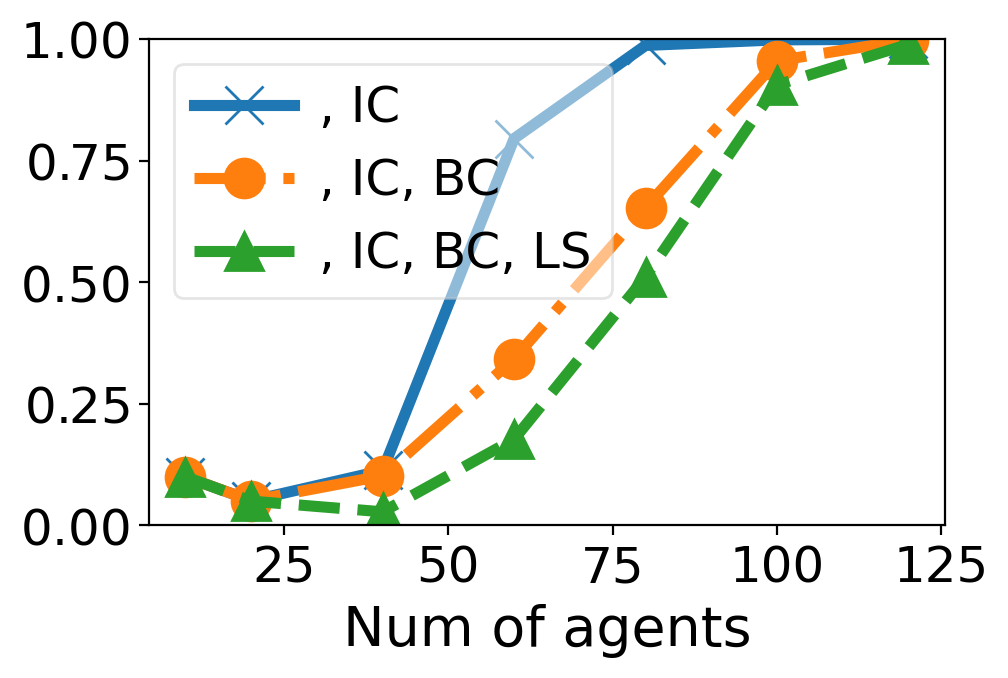}}
\end{minipage}
\hfill
\begin{minipage}{.15\linewidth}
  \centerline{\includegraphics[width=2.1cm]{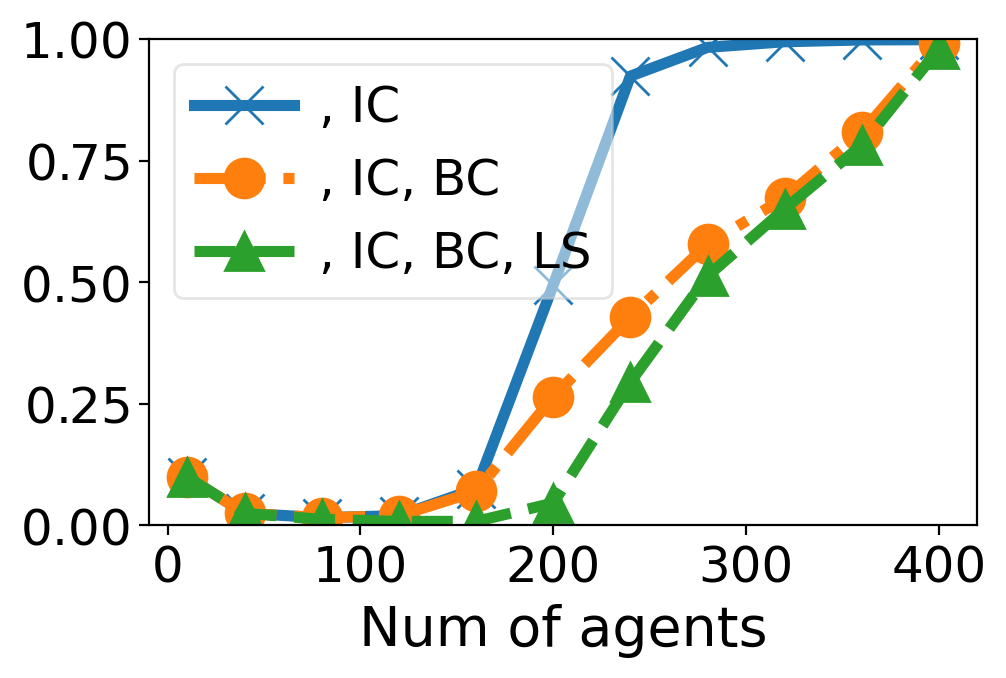}}
\end{minipage}
\hfill
\begin{minipage}{.15\linewidth}
  \centerline{\includegraphics[width=2.1cm]{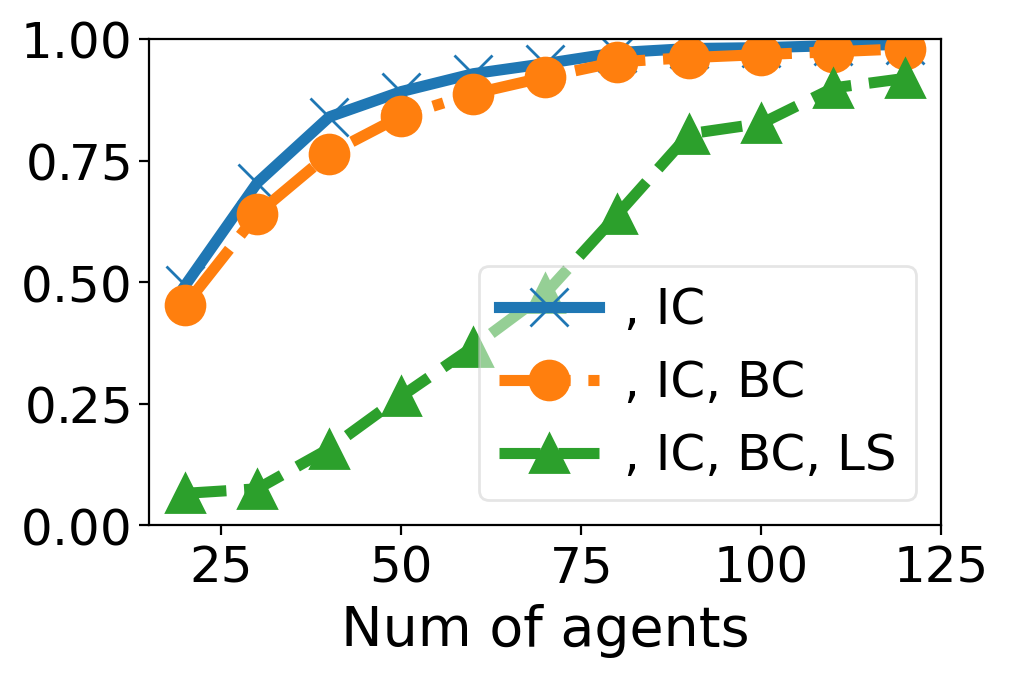}}
\end{minipage}
\hfill
\begin{minipage}{.15\linewidth}
  \centerline{\includegraphics[width=2.1cm]{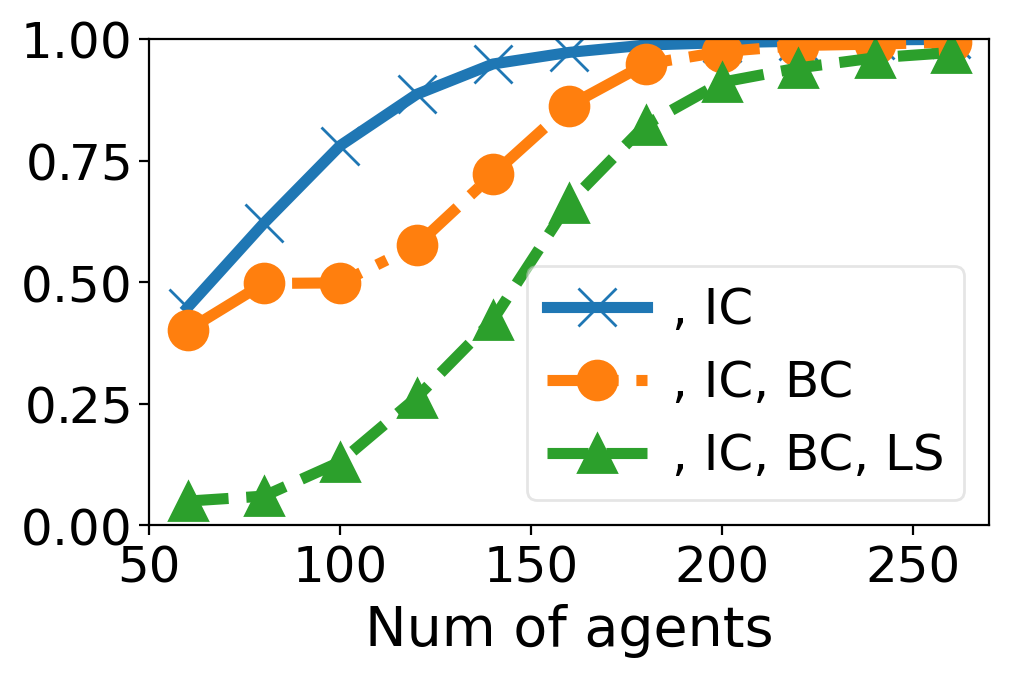}}
\end{minipage}
\hfill
\begin{minipage}{.15\linewidth}
  \centerline{\includegraphics[width=2.1cm]{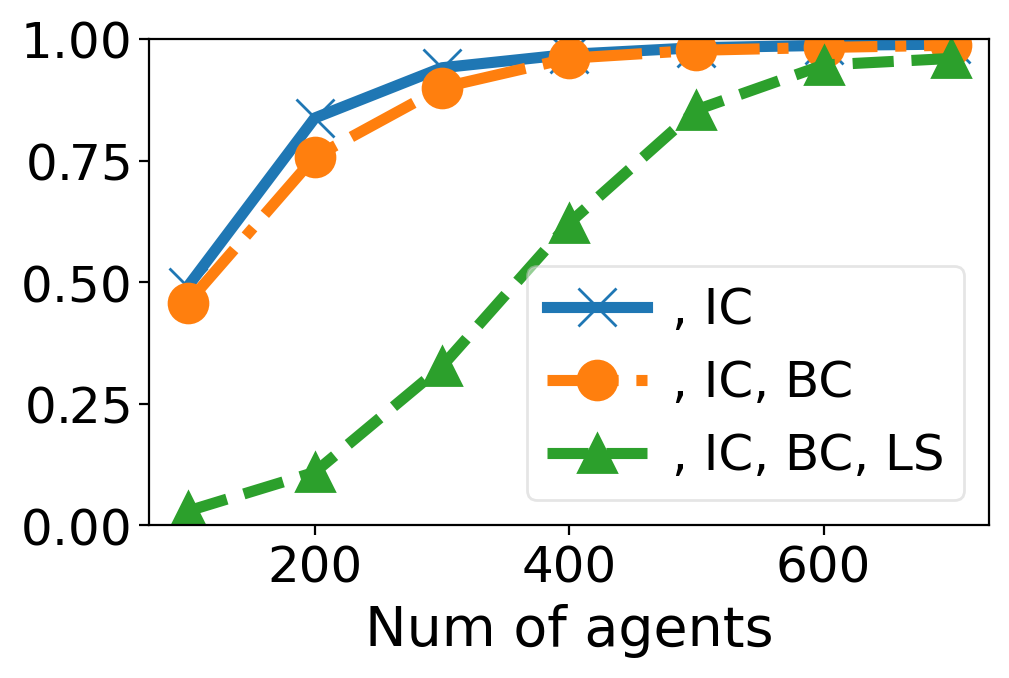}}
\end{minipage}
\hfill
\begin{minipage}{.15\linewidth}
  \centerline{\includegraphics[width=2.1cm]{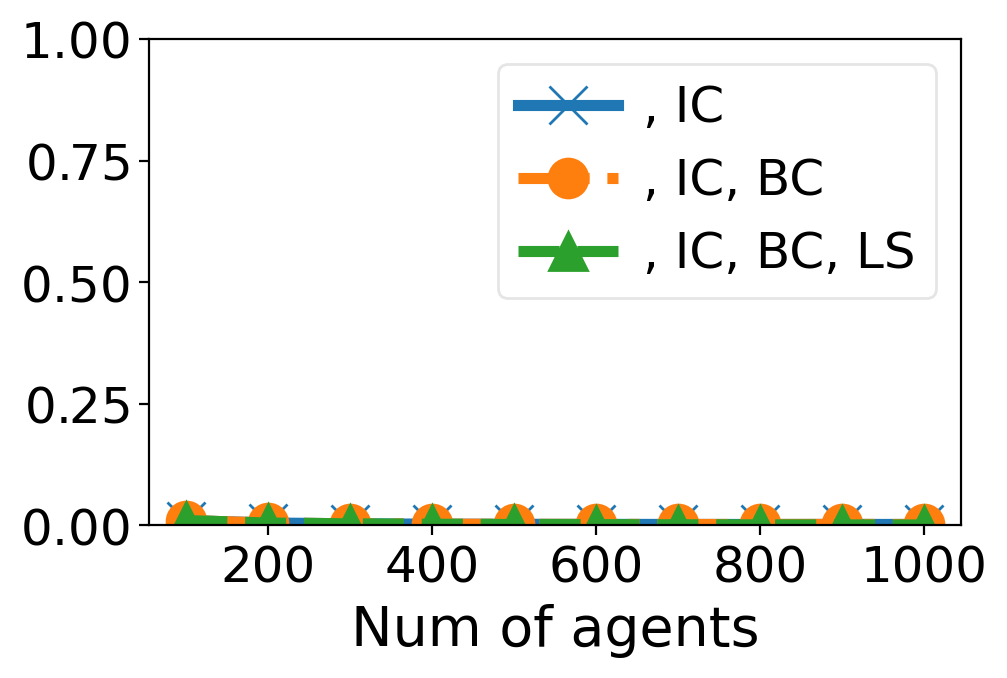}}
\end{minipage}
\vfill

\begin{minipage}{.04\linewidth}
  \rotatebox{90}{\scalebox{0.8}{subproblems}}
\end{minipage}
\hfill
\begin{minipage}{.15\linewidth}
  \centerline{\includegraphics[width=2.1cm]{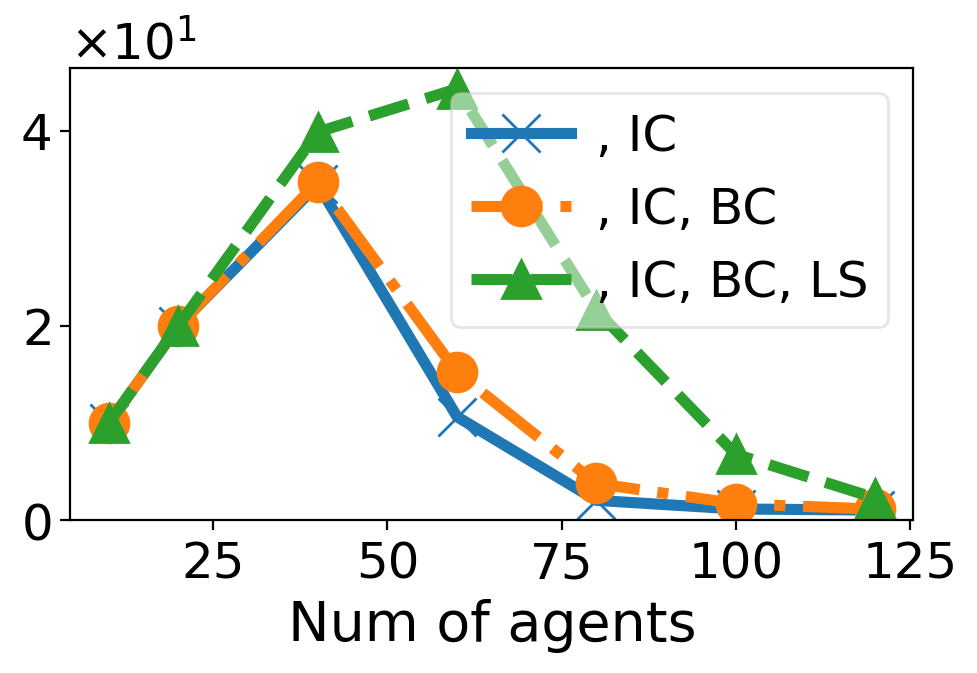}}
\end{minipage}
\hfill
\begin{minipage}{.15\linewidth}
  \centerline{\includegraphics[width=2.1cm]{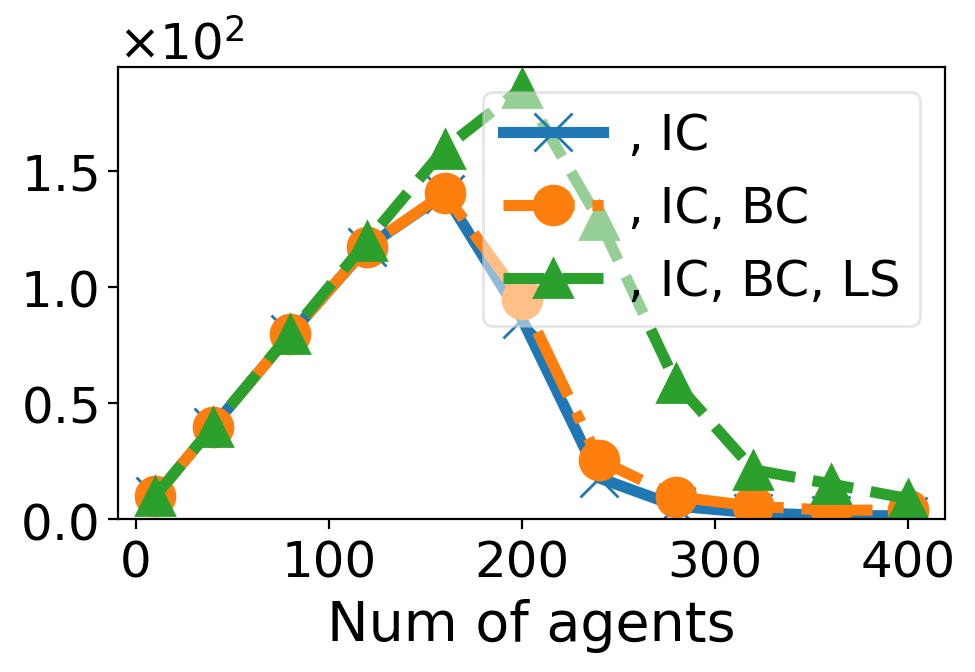}}
\end{minipage}
\hfill
\begin{minipage}{.15\linewidth}
  \centerline{\includegraphics[width=2.1cm]{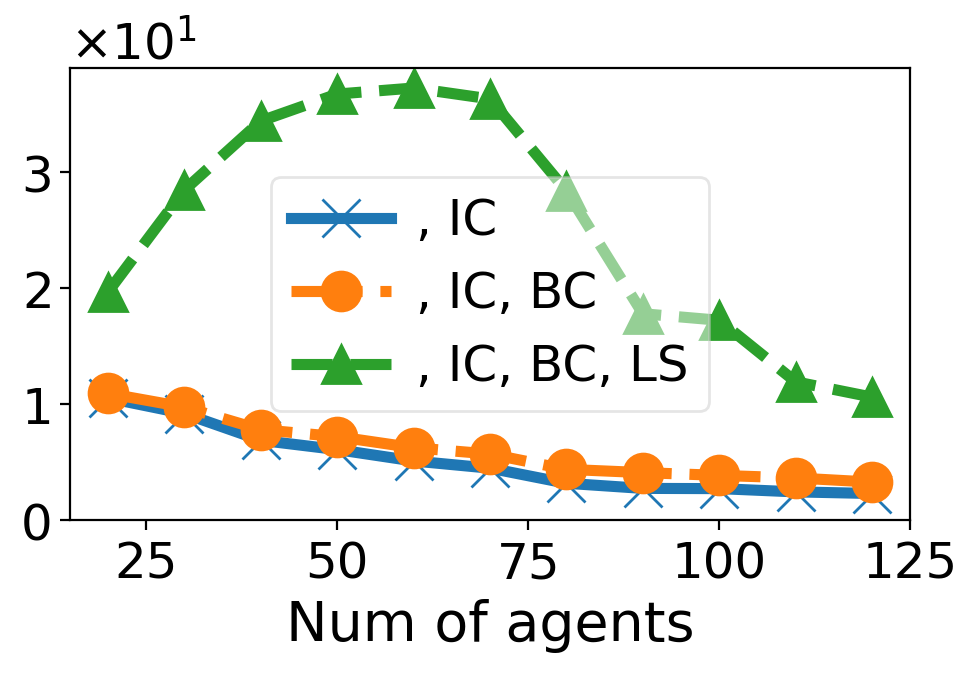}}
\end{minipage}
\hfill
\begin{minipage}{.15\linewidth}
  \centerline{\includegraphics[width=2.1cm]{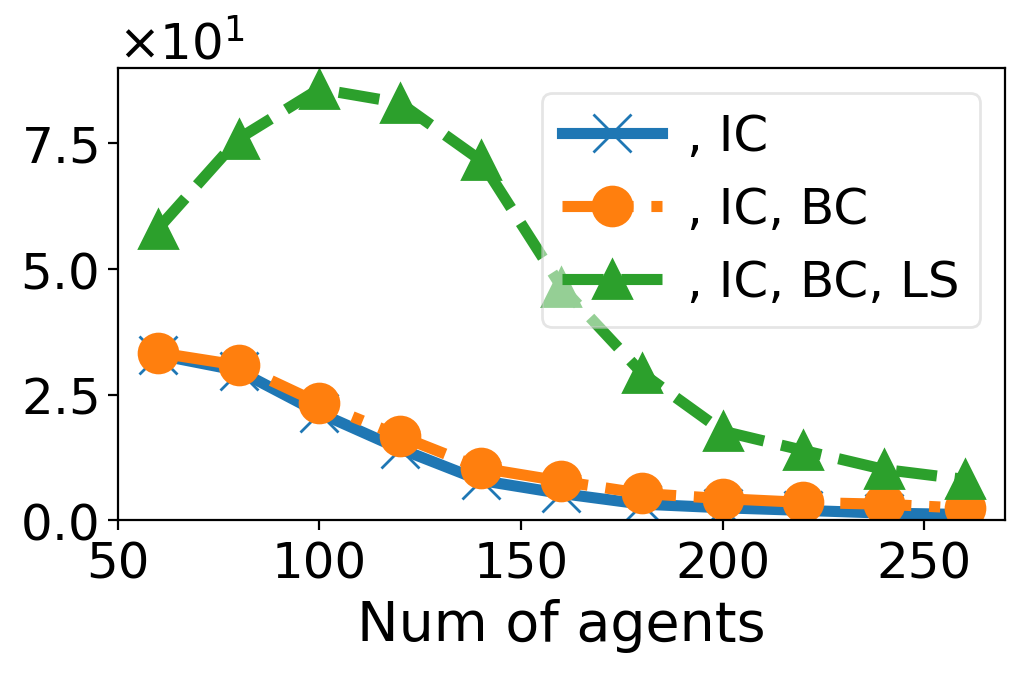}}
\end{minipage}
\hfill
\begin{minipage}{.15\linewidth}
  \centerline{\includegraphics[width=2.1cm]{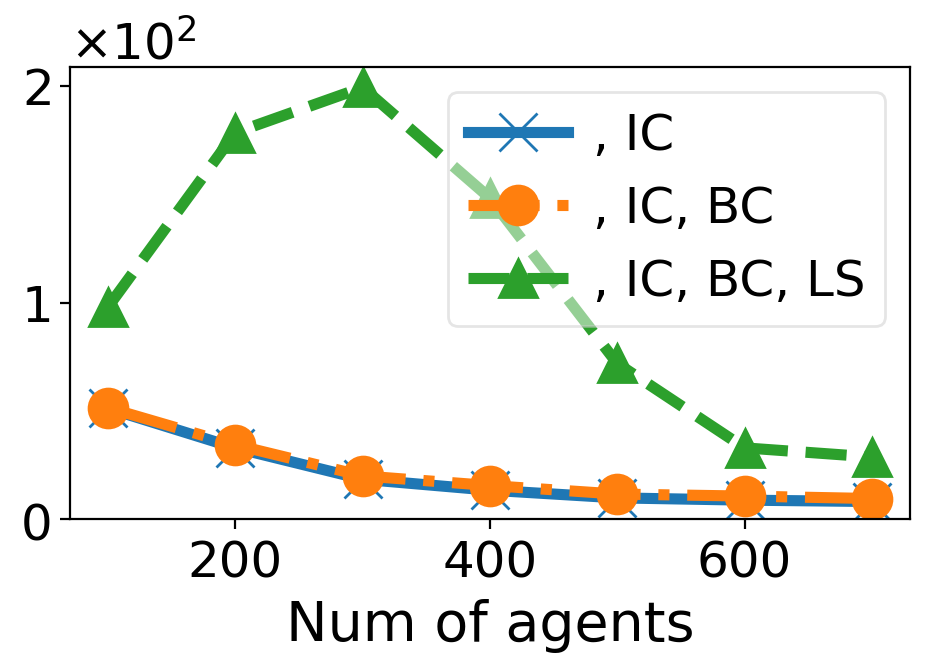}}
\end{minipage}
\hfill
\begin{minipage}{.15\linewidth}
  \centerline{\includegraphics[width=2.1cm]{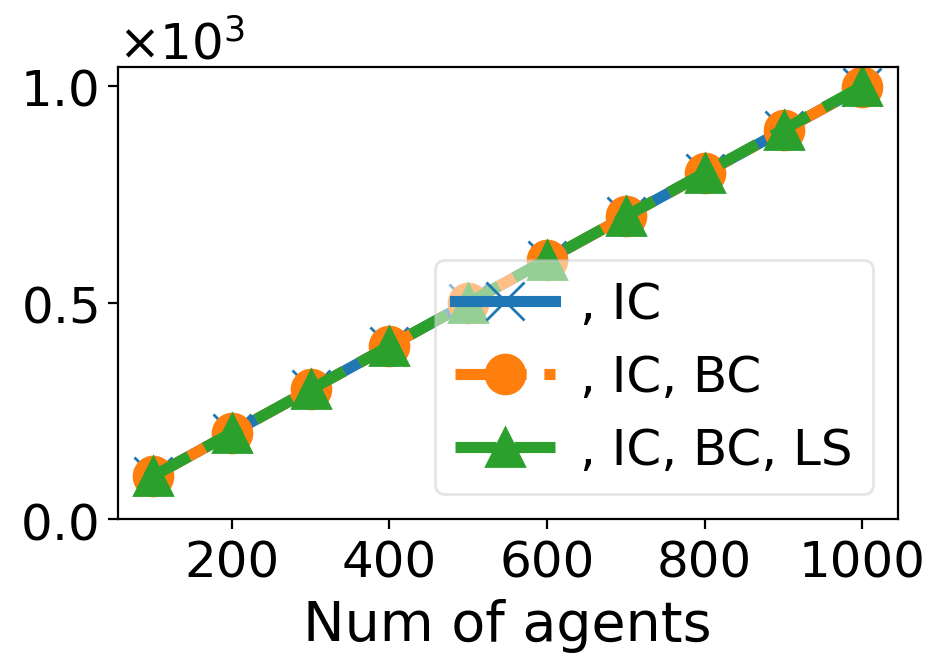}}
\end{minipage}
\vfill

\begin{minipage}{.04\linewidth}
  \rotatebox{90}{\scalebox{0.8}{time cost (ms)}}
\end{minipage}
\hfill
\begin{minipage}{.15\linewidth}
  \centerline{\includegraphics[width=2.1cm]{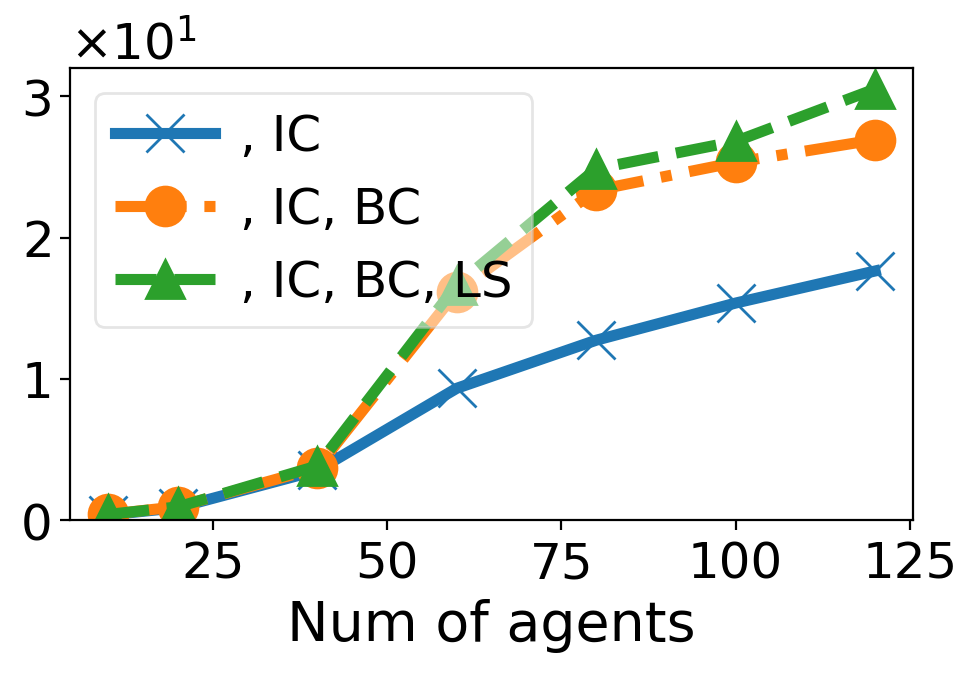}}
\end{minipage}
\hfill
\begin{minipage}{.15\linewidth}
  \centerline{\includegraphics[width=2.1cm]{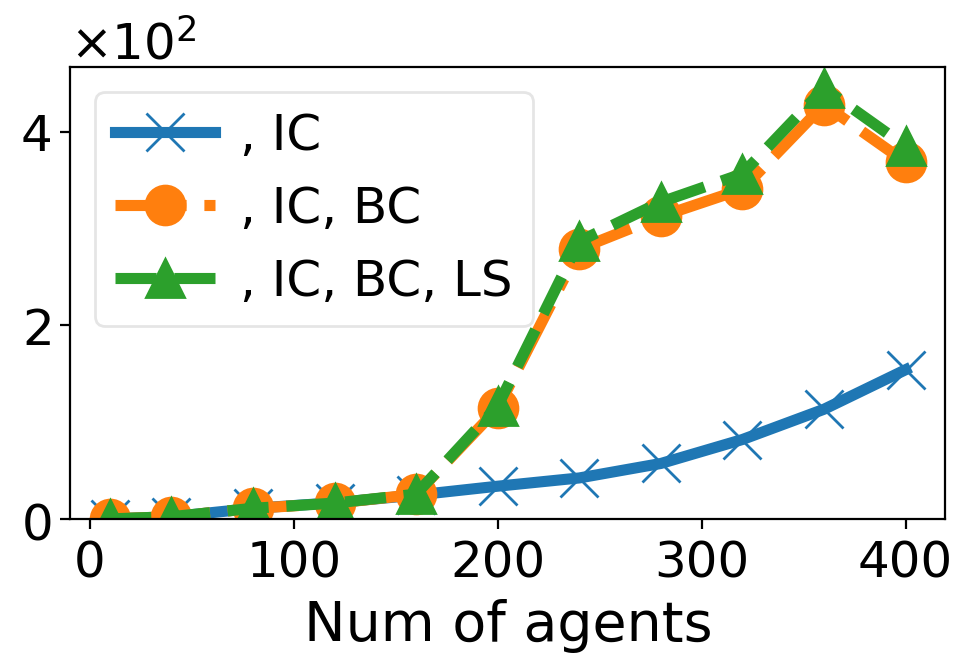}}
\end{minipage}
\hfill
\begin{minipage}{.15\linewidth}
  \centerline{\includegraphics[width=2.1cm]{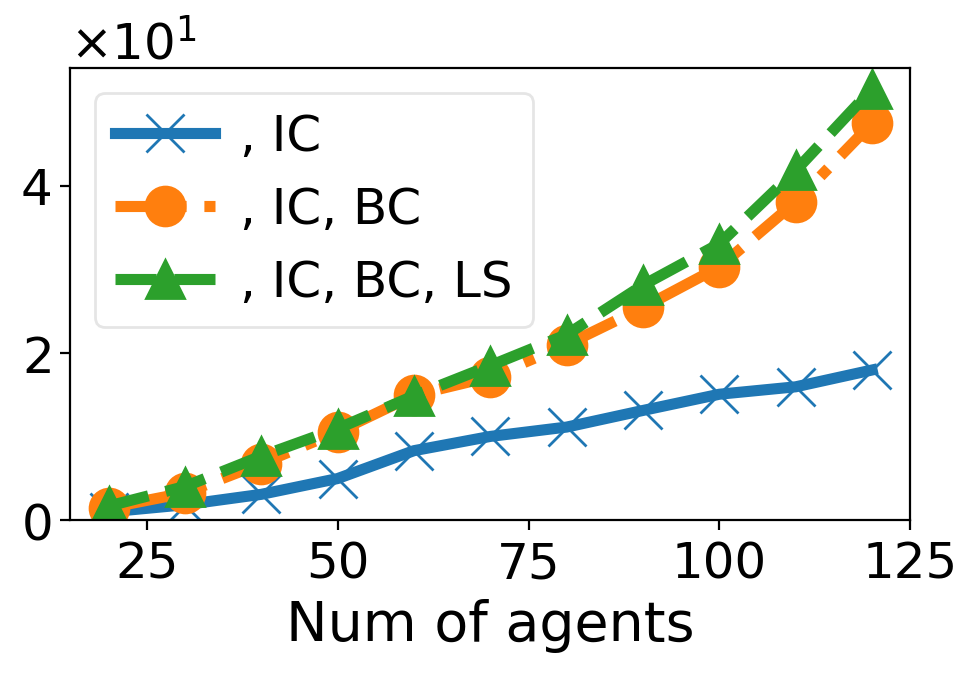}}
\end{minipage}
\hfill
\begin{minipage}{.15\linewidth}
  \centerline{\includegraphics[width=2.1cm]{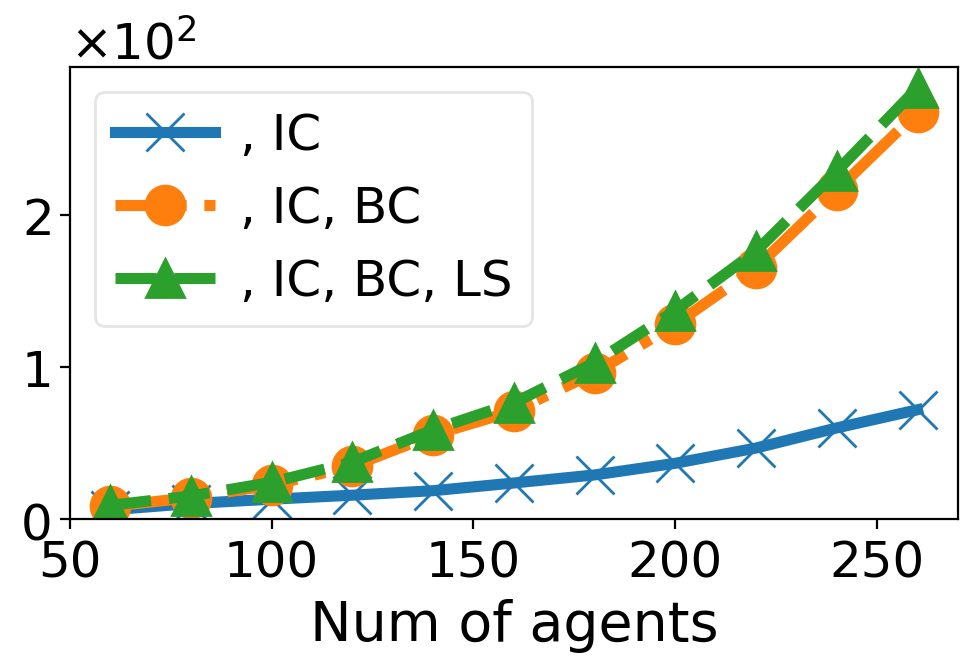}}
\end{minipage}
\hfill
\begin{minipage}{.15\linewidth}
  \centerline{\includegraphics[width=2.1cm]{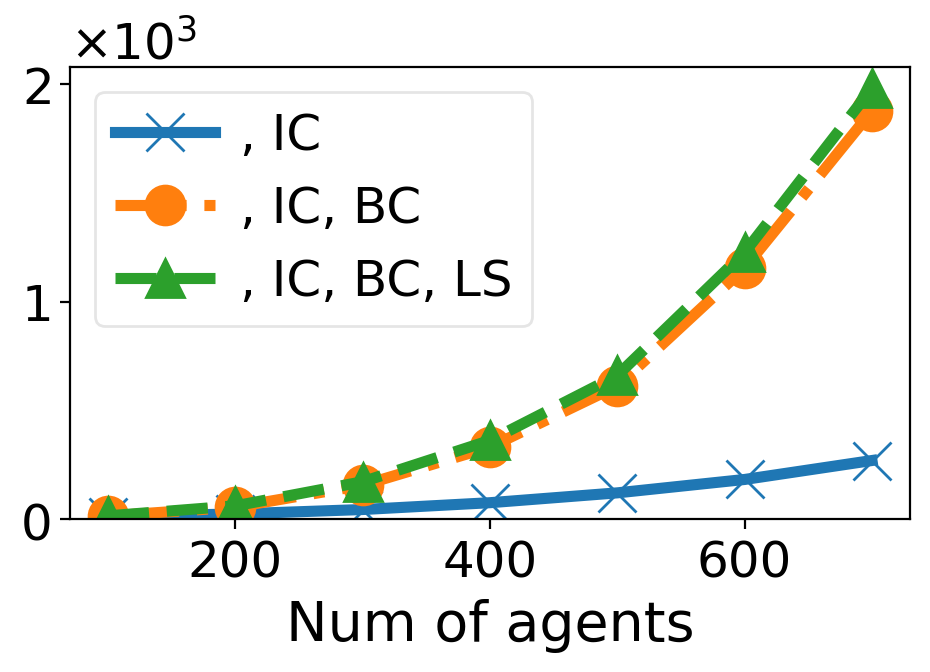}}
\end{minipage}
\hfill
\begin{minipage}{.15\linewidth}
  \centerline{\includegraphics[width=2.1cm]{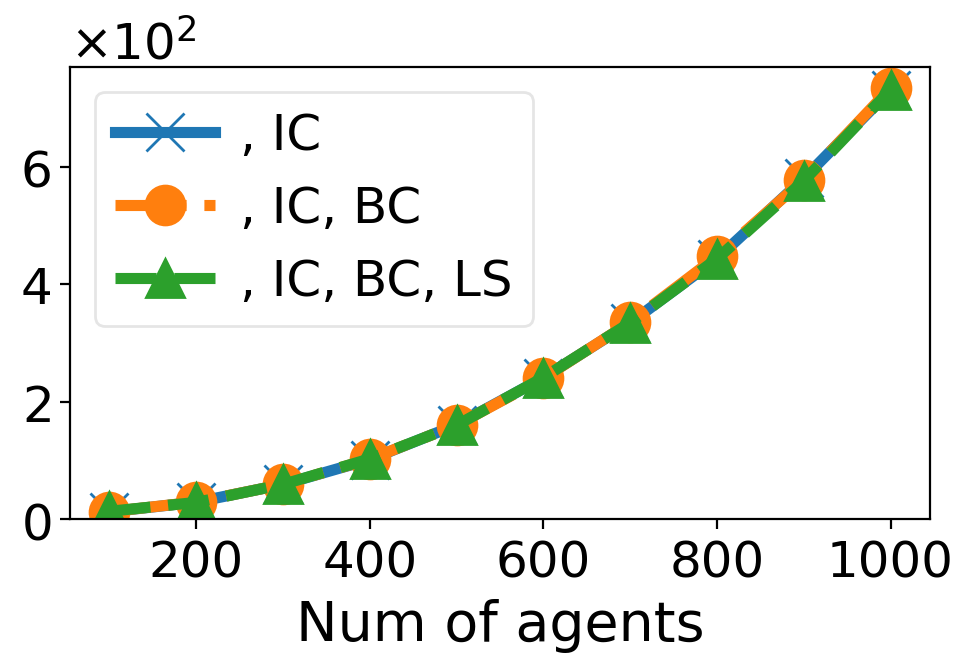}}
\end{minipage}
\vfill

\begin{minipage}{.04\linewidth}
  \rotatebox{90}{\scalebox{0.8}{memory usage (MB)}}
\end{minipage}
\hfill
\begin{minipage}{.15\linewidth}
  \centerline{\includegraphics[width=2.1cm]{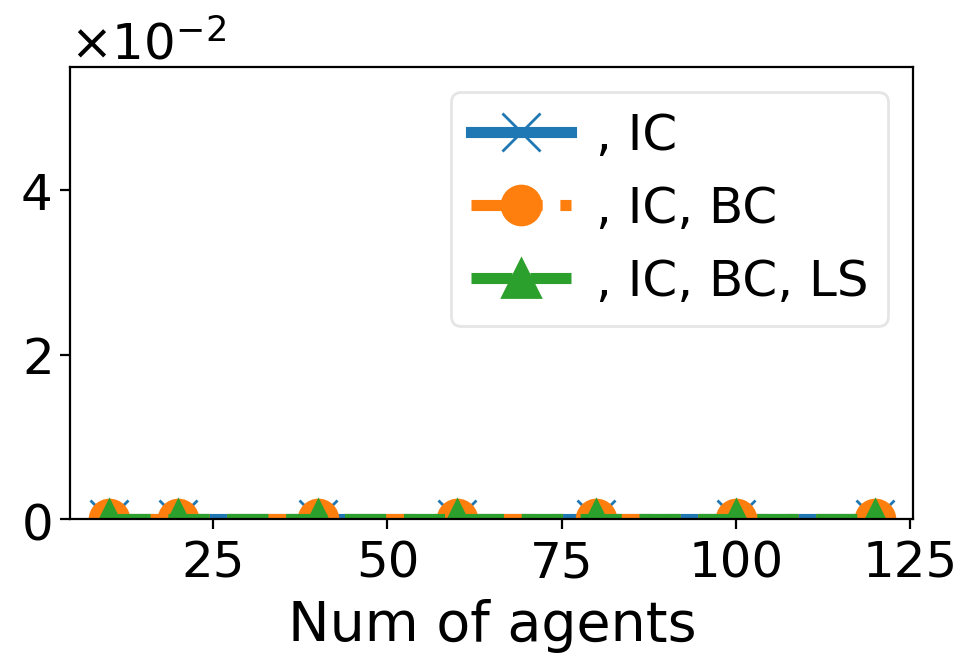}}
\end{minipage}
\hfill
\begin{minipage}{.15\linewidth}
  \centerline{\includegraphics[width=2.1cm]{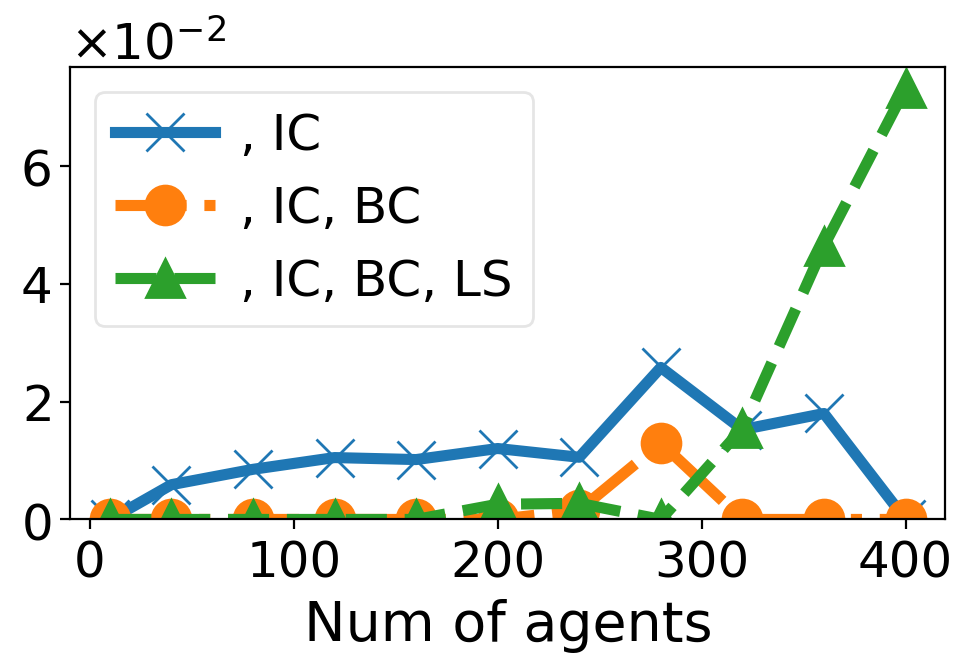}}
\end{minipage}
\hfill
\begin{minipage}{.15\linewidth}
  \centerline{\includegraphics[width=2.1cm]{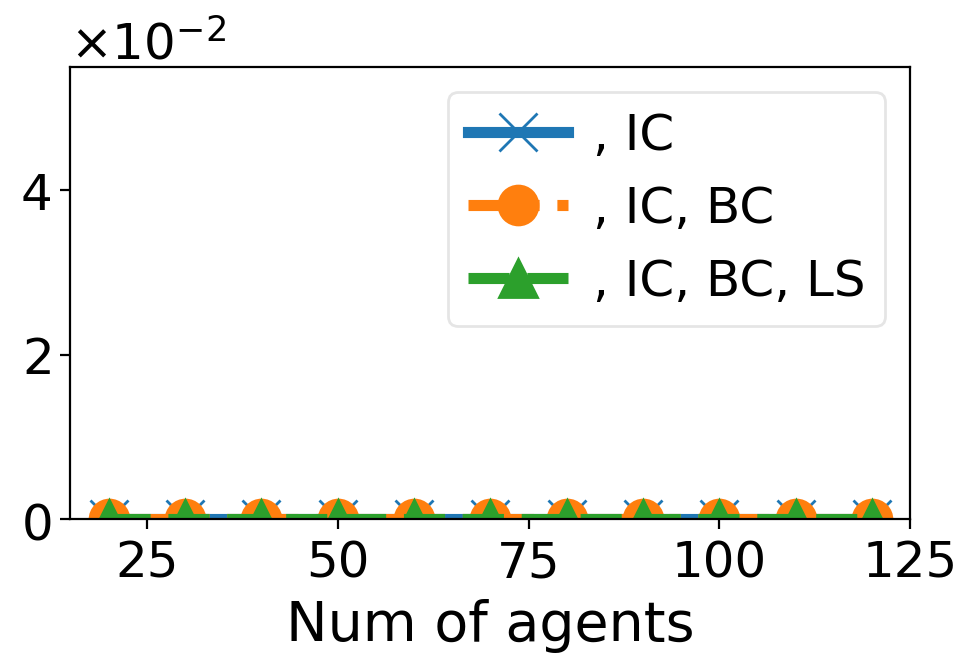}}
\end{minipage}
\hfill
\begin{minipage}{.15\linewidth}
  \centerline{\includegraphics[width=2.1cm]{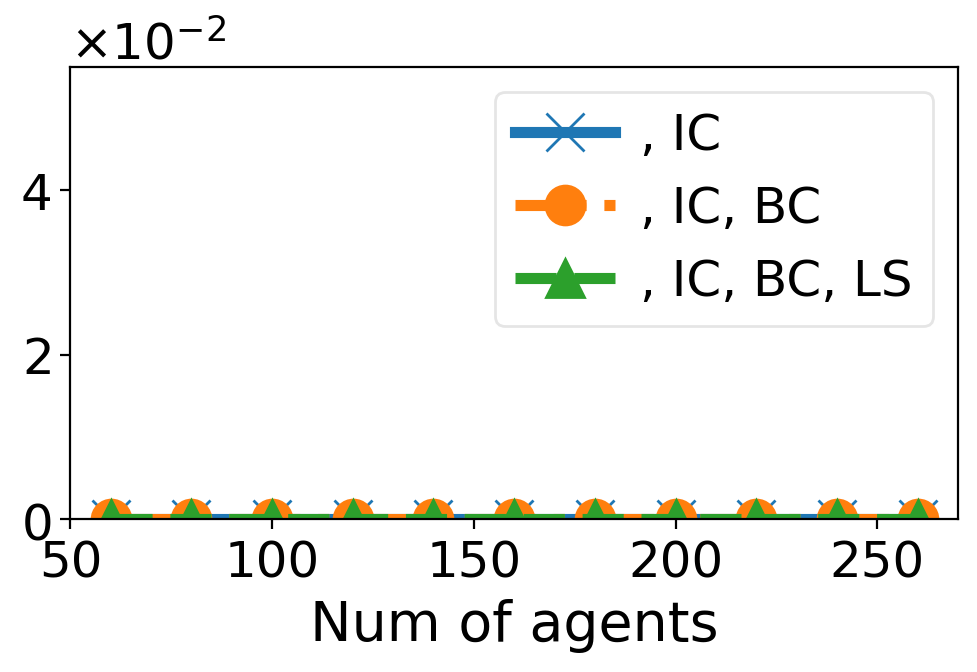}}
\end{minipage}
\hfill
\begin{minipage}{.15\linewidth}
  \centerline{\includegraphics[width=2.1cm]{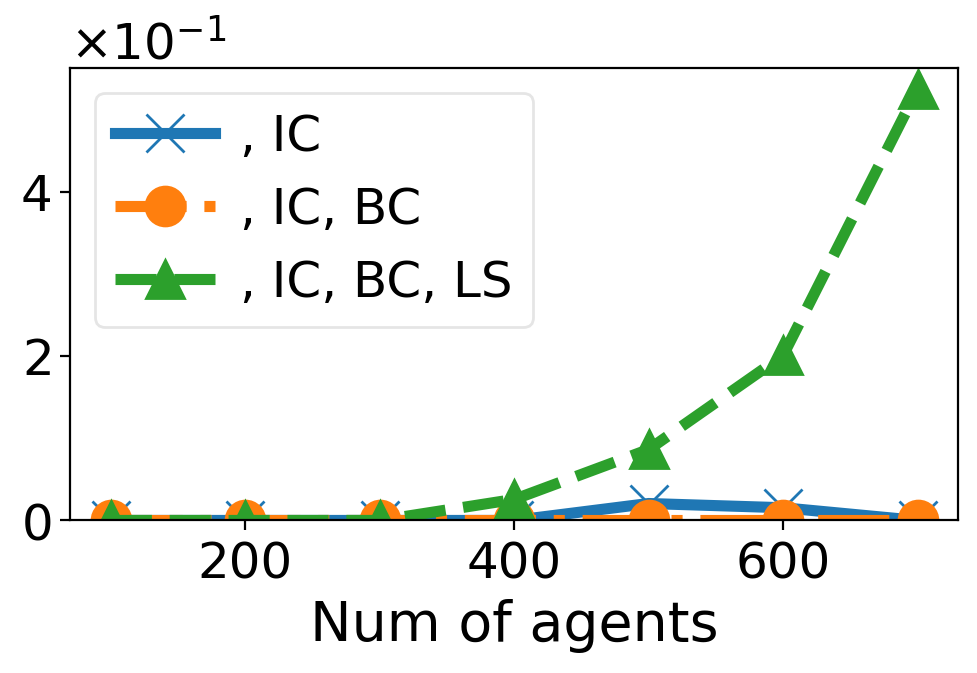}}
\end{minipage}
\hfill
\begin{minipage}{.15\linewidth}
  \centerline{\includegraphics[width=2.1cm]{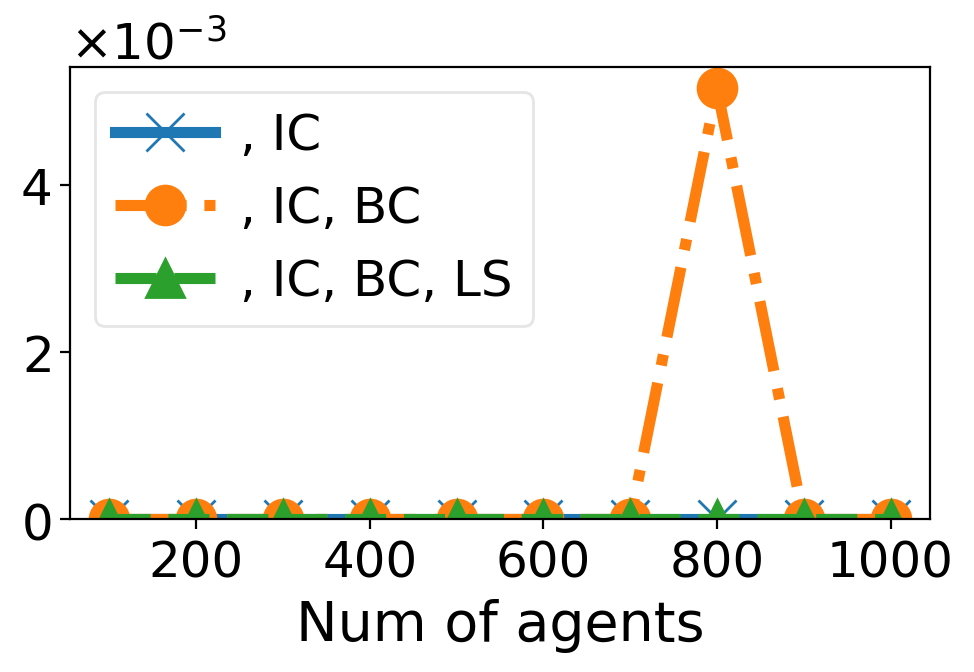}}
\end{minipage}
\vfill

\begin{minipage}{.04\linewidth}
\centerline{ }
\end{minipage}
\hfill
\begin{minipage}{.11\linewidth}
\leftline{\scalebox{0.8}{7.den312d}}
\leftline{\scalebox{0.8}{65*85 (2,445)}}
\end{minipage}
\hfill
\begin{minipage}{.04\linewidth}
\leftline{\includegraphics[width=.5cm]{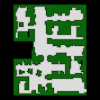}}
\end{minipage}
\hfill
\begin{minipage}{.11\linewidth}
\leftline{\scalebox{0.8}{8.den520d}}
\leftline{\scalebox{0.8}{256*257 (28,178)}}
\end{minipage}
\hfill
\begin{minipage}{.04\linewidth}
\rightline{\includegraphics[width=.5cm]{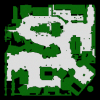}}
\end{minipage}
\hfill
\begin{minipage}{.11\linewidth}
\leftline{\scalebox{0.8}{9.Berlin\_1\_256}}
\leftline{\scalebox{0.8}{256*256 (47,540)}}
\end{minipage}
\hfill
\begin{minipage}{.04\linewidth}
\leftline{\includegraphics[width=.5cm]{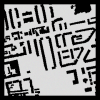}}
\end{minipage}
\hfill
\begin{minipage}{.11\linewidth}
\leftline{\scalebox{0.8}{10.Paris\_1\_256}}
\leftline{\scalebox{0.8}{256*256 (47,240)}}
\end{minipage}
\hfill
\begin{minipage}{.04\linewidth}
\leftline{\includegraphics[width=.5cm]{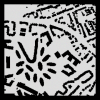}}
\end{minipage}
\hfill
\begin{minipage}{.11\linewidth}
\leftline{\scalebox{0.8}{11.ht\_chantry}}
\leftline{\scalebox{0.8}{162*141 (7,461)}}
\end{minipage}
\hfill
\begin{minipage}{.04\linewidth}
\leftline{\includegraphics[width=.5cm]{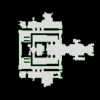}}
\end{minipage}
\hfill
\begin{minipage}{.11\linewidth}
\leftline{\scalebox{0.8}{12.lak303d}}
\leftline{\scalebox{0.8}{194*194 (14,784)}}
\end{minipage}
\hfill
\begin{minipage}{.04\linewidth}
\leftline{\includegraphics[width=.5cm]{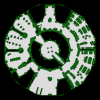}}
\end{minipage}
\vfill

\begin{minipage}{.04\linewidth}
  \rotatebox{90}{\scalebox{0.8}{decomposition rate}}
\end{minipage}
\hfill
\begin{minipage}{.15\linewidth}
  \centerline{\includegraphics[width=2.1cm]{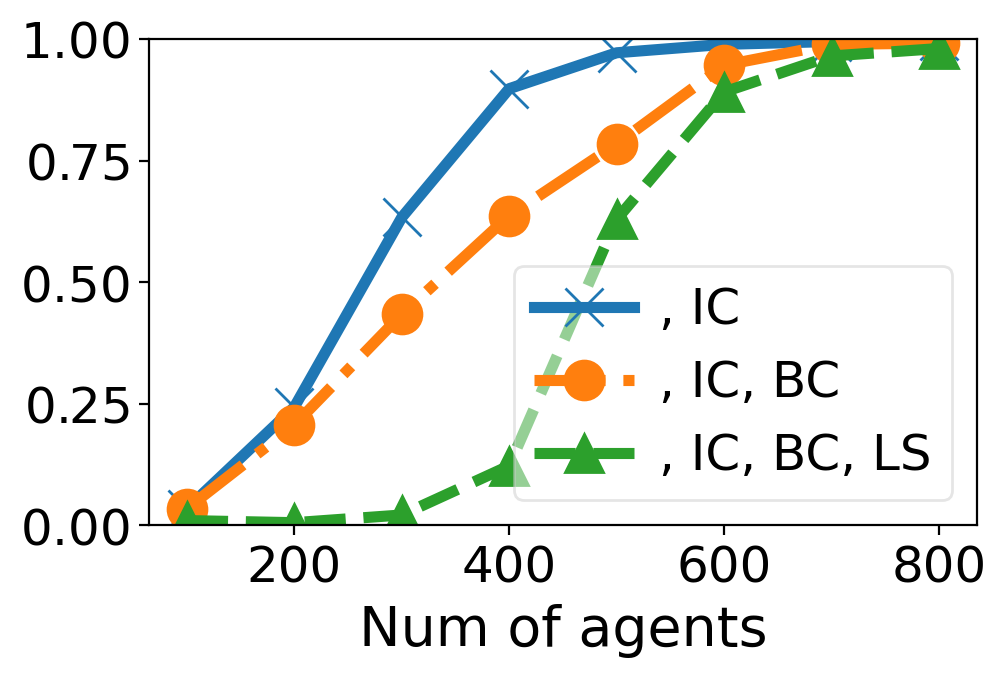}}
\end{minipage}
\hfill
\begin{minipage}{.15\linewidth}
  \centerline{\includegraphics[width=2.1cm]{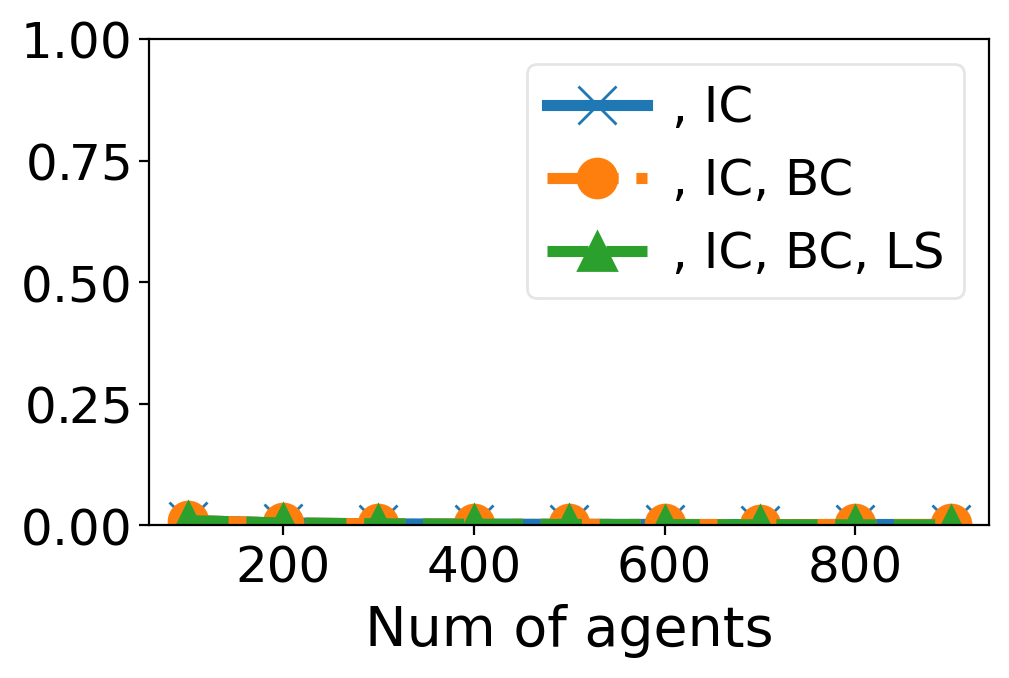}}
\end{minipage}
\hfill
\begin{minipage}{.15\linewidth}
  \centerline{\includegraphics[width=2.1cm]{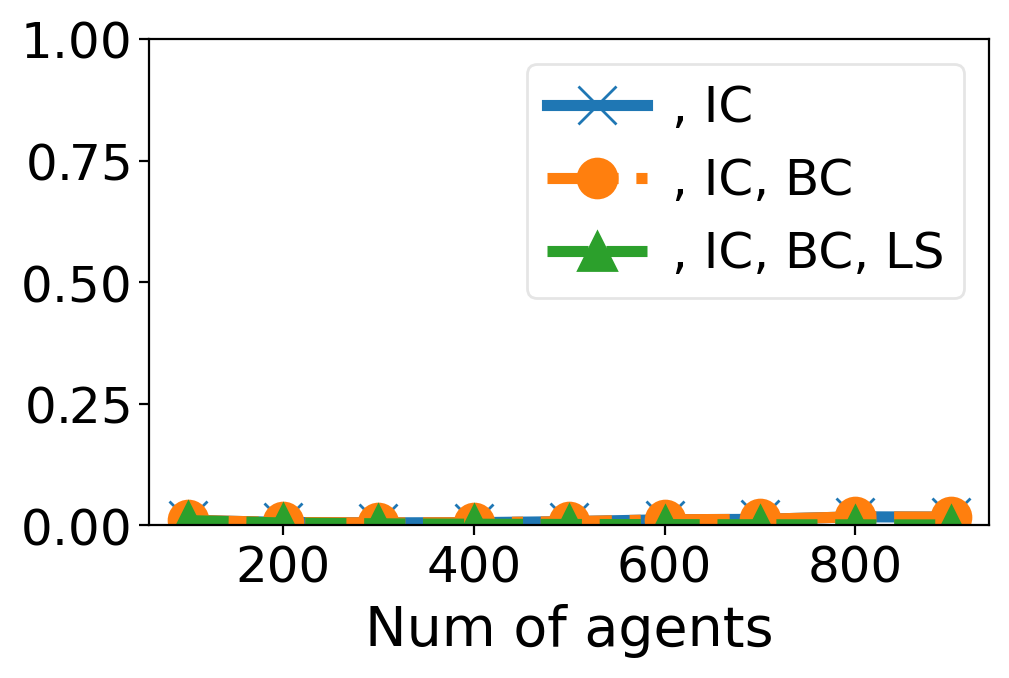}}
\end{minipage}
\hfill
\begin{minipage}{.15\linewidth}
  \centerline{\includegraphics[width=2.1cm]{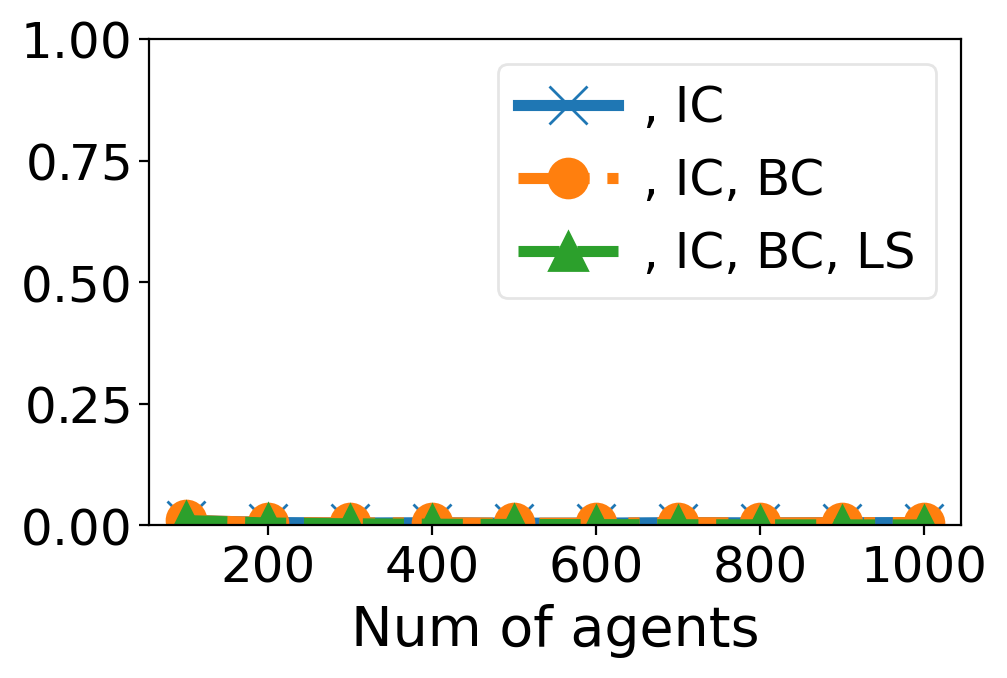}}
\end{minipage}
\hfill
\begin{minipage}{.15\linewidth}
  \centerline{\includegraphics[width=2.1cm]{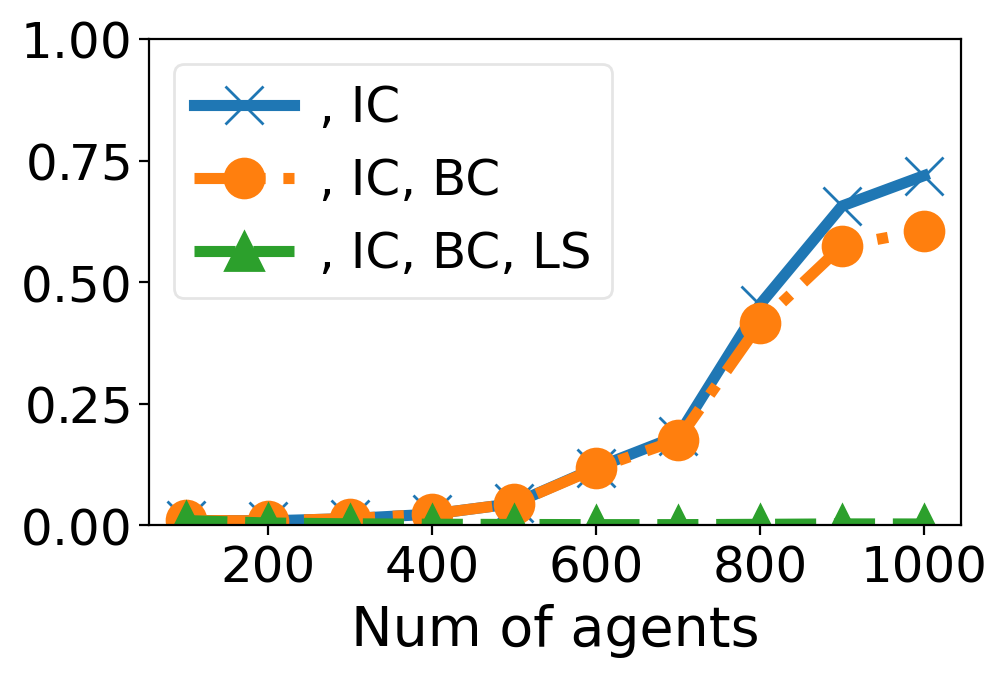}}
\end{minipage}
\hfill
\begin{minipage}{.15\linewidth}
  \centerline{\includegraphics[width=2.1cm]{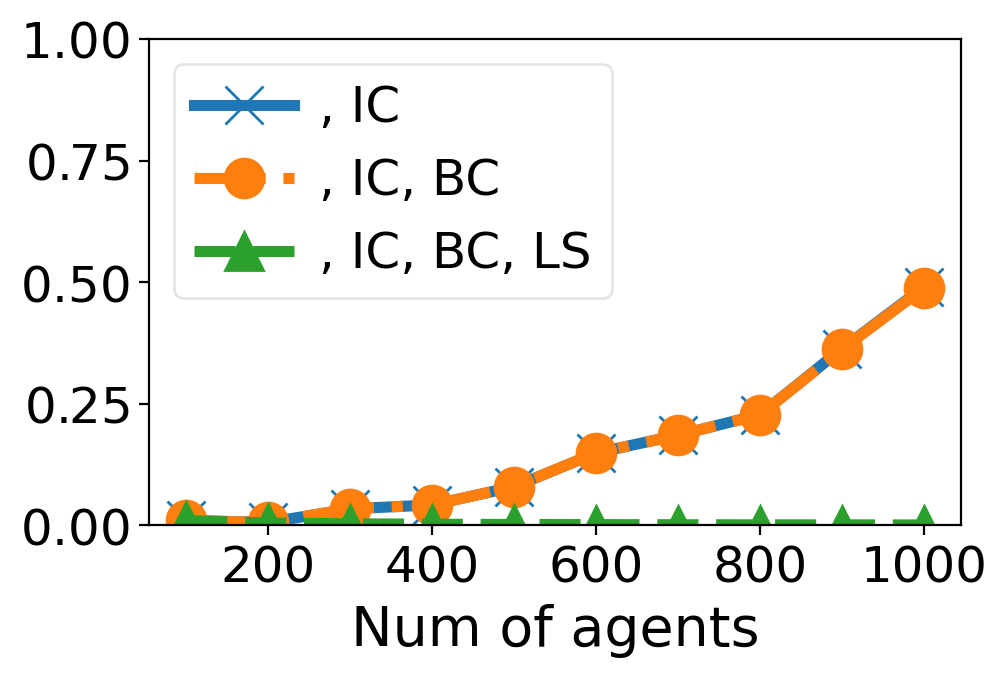}}
\end{minipage}
\vfill

\begin{minipage}{.04\linewidth}
  \rotatebox{90}{\scalebox{0.8}{subproblems}}
\end{minipage}
\hfill
\begin{minipage}{.15\linewidth}
  \centerline{\includegraphics[width=2.1cm]{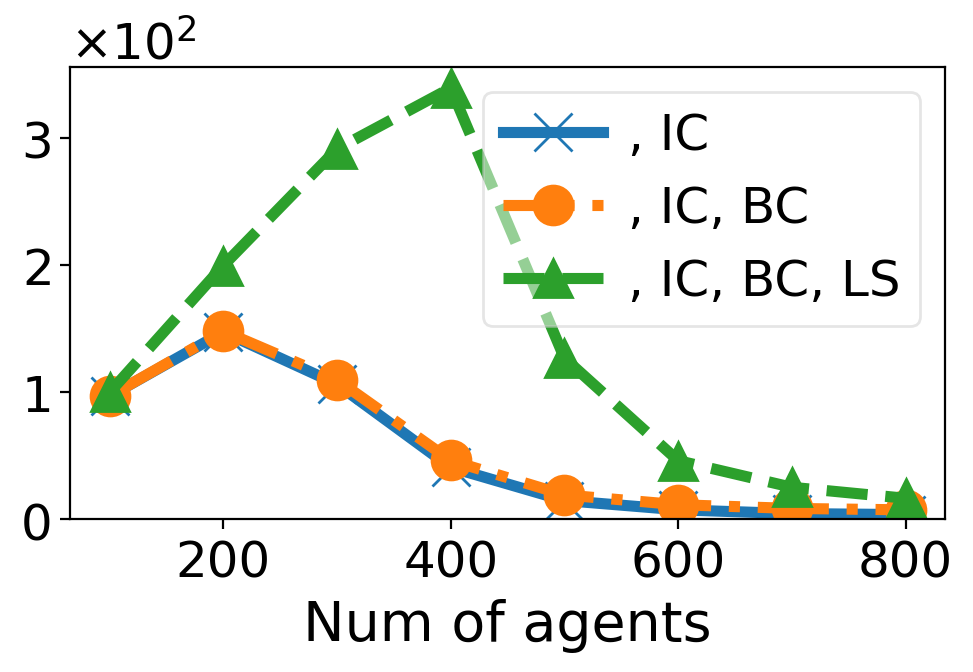}}
\end{minipage}
\hfill
\begin{minipage}{.15\linewidth}
  \centerline{\includegraphics[width=2.1cm]{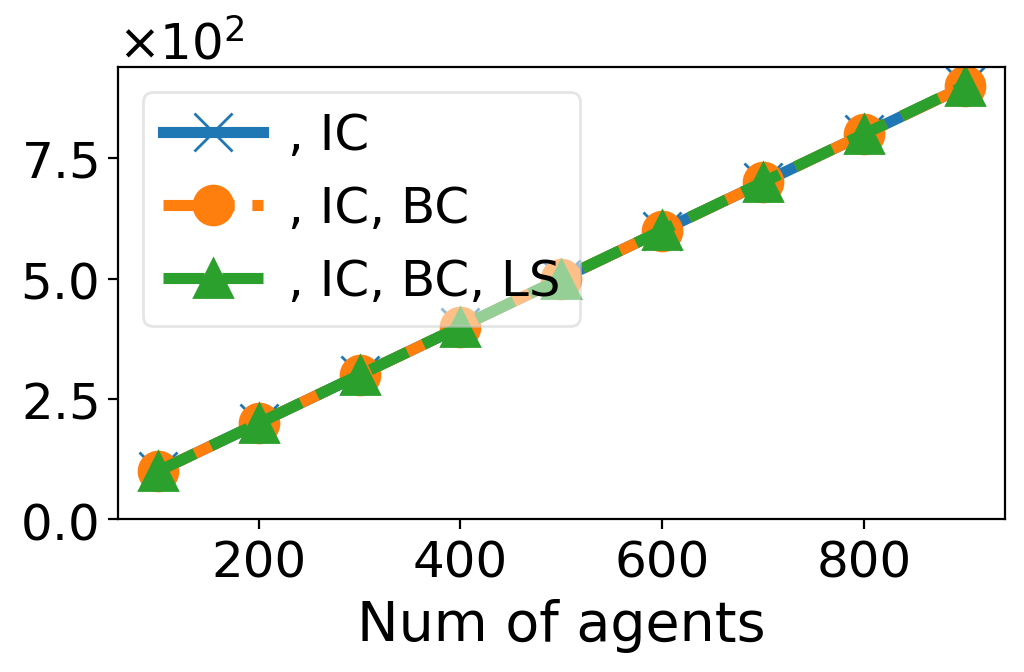}}
\end{minipage}
\hfill
\begin{minipage}{.15\linewidth}
  \centerline{\includegraphics[width=2.1cm]{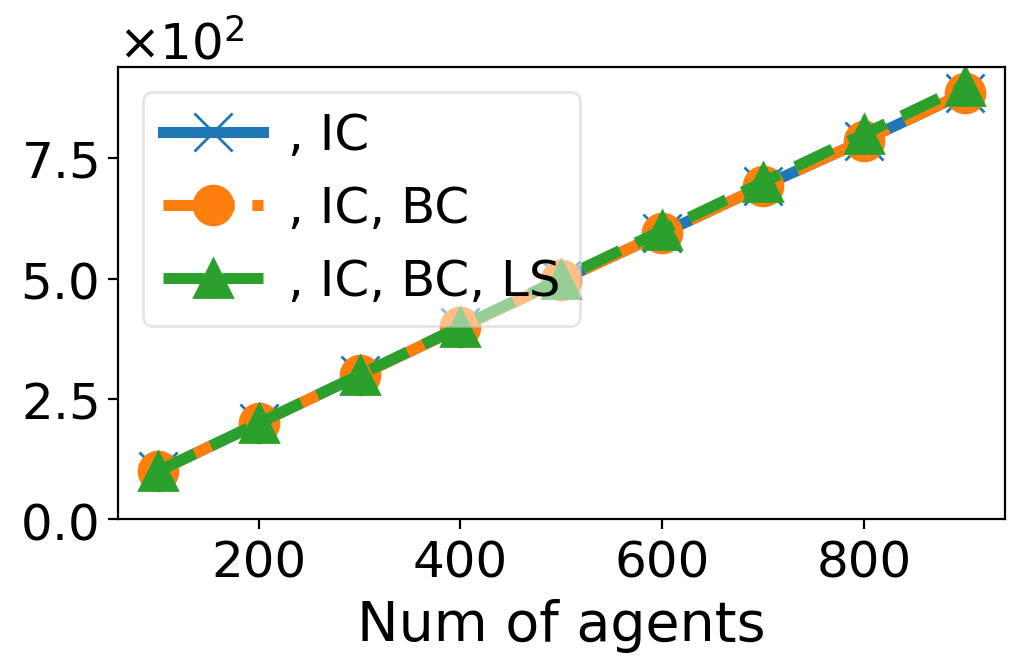}}
\end{minipage}
\hfill
\begin{minipage}{.15\linewidth}
  \centerline{\includegraphics[width=2.1cm]{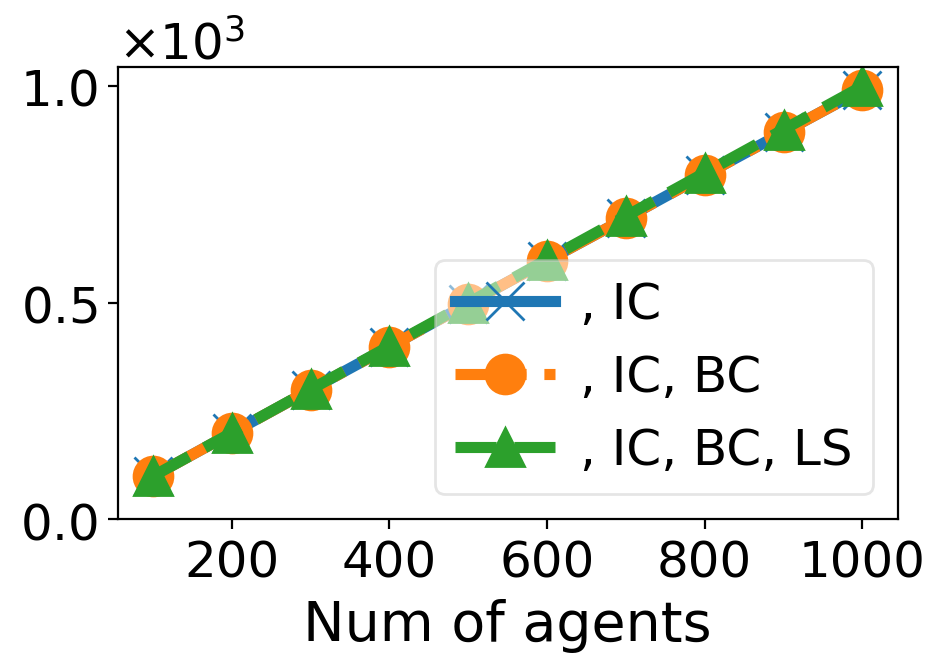}}
\end{minipage}
\hfill
\begin{minipage}{.15\linewidth}
  \centerline{\includegraphics[width=2.1cm]{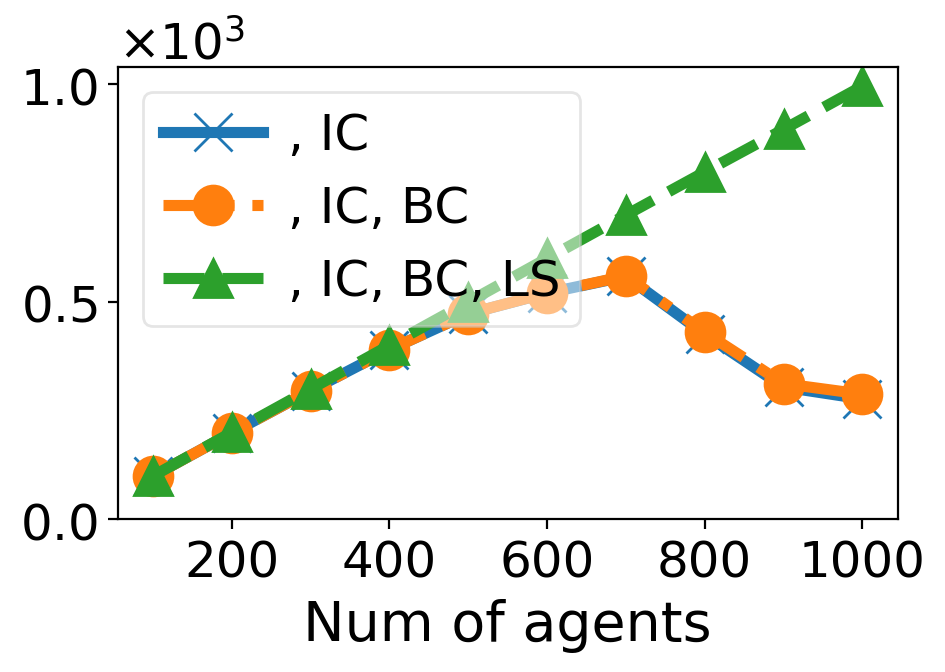}}
\end{minipage}
\hfill
\begin{minipage}{.15\linewidth}
  \centerline{\includegraphics[width=2.1cm]{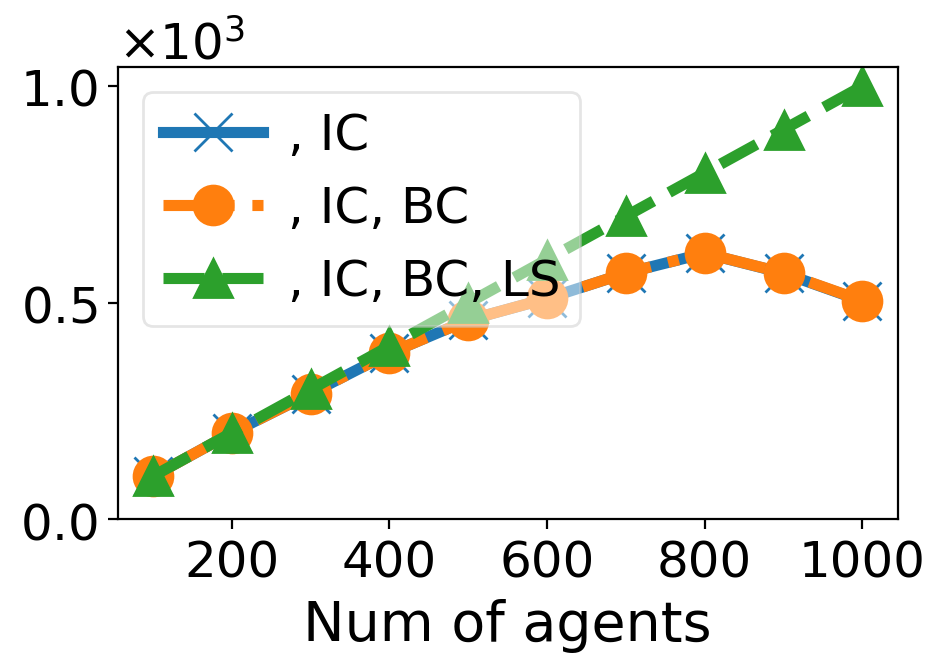}}
\end{minipage}
\vfill

\begin{minipage}{.04\linewidth}
  \rotatebox{90}{\scalebox{0.8}{time cost (ms)}}
\end{minipage}
\hfill
\begin{minipage}{.15\linewidth}
  \centerline{\includegraphics[width=2.1cm]{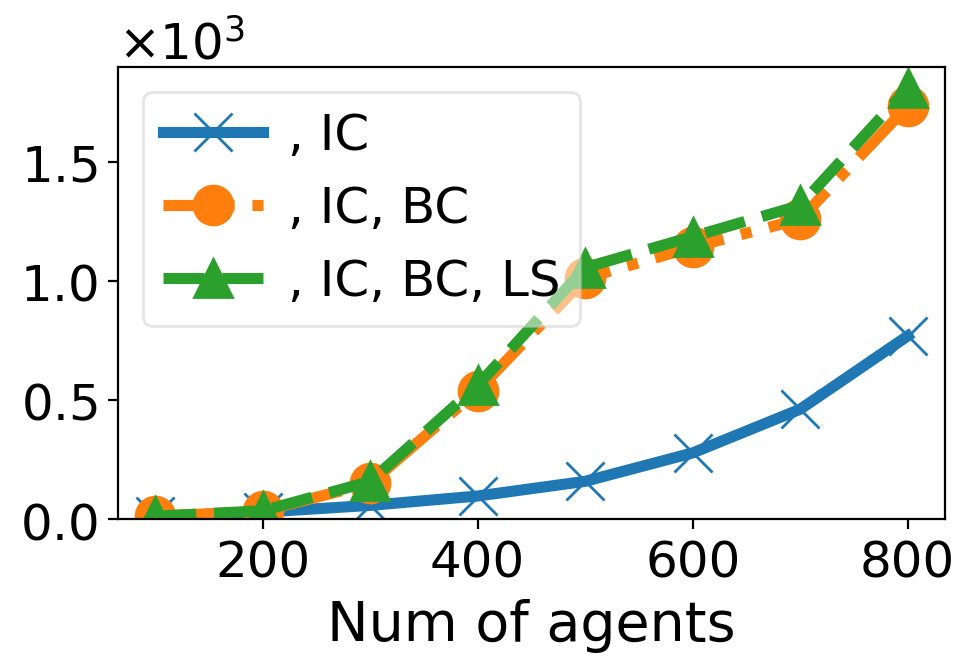}}
\end{minipage}
\hfill
\begin{minipage}{.15\linewidth}
  \centerline{\includegraphics[width=2.1cm]{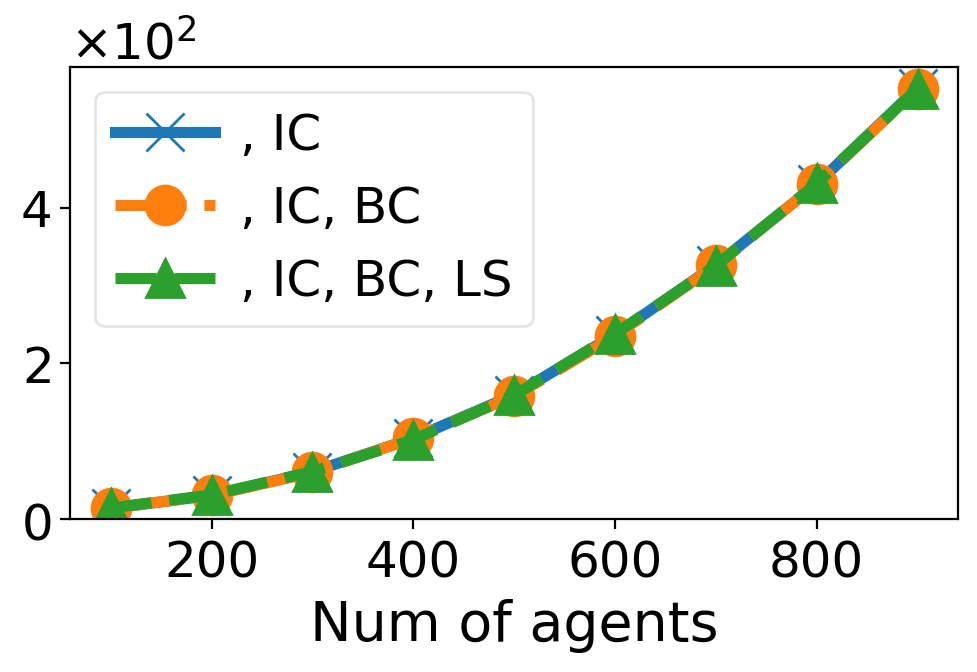}}
\end{minipage}
\hfill
\begin{minipage}{.15\linewidth}
  \centerline{\includegraphics[width=2.1cm]{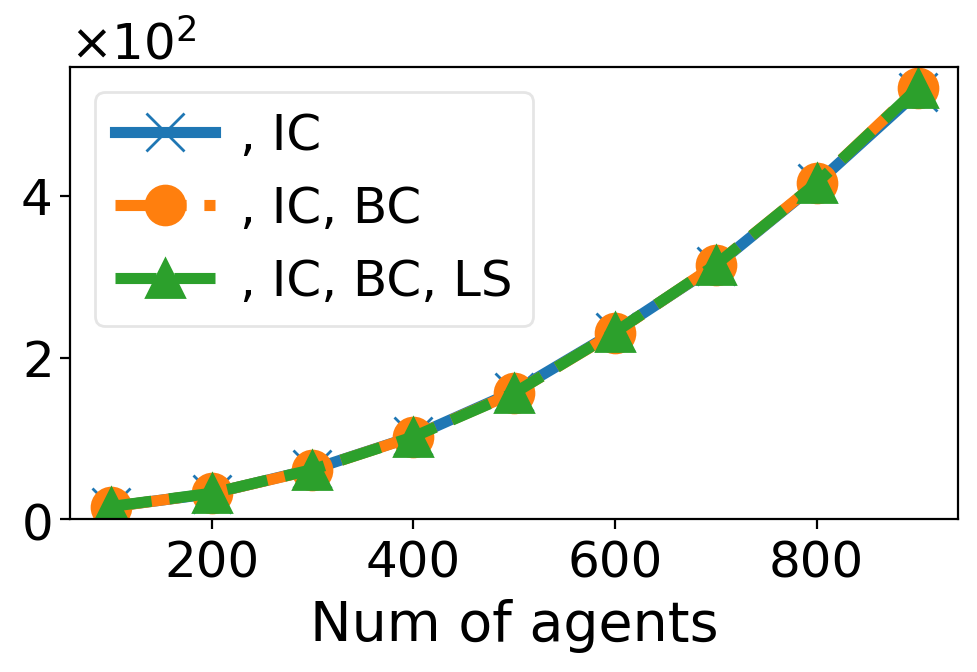}}
\end{minipage}
\hfill
\begin{minipage}{.15\linewidth}
  \centerline{\includegraphics[width=2.1cm]{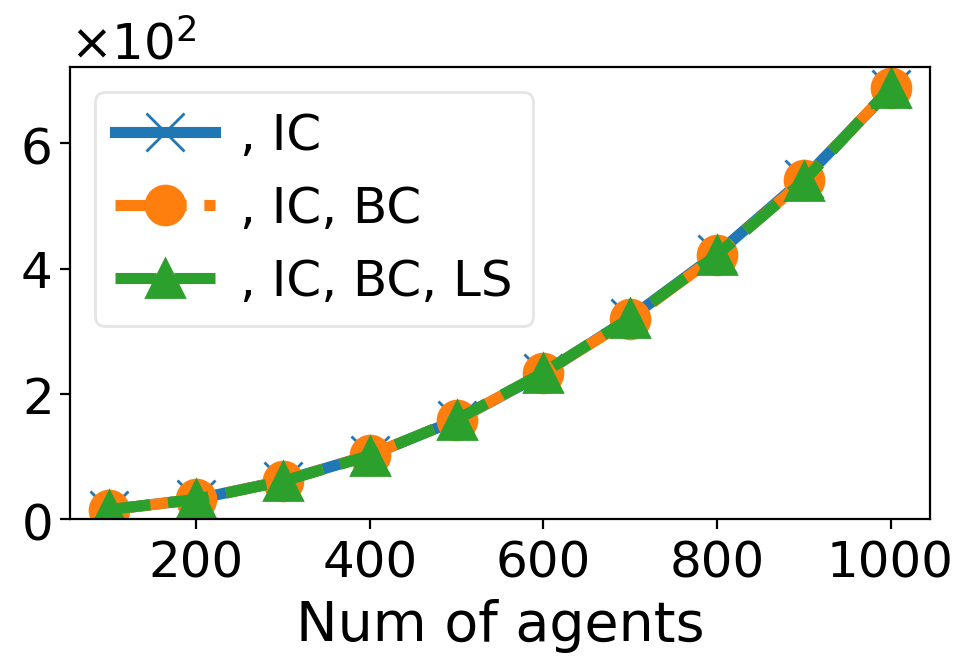}}
\end{minipage}
\hfill
\begin{minipage}{.15\linewidth}
  \centerline{\includegraphics[width=2.1cm]{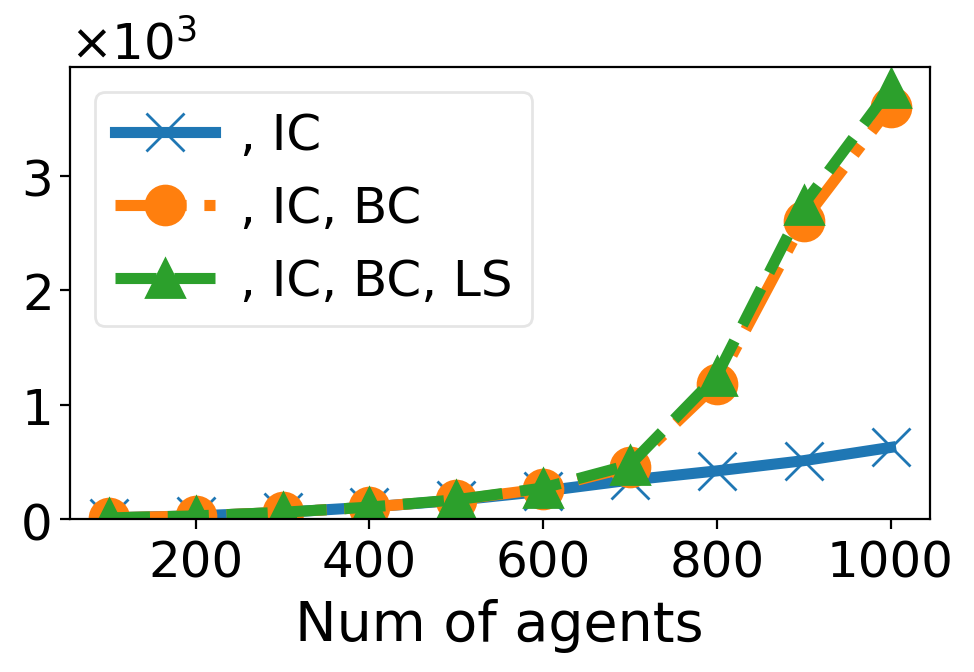}}
\end{minipage}
\hfill
\begin{minipage}{.15\linewidth}
  \centerline{\includegraphics[width=2.1cm]{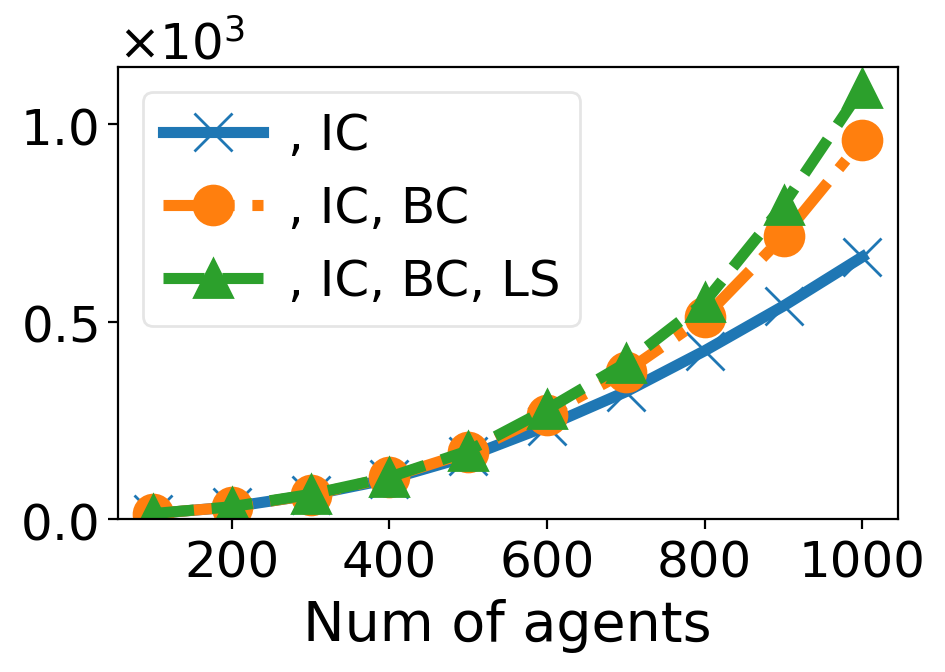}}
\end{minipage}
\vfill

\begin{minipage}{.04\linewidth}
  \rotatebox{90}{\scalebox{0.8}{memory usage (MB)}}
\end{minipage}
\hfill
\begin{minipage}{.15\linewidth}
  \centerline{\includegraphics[width=2.1cm]{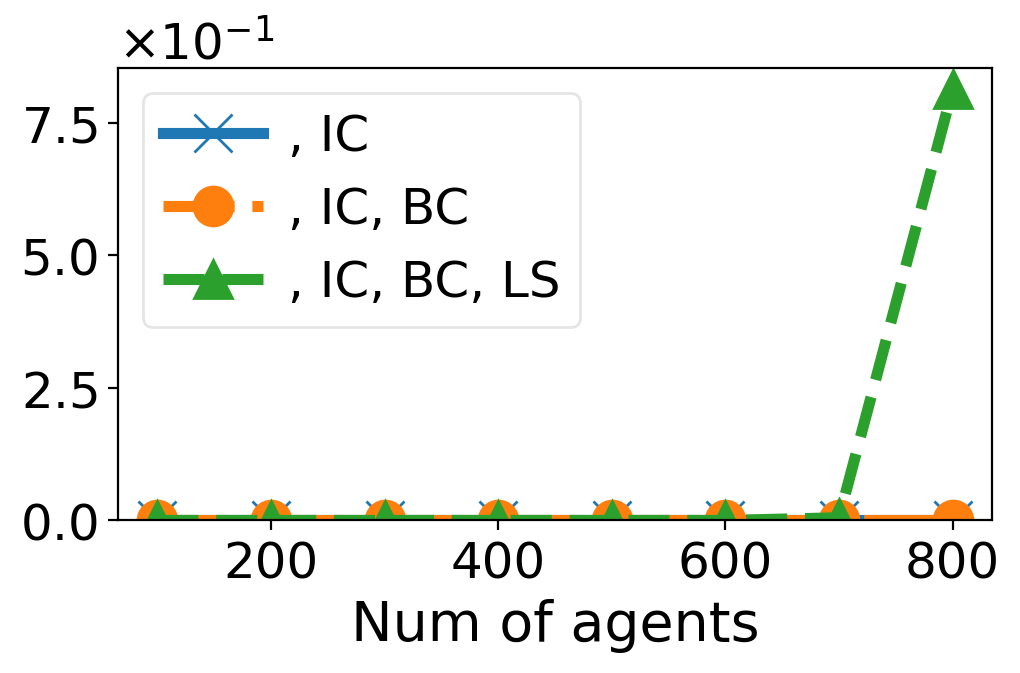}}
\end{minipage}
\hfill
\begin{minipage}{.15\linewidth}
  \centerline{\includegraphics[width=2.1cm]{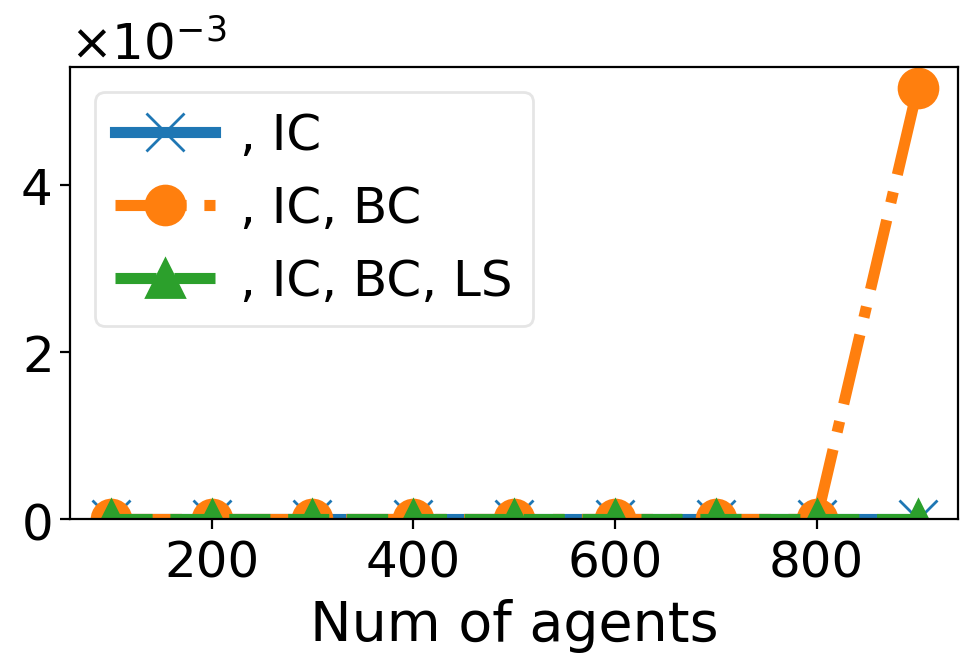}}
\end{minipage}
\hfill
\begin{minipage}{.15\linewidth}
  \centerline{\includegraphics[width=2.1cm]{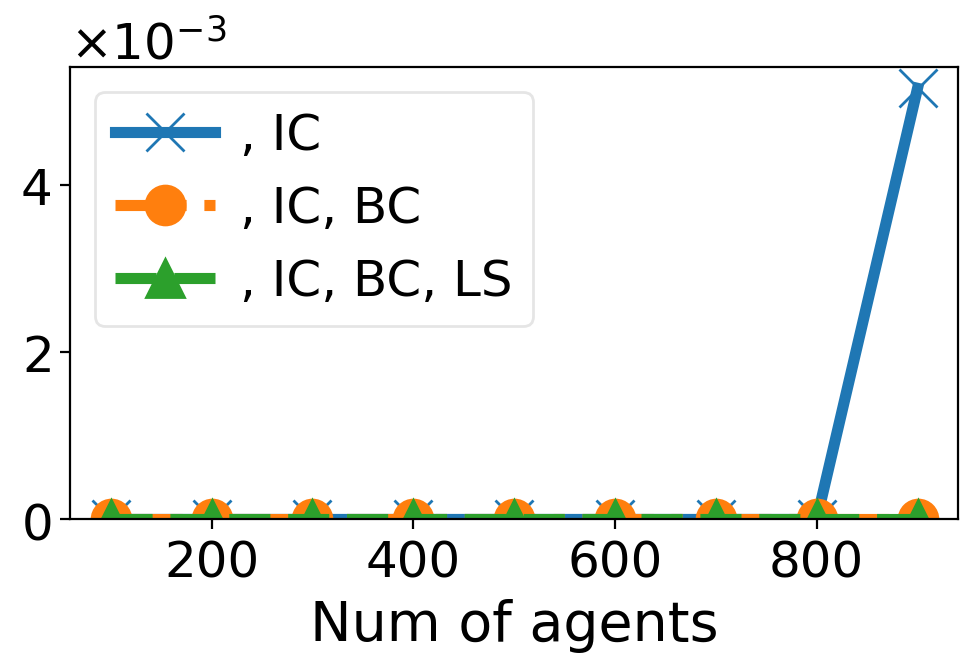}}
\end{minipage}
\hfill
\begin{minipage}{.15\linewidth}
  \centerline{\includegraphics[width=2.1cm]{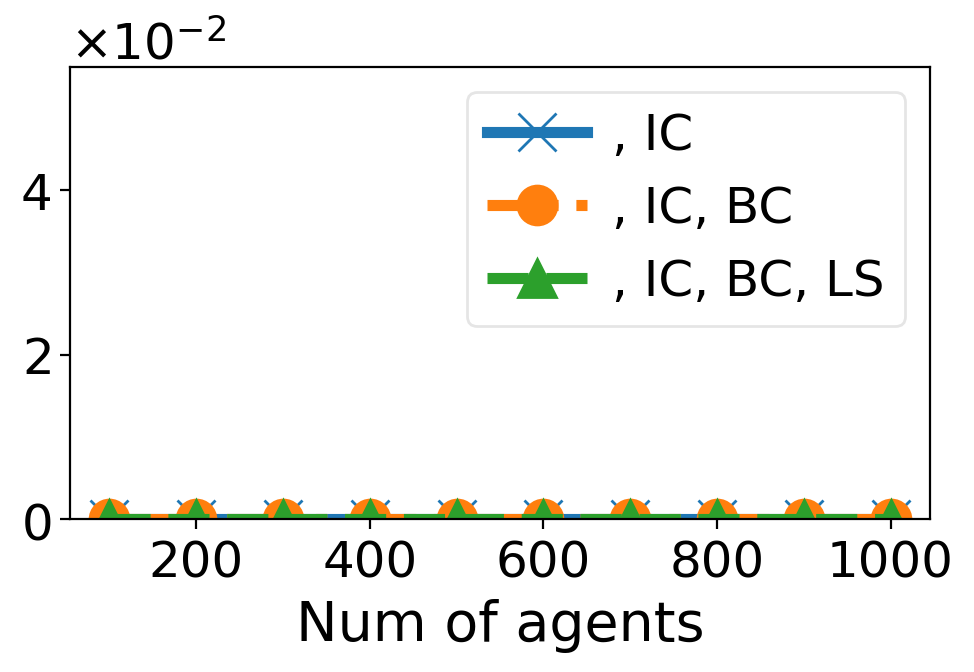}}
\end{minipage}
\hfill
\begin{minipage}{.15\linewidth}
  \centerline{\includegraphics[width=2.1cm]{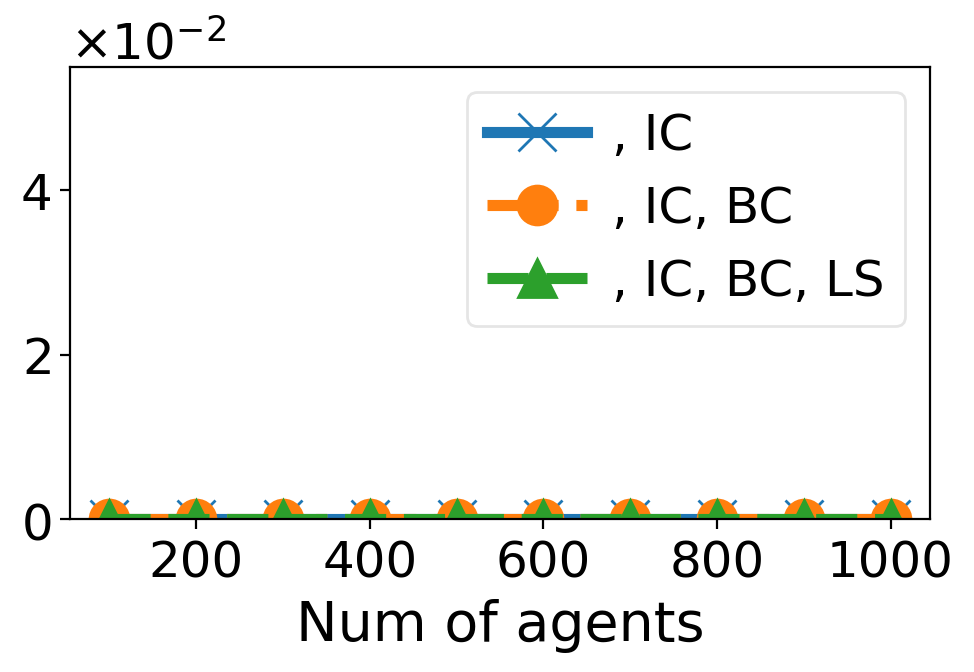}}
\end{minipage}
\hfill
\begin{minipage}{.15\linewidth}
  \centerline{\includegraphics[width=2.1cm]{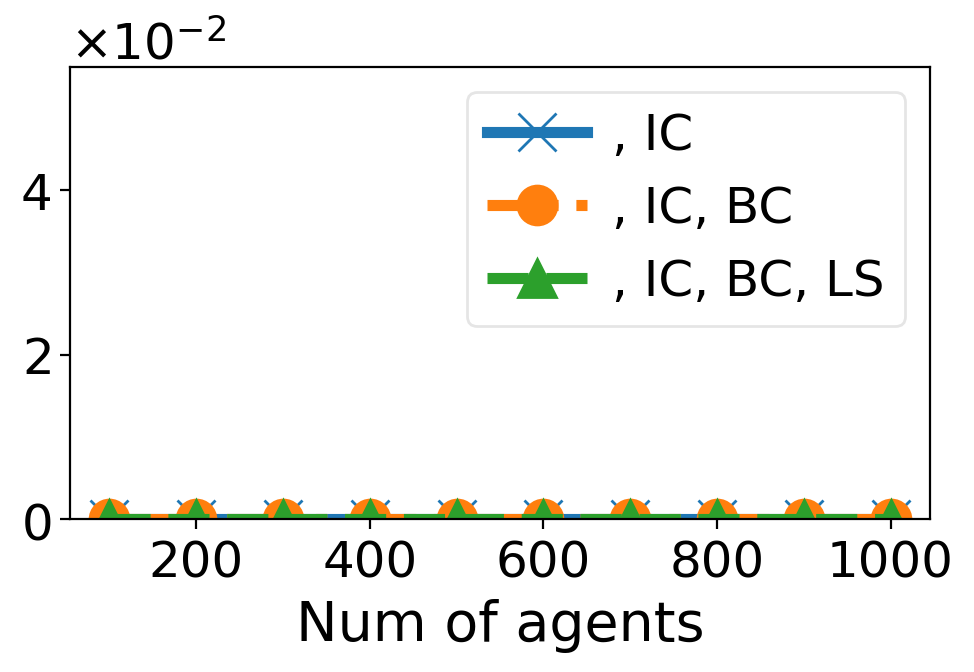}}
\end{minipage}
\vfill

\caption{These figures show our decomposition's performance under various of maps, as number of agents increases, in terms of decomposition rate, number of subproblems time cost and memory usage. We also present scale and number of passable cells (in brackets). The map index is also used in the subsequent figures. }
\label{decomposition1}
\end{figure}

\begin{figure}[h] \tiny

\begin{minipage}{.04\linewidth}
\centerline{ }
\end{minipage}
\hfill
\begin{minipage}{.11\linewidth}
\leftline{\scalebox{0.8}{13.random-32-32}}
\leftline{\scalebox{0.8}{-20}}
\leftline{\scalebox{0.8}{32x32 (819)}}
\end{minipage}
\hfill
\begin{minipage}{.04\linewidth}
\leftline{\includegraphics[width=.5cm]{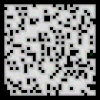}}
\end{minipage}
\hfill
\begin{minipage}{.11\linewidth}
\leftline{\scalebox{0.8}{14.random-64-64}}
\leftline{\scalebox{0.8}{-10}}
\leftline{\scalebox{0.8}{64x64 (3,687)}}
\end{minipage}
\hfill
\begin{minipage}{.04\linewidth}
\rightline{\includegraphics[width=.5cm]{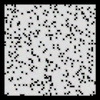}}
\end{minipage}
\hfill
\begin{minipage}{.11\linewidth}
\leftline{\scalebox{0.8}{15.room-32-32-4}}
\leftline{\scalebox{0.8}{32x32 (682)}}
\end{minipage}
\hfill
\begin{minipage}{.04\linewidth}
\leftline{\includegraphics[width=.5cm]{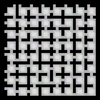}}
\end{minipage}
\hfill
\begin{minipage}{.11\linewidth}
\leftline{\scalebox{0.8}{16.room-64-64-8}}
\leftline{\scalebox{0.8}{64x64 (3,232)}}
\end{minipage}
\hfill
\begin{minipage}{.04\linewidth}
\leftline{\includegraphics[width=.5cm]{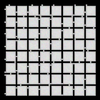}}
\end{minipage}
\hfill
\begin{minipage}{.11\linewidth}
\leftline{\scalebox{0.8}{17.room-64-64-16}}
\leftline{\scalebox{0.8}{64x64 (3,646)}}
\end{minipage}
\hfill
\begin{minipage}{.04\linewidth}
\leftline{\includegraphics[width=.5cm]{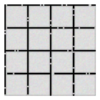}}
\end{minipage}
\hfill
\begin{minipage}{.11\linewidth}
\leftline{\scalebox{0.8}{18.warehouse-10-}}
\leftline{\scalebox{0.8}{20-10-2-1}}
\leftline{\scalebox{0.8}{161x63 (5,699)}}
\end{minipage}
\hfill
\begin{minipage}{.04\linewidth}
\leftline{\includegraphics[width=.5cm]{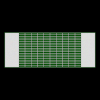}}
\end{minipage}
\vfill

\begin{minipage}{.04\linewidth}
  \rotatebox{90}{\scalebox{0.8}{decomposition rate}}
\end{minipage}
\hfill
\begin{minipage}{.15\linewidth}
  \centerline{\includegraphics[width=2.1cm]{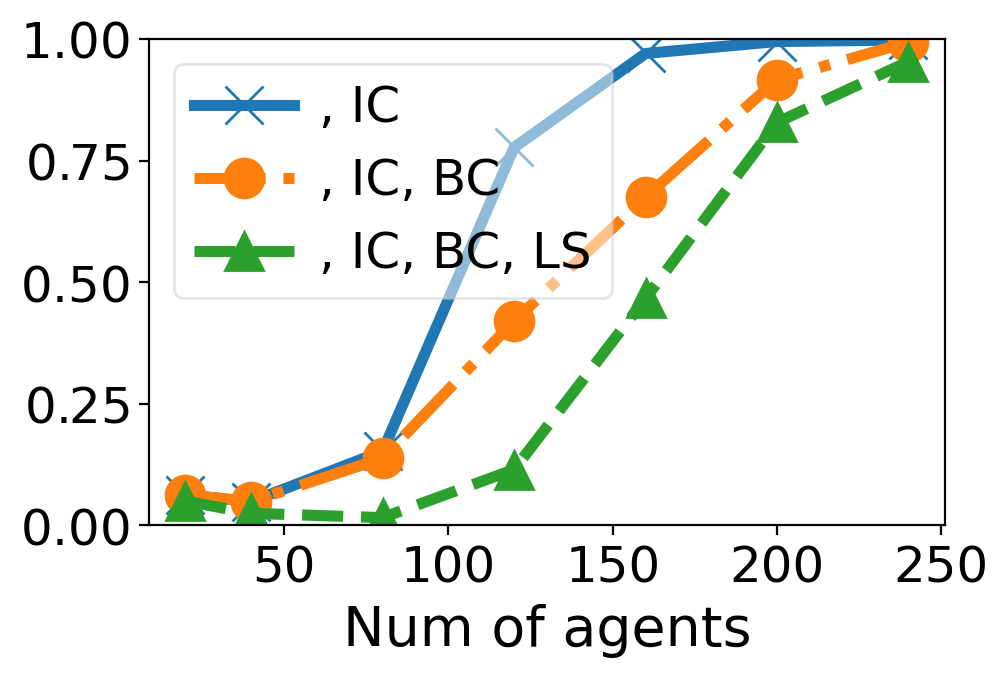}}
\end{minipage}
\hfill
\begin{minipage}{.15\linewidth}
  \centerline{\includegraphics[width=2.1cm]{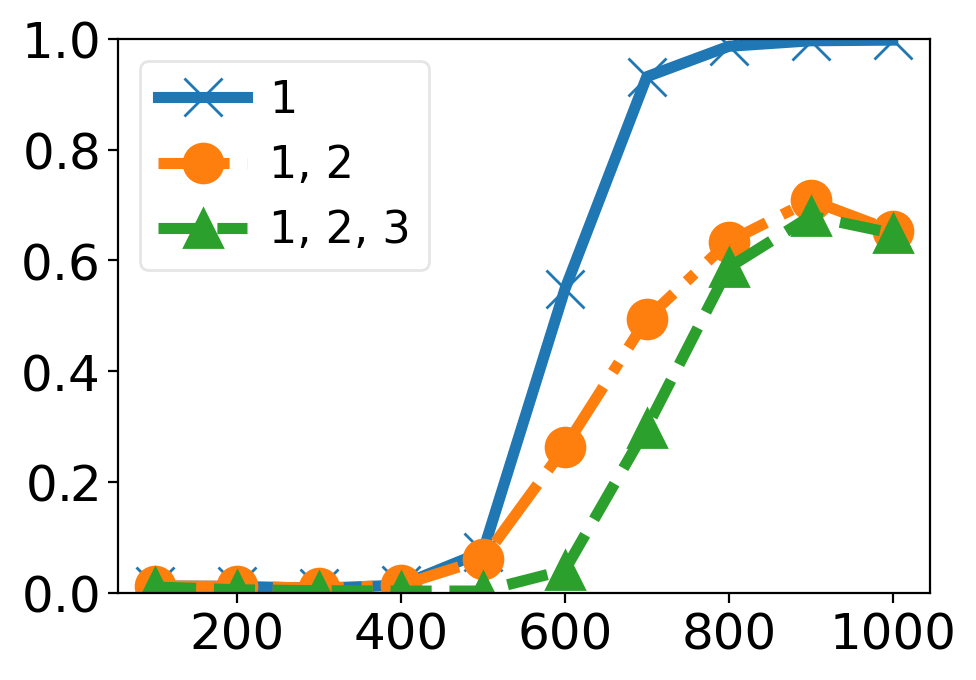}}
\end{minipage}
\hfill
\begin{minipage}{.15\linewidth}
  \centerline{\includegraphics[width=2.1cm]{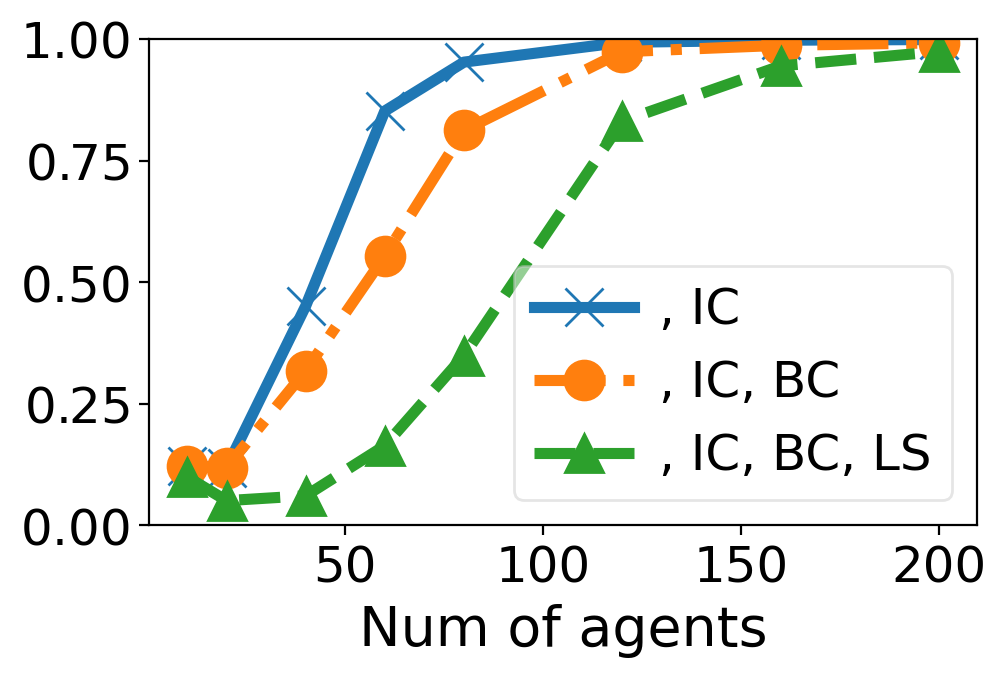}}
\end{minipage}
\hfill
\begin{minipage}{.15\linewidth}
  \centerline{\includegraphics[width=2.1cm]{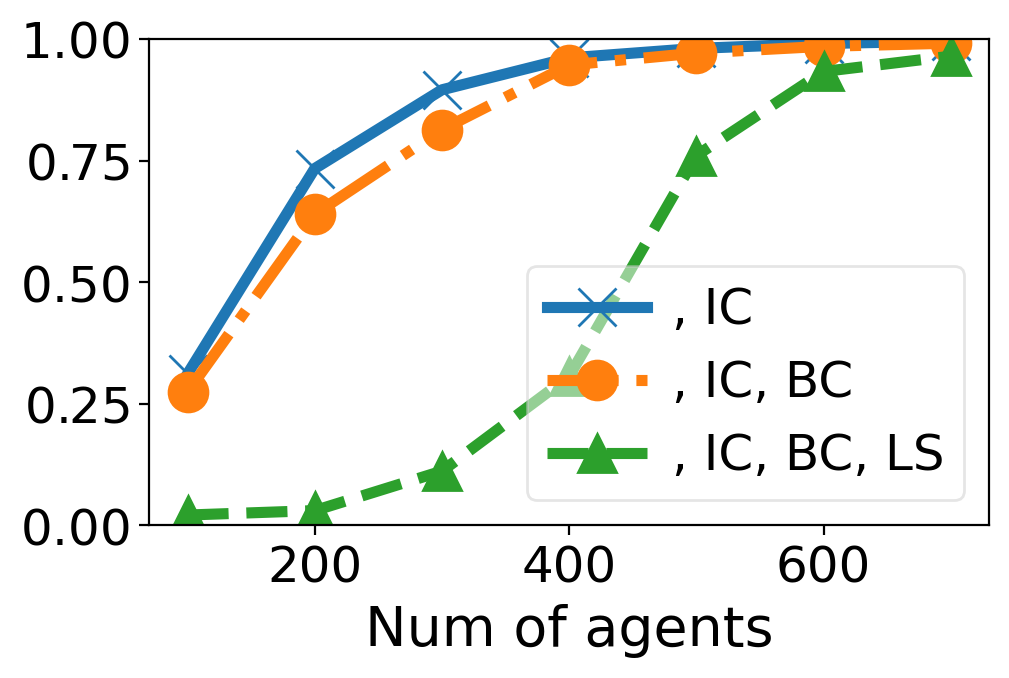}}
\end{minipage}
\hfill
\begin{minipage}{.15\linewidth}
  \centerline{\includegraphics[width=2.1cm]{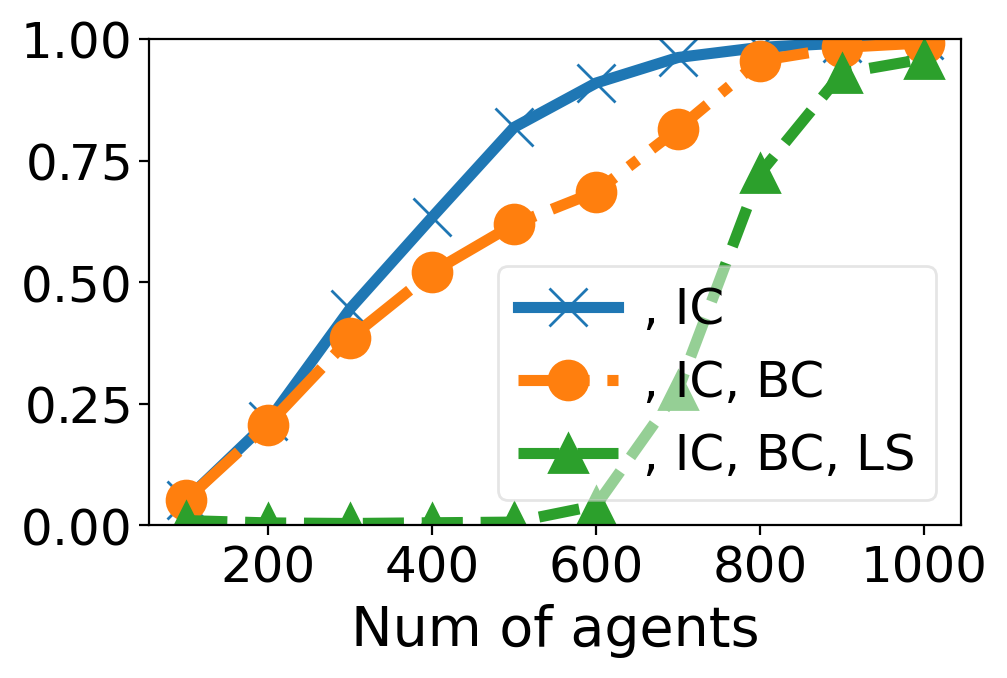}}
\end{minipage}
\hfill
\begin{minipage}{.15\linewidth}
  \centerline{\includegraphics[width=2.1cm]{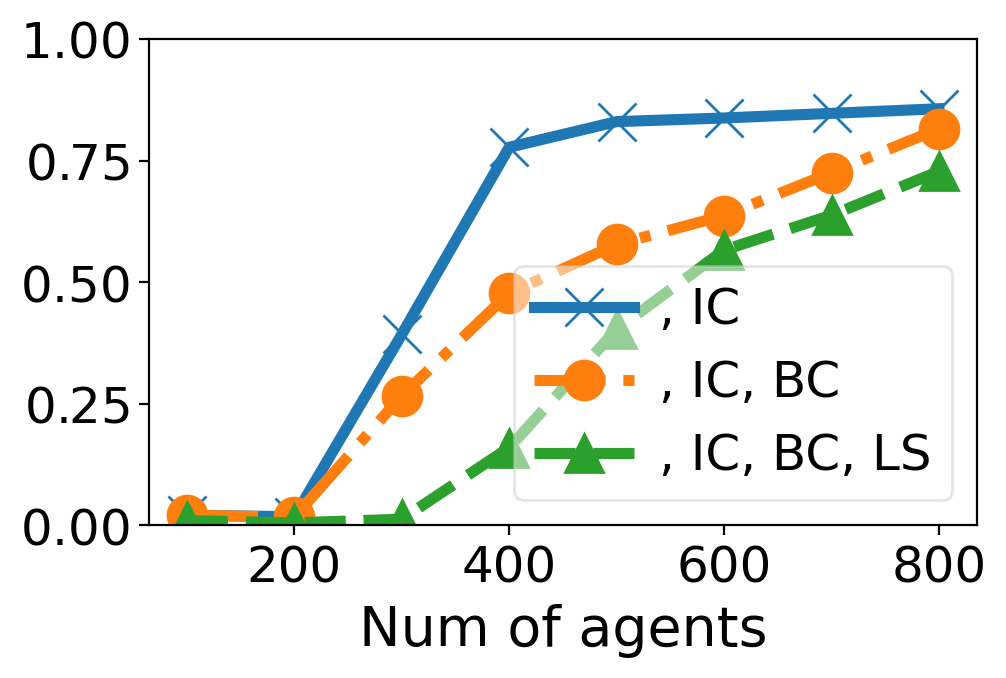}}
\end{minipage}
\vfill

\begin{minipage}{.04\linewidth}
  \rotatebox{90}{\scalebox{0.8}{subproblems}}
\end{minipage}
\hfill
\begin{minipage}{.15\linewidth}
  \centerline{\includegraphics[width=2.1cm]{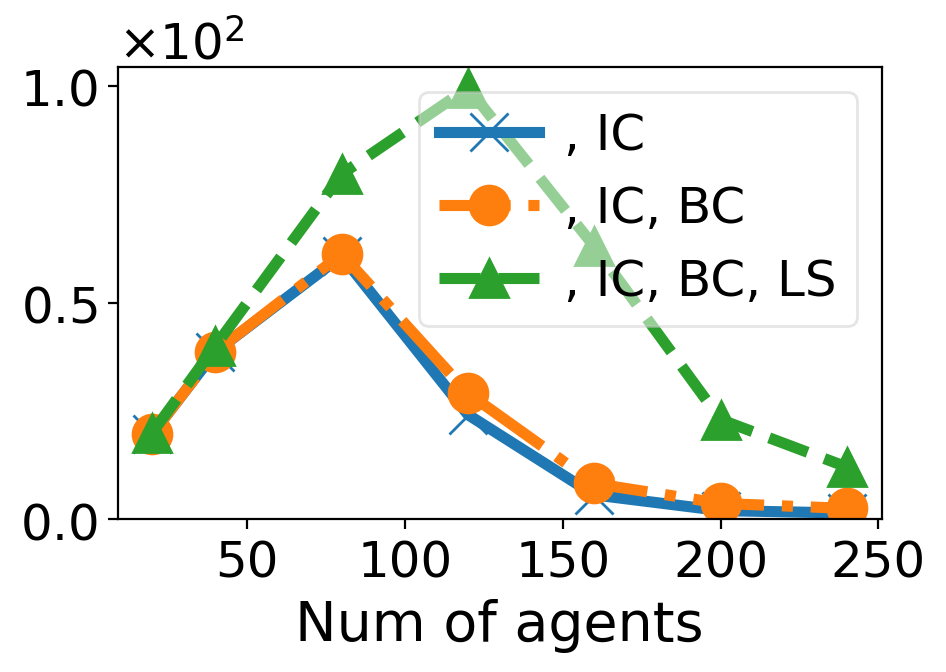}}
\end{minipage}
\hfill
\begin{minipage}{.15\linewidth}
  \centerline{\includegraphics[width=2.1cm]{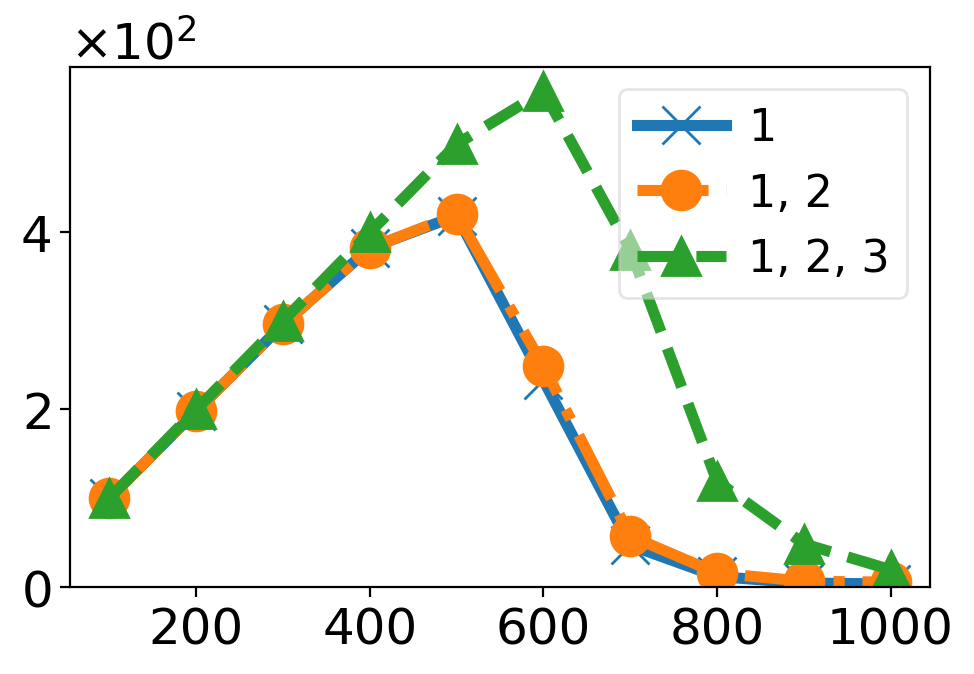}}
\end{minipage}
\hfill
\begin{minipage}{.15\linewidth}
  \centerline{\includegraphics[width=2.1cm]{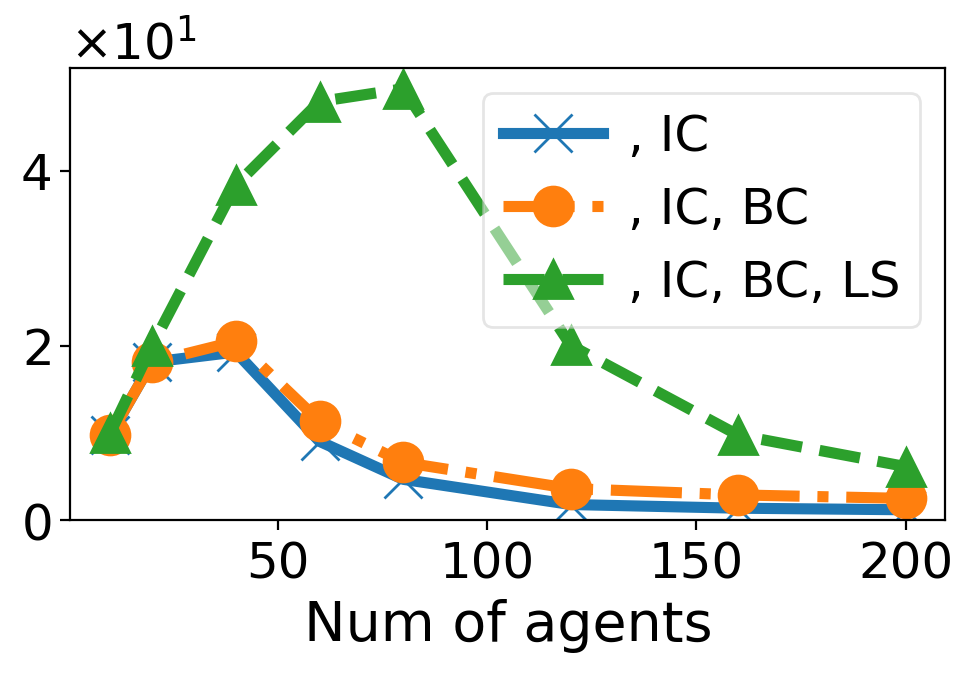}}
\end{minipage}
\hfill
\begin{minipage}{.15\linewidth}
  \centerline{\includegraphics[width=2.1cm]{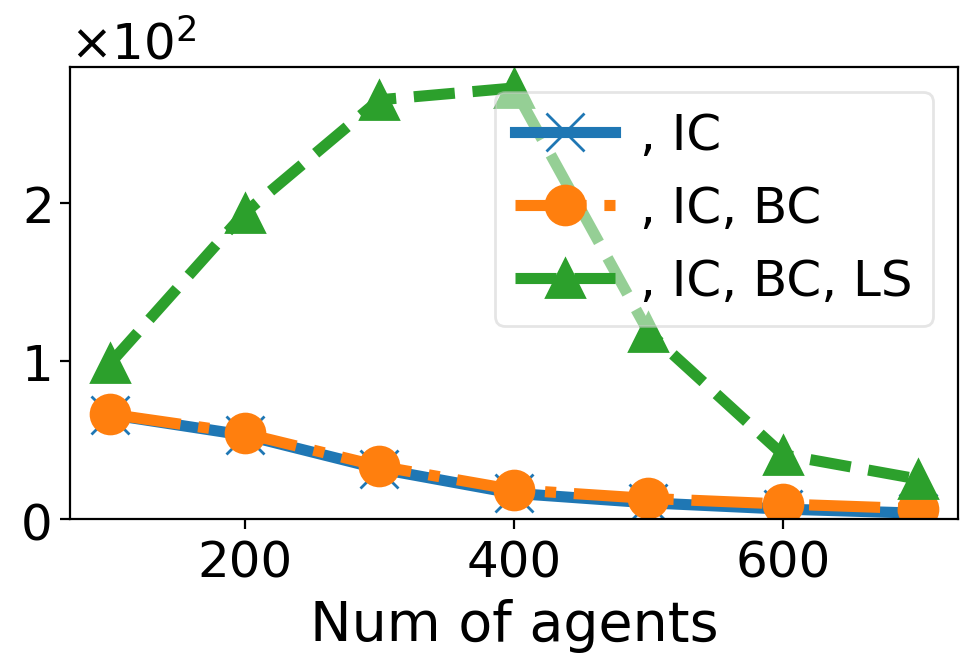}}
\end{minipage}
\hfill
\begin{minipage}{.15\linewidth}
  \centerline{\includegraphics[width=2.1cm]{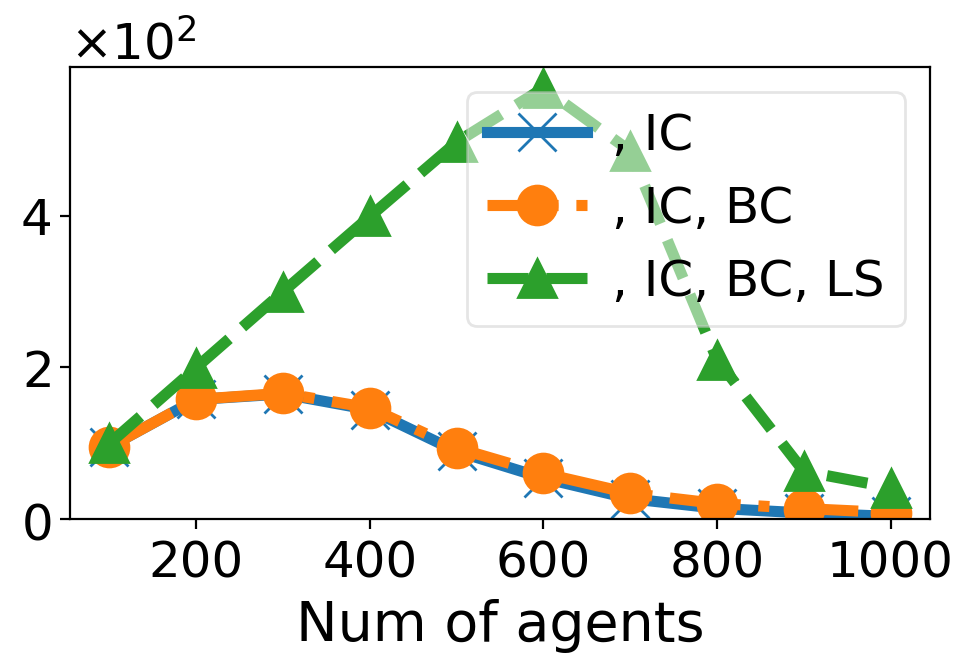}}
\end{minipage}
\hfill
\begin{minipage}{.15\linewidth}
  \centerline{\includegraphics[width=2.1cm]{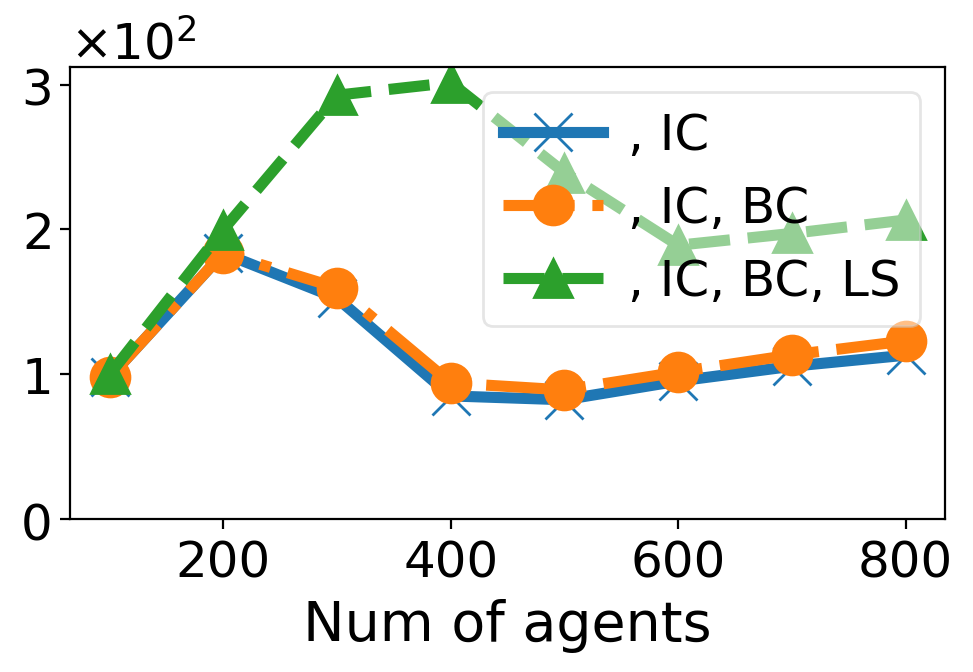}}
\end{minipage}
\vfill

\begin{minipage}{.04\linewidth}
  \rotatebox{90}{\scalebox{0.8}{time cost (ms)}}
\end{minipage}
\hfill
\begin{minipage}{.15\linewidth}
  \centerline{\includegraphics[width=2.1cm]{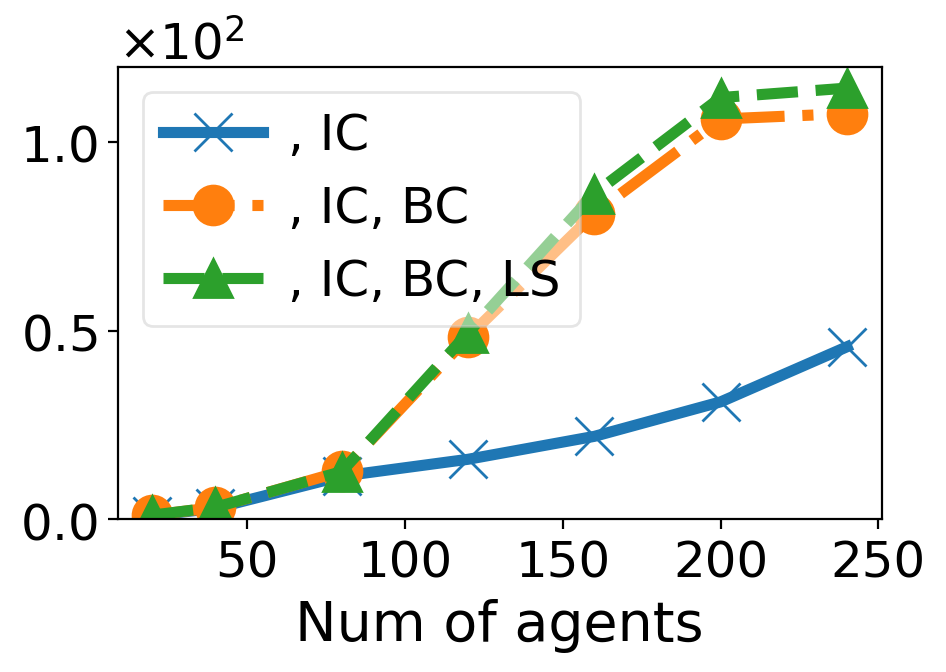}}
\end{minipage}
\hfill
\begin{minipage}{.15\linewidth}
  \centerline{\includegraphics[width=2.1cm]{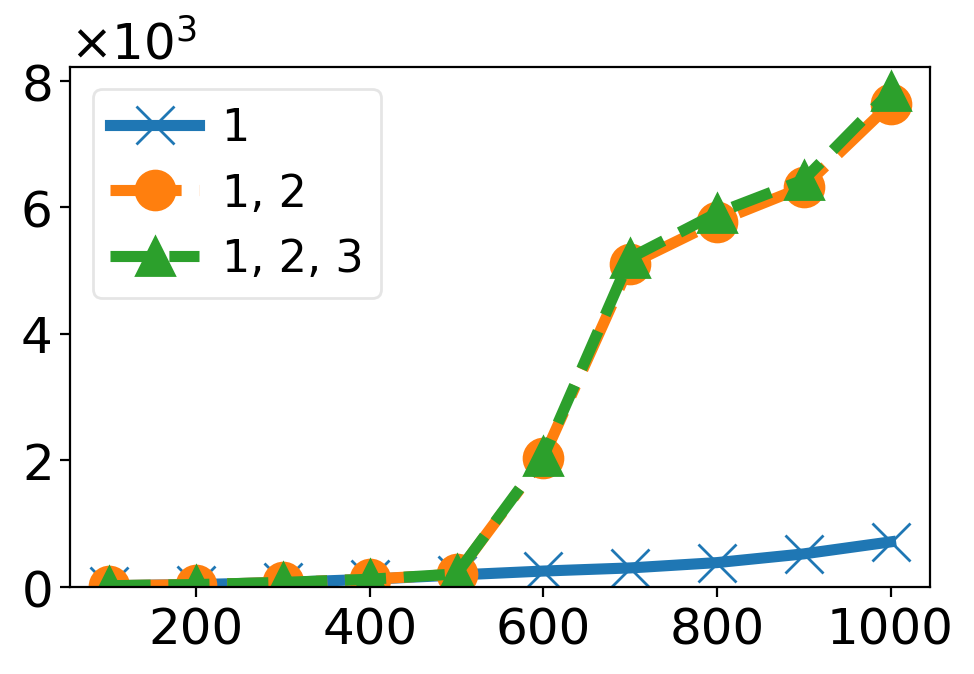}}
\end{minipage}
\hfill
\begin{minipage}{.15\linewidth}
  \centerline{\includegraphics[width=2.1cm]{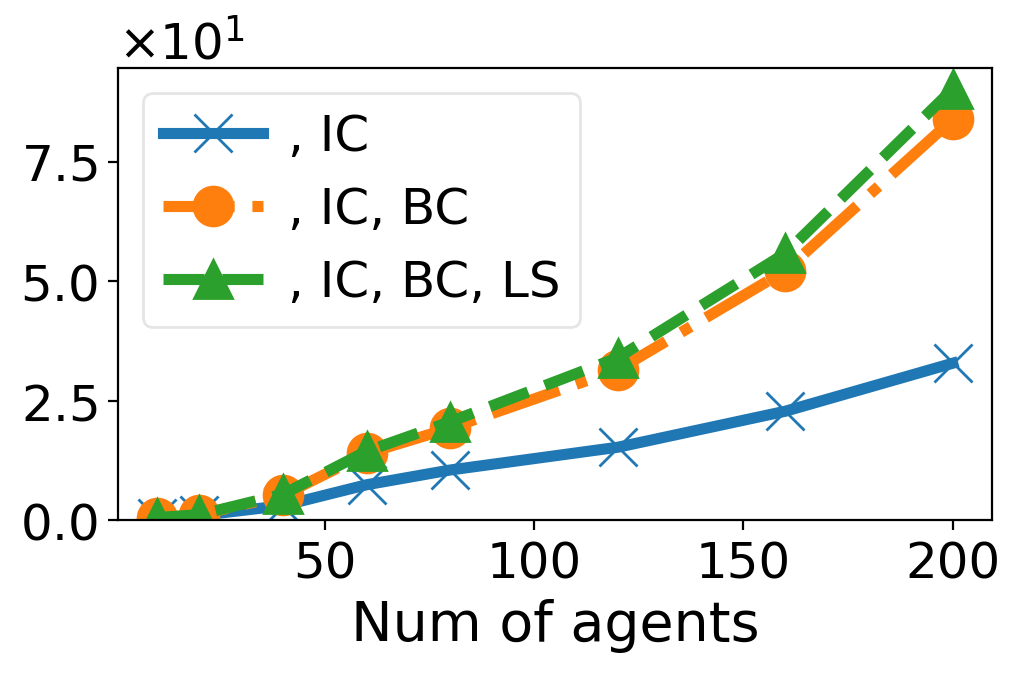}}
\end{minipage}
\hfill
\begin{minipage}{.15\linewidth}
  \centerline{\includegraphics[width=2.1cm]{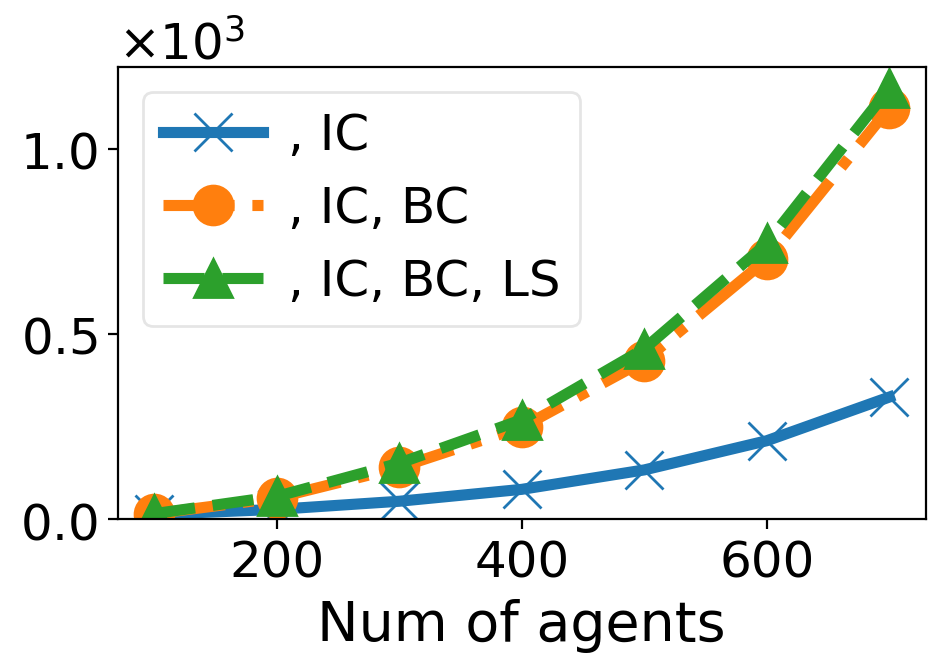}}
\end{minipage}
\hfill
\begin{minipage}{.15\linewidth}
  \centerline{\includegraphics[width=2.1cm]{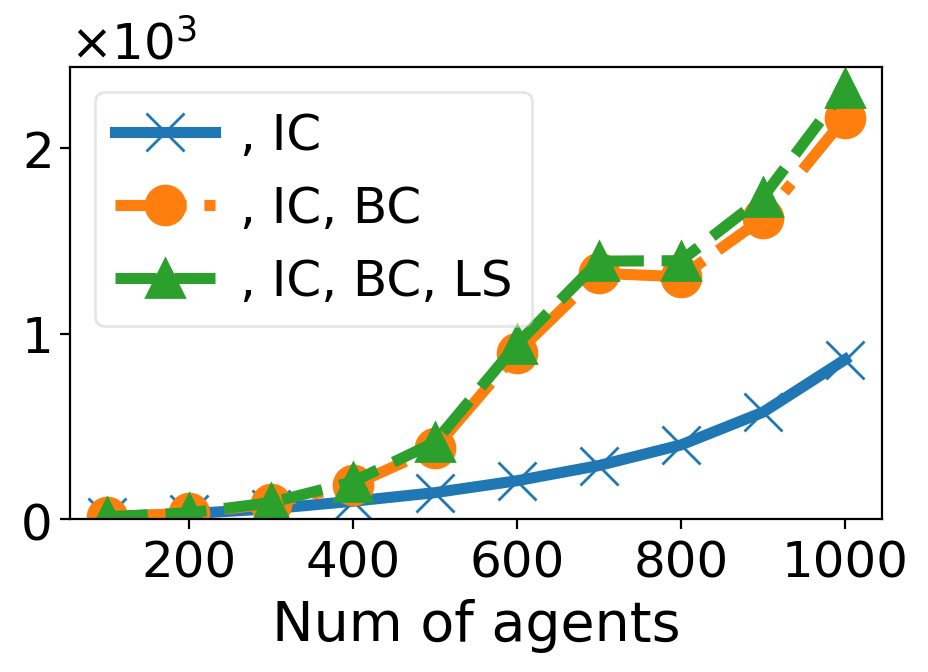}}
\end{minipage}
\hfill
\begin{minipage}{.15\linewidth}
  \centerline{\includegraphics[width=2.1cm]{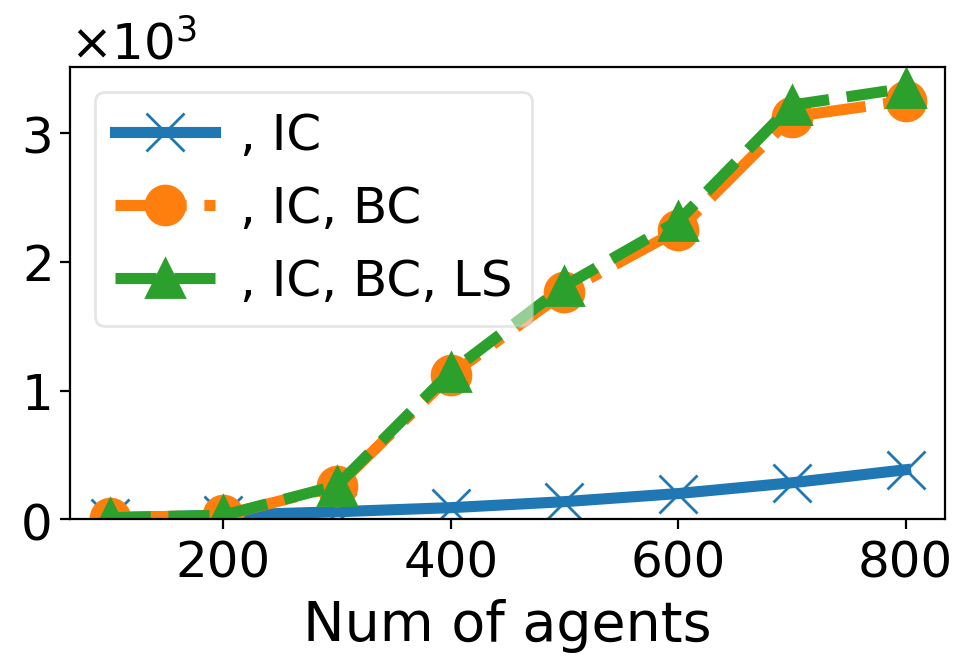}}
\end{minipage}
\vfill

\begin{minipage}{.04\linewidth}
  \rotatebox{90}{\scalebox{0.8}{memory usage (MB)}}
\end{minipage}
\hfill
\begin{minipage}{.15\linewidth}
  \centerline{\includegraphics[width=2.1cm]{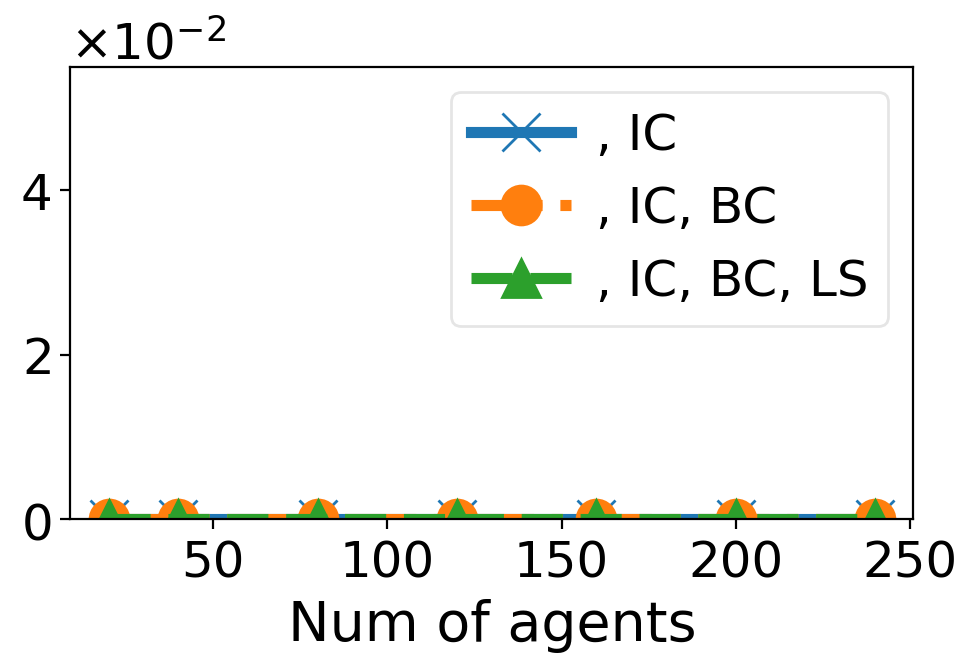}}
\end{minipage}
\hfill
\begin{minipage}{.15\linewidth}
  \centerline{\includegraphics[width=2.1cm]{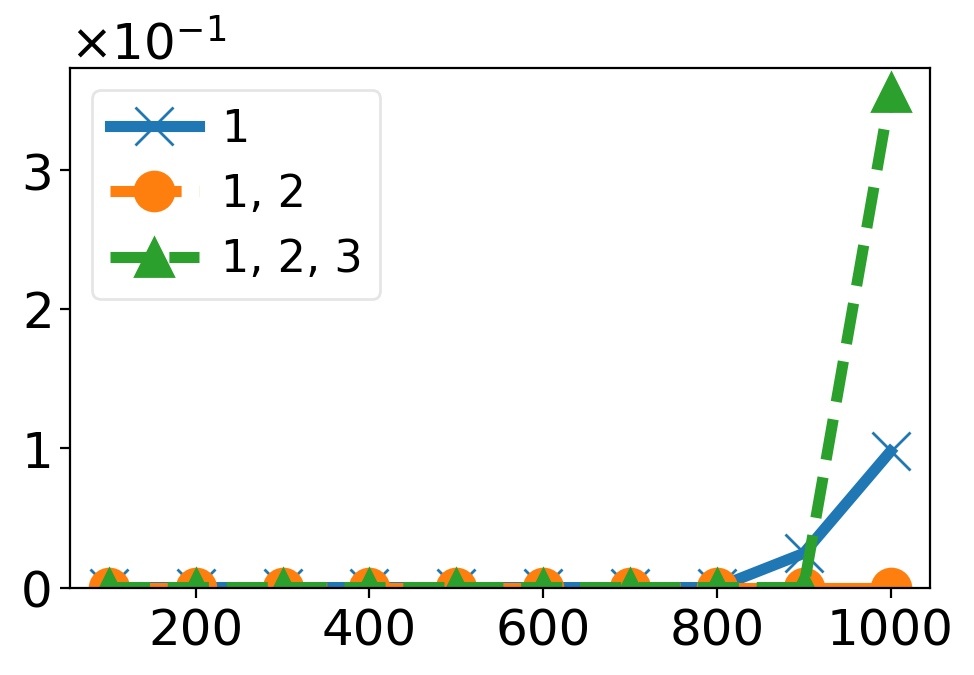}}
\end{minipage}
\hfill
\begin{minipage}{.15\linewidth}
  \centerline{\includegraphics[width=2.1cm]{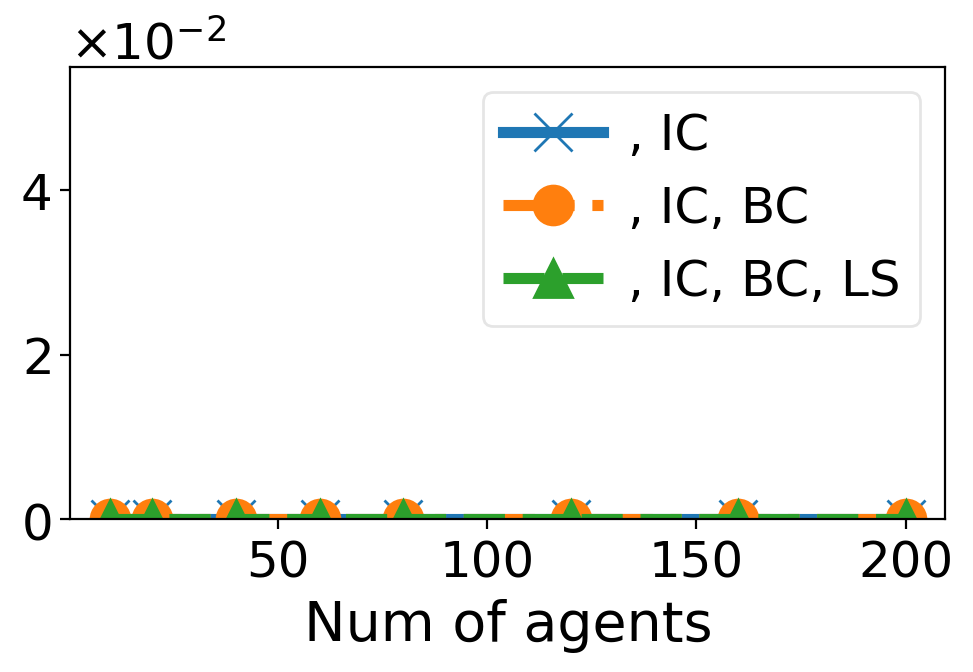}}
\end{minipage}
\hfill
\begin{minipage}{.15\linewidth}
  \centerline{\includegraphics[width=2.1cm]{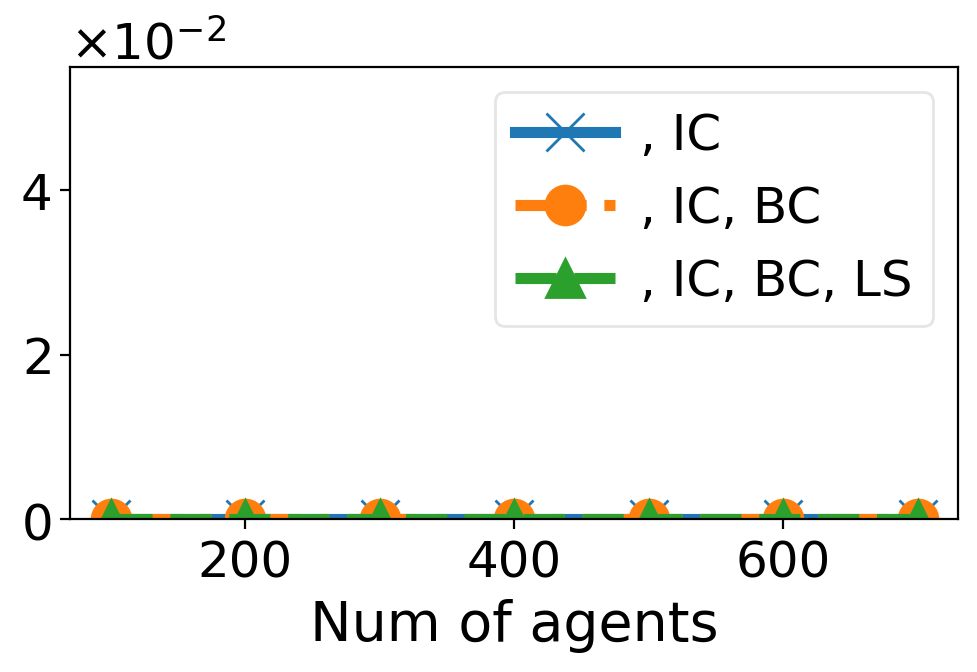}}
\end{minipage}
\hfill
\begin{minipage}{.15\linewidth}
  \centerline{\includegraphics[width=2.1cm]{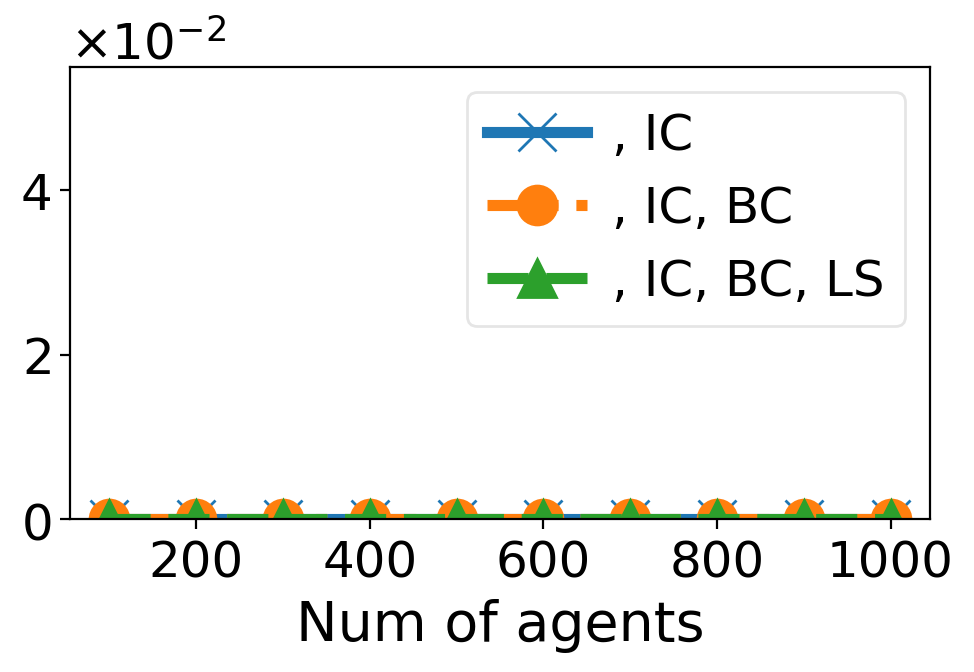}}
\end{minipage}
\hfill
\begin{minipage}{.15\linewidth}
  \centerline{\includegraphics[width=2.1cm]{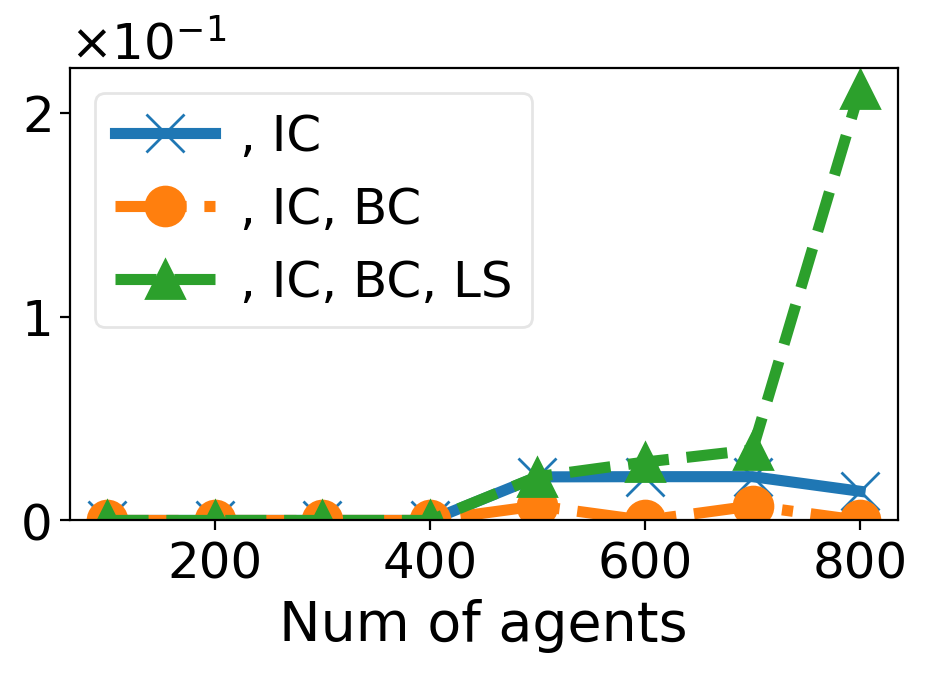}}
\end{minipage}
\vfill

\begin{minipage}{.04\linewidth}
\centerline{ }
\end{minipage}
\hfill
\begin{minipage}{.11\linewidth}
\leftline{\scalebox{0.8}{19.warehouse-10-}}
\leftline{\scalebox{0.8}{20-10-2-2}}
\leftline{\scalebox{0.8}{170x84 (9,776)}}
\end{minipage}
\hfill
\begin{minipage}{.04\linewidth}
\leftline{\includegraphics[width=.5cm]{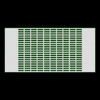}}
\end{minipage}
\hfill
\begin{minipage}{.11\linewidth}
\leftline{\scalebox{0.8}{20.warehouse-20-}}
\leftline{\scalebox{0.8}{40-10-2-1}}
\leftline{\scalebox{0.8}{321x123 (22,599)}}
\end{minipage}
\hfill
\begin{minipage}{.04\linewidth}
\rightline{\includegraphics[width=.5cm]{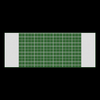}}
\end{minipage}
\hfill
\begin{minipage}{.10\linewidth}
\leftline{\scalebox{0.8}{21.warehouse-}}
\leftline{\scalebox{0.8}{20-40-10-2-2}}
\leftline{\scalebox{0.8}{340x164(38,756)}}
\end{minipage}
\hfill
\begin{minipage}{.04\linewidth}
\leftline{\includegraphics[width=.5cm]{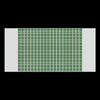}}
\end{minipage}
\hfill
\begin{minipage}{.11\linewidth}
\leftline{\scalebox{0.8}{22.Boston\_0\_256}}
\leftline{\scalebox{0.8}{256x256 (47,768)}}
\end{minipage}
\hfill
\begin{minipage}{.04\linewidth}
\leftline{\includegraphics[width=.5cm]{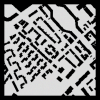}}
\end{minipage}
\hfill
\begin{minipage}{.11\linewidth}
\leftline{\scalebox{0.8}{23.lt\_gallowstemp}}
\leftline{\scalebox{0.8}{lar\_n}}
\leftline{\scalebox{0.8}{251x180 (10,021)}}
\end{minipage}
\hfill
\begin{minipage}{.04\linewidth}
\rightline{\includegraphics[width=.5cm]{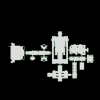}}
\end{minipage}
\hfill
\begin{minipage}{.11\linewidth}
\leftline{\scalebox{0.8}{24.ost003d}}
\leftline{\scalebox{0.8}{194x194 (13,214)}}
\end{minipage}
\hfill
\begin{minipage}{.04\linewidth}
\leftline{\includegraphics[width=.5cm]{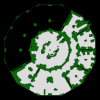}}
\end{minipage}
\vfill

\begin{minipage}{.04\linewidth}
  \rotatebox{90}{\scalebox{0.8}{decomposition rate}}
\end{minipage}
\hfill
\begin{minipage}{.15\linewidth}
  \centerline{\includegraphics[width=2.1cm]{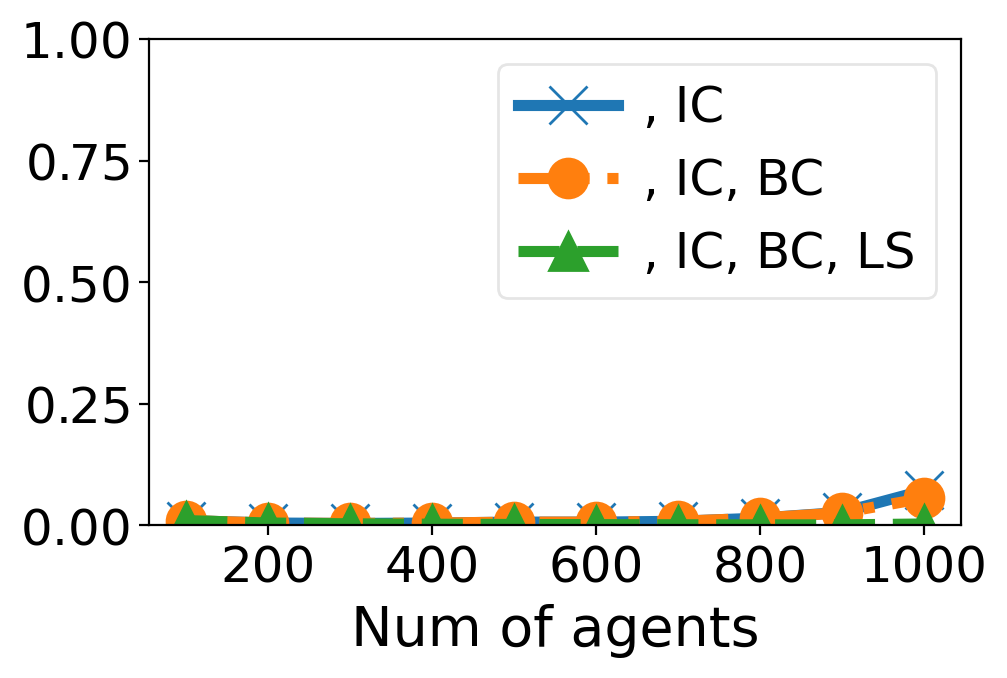}}
\end{minipage}
\hfill
\begin{minipage}{.15\linewidth}
  \centerline{\includegraphics[width=2.1cm]{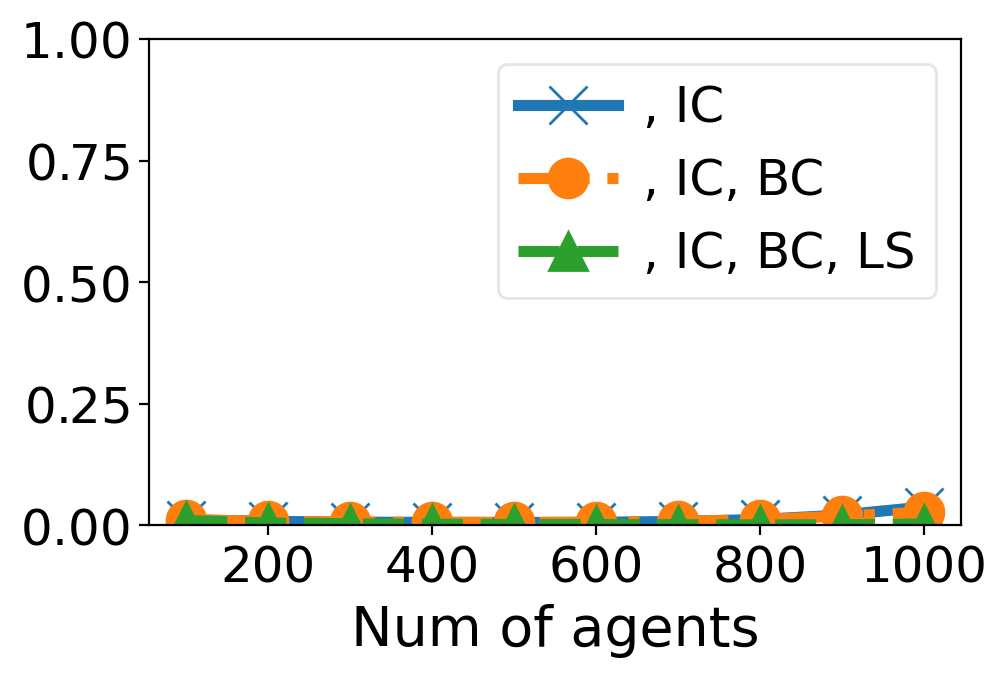}}
\end{minipage}
\hfill
\begin{minipage}{.15\linewidth}
  \centerline{\includegraphics[width=2.1cm]{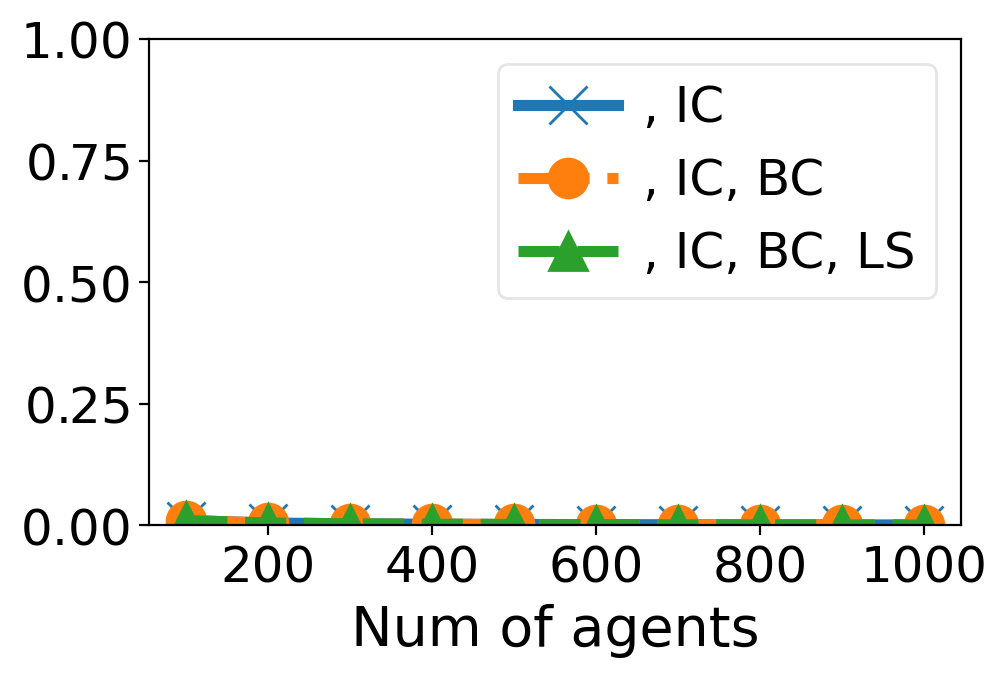}}
\end{minipage}
\hfill
\begin{minipage}{.15\linewidth}
  \centerline{\includegraphics[width=2.1cm]{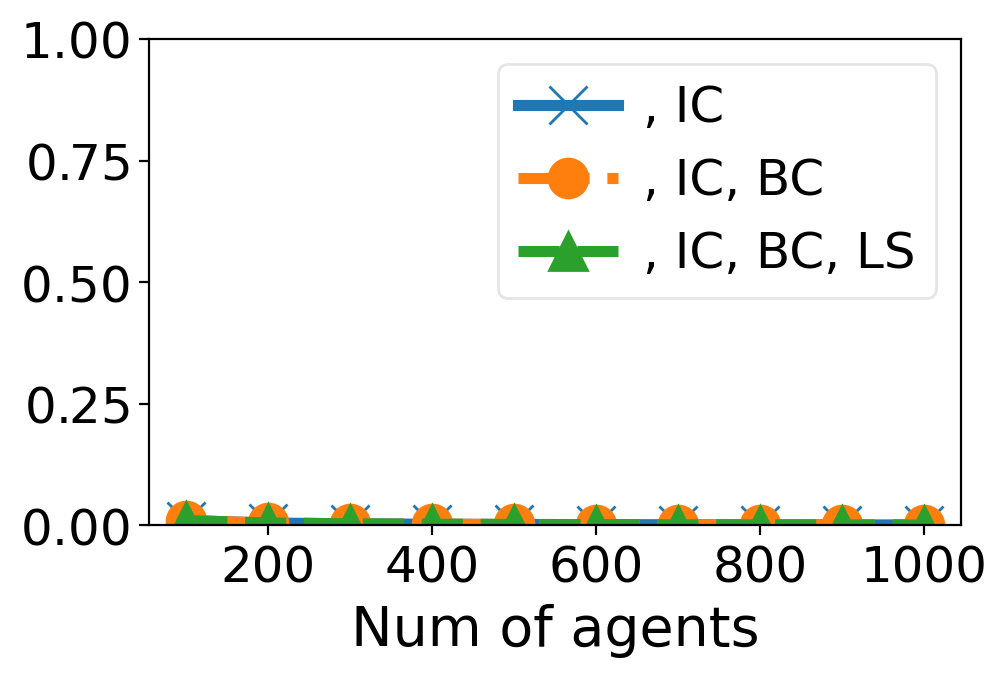}}
\end{minipage}
\hfill
\begin{minipage}{.15\linewidth}
  \centerline{\includegraphics[width=2.1cm]{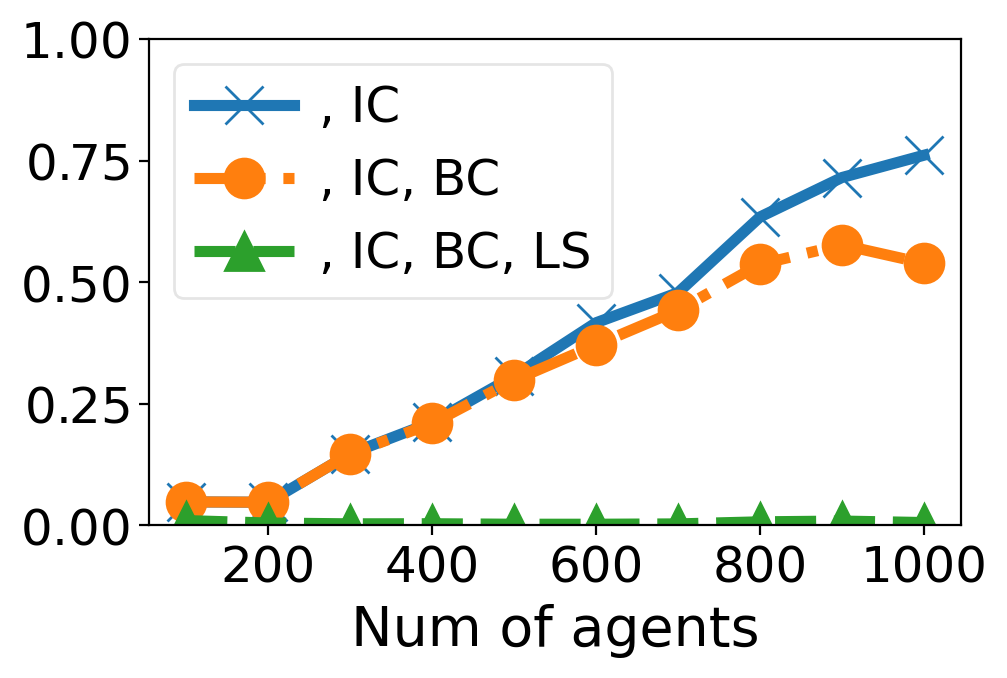}}
\end{minipage}
\hfill
\begin{minipage}{.15\linewidth}
  \centerline{\includegraphics[width=2.1cm]{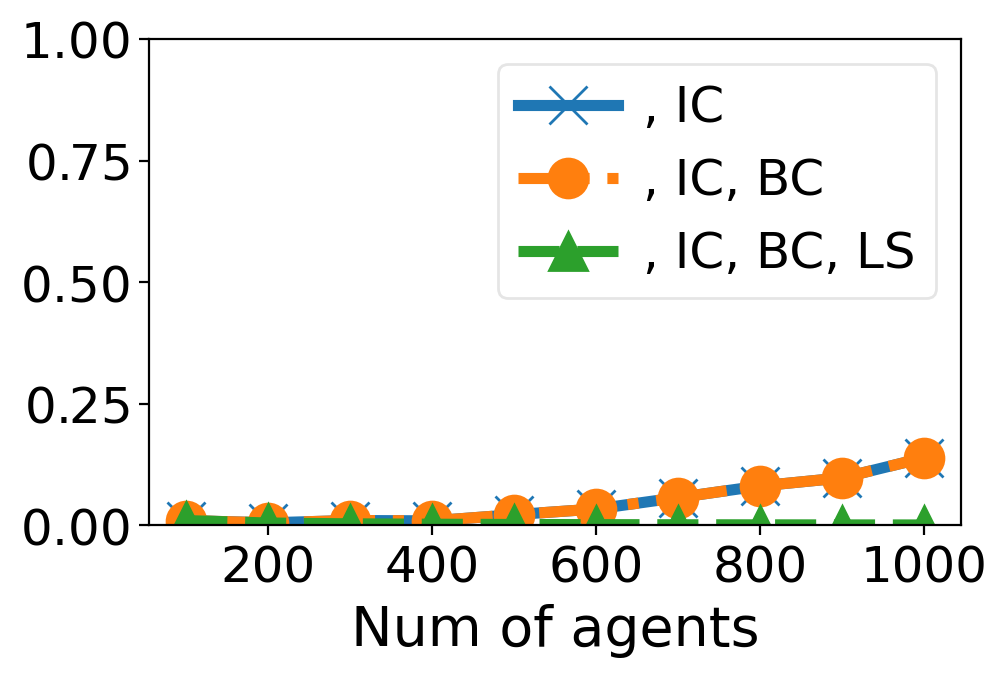}}
\end{minipage}
\vfill

\begin{minipage}{.04\linewidth}
  \rotatebox{90}{\scalebox{0.8}{subproblems}}
\end{minipage}
\hfill
\begin{minipage}{.15\linewidth}
  \centerline{\includegraphics[width=2.1cm]{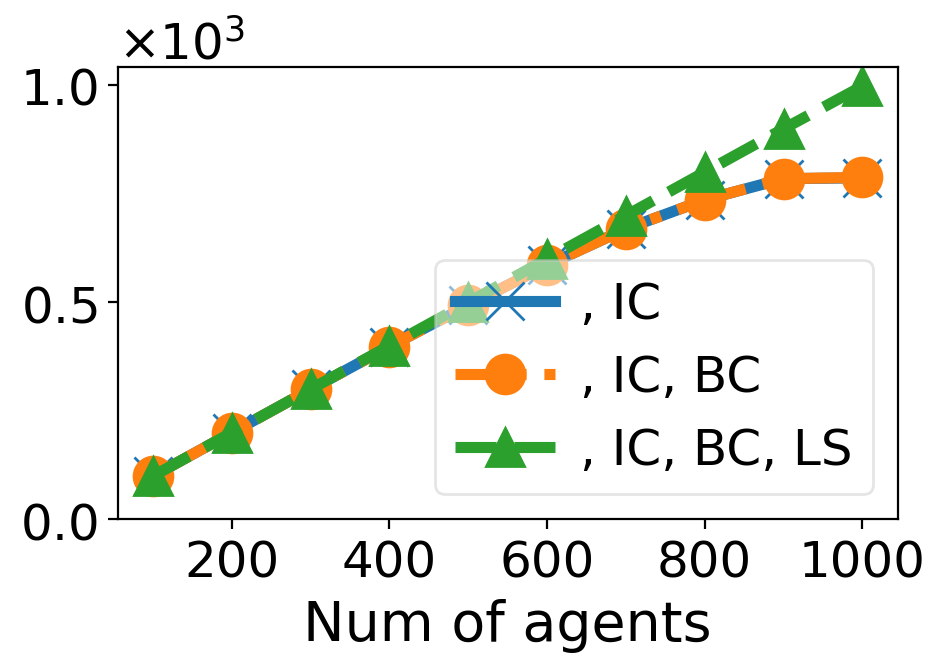}}
\end{minipage}
\hfill
\begin{minipage}{.15\linewidth}
  \centerline{\includegraphics[width=2.1cm]{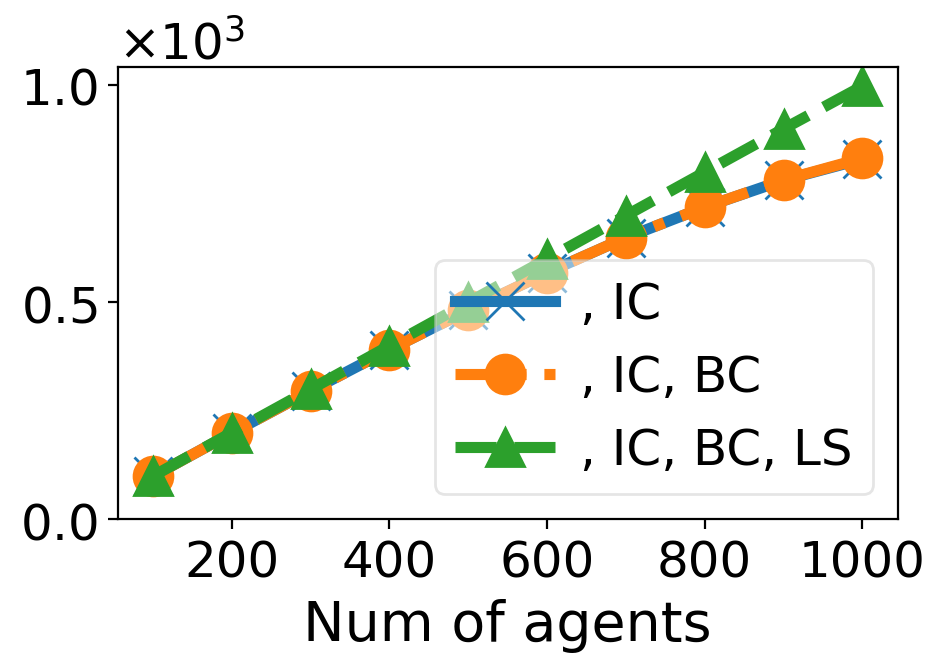}}
\end{minipage}
\hfill
\begin{minipage}{.15\linewidth}
  \centerline{\includegraphics[width=2.1cm]{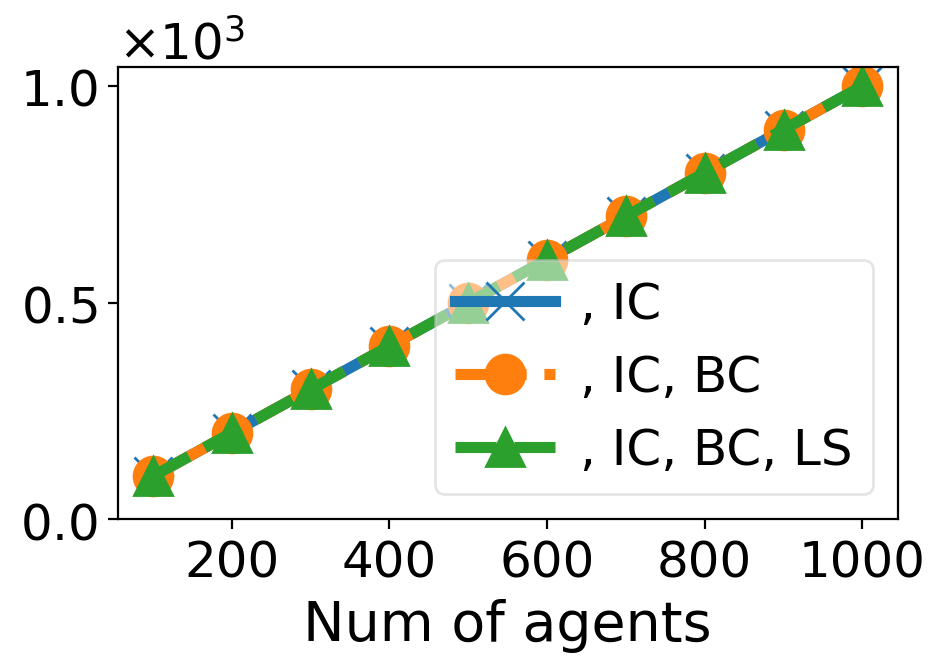}}
\end{minipage}
\hfill
\begin{minipage}{.15\linewidth}
  \centerline{\includegraphics[width=2.1cm]{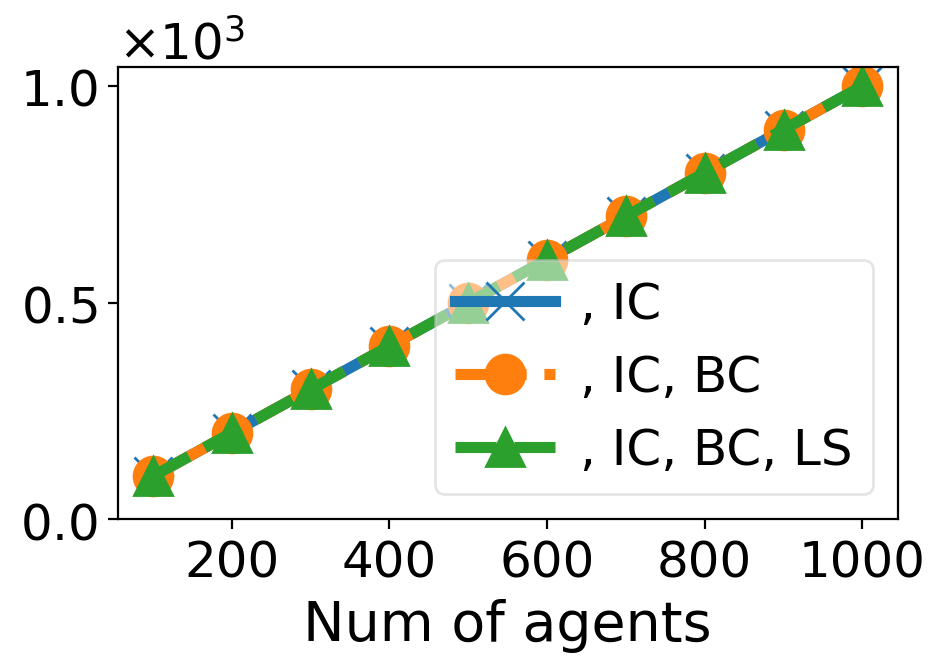}}
\end{minipage}
\hfill
\begin{minipage}{.15\linewidth}
  \centerline{\includegraphics[width=2.1cm]{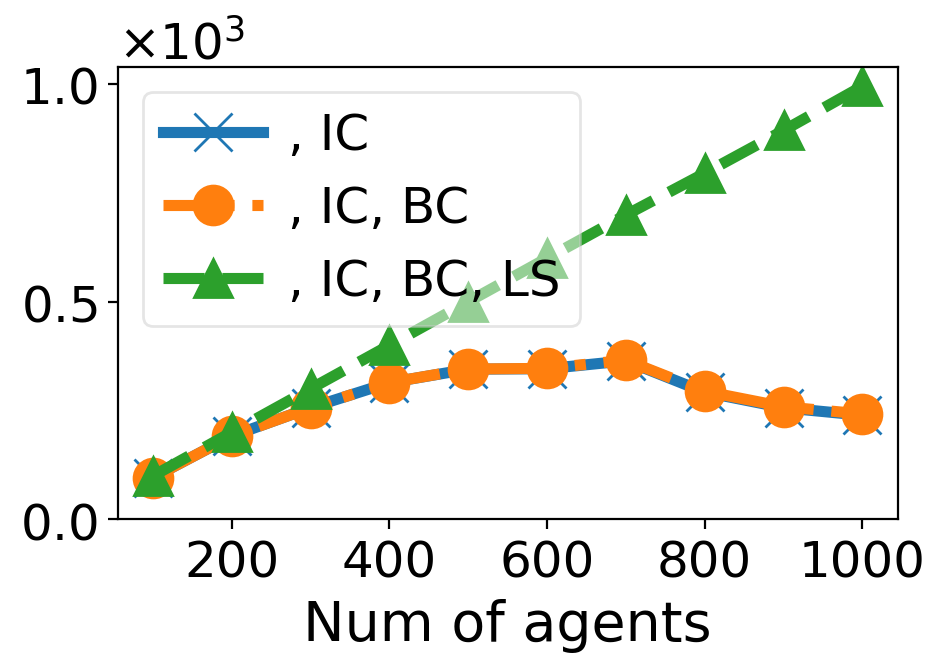}}
\end{minipage}
\hfill
\begin{minipage}{.15\linewidth}
  \centerline{\includegraphics[width=2.1cm]{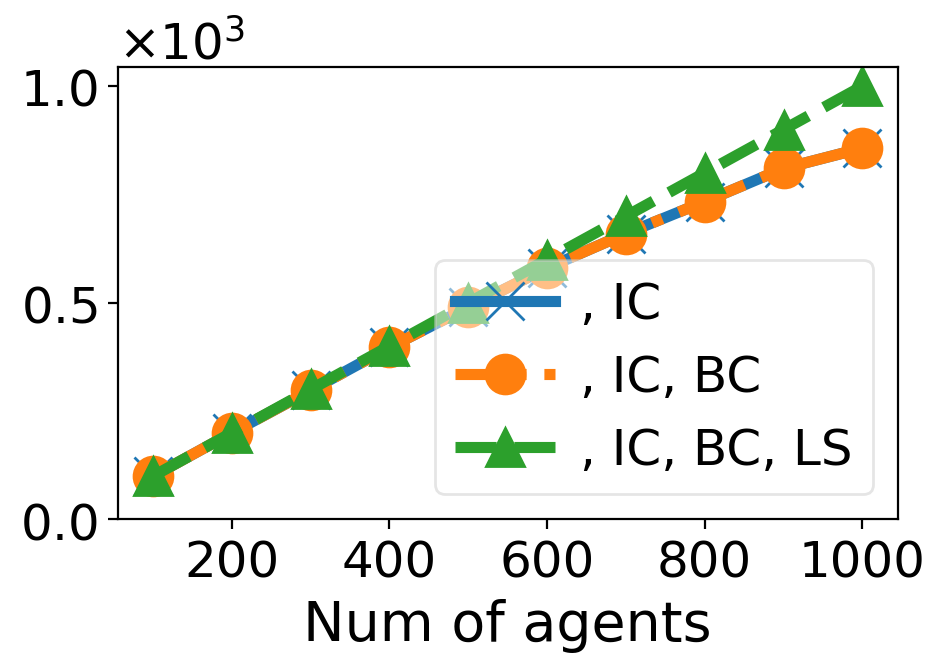}}
\end{minipage}
\vfill

\begin{minipage}{.04\linewidth}
  \rotatebox{90}{\scalebox{0.8}{time cost (ms)}}
\end{minipage}
\hfill
\begin{minipage}{.15\linewidth}
  \centerline{\includegraphics[width=2.1cm]{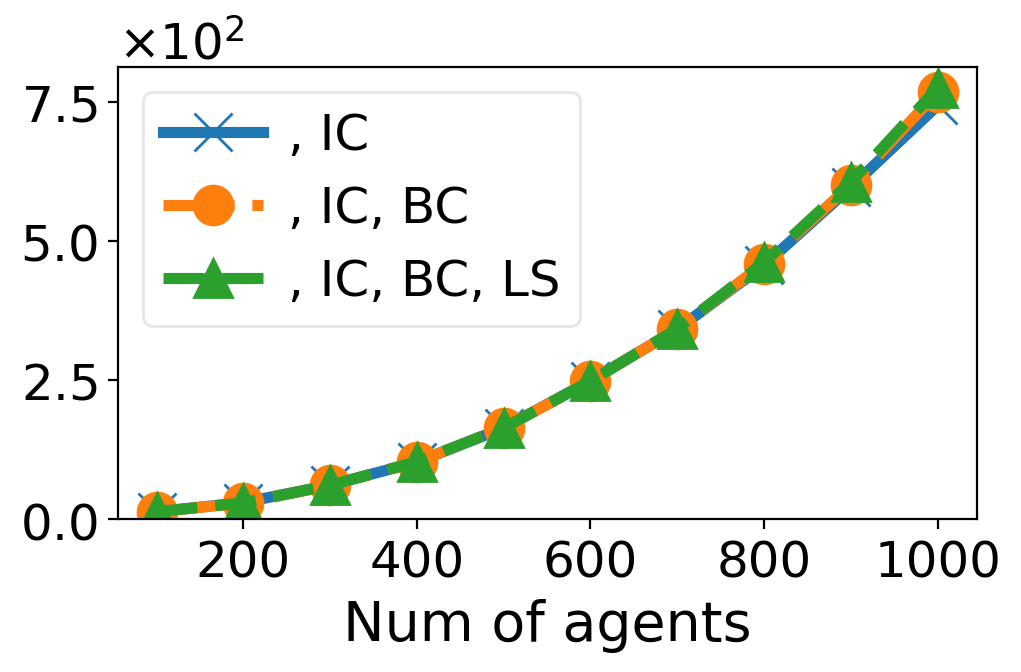}}
\end{minipage}
\hfill
\begin{minipage}{.15\linewidth}
  \centerline{\includegraphics[width=2.1cm]{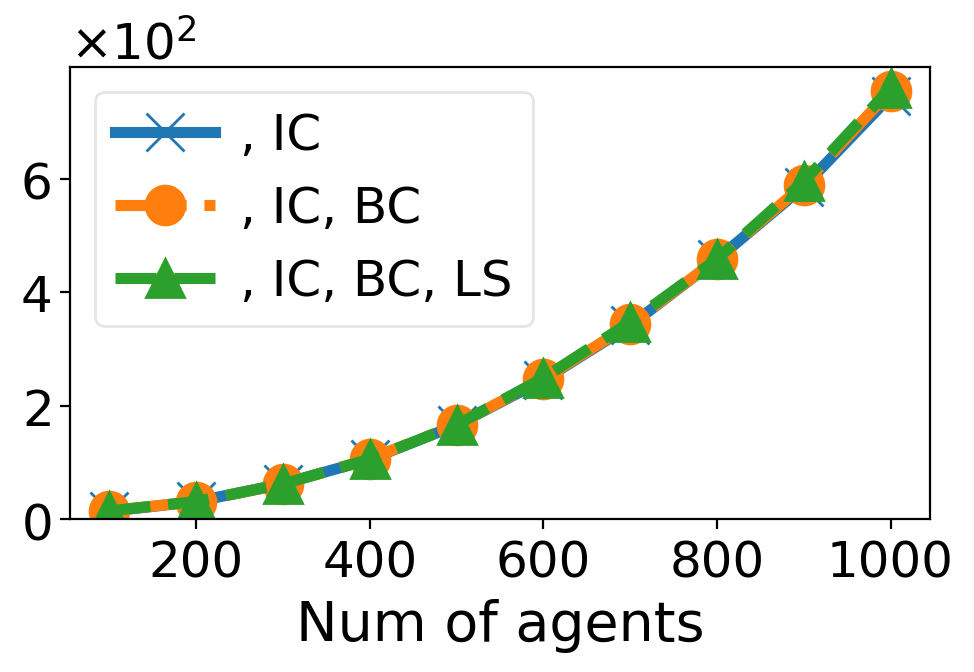}}
\end{minipage}
\hfill
\begin{minipage}{.15\linewidth}
  \centerline{\includegraphics[width=2.1cm]{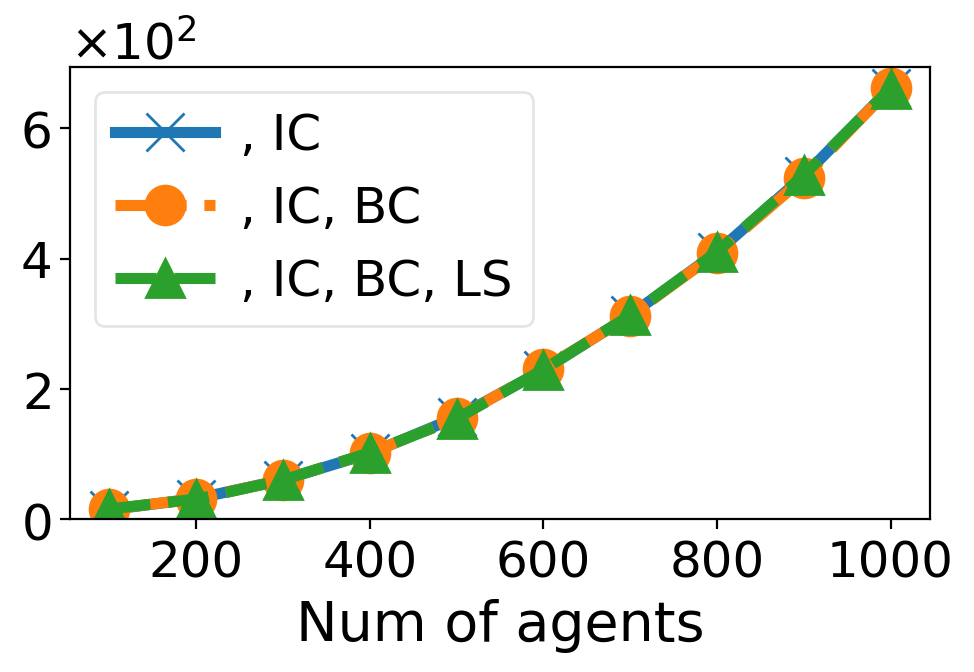}}
\end{minipage}
\hfill
\begin{minipage}{.15\linewidth}
  \centerline{\includegraphics[width=2.1cm]{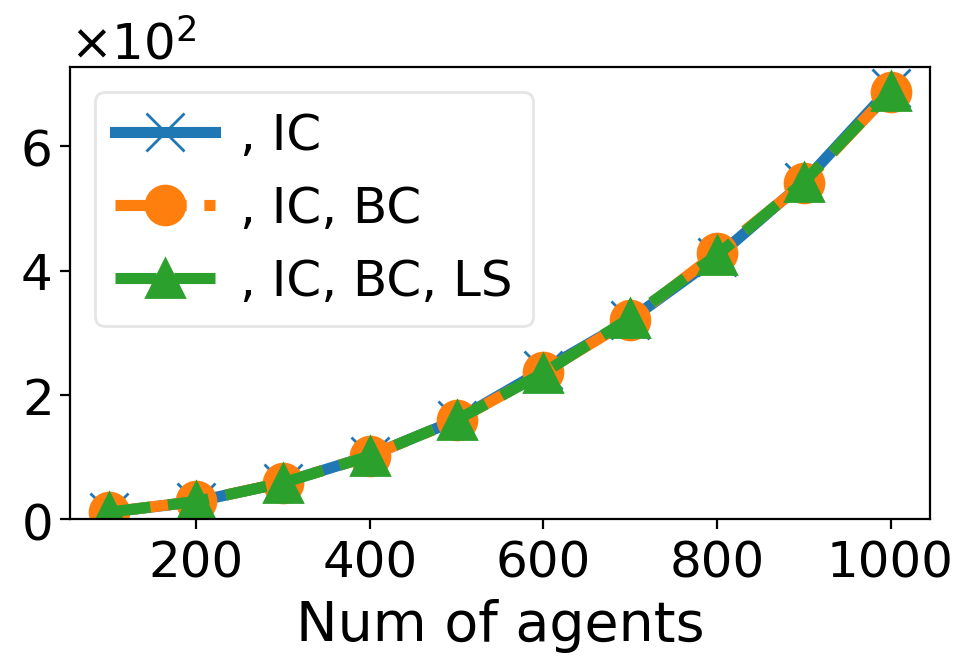}}
\end{minipage}
\hfill
\begin{minipage}{.15\linewidth}
  \centerline{\includegraphics[width=2.1cm]{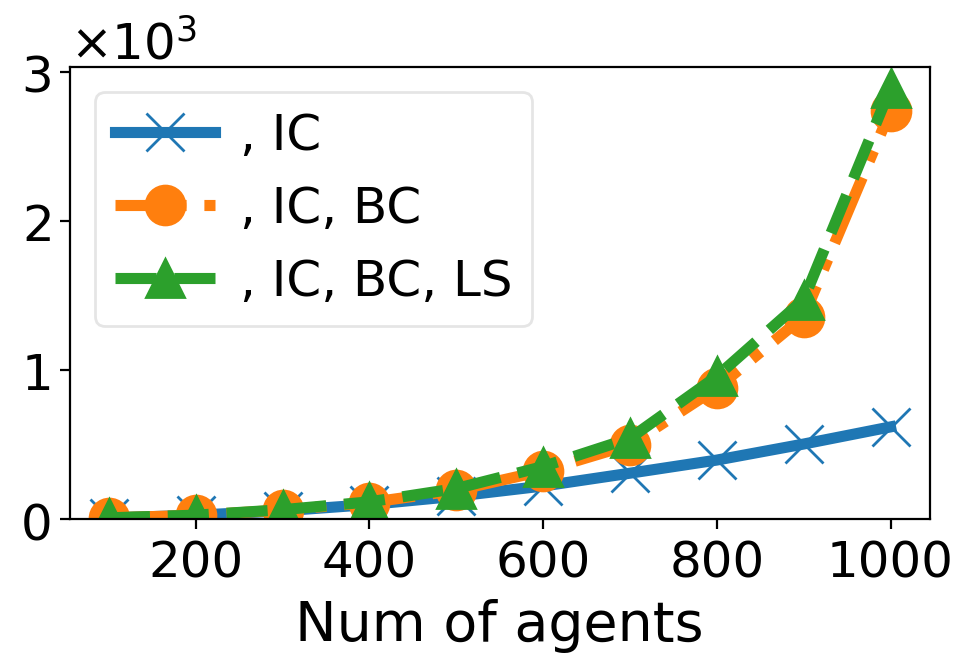}}
\end{minipage}
\hfill
\begin{minipage}{.15\linewidth}
  \centerline{\includegraphics[width=2.1cm]{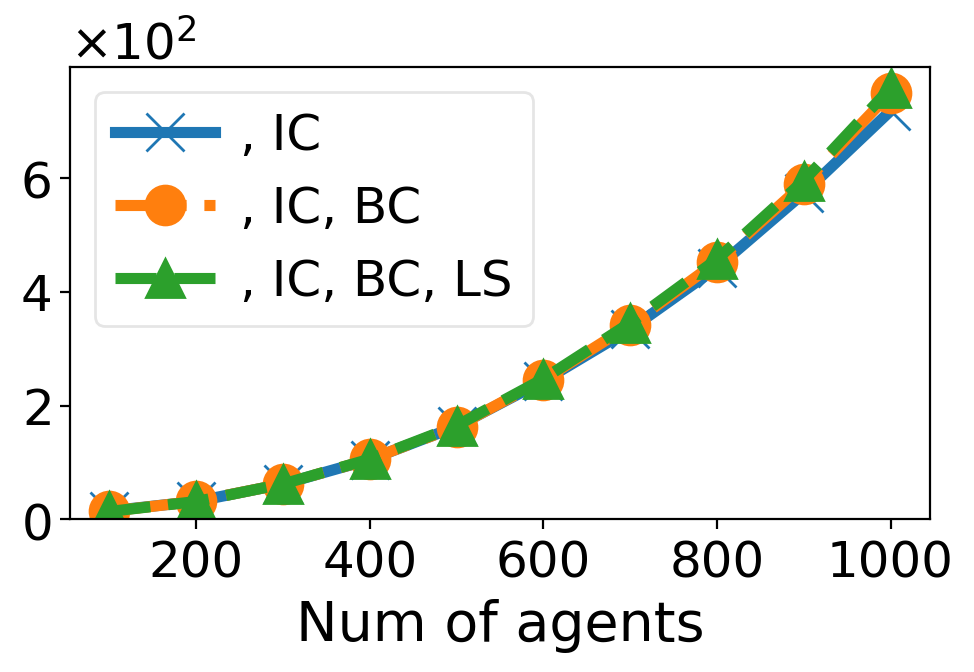}}
\end{minipage}
\vfill

\begin{minipage}{.04\linewidth}
  \rotatebox{90}{\scalebox{0.8}{memory usage (MB)}}
\end{minipage}
\hfill
\begin{minipage}{.15\linewidth}
  \centerline{\includegraphics[width=2.1cm]{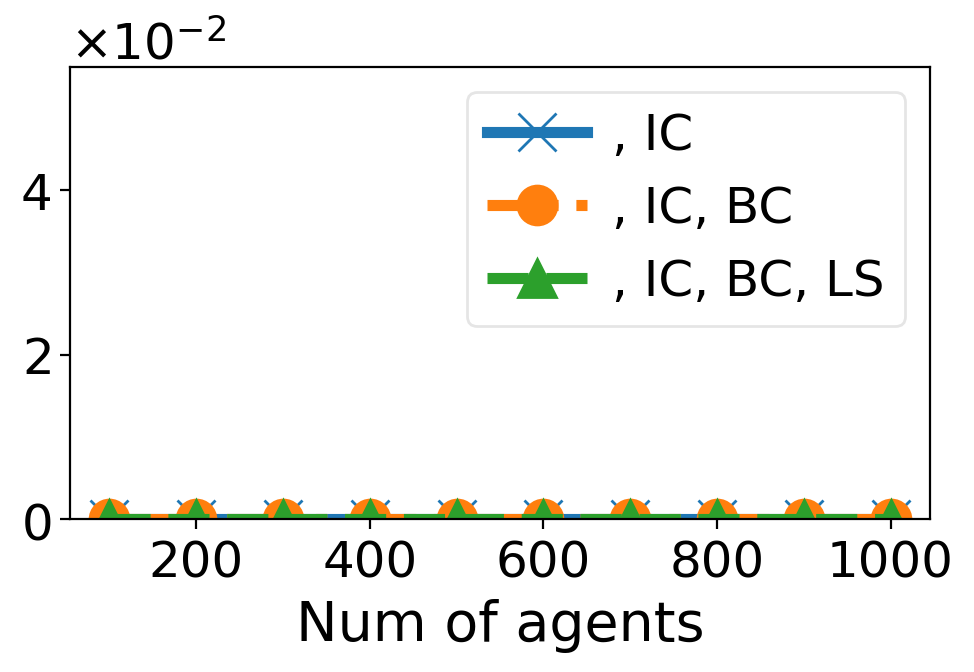}}
\end{minipage}
\hfill
\begin{minipage}{.15\linewidth}
  \centerline{\includegraphics[width=2.1cm]{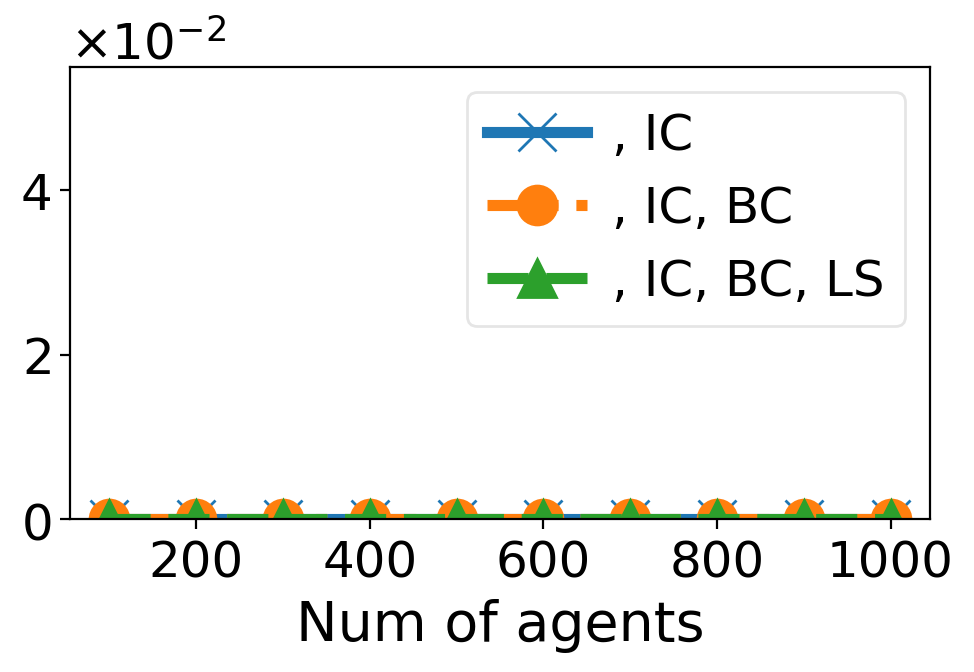}}
\end{minipage}
\hfill
\begin{minipage}{.15\linewidth}
  \centerline{\includegraphics[width=2.1cm]{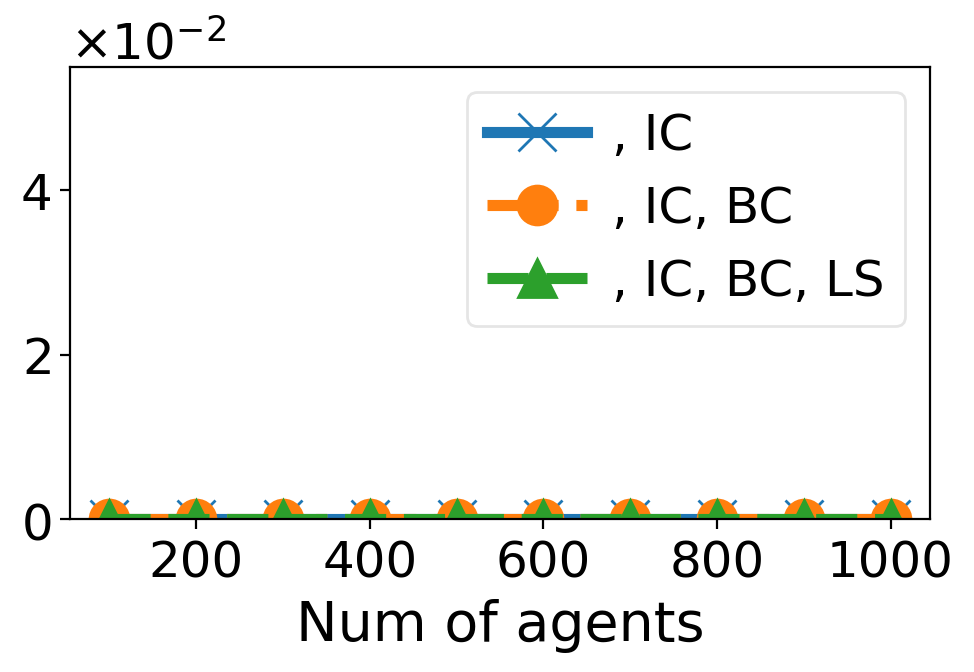}}
\end{minipage}
\hfill
\begin{minipage}{.15\linewidth}
  \centerline{\includegraphics[width=2.1cm]{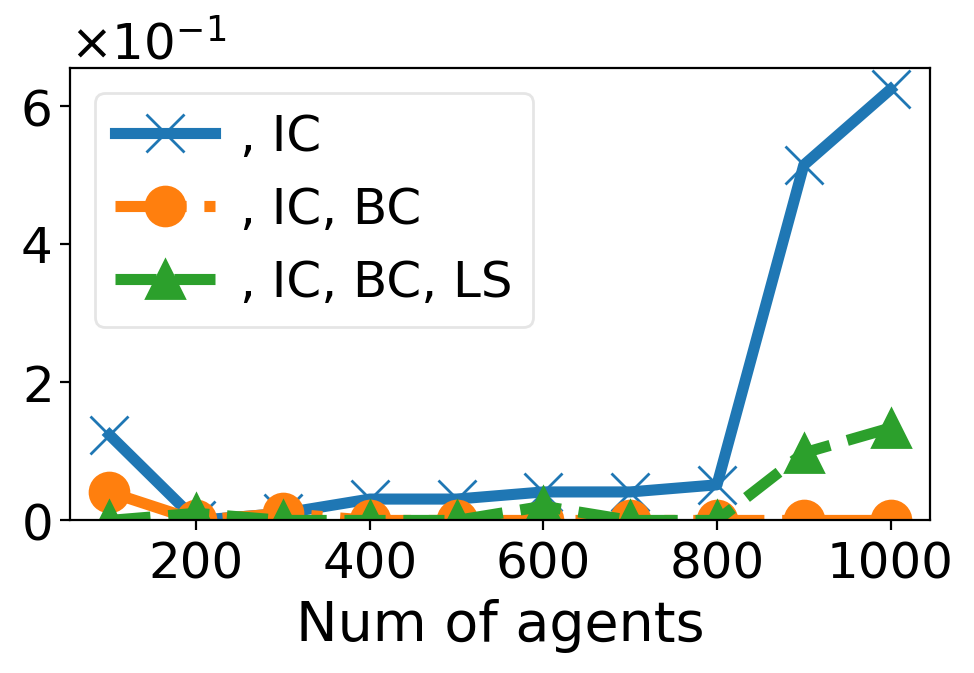}}
\end{minipage}
\hfill
\begin{minipage}{.15\linewidth}
  \centerline{\includegraphics[width=2.1cm]{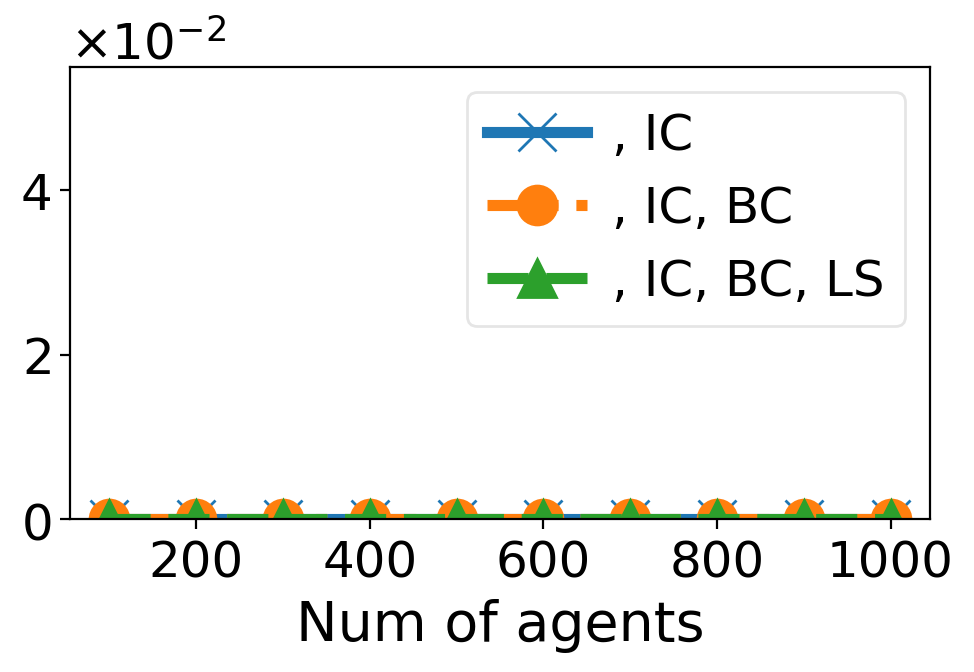}}
\end{minipage}
\hfill
\begin{minipage}{.15\linewidth}
  \centerline{\includegraphics[width=2.1cm]{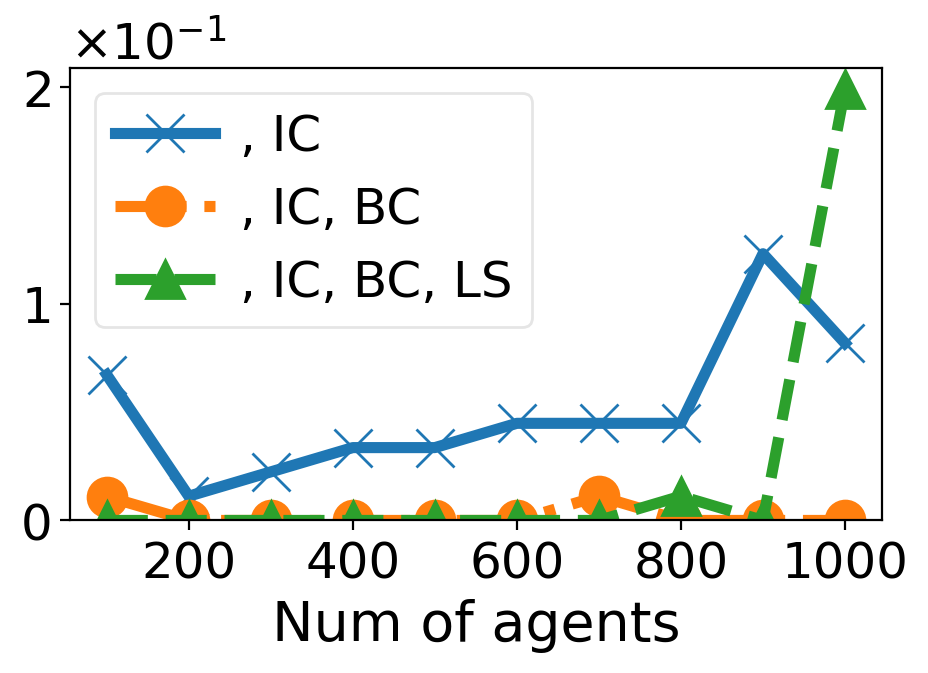}}
\end{minipage}
\vfill

\caption{These figures illustrate the performance of our decomposition method across various maps, as number of agents increases, focusing on decomposition rate, number of subproblems, time cost, and memory usage. Additionally, we provide information on the scale and number of passable cells (in brackets). The map index is also used in the subsequent figures. }
\label{decomposition2}
\end{figure}

The results of our experiments are presented in Fig. \ref{decomposition1} and Fig. \ref{decomposition2}.

\subsection{Decomposition rate}

The decomposition rate is closely correlated with the time cost and memory usage in solving a MAPF instance, as both time cost and memory usage are primarily determined by the size of the largest subproblem.

\subsubsection{Performance under different maps}

As depicted in Fig. \ref{decomposition1}, the final decomposition rate (i.e., the decomposition rate after step 3) demonstrates a consistent increase across almost all maps, with the exception of maps containing a high number of passable cells (e.g., den520, Berlin\_1\_256). In these maps, the number of passable cells closely aligns with the number of agents, resulting in fewer opportunities to alter dependence paths and decompose the instance. Consequently, the size of the largest subproblem approaches that of the original MAPF instance. This limitation becomes increasingly significant as the number of agents grows, ultimately leading to a decomposition rate close to 1, indicating that decomposition becomes ineffective when agents are densely packed.

Conversely, in maps with a surplus of passable cells compared to the number of agents, such as den520 and Berlin\_1\_256, decomposition remains highly effective despite increases in the number of agents. In these scenarios, the abundance of passable cells results in numerous free cell groups, facilitating the generation of dependence paths that no need to traverse multiple agents' start and target locations.

\subsubsection{How each step contribute to decomposition}

As shown in Fig. \ref{decomposition1}, we observe that each step of the decomposition process contributes to a decrease in the decomposition rate (i.e., a reduction in the maximum size of subproblems) overall. However, the extent of their contributions varies across different maps.

In maps characterized by a surplus of passable cells, exceeding the number of agents (such as Berlin\_1\_256, Paris\_1\_256, and Den520d), the maximum size of clusters is reduced to merely 1 after IC, even with hundreds of agents. This is due to the high likelihood of an agent's initial dependence path avoiding passing through other agents' start or target locations. Consequently, subsequent BC and LS have no further room for decomposition, resulting in similar time costs after each step in these maps.

Conversely, in maps where the number of passable cells is close to the number of agents (e.g., empty\_16\_16, empty\_32\_32, random-32-32-20, and maze-32-32-4), BC and LS contribute more to decomposition than IC as the number of agents increases. This is because there is a lower likelihood of an agent's initial dependence path avoiding passing through other agents' start or target locations. Consequently, BC and LS have the opportunity to decompose subproblems by updating dependence paths.

\subsection{Number of subproblems}


The count of subproblems generally follows a pattern of initial increase followed by a decrease as the number of agents increases. Initially, decomposition easily generates small subproblems, leading to an increase in their count as the number of agents grows. However, as agents become denser, decomposing becomes more challenging, resulting in an increase in the decomposition rate and a subsequent decrease in the number of subproblems until it equals 1, indicating the raw MAPF instance.

Special cases arise in maps with numerous passable cells, exceeding the number of agents, such as den520d and Berlin\_1\_256. In these instances, decomposition divides the raw instance into subproblems, each containing only one agent, resulting in the number of subproblems equaling the number of agents in the raw MAPF instance. However, it is predictable that as the number of agents increases, decomposition will become increasingly ineffective, ultimately reducing the number of subproblems to 1.

\subsection{Time cost and memory usage}



Time cost and memory usage are crucial factors in the application of decomposition for MAPF. While theoretically, decomposing a MAPF instance should reduce the time cost and memory usage of solving the problem, if the decomposition process itself consumes too much time or memory space, it may not effectively reduce the total cost of solving the MAPF instance. So we analysis how many resources decomposition cost under various of maps.

Generally, as depicted in the Fig. \ref{decomposition1} and Fig. \ref{decomposition2}, the time cost of decomposition is less than 1 second in most maps and less than 3 seconds in the worst-case scenario (such as 800 to 1000 dense agents), which is deemed acceptable for practical application. In the majority of cases, all three steps contribute to the time cost. However, there are differences in the time cost of steps across different maps. In maps with an abundance of passable cells exceeding the number of agents, such as den512d, Berlin\_1\_256, and Paris\_1\_256, IC consumes almost all of the time cost, resulting in overlapping time costs after IC, BC, and LS. Conversely, in maps with few passable cells close to the number of agents, all steps contribute to the time cost, as these scenarios require BC and LS to decompose initial subproblems into smaller subproblems.

Regarding memory usage, decomposition for almost all maps consumes less than 1 MB, which is considered acceptable for common platforms such as laptops or industrial process computers.

\subsection{Comparison with Independence Detection}
\begin{figure}[h] \tiny


\begin{minipage}{.24\linewidth}
  \centerline{\includegraphics[width=3.2cm]{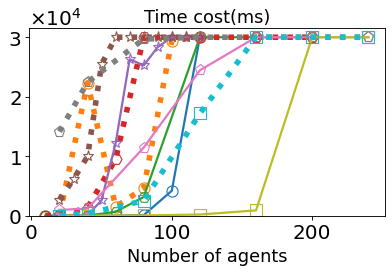}}
\end{minipage}
\hfill
\begin{minipage}{.24\linewidth}
  \centerline{\includegraphics[width=3.2cm]{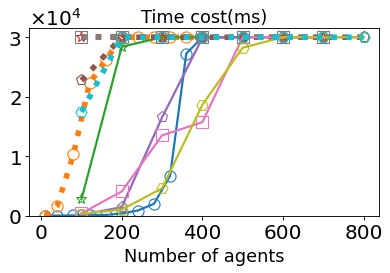}}
\end{minipage}
\hfill
\begin{minipage}{.24\linewidth}
  \centerline{\includegraphics[width=3.2cm]{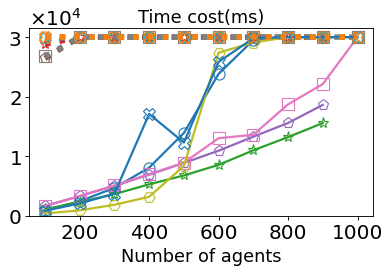}}
\end{minipage}
\hfill
\begin{minipage}{.24\linewidth}
  \centerline{\includegraphics[width=3.2cm]{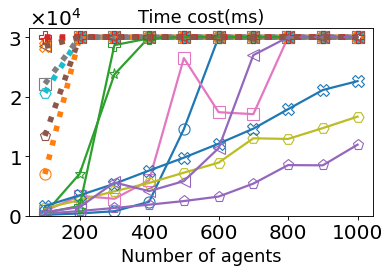}}
\end{minipage}
\vfill

\begin{minipage}{.24\linewidth}
  \centerline{\includegraphics[width=3.2cm]{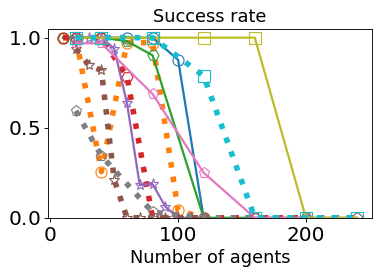}}
\end{minipage}
\hfill
\begin{minipage}{.24\linewidth}
  \centerline{\includegraphics[width=3.2cm]{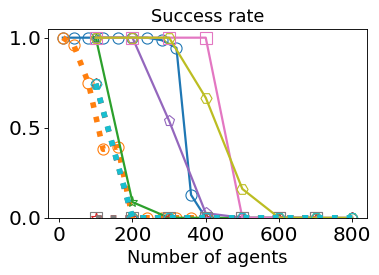}}
\end{minipage}
\hfill
\begin{minipage}{.24\linewidth}
  \centerline{\includegraphics[width=3.2cm]{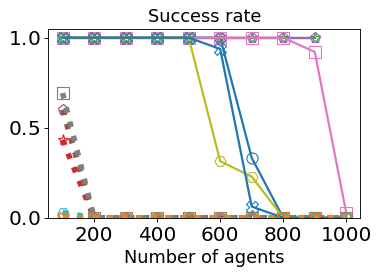}}
\end{minipage}
\hfill
\begin{minipage}{.24\linewidth}
  \centerline{\includegraphics[width=3.2cm]{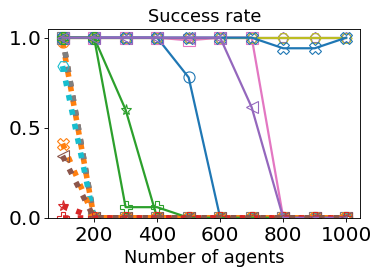}}
\end{minipage}
\vfill

\begin{minipage}{.24\linewidth}
  \centerline{\includegraphics[width=3.2cm]{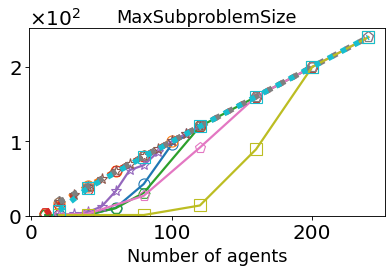}}
\end{minipage}
\hfill
\begin{minipage}{.24\linewidth}
  \centerline{\includegraphics[width=3.2cm]{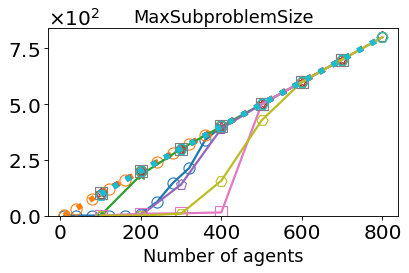}}
\end{minipage}
\hfill
\begin{minipage}{.24\linewidth}
  \centerline{\includegraphics[width=3.2cm]{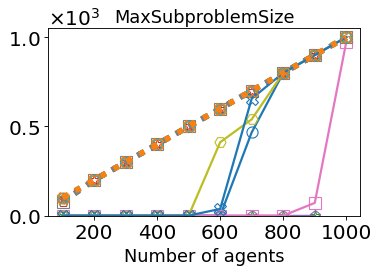}}
\end{minipage}
\hfill
\begin{minipage}{.24\linewidth}
  \centerline{\includegraphics[width=3.2cm]{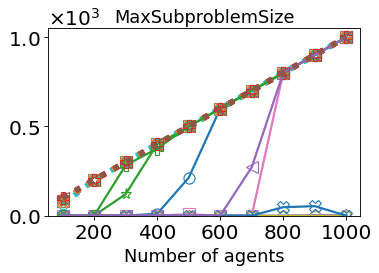}}
\end{minipage}
\vfill

\begin{minipage}{.24\linewidth}
  \centerline{\includegraphics[width=3.2cm]{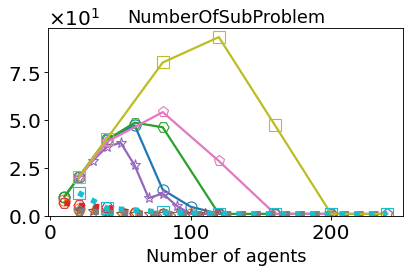}}
\end{minipage}
\hfill
\begin{minipage}{.24\linewidth}
  \centerline{\includegraphics[width=3.2cm]{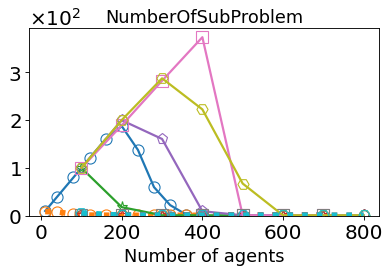}}
\end{minipage}
\hfill
\begin{minipage}{.24\linewidth}
  \centerline{\includegraphics[width=3.2cm]{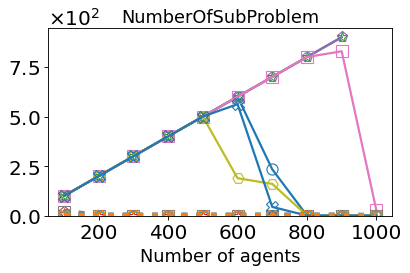}}
\end{minipage}
\hfill
\begin{minipage}{.24\linewidth}
  \centerline{\includegraphics[width=3.2cm]{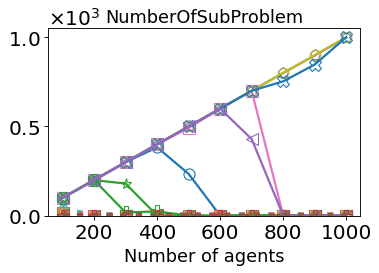}}
\end{minipage}
\vfill

\begin{minipage}{.24\linewidth}
  \centerline{\includegraphics[width=3.2cm]{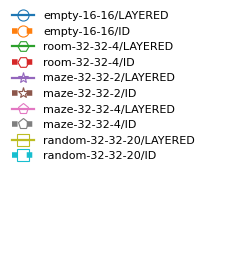}}
\end{minipage}
\hfill
\begin{minipage}{.24\linewidth}
  \centerline{\includegraphics[width=3.2cm]{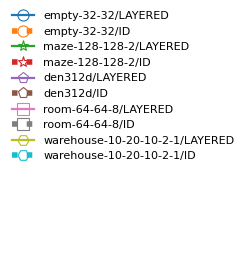}}
\end{minipage}
\hfill
\begin{minipage}{.24\linewidth}
  \centerline{\includegraphics[width=3.2cm]{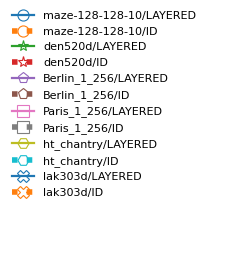}}
\end{minipage}
\hfill
\begin{minipage}{.24\linewidth}
  \centerline{\includegraphics[width=3.2cm]{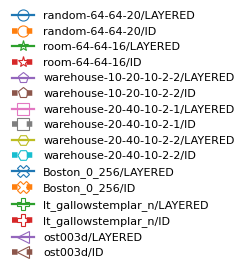}}
\end{minipage}
\vfill

\caption{These figures show how Layered CBS and ID using CBS perform across various maps. Their performance is evaluated in terms of time cost, success rate, maximum subproblem size, and total number of subproblems. The data for Layered CBS is shown as solid lines, while the data for ID (using CBS) is shown as dotted lines. More details about the maps can be found in Fig. \ref{decomposition1} and Fig. \ref{decomposition2}. The last row contains the legend for the upper figures; figures in the same column share the same legend, and the same applies hereinafter.
} 
\label{compare_with_id_cbs}
\end{figure}

\begin{figure}[h] \tiny


\begin{minipage}{.24\linewidth}
  \centerline{\includegraphics[width=3.2cm]{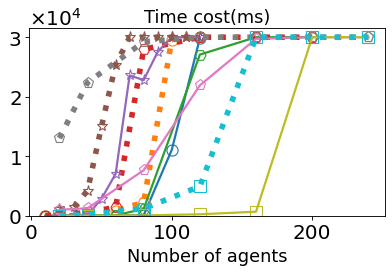}}
\end{minipage}
\hfill
\begin{minipage}{.24\linewidth}
  \centerline{\includegraphics[width=3.2cm]{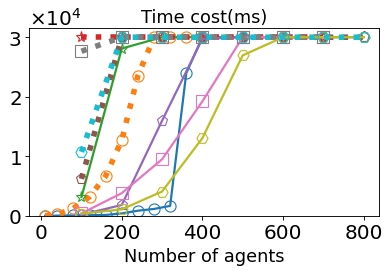}}
\end{minipage}
\hfill
\begin{minipage}{.24\linewidth}
  \centerline{\includegraphics[width=3.2cm]{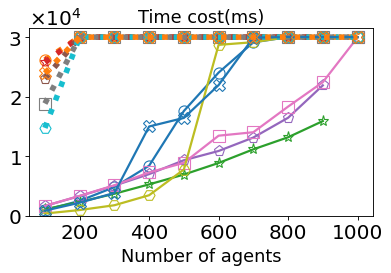}}
\end{minipage}
\hfill
\begin{minipage}{.24\linewidth}
  \centerline{\includegraphics[width=3.2cm]{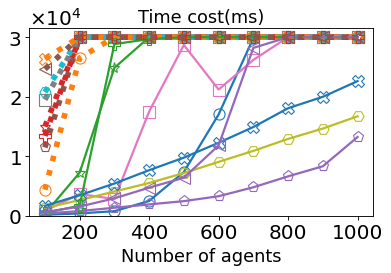}}
\end{minipage}
\vfill

\begin{minipage}{.24\linewidth}
  \centerline{\includegraphics[width=3.2cm]{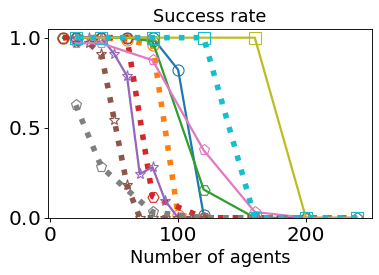}}
\end{minipage}
\hfill
\begin{minipage}{.24\linewidth}
  \centerline{\includegraphics[width=3.2cm]{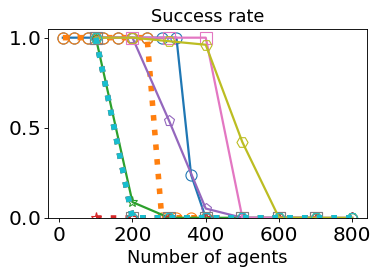}}
\end{minipage}
\hfill
\begin{minipage}{.24\linewidth}
  \centerline{\includegraphics[width=3.2cm]{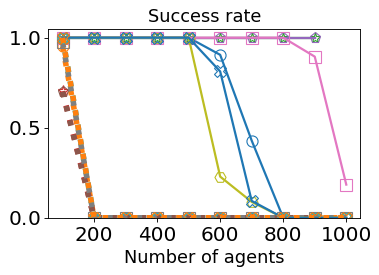}}
\end{minipage}
\hfill
\begin{minipage}{.24\linewidth}
  \centerline{\includegraphics[width=3.2cm]{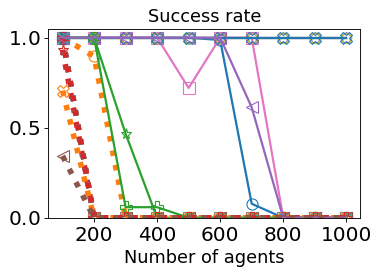}}
\end{minipage}
\vfill

\begin{minipage}{.24\linewidth}
  \centerline{\includegraphics[width=3.2cm]{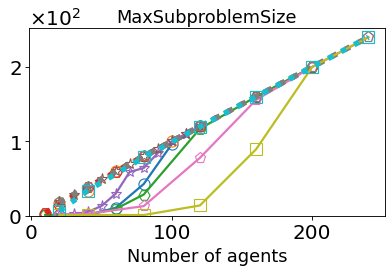}}
\end{minipage}
\hfill
\begin{minipage}{.24\linewidth}
  \centerline{\includegraphics[width=3.2cm]{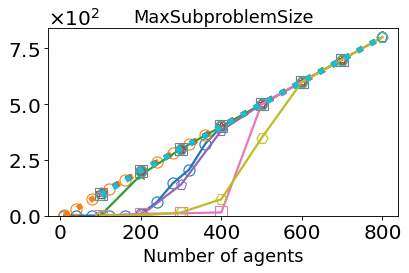}}
\end{minipage}
\hfill
\begin{minipage}{.24\linewidth}
  \centerline{\includegraphics[width=3.2cm]{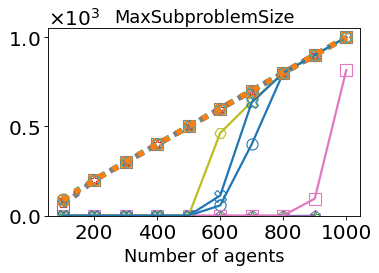}}
\end{minipage}
\hfill
\begin{minipage}{.24\linewidth}
  \centerline{\includegraphics[width=3.2cm]{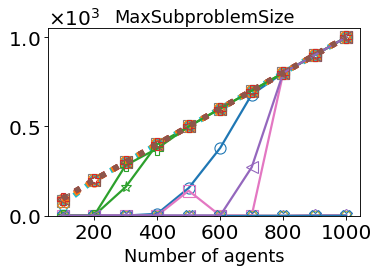}}
\end{minipage}
\vfill

\begin{minipage}{.24\linewidth}
  \centerline{\includegraphics[width=3.2cm]{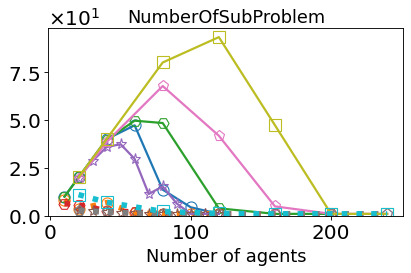}}
\end{minipage}
\hfill
\begin{minipage}{.24\linewidth}
  \centerline{\includegraphics[width=3.2cm]{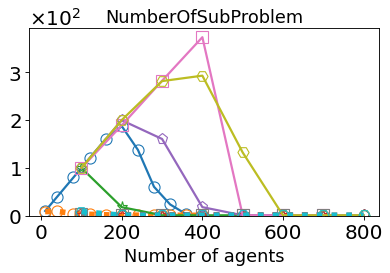}}
\end{minipage}
\hfill
\begin{minipage}{.24\linewidth}
  \centerline{\includegraphics[width=3.2cm]{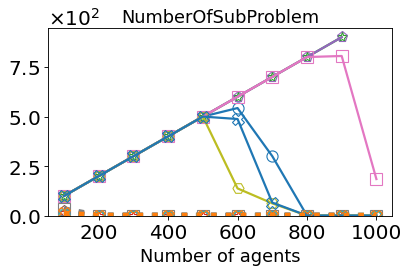}}
\end{minipage}
\hfill
\begin{minipage}{.24\linewidth}
  \centerline{\includegraphics[width=3.2cm]{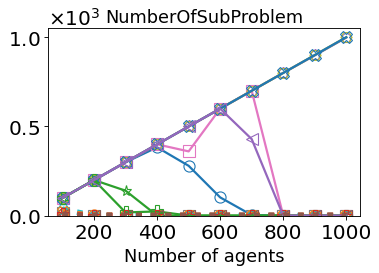}}
\end{minipage}
\vfill

\begin{minipage}{.24\linewidth}
  \centerline{\includegraphics[width=3.2cm]{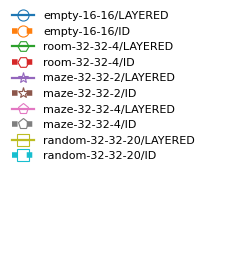}}
\end{minipage}
\hfill
\begin{minipage}{.24\linewidth}
  \centerline{\includegraphics[width=3.2cm]{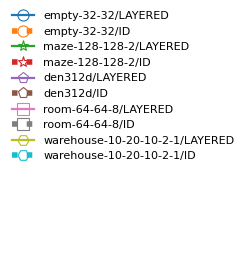}}
\end{minipage}
\hfill
\begin{minipage}{.24\linewidth}
  \centerline{\includegraphics[width=3.2cm]{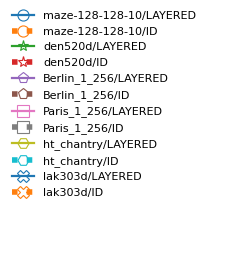}}
\end{minipage}
\hfill
\begin{minipage}{.24\linewidth}
  \centerline{\includegraphics[width=3.2cm]{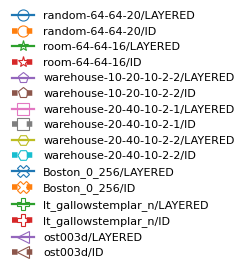}}
\end{minipage}
\vfill

\caption{These figures show how Layered EECBS and ID using EECBS perform across various maps. Their performance is evaluated in terms of time cost, success rate, maximum subproblem size, and total number of subproblems. The data for Layered EECBS is shown as solid lines, while the data for ID (using EECBS) is shown as dotted lines. More details about the maps can be found in Fig. \ref{decomposition1} and Fig. \ref{decomposition2}.
} 
\label{compare_with_id_eecbs}
\end{figure}

Similar to Layered MAPF, Independence Detection (ID) \cite{Standley2011CompleteAF} is a technique used to decompose a group of agents into the smallest possible groups. Unlike Layered MAPF, ID assigns each agent to its own unique group and then merges two groups if one group cannot find a solution (using MAPF methods) that avoids conflict with the solution of another group. ID is limited to methods that can take external paths as constraints to avoid conflicts, so not every MAPF method is applicable to ID. In this section, we compare Layered MAPF and ID using CBS and EECBS. The related results are shown in Fig. \ref{compare_with_id_cbs} and Fig. \ref{compare_with_id_eecbs}. For the implementation, we use the code from \cite{surynek2018variants}\footnote{https://github.com/svancaj/HybridMAPF} for ID.

In this section, we compare Layered MAPF with ID in terms of maximum subproblem size, number of subproblems, time cost, and success rate in finding a solution. Since MAPF methods are embedded in ID, we compare the total cost of Layered MAPF (including the time cost of decomposition and solving subproblems) with the time cost of ID. To maintain consistency, we use the same MAPF instances for the decomposition of each instance.

As a common rule in solving MAPF instances, we set an upper bound on the time cost (30 seconds). Methods that are complete but fail to find a solution within the allotted time are considered to have failed. When ID fails to find collision-free solutions for an instance, we use the number of agents in the instance as its maximum subproblem size. To maintain consistency, this rule is also applied to Layered MAPF, even though our method decomposes the instance into multiple subproblems.

\begin{figure}[h] \tiny

\begin{minipage}{.24\linewidth}
  \centerline{\includegraphics[width=3.3cm]{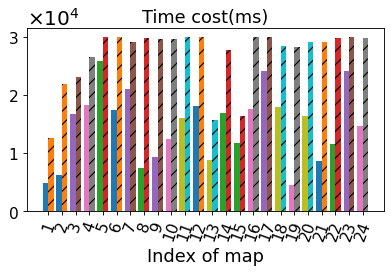}}
\end{minipage}
\hfill
\begin{minipage}{.24\linewidth}
  \centerline{\includegraphics[width=3.2cm]{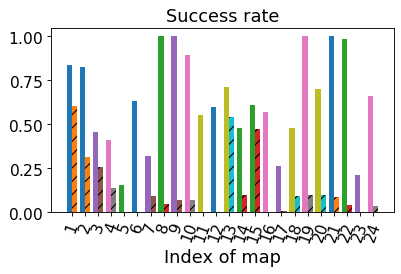}}
\end{minipage}
\hfill
\begin{minipage}{.24\linewidth}
  \centerline{\includegraphics[width=3.2cm]{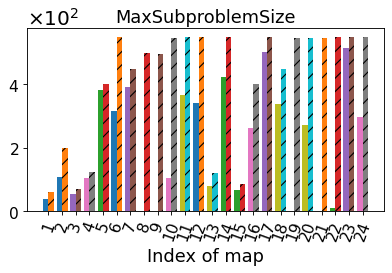}}
\end{minipage}
\hfill
\begin{minipage}{.24\linewidth}
  \centerline{\includegraphics[width=3.2cm]{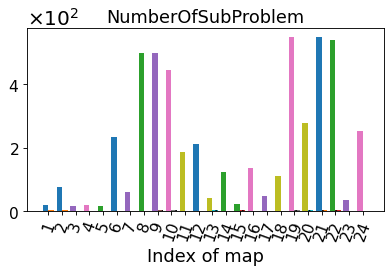}}
\end{minipage}
\vfill

\caption{These figures show the average time cost, success rate, maximum subproblem size, and number of subproblems for Layered CBS and ID (using CBS) across various maps. These figures summarize the data from Fig. \ref{compare_with_id_cbs}. The map indices correspond to those in Fig. \ref{decomposition1} and Fig. \ref{decomposition2}, and the same applies hereinafter.
} 
\label{compare_with_id_cbs_sum}
\end{figure}

\begin{figure}[h] \tiny

\begin{minipage}{.24\linewidth}
  \centerline{\includegraphics[width=3.3cm]{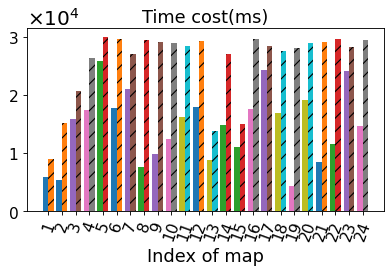}}
\end{minipage}
\hfill
\begin{minipage}{.24\linewidth}
  \centerline{\includegraphics[width=3.2cm]{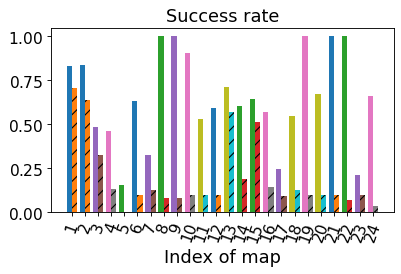}}
\end{minipage}
\hfill
\begin{minipage}{.24\linewidth}
  \centerline{\includegraphics[width=3.2cm]{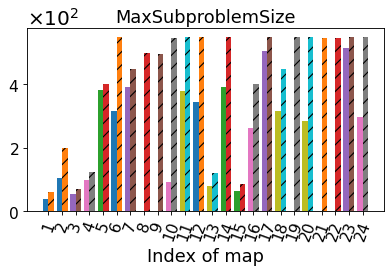}}
\end{minipage}
\hfill
\begin{minipage}{.24\linewidth}
  \centerline{\includegraphics[width=3.2cm]{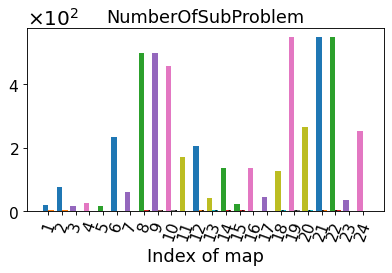}}
\end{minipage}
\vfill

\caption{These figures show the average time cost, success rate, maximum subproblem size, and number of subproblems for Layered EECBS and ID (using EECBS) across various maps (please refer to Fig. \ref{decomposition2} for the maps corresponding to the map indices). These figures summarize the data from Fig. \ref{compare_with_id_eecbs}.
} 
\label{compare_with_id_eecbs_sum}
\end{figure}

As shown in Fig. \ref{compare_with_id_cbs} and Fig. \ref{compare_with_id_eecbs}, for both CBS and EECBS, Layered MAPF's maximum subproblem size increases significantly slower than ID as the number of agents increases. Additionally, Layered MAPF decomposes the MAPF instance into more subproblems compared to ID. As a result, Layered MAPF has a lower time cost (for CBS, 14.5s $<$ 27.4s; for EECBS, 14.4s $<$ 26.0s), a higher success rate (for CBS, 0.64 $>$ 0.12; for EECBS, 0.65 $>$ 0.18), a smaller maximum subproblem size (for CBS, 206.8 $<$ 426.5; for EECBS, 206.9 $<$ 426.9), and more subproblems (for CBS, 218.7 $>$ 1.9; for EECBS, 218.7 $>$ 2.31), on average, compared to ID, as shown in Fig. \ref{compare_with_id_cbs_sum} and Fig. \ref{compare_with_id_eecbs_sum}.

There are several reasons that cause this phenomenon:

1, ID does not consider the order of solving subproblems, while Layered MAPF does. Therefore, Layered MAPF has a higher probability of decomposing a MAPF instance into smaller subproblems;

2, when ID tries to check whether a group (i.e., a subproblem) can avoid conflicts with another group, it does not consider avoiding groups that it has already avoided in the past. As a result, ID can sometimes fall into infinite loops. For example, suppose group A avoids group B. Then, ID detects conflicts between group A and group C and finds a new solution for group A that avoids conflicts with group C. However, this new solution for group A conflicts with group B. ID will then repeat the process of making group A avoid group B and group C until it runs out of time. This phenomenon is less frequent when there are a few sparse agents but becomes more common when there are dense agents; 

3, in Layered MAPF, each agent participates in a multi-agent pathfinding process only once. In contrast, in ID, an agent may participate multiple times, as it can be merged into different groups at different stages. This causes ID to have a higher time cost than Layered MAPF.  

\subsection{Summary}



\begin{figure}[h] 
\centerline{\includegraphics[width=7.0cm]{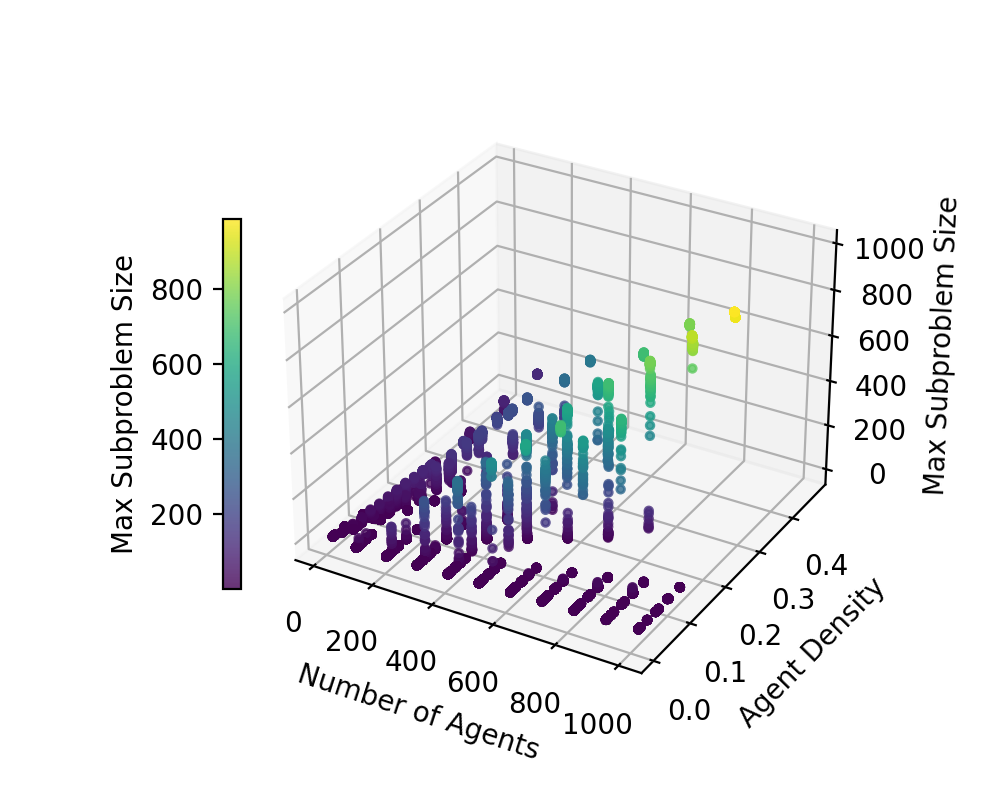}}
\caption{
This figure illustrates how the maximum subproblem size changes as the density of agents and the number of agents in the instance change. The data source is the same as in Fig. \ref{decomposition1} and Fig. \ref{decomposition2}.
}
\label{decom_summary}
\end{figure} 

Our method demonstrates high effectiveness for MAPF instances with cells exceeding two times the number of agents, while its effectiveness diminishes as the number of passable cells approaches two times the number of agents. This is because a higher number of passable cells increases the likelihood of decomposing the instance into smaller subproblems. Conversely, for the same map, the effectiveness of decomposition decreases with an increase in the number of agents in the MAPF instance.

Regarding subproblems, in most maps, the number of subproblems initially increases and then decreases as the number of agents increases. Initially, with only a few agents, decomposition is effective, but as the number of agents increases, decomposition becomes less effective.

In terms of costs, the average time cost is less than 1 second in average, with a maximum of less than 3 seconds in the worst cases (such as 800 to 1000 dense agents), and memory usage remains below 1 MB. Comparatively, maps with more passable cells than the number of agents require fewer computations and less memory space.

We also studied how the decomposition of an instance changes as the number of agents and the agent density change (defined as the ratio between the number of agents and the number of passable cells), as shown in Fig. \ref{decom_summary}. We can observe that the maximum subproblem size is large only when both the agent density and the number of agents are high. In other words, instance decomposition can be effective even with a large number of agents, as long as the agent density is low.

\section{Results of decomposition's application}
\label{application}

\begin{figure}[h] \tiny

\begin{minipage}{.02\linewidth}
\centerline{ }
\end{minipage}
\hfill
\begin{minipage}{.185\linewidth}
  \centerline{\scalebox{0.8}{Time cost (ms)}}
\end{minipage}
\hfill
\begin{minipage}{.185\linewidth}
  \centerline{\scalebox{0.8}{Memory Usage (MB)}}
\end{minipage}
\hfill
\begin{minipage}{.185\linewidth}
  \centerline{\scalebox{0.8}{Success rate}}
\end{minipage}
\hfill
\begin{minipage}{.185\linewidth}
  \centerline{\scalebox{0.8}{Sum of cost}}
\end{minipage}
\hfill
\begin{minipage}{.185\linewidth}
  \centerline{\scalebox{0.8}{Makespan}}
\end{minipage}
\vfill

\begin{minipage}{.02\linewidth}
  \rotatebox{90}{\scalebox{0.8}{EECBS}}
\end{minipage}
\hfill
\begin{minipage}{.185\linewidth}
  \centerline{\includegraphics[width=2.5cm, cframe=red 0.2mm]{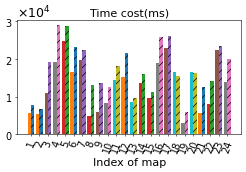}}
\end{minipage}
\hfill
\begin{minipage}{.185\linewidth}
  \centerline{\includegraphics[width=2.5cm, cframe=red 0.2mm]{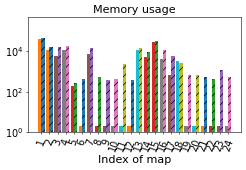}}
\end{minipage}
\hfill
\begin{minipage}{.185\linewidth}
  \centerline{\includegraphics[width=2.5cm, cframe=red 0.2mm]{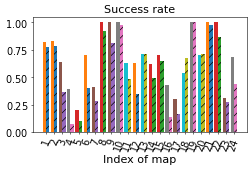}}
\end{minipage}
\hfill
\begin{minipage}{.185\linewidth}
  \centerline{\includegraphics[width=2.5cm, cframe=red 0.2mm]{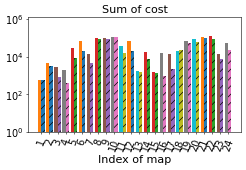}}
\end{minipage}
\hfill
\begin{minipage}{.185\linewidth}
  \centerline{\includegraphics[width=2.5cm, cframe=red 0.2mm]{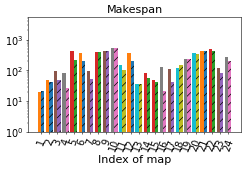}}
\end{minipage}
\vfill

\begin{minipage}{.02\linewidth}
  \rotatebox{90}{\scalebox{0.8}{PBS}}
\end{minipage}
\hfill
\begin{minipage}{.185\linewidth}
  \centerline{\includegraphics[width=2.5cm, cframe=red 0.2mm]{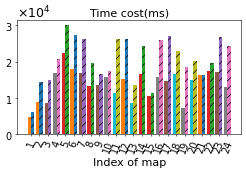}}
\end{minipage}
\hfill
\begin{minipage}{.185\linewidth}
  \centerline{\includegraphics[width=2.5cm, cframe=red 0.2mm]{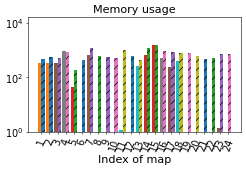}}
\end{minipage}
\hfill
\begin{minipage}{.185\linewidth}
  \centerline{\includegraphics[width=2.5cm, cframe=red 0.2mm]{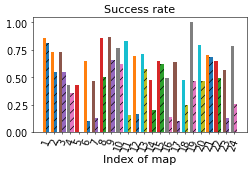}}
\end{minipage}
\hfill
\begin{minipage}{.185\linewidth}
  \centerline{\includegraphics[width=2.5cm, cframe=red 0.2mm]{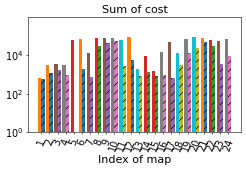}}
\end{minipage}
\hfill
\begin{minipage}{.185\linewidth}
  \centerline{\includegraphics[width=2.5cm, cframe=red 0.2mm]{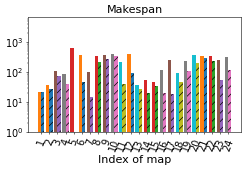}}
\end{minipage}
\vfill

\begin{minipage}{.02\linewidth}
  \rotatebox{90}{\scalebox{0.8}{LNS2}}
\end{minipage}
\hfill
\begin{minipage}{.185\linewidth}
  \centerline{\includegraphics[width=2.5cm, cframe=red 0.2mm]{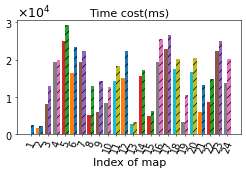}}
\end{minipage}
\hfill
\begin{minipage}{.185\linewidth}
  \centerline{\includegraphics[width=2.5cm, cframe=red 0.2mm]{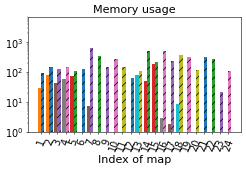}}
\end{minipage}
\hfill
\begin{minipage}{.185\linewidth}
  \centerline{\includegraphics[width=2.5cm, cframe=red 0.2mm]{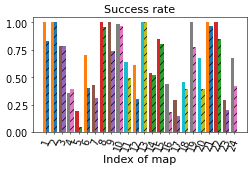}}
\end{minipage}
\hfill
\begin{minipage}{.18\linewidth}
  \centerline{\includegraphics[width=2.5cm, cframe=red 0.2mm]{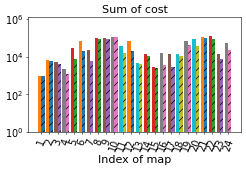}}
\end{minipage}
\hfill
\begin{minipage}{.185\linewidth}
  \centerline{\includegraphics[width=2.5cm, cframe=red 0.2mm]{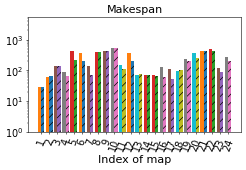}}
\end{minipage}
\vfill

\begin{minipage}{.02\linewidth}
  \rotatebox{90}{\scalebox{0.8}{PushAndSwap}}
\end{minipage}
\hfill
\begin{minipage}{.185\linewidth}
  \centerline{\includegraphics[width=2.5cm, cframe=red 0.2mm]{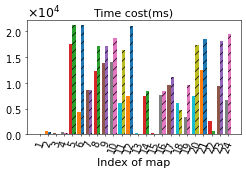}}
\end{minipage}
\hfill
\begin{minipage}{.185\linewidth}
  \centerline{\includegraphics[width=2.5cm, cframe=red 0.2mm]{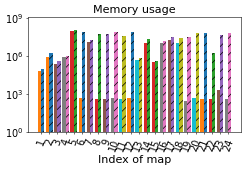}}
\end{minipage}
\hfill
\begin{minipage}{.185\linewidth}
  \centerline{\includegraphics[width=2.5cm, cframe=red 0.2mm]{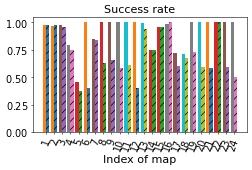}}
\end{minipage}
\hfill
\begin{minipage}{.185\linewidth}
  \centerline{\includegraphics[width=2.5cm, cframe=red 0.2mm]{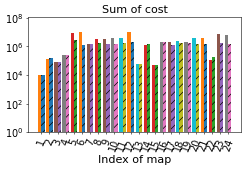}}
\end{minipage}
\hfill
\begin{minipage}{.185\linewidth}
  \centerline{\includegraphics[width=2.5cm, cframe=red 0.2mm]{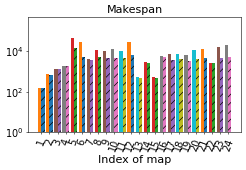}}
\end{minipage}
\vfill

\begin{minipage}{.02\linewidth}
  \rotatebox{90}{\scalebox{0.8}{HCA$\ast$}}
\end{minipage}
\hfill
\begin{minipage}{.185\linewidth}
  \centerline{\includegraphics[width=2.5cm, cframe=red 0.2mm]{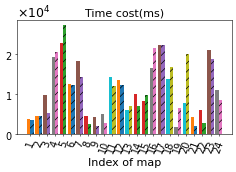}}
\end{minipage}
\hfill
\begin{minipage}{.185\linewidth}
  \centerline{\includegraphics[width=2.5cm, cframe=red 0.2mm]{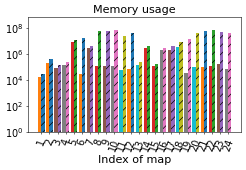}}
\end{minipage}
\hfill
\begin{minipage}{.185\linewidth}
  \centerline{\includegraphics[width=2.5cm, cframe=red 0.2mm]{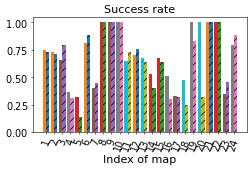}}
\end{minipage}
\hfill
\begin{minipage}{.185\linewidth}
  \centerline{\includegraphics[width=2.5cm, cframe=red 0.2mm]{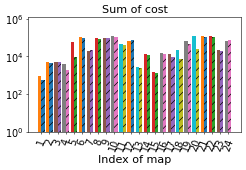}}
\end{minipage}
\hfill
\begin{minipage}{.185\linewidth}
  \centerline{\includegraphics[width=2.5cm, cframe=red 0.2mm]{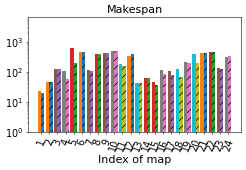}}
\end{minipage}
\vfill

\begin{minipage}{.02\linewidth}
  \rotatebox{90}{\scalebox{0.8}{PIBT+}}
\end{minipage}
\hfill
\begin{minipage}{.185\linewidth}
  \centerline{\includegraphics[width=2.5cm, cframe=red 0.2mm]{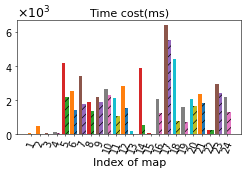}}
\end{minipage}
\hfill
\begin{minipage}{.185\linewidth}
  \centerline{\includegraphics[width=2.5cm, cframe=red 0.2mm]{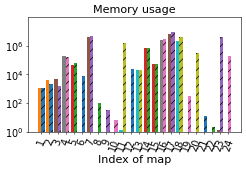}}
\end{minipage}
\hfill
\begin{minipage}{.185\linewidth}
  \centerline{\includegraphics[width=2.5cm, cframe=red 0.2mm]{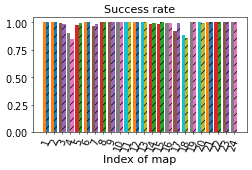}}
\end{minipage}
\hfill
\begin{minipage}{.185\linewidth}
  \centerline{\includegraphics[width=2.5cm, cframe=red 0.2mm]{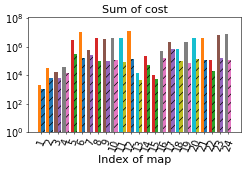}}
\end{minipage}
\hfill
\begin{minipage}{.185\linewidth}
  \centerline{\includegraphics[width=2.5cm, cframe=red 0.2mm]{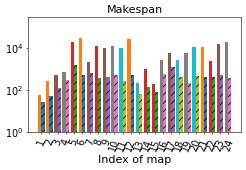}}
\end{minipage}
\vfill

\begin{minipage}{.02\linewidth}
  \rotatebox{90}{\scalebox{0.8}{LaCAM}}
\end{minipage}
\hfill
\begin{minipage}{.185\linewidth}
  \centerline{\includegraphics[width=2.5cm, cframe=red 0.2mm]{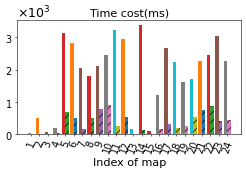}}
\end{minipage}
\hfill
\begin{minipage}{.185\linewidth}
  \centerline{\includegraphics[width=2.5cm, cframe=red 0.2mm]{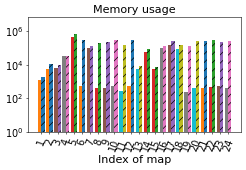}}
\end{minipage}
\hfill
\begin{minipage}{.185\linewidth}
  \centerline{\includegraphics[width=2.5cm, cframe=red 0.2mm]{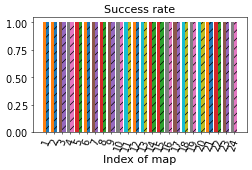}}
\end{minipage}
\hfill
\begin{minipage}{.185\linewidth}
  \centerline{\includegraphics[width=2.5cm, cframe=red 0.2mm]{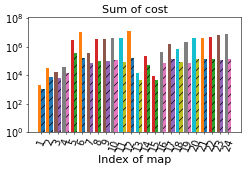}}
\end{minipage}
\hfill
\begin{minipage}{.185\linewidth}
  \centerline{\includegraphics[width=2.5cm, cframe=red 0.2mm]{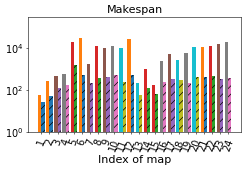}}
\end{minipage}
\vfill

\caption{These figures serve as a detailed version of Fig. \ref{method_summary}, offering insights into the performance of both raw MAPF methods and Layered MAPF methods under specific maps.
}
\label{method_map_summary}
\end{figure}

\begin{figure}[h] \scriptsize

\begin{minipage}{.18\linewidth}
  \centerline{\includegraphics[width=2.6cm]{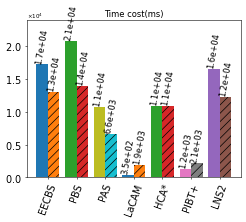}}
\end{minipage}
\hfill
\begin{minipage}{.18\linewidth}
  \centerline{\includegraphics[width=2.6cm]{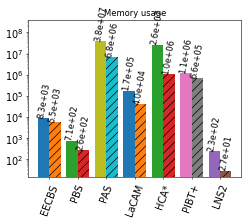}}
\end{minipage}
\hfill
\begin{minipage}{.18\linewidth}
  \centerline{\includegraphics[width=2.6cm]{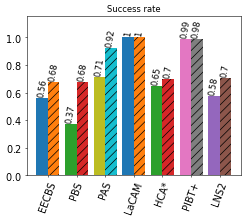}}
\end{minipage}
\hfill
\begin{minipage}{.18\linewidth}
  \centerline{\includegraphics[width=2.6cm]{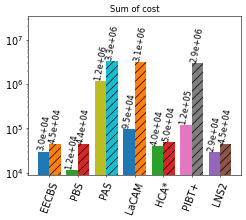}}
\end{minipage}
\hfill
\begin{minipage}{.18\linewidth}
  \centerline{\includegraphics[width=2.6cm]{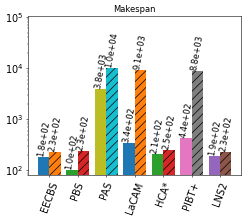}}
\end{minipage}
\vfill

\caption{These figures provide a summary comparison between the raw MAPF method and the Layered MAPF method, combining the results across all maps. In the visualizations, crossed bars represent the Layered version of the MAPF method, while empty bars represent the raw MAPF method.
}
\label{method_summary}
\end{figure}

In this section, we evaluate the influence of decomposing MAPF instances on various MAPF methods, including EECBS, PBS, LNS, HCA$\ast$ (serial MAPF method), and Push and Swap, PIBT$+$, LaCAM2 (parallel MAPF method). We refer to the methods with decomposition of MAPF instances as the ``Layered" versions of the raw MAPF methods. To maintain consistency, we utilize the same MAPF instances as in Section \ref{Resultsofdecomposition} to analyze how decomposition affects their application. We assess decomposition's impact in terms of time cost, memory usage, success rate, sum of cost, and makespan. The default configurations of these MAPF methods are used in our experiments. In particular, due to the memory usage not being immediately released after the code is executed on an Ubuntu system, the statistics of memory usage (in GB, MB, or kB) experience a delay, causing imprecision when comparing Layered MAPF and raw MAPF. Therefore, we use the count of key data elements that play a major role in memory usage as an indicator of memory usage, such as the number of high-level nodes for LaCAM2 or the size of the constraint tree for EECBS.

As a common rule in solving MAPF instances, we set an upper bound on the time cost (30 seconds). Methods that are complete but fail to find a solution within the limited time are considered as failed. Additionally, methods that run out of memory space are also considered as failed.

In this section, experiments were conducted on a computer running Ubuntu 20.04, equipped with an Intel Xeon(R) CPU (2.1 GHz) and 64 GB of memory. All code was implemented in C++.

%




\subsection{Explicit Estimation CBS}

Explicit Estimation CBS (EECBS) \cite{li2021eecbs} stands out as one of the state-of-the-art CBS-based MAPF methods, characterized by its completeness and bounded suboptimality. Further details can be found in Section \ref{Relatedworks}. EECBS's code \footnote{https://github.com/Jiaoyang-Li/EECBS} operates by searching each agent's path separately and provides interfaces to avoid conflicts with external paths, rendering it a serial MAPF method. In its Layered version, which we refer to as ``Layered EECBS", previous subproblems' paths and subsequent subproblems' starts are set as constraints. We use the number of constraint nodes as an estimate of the memory usage of EECBS, as it is the only variable that increases during its execution. 

The average time cost, success rate, memory usage, makespan, and total cost of Layered EECBS and raw EECBS are shown in Fig. \ref{method_summary}. How Layered EECBS and raw EECBS perform under various maps as the number of agents increases is shown in Fig. \ref{compare_with_raw_eecbs} in the Appendix. As shown in Fig. \ref{compare_with_raw_eecbs}, Layered EECBS has lower time cost and thus higher success rates compared to raw EECBS when there are few agents. However, these advantages slowly diminish as the number of agents increases, as does Layered EECBS's advantage in memory usage. Despite this, Layered EECBS still has an advantage in most cases. In terms of path quality, the solutions of Layered EECBS are slightly worse than those of raw EECBS. A summary of the performance of Layered EECBS and raw EECBS can be found in Fig. \ref{method_summary}.

As illustrated in Fig. \ref{method_summary}, on average, Layered EECBS exhibits lower time costs ($1.2\times10^4$ ms $<$ $1.7\times10^4$ ms), leading to a higher success rate (0.68 $>$ 0.57). The time cost of decomposition is relatively small compared to the time cost of solving the MAPF instance. Regarding memory usage, Layered EECBS consumes less memory space compared to raw EECBS (4839.0 constraint nodes $<$ 7542.8 constraint nodes). Decomposition of the MAPF instance significantly reduces the growth of the constraint tree's size by decreasing the number of agents solved simultaneously.

In terms of path quality, Layered EECBS yields a larger makespan (Layered EECBS: $2.3\times10^2$, raw EECBS: $1.9\times10^2$), resulting in a larger sum of costs (Layered EECBS: $4.7\times10^4$, raw EECBS: $3.2\times10^4$). In maps where both raw EECBS and Layered EECBS achieve a success rate close to 1 (e.g., map 1, 8, 10, 19, 21 and 22), both methods exhibit similar makespan and sum of costs. However, in maps where raw EECBS has a lower success rate compared to Layered EECBS (e.g., map 4, 16, 17 and 24), Layered EECBS demonstrates a larger makespan and sum of costs, as depicted in Fig. \ref{method_map_summary}.

This phenomenon can be attributed to two factors: Firstly, solving all agents together provides more opportunities to find shorter solutions than solving them separately. Secondly, raw EECBS struggles to find longer solutions, particularly in maps where its success rate is lower.

In summary, Layered EECBS boasts explicit advantages in terms of time cost, memory usage, and success rate compared to raw EECBS. Additionally, in maps where both methods achieve a high success rate, their solutions tend to be similar. However, Layered EECBS outperforms raw EECBS in finding longer solutions when the latter fails to do so.

\subsection{Priority Based Search}

Priority Based Search (PBS) \cite{ma2019searching} is an incomplete, suboptimal, and priority-based MAPF method. PBS searches each agent's path separately and it's code\footnote{\url{https://github.com/Jiaoyang-Li/PBS}} provides interfaces to avoid conflicts with external paths, making it a serial MAPF method. In its Layered version, which we refer to as ``Layered PBS", we set previous subproblems' paths and subsequent subproblems' starts as constraints. We use the number of priority nodes as an estimate of the memory usage of PBS, as it is the only variable that increases during its execution.

The average time cost, success rate, memory usage, makespan, and total cost of Layered PBS and raw PBS are shown in Fig. \ref{method_summary}. How Layered PBS and raw PBS perform under various maps as the number of agents increases is shown in Fig. \ref{compare_with_raw_pbs} in the Appendix. As shown in Fig. \ref{compare_with_raw_pbs}, Layered PBS has lower time cost and thus higher success rates compared to raw PBS when there are few agents. However, these advantages slowly diminish as the number of agents increases, as does Layered PBS's advantage in memory usage. Despite this, Layered PBS still has an advantage in most cases. In terms of path quality, the solutions of Layered PBS are slightly worse than those of raw PBS. A summary of the performance of Layered PBS and raw PBS can be found in Fig. \ref{method_summary}.

As shown in Fig. \ref{method_summary}, on average, Layered PBS has a lower time cost ($1.3\times10^4$ ms $<$ $2.0\times10^4$ ms) and thus a higher success rate (0.68 $>$ 0.37). The time cost of decomposition is relatively small compared to the time cost of solving the MAPF instance. In terms of memory usage, Layered PBS has lower memory usage compared to raw PBS (234.4 priority nodes $<$ 706.8 priority nodes) because decomposition of the MAPF instance reduces the complexity of the priority tree significantly.

When it comes to path quality, Layered PBS has a larger makespan ($2.3\times10^2$ $>$ $1.0\times10^2$) and thus a larger SOC ($4.5\times10^4$ $>$ $1.3\times10^4$). Similar to EECBS, raw PBS finds it difficult to find longer solutions and can only finds shorter solutions, resulting in lower success rates for maps where raw PBS struggles (e.g., map 4, 6, 16 and 17). For maps where both methods have high success rates (e.g., map 1, 9 and 10), they have similar makespan and SOC, as shown in Fig. \ref{method_map_summary}.

In summary, Layered PBS has explicit advantages in time cost, memory usage, and success rate compared to raw PBS. Additionally, Layered PBS produces solutions similar to raw PBS's solutions on maps where both methods have success rates close to 1. Furthermore, Layered PBS achieves a higher success rate in finding long solutions when raw PBS fails.

\subsection{Large Neighborhood Search 2}

Large Neighborhood Search 2 (LNS2) \cite{li2022mapf} is a suboptimal MAPF method that starts from a set of paths with collisions and repeatedly replans subsets of paths to reduce the overall number of collisions until the paths become collision-free. LNS2 updates each agent's path separately, and its code\footnote{\url{https://github.com/Jiaoyang-Li/MAPF-LNS2}} provides interfaces to avoid conflicts with external paths, making it a serial MAPF method. In LNS2's implementation, LNS2 considers all agents in the initial planning phase, so it generally has the highest memory usage during this phase. LNS2 provides several classic MAPF methods to generate the initial solution, including EECBS, PP, PPS, CBS, PIBT, and winPIBT. To obtain a conflict-free solution as quickly as possible, we choose EECBS to generate the initial solutions. Therefore, we use the number of constraint nodes in EECBS during the initial planning phase as an estimate of the memory usage of LNS2, as this is the variable that contributes most to memory usage during its execution.

The average time cost, success rate, memory usage, makespan, and total cost of Layered LNS2 and raw LNS2 are shown in Fig. \ref{method_summary}. How Layered LNS2 and raw LNS2 perform under various maps as the number of agents increases is shown in Fig. \ref{compare_with_raw_lns} in the Appendix. As shown in Fig. \ref{compare_with_raw_lns}, Layered LNS2 has lower time cost and thus a higher success rate when there are few agents compared to raw LNS2. However, these advantages slowly diminish as the number of agents increases, as does Layered LNS2's advantage in memory usage. Despite this, Layered LNS2 still has an advantage in most cases. In terms of path quality, the solutions from Layered LNS2 are slightly worse than those from raw LNS2. A summary of the performance of Layered LNS2 and raw LNS2 can be found in Fig. \ref{method_summary}.

As shown in Fig. \ref{method_summary}, on average, Layered LNS2 has a lower time cost ($1.2\times10^4$ ms $<$ $1.6\times10^4$ ms) and thus a higher success rate (0.72 $>$ 0.52). In terms of memory usage, Layered LNS2 has lower memory usage compared to raw LNS2 (24.9 constraint nodes $<$ 228.8 constraint nodes). Due to the decomposition of the MAPF instance, other agents' paths are considered as dynamic obstacles and do not join conflict resolve, resulting in Layered LNS2 requiring fewer repair actions to fix conflicts between agents' paths.

Regarding path quality, Layered LNS2 has a larger makespan ($2.4\times10^2$ $>$ $1.9\times10^2$) and SOC ($4.7\times10^4$ $>$ $3.0\times10^4$). Similar to EECBS and PBS, raw LNS2 struggles to find longer solutions and can only discovers shorter solutions, particularly in maps where raw LNS2 has a lower success rate (e.g., map 12 and 16). For maps where both methods have similar success rates (e.g., map 1, 2, 8, 21 and 22), raw LNS2 and Layered LNS2 have similar makespan and SOC, as shown in Fig. \ref{method_map_summary}.

In summary, Layered LNS2 exhibits explicit advantages in time cost, memory usage, and success rate compared to raw LNS2. Additionally, Layered LNS2 produces solutions similar to raw LNS2's solutions in maps where both methods have success rates close to 1. Furthermore, Layered LNS2 achieves a higher success rate in finding long solutions when raw LNS2 cannot.

\subsection{Push and Swap}

Push and Swap (PAS) \cite{luna2011push} is a suboptimal and complete rule-based MAPF method. Push and Swap updates agents' paths simultaneously (via an action called ``MULTIPUSH'', which moves a set of adjacent agents simultaneously), and its code\footnote{\url{https://github.com/Kei18/pibt2}} provides no interfaces to avoid conflicts with external paths and ensures that the same state of agents does not occur twice in the solution, making it a parallel MAPF method. In its Layered version, referred to as ``Layered Push and Swap" or ``Layered PAS," we set previous subproblems' targets and subsequent subproblems' starts as static obstacles. Subsequently, we merge subproblems' solutions by adding wait actions after solving all subproblems. 
In the implementation of Push and Swap, the variable that occupies the most memory is a two-dimensional matrix that stores each agent's location at all times. Therefore, we use the number of elements in the matrix as an estimate of the memory usage of Push and Swap, which is calculated as the makespan of the solution multiplied by the number of agents.

The average time cost, success rate, memory usage, makespan, and sum of cost of Layered Push and Swap (PAS) and raw PAS are shown in Fig. \ref{method_summary}. How Layered PAS and raw PAS perform under various maps as the number of agents increases is shown in Fig. \ref{compare_with_raw_pas} in the Appendix.

As shown in Fig. \ref{compare_with_raw_pas}, in some cases, Layered PAS takes more time than raw PAS, while in other cases, it does not, as PAS's time cost is close to the time cost of instance decomposition. However, when there are many agents, Layered PAS takes less time than raw PAS. Consequently, Layered PAS achieves a higher success rate than raw PAS as the number of agents increases in most cases.

In terms of memory usage, Layered PAS consistently requires less memory than raw PAS, even as the number of agents increases.

Regarding path quality, Layered PAS's solutions are slightly worse than raw PAS's solutions. A summary of Layered PAS and raw PAS's performance can be found in Fig. \ref{method_summary}.

As shown in Fig. \ref{method_summary}, on average, Layered PAS consumes less time ($6.8\times10^3$ ms $<$ $1.15\times10^4$ ms) and memory space ($6.0\times10^6$ constraint nodes $<$ $4.0\times10^7$ constraint nodes) than raw PAS, resulting in a higher success rate (0.92 $>$ 0.69). Due to the number of ``push'' and ``swap'' actions grows exponentially as the number of agents increases, decomposing the MAPF instance enables Layered Push and Swap to solve subproblems separately, requiring fewer ``push'' and ``swap'' actions.

Regarding path quality, as PAS is a parallel MAPF method, the addition of wait actions to merge subproblems' solutions, Layered PAS has a larger makespan ($1.0\times10^4$ $>$ $4.0\times10^3$) and SOC ($3.5\times10^6$ $>$ $1.2\times10^6$) compared to raw PAS.

In summary, Layered Push and Swap consume less time and memory space than raw Push and Swap, resulting in a higher success rate. However, the introduction of wait actions to merge subproblems' solutions leads to a larger makespan and SOC compared to raw Push and Swap.

\subsection{Hierarchical Cooperative A$\ast$}

Hierarchical Cooperative A* (HCA*) \cite{silver2005cooperative} is a suboptimal and incomplete prioritized MAPF algorithm that decouples into a series of single-agent searches. HCA* calculates a distance table for each agent to determine their priority. The distance table stores the distance from every location on the map to the target state. Since HCA* searches one path at a time, and after reviewing its code, we found that there are no other variables consuming significant memory space aside from the distance table, we conclude that HCA*'s memory usage is primarily determined by the size of these distance tables. Specifically, it depends on the number of agents multiplied by the size of the map. Thus, we use the product of the number of agents and the size of the map as an estimate for HCA*'s memory usage.

HCA* searches each agent's path separately, making it a serial MAPF method. But the code\footnote{\url{https://github.com/Kei18/pibt2}} for HCA* provides no interfaces to set external paths as dynamic obstacles to avoid. We updated its code to enable it to set external paths as dynamic obstacles.

The average time cost, success rate, memory usage, makespan, and sum of costs for Layered HCA* and raw HCA* are shown in Fig. \ref{method_summary}. A comparison of how Layered HCA* and raw HCA* perform under various maps as the number of agents increases is shown in Fig. \ref{compare_with_raw_hca} in the Appendix. As illustrated in Fig. \ref{compare_with_raw_hca}, the time cost of Layered HCA* is very close to that of raw HCA*, as HCA* searches each agent's path separately. However, Layered HCA* has lower memory usage compared to raw HCA*, as it does not need to store the heuristic table for all agents.

In terms of success rate, Layered HCA* has a slightly higher success rate, as HCA* is incomplete, and solving the instance separately may increase its chances of finding a solution (e.g., instances are decomposed into subproblems that involve only one agent).

Regarding path quality, the solution of Layered HCA* is slightly worse than that of raw HCA*. A summary of the performance of Layered HCA* and raw HCA* can be found in Fig. \ref{method_summary}.

On average, Layered HCA* has a slightly higher time cost than raw HCA* ($1.06\times10^4$ ms $>$ $1.05\times10^4$ ms) and a similar success rate (0.71 $>$ 0.66) as HCA* solves each agent separately, and decomposition does not improve the efficiency of HCA*. However, Layered HCA* requires less memory space than raw HCA* ($9.1\times10^5$ $<$ $2.6\times10^7$), as Layered HCA* only stores the distance table of the current subproblem's agents, while raw HCA* needs to store the distance table of all agents.

Regarding path quality, Layered HCA* has a slightly larger makespan ($2.5\times10^2$ $>$ $2.2\times10^2$) and SOC ($5.2\times10^4$ $>$ $4.2\times10^4$). Because HCA* solves agents in order of decreasing distance to the target, raw HCA* solves agents with longer paths first, limiting the makespan of these agents' solutions, while Layered HCA* breaks this order as agents with long paths may not be in the same subproblem and may not be solved first.

In summary, Layered HCA* shows little improvement compared to raw HCA* and has slightly worse solutions because HCA* has a fixed priority and solves each agent separately. However, Layered HCA* has an advantage in memory usage, as it only needs to store the distance table of the current subproblem's agents rather than all agents.

\subsection{Priority Inheritance with Backtracking$+$}

Priority Inheritance with Backtracking$+$ (PIBT$+$) \cite{okumura2022priority} is an incomplete and suboptimal MAPF algorithm. PIBT$+$ avoids the occurrence of the same state of agents twice in the solution. And the code\footnote{\url{https://github.com/Kei18/pibt2}} for PIBT$+$ provides no interfaces to avoid conflicts with external paths, making it a parallel MAPF method. In the implementation of PIBT$+$, a function called $funcPIBT$ is used recursively to move all agents one step at a time without any collisions. The function is called repeatedly until all agents reach their targets. In computer systems (e.g., Windows or Linux), function recursion relies on the stack to manage the context of function calls. Each recursive call allocates a new stack frame, and the depth of recursion is limited by the size of the stack. Therefore, the memory usage of PIBT$+$ is primarily determined by the recursion depth of $funcPIBT$. We use the maximum recursion depth as an estimate of the memory usage for PIBT$+$.

The average time cost, success rate, memory usage, makespan, and sum of costs for Layered PIBT$+$ and raw PIBT$+$ are shown in Fig. \ref{method_summary}. The performance of Layered PIBT$+$ and raw PIBT$+$ under various maps as the number of agents increases is presented in Fig. \ref{compare_with_raw_PIBT} in the Appendix. \\ 
As shown in Fig. \ref{compare_with_raw_PIBT}, in some cases, Layered PIBT$+$ takes more time compared to raw PIBT$+$, while in other cases, it does not. This is because the time cost of PIBT$+$ is closely tied to the time cost of instance decomposition. Both Layered PIBT$+$ and raw PIBT$+$ have a high success rate, but neither achieves a 100\% success rate, as PIBT$+$ is an incomplete MAPF method. In terms of memory usage, Layered PIBT$+$ consistently requires less memory than raw PIBT$+$, even as the number of agents increases. \\
Regarding path quality, the solutions found by Layered PIBT$+$ are generally worse than those found by raw PIBT$+$, though in some cases, both Layered PIBT$+$ and raw PIBT$+$ solutions see a decrease in SOC and makespan as the number of agents increases. This is because, with more agents, there are fewer opportunities for PIBT$+$ to find longer paths. A summary of Layered PIBT$+$ and raw PIBT$+$ performance can be found in Fig. \ref{method_summary}.
 
In general, Layered PIBT$+$ takes more time than raw PIBT$+$ ($2.2\times10^3$ ms $>$ $1.9\times10^3$ ms). As mentioned in Section \ref{Resultsofdecomposition}, the decomposition of MAPF instances with 800 to 1000 agents may cost 1s to 3s, and these instances play a major role in contributing to the mean time cost of the MAPF method. Although decomposition reduces the time cost of PIBT$+$, the time cost of decomposition offsets the reduced time cost and thus increases the overall time cost.

Layered PIBT$+$ has a similar success rate compared to raw PIBT$+$ (Layered PIBT$+$: 0.98, raw PIBT$+$: 0.98), as both Layered PIBT$+$ and raw PIBT$+$ typically take less than 30s in most cases. In terms of memory usage, Layered PIBT$+$ requires less memory space than raw PIBT$+$ ($6.5\times10^5$ $<$ $1.1\times10^6$), as decomposition reduces the number of intermediate states by reducing the number of agents solved simultaneously.

Regarding solution quality, as PIBT$+$ is a parallel MAPF method, Layered PIBT$+$ introduces extra wait actions, resulting in solutions with higher makespan ($9.4\times10^3$ $>$ $4.3\times10^2$) and SOC ($3.2\times10^6$ $>$ $1.3\times10^5$) compared to raw PIBT$+$'s solutions.

In summary, Layered PIBT$+$ shows little improvement compared to raw PIBT$+$ and has worse solutions due to the introduction of wait actions when merging solutions. However, Layered PIBT$+$ has an advantage in memory usage, as it reduces the number of intermediate states by reducing the number of agents solved simultaneously.

\subsection{Lazy Constraints Addition for MAPF 2}

Lazy Constraints Addition for MAPF 2 (LaCAM2) \cite{okumura2023improving} is a complete and suboptimal method. More details about LaCAM2 can be found in Section \ref{Relatedworks}. LaCAM avoids the occurrence of the same state of agents twice in the solution. Compared to LaCAM, LaCAM2 \cite{okumura2023improving} introduces a swap action to the successor generator, allowing for the quick generation of solutions. The code\footnote{\url{https://github.com/Kei18/lacam2}} for LaCAM2 provides no interfaces to avoid conflicts with external paths, making it a parallel MAPF method. In the implementation, LaCAM2 stores all high-level nodes during planning, which increases rapidly as the number of agents grows. Therefore, LaCAM2's memory usage is primarily determined by the maximum number of high-level nodes. We use this maximum number as an estimation of LaCAM2's memory usage.

The average time cost, success rate, memory usage, makespan, and sum of costs for both Layered LaCAM2 and raw LaCAM2 are shown in Fig. \ref{method_summary}. The performance of Layered LaCAM2 and raw LaCAM2 across various maps, as the number of agents increases, is presented in Fig. \ref{compare_with_raw_LaCAM} in the Appendix. \\
As shown in Fig. \ref{compare_with_raw_LaCAM}, in some cases, Layered LaCAM2 takes more time than raw LaCAM2, while in other cases it does not, as LaCAM2's time cost is close to the decomposition time cost of the instance. Both Layered LaCAM2 and raw LaCAM2 achieve a 100\% success rate, as LaCAM2 is a complete and highly efficient MAPF method. \\
In terms of memory usage, Layered LaCAM2 consistently requires less memory than raw LaCAM2, even as the number of agents increases. \\
Regarding path quality, the solutions generated by Layered LaCAM2 are slightly worse than those generated by raw LaCAM2. A summary of the performance of both Layered LaCAM2 and raw LaCAM2 can be found in Fig. \ref{method_summary}.

In average, Layered LaCAM2 has larger time cost than raw LaCAM2 ($9.6\times10^2$ ms $>$ $7.3\times10^2$ ms). Similar to PIBT$+$, the decomposition of MAPF instances with 800 to 1000 agents may cost 1s to 3s, and these instances play a major role in determining the time cost of the MAPF method. Although decomposition reduces the time cost of LaCAM2, the time cost of decomposition offsets the reduced time cost, resulting in no significant difference in the overall time cost.

Both Layered LaCAM2 and raw LaCAM2 achieve the same success rate (Layered LaCAM2: 1.0, raw LaCAM2: 1.0), indicating that both methods successfully find all solutions within 30 seconds. This suggests that the decomposition of the instance does not sacrifice its solvability during the experiments. If there had been any loss of solvability, Layered LaCAM2's success rate would have been lower than 1.0. Based on this data, we speculate that the likelihood of our decomposition causing a loss of solvability is less than 1\%.  In terms of memory usage, Layered LaCAM2 requires significantly less memory space than raw LaCAM2 ($3.4\times10^4$ configs $<$ $1.7\times10^5$ configs), as decomposition reduces the number of intermediate states by reducing the number of agents solved simultaneously.

Regarding solution quality, Layered LaCAM2 introduces extra wait actions, resulting in solutions with higher makespan ($9.5\times10^3$ $>$ $3.4\times10^2$) and SOC ($3.2\times10^6$ $>$ $9.5\times10^4$) compared to raw LaCAM2's solutions.

LaCAM2 is the only algorithm among the seven that achieves a 100\% success rate, making it a useful benchmark for estimating the loss of solvability caused by our decomposition of the instance. Since Layered LaCAM2 also maintains a 100\% success rate, we speculate that the loss of solvability due to our decomposition is less than 1\%.

In summary, Layered LaCAM2 shows little improvement compared to raw LaCAM2 and has worse solutions due to the introduction of wait actions when merging solutions. However, Layered LaCAM2 has an advantage in memory usage, as it reduces the number of intermediate states by reducing the number of agents solved simultaneously.

\subsection{Summary}
On average, decomposition of MAPF instances significantly reduces the memory usage of all seven methods and also decreases the time cost for EECBS, PBS, LNS2, Push and Swap. Layered MAPF methods exhibit higher success rates compared to raw MAPF methods, particularly for serial MAPF methods, with the exception of HCA*, an incomplete prioritized method that solves agents one-by-one.

Regarding solution quality, the Layered version of serial MAPF methods generally exhibits similar quality compared to the raw version. However, the Layered version of parallel MAPF methods tends to produce worse solutions due to the insertion of numerous wait actions during solution merging. This disparity arises from the fact that serial MAPF methods treat external paths as dynamic obstacles to avoid, while parallel MAPF methods do not.

In conclusion, decomposition of MAPF instances is most advantageous for serial MAPF methods, resulting in reduced time cost and memory usage without significantly sacrificing solution quality. For parallel MAPF methods, decomposition reduce memory usage significantly but often leads to worsened solutions without notable improvements in time cost.

Moreover, serial MAPF methods (e.g., EECBS, LNS2) generally demand more time compared to parallel MAPF methods (e.g., LaCAM and PIBT+), yet they yield higher-quality solutions. Notably, serial MAPF methods frequently fail due to time constraints, while parallel MAPF methods tend to fail due to memory limitations.

The observed discrepancy can be attributed to the differing search algorithms utilized: serial methods repeatedly search for complete paths. However, they employ memory-efficient data structures such as priority trees or constraint trees to store knowledge about resolving conflicts. In contrast, parallel methods store previous conflict-solving knowledge as intermediate states to avoid searching for the complete path of agents repeatedly. However, they may exhaust memory resources under complex MAPF instances.

\section{Conclusion}
\label{Conclusion}

MAPF is extensively utilized in autonomous systems, such as automated warehouses, UAV traffic management and multi-agent exploration. Motivated by the exponential growth in the cost of solving MAPF instances (in terms of time and memory usage) as the number of agents increases, we proposed Layered MAPF as a solution to reduce the computational burden. This approach decomposes a MAPF instance into multiple smaller subproblemsprogressively, and minimize the possibility of loss of solvability by introducing legality check during each step of decomposition. In terms of time complexity, in the worst case, time complexity of decomposition is $\mathcal{O}(k^5)$; while in the best case, its time complexity is $\mathcal{O}(k^3)$, where $k$ represents the number of agents.

In the results of our decomposition of MAPF instances (Section \ref{Resultsofdecomposition}), we observed that our method is highly effective for MAPF instances with passable cells exceeding twice the number of agents. On average, the time cost is around 1s and never exceeds 3s, even for dense instances with 800 to 1000 agents. Memory usage remains below 1MB, with fewer computations and memory space required for maps with more passable cells than agents.

When applied to the state-of-the-art methods (EECBS, PBS, LNS2, HCA*, Push and Swap, PIBT+, LaCAM2), Layered MAPF significantly reduces memory usage and time cost, particularly for serial MAPF methods. Consequently, Layered MAPF methods achieve higher success rates than raw MAPF methods, especially for serial MAPF. And the quality of solution for the Layered version of serial MAPF methods is similar to the raw version, while the Layered version of parallel MAPF methods produces inferior solutions due to the introduction of numerous wait actions during solution merging.

In conclusion, decomposition of MAPF instances is most beneficial for serial MAPF methods, resulting in reduced time cost and memory usage without sacrificing solution quality significantly. However, for parallel MAPF methods, decomposition may reduce memory usage but often worsens the solution without notable improvements in time cost.

Despite its effectiveness, Layered MAPF has limitations: it becomes less effective as the number of agents increases in dense instances, and its application to parallel MAPF methods introduces numerous wait actions during solution merging.

In the future, we plan to propose new merging solution techniques for parallel methods without compromising solution quality. Essentially, the decomposition of the instance is a trade-off between solution quality and efficiency, and a systematic exploration of this trade-off between runtime and solution quality is necessary. Given that the initial state of the cluster plays a crucial role in the decomposition process, a randomized restart approach could benefit the decomposition. We plan to generate different initial clusters during the bipartition phase and select the one that results in the best decomposition. While this process introduces some extra time cost, it may lead to better decomposition outcomes. Additionally, we aim to generalize the idea of decomposing MAPF instances to address extensions of MAPF problems, such as considering the shape of agents.

\bibliography{mybibfile}


\begin{thebibliography}{32}
\ifx \bisbn   \undefined \def \bisbn  #1{ISBN #1}\fi
\ifx \binits  \undefined \def \binits#1{#1}\fi
\ifx \bauthor  \undefined \def \bauthor#1{#1}\fi
\ifx \batitle  \undefined \def \batitle#1{#1}\fi
\ifx \bjtitle  \undefined \def \bjtitle#1{#1}\fi
\ifx \bvolume  \undefined \def \bvolume#1{\textbf{#1}}\fi
\ifx \byear  \undefined \def \byear#1{#1}\fi
\ifx \bissue  \undefined \def \bissue#1{#1}\fi
\ifx \bfpage  \undefined \def \bfpage#1{#1}\fi
\ifx \blpage  \undefined \def \blpage #1{#1}\fi
\ifx \burl  \undefined \def \burl#1{\textsf{#1}}\fi
\ifx \doiurl  \undefined \def \doiurl#1{\url{https://doi.org/#1}}\fi
\ifx \betal  \undefined \def \betal{\textit{et al.}}\fi
\ifx \binstitute  \undefined \def \binstitute#1{#1}\fi
\ifx \binstitutionaled  \undefined \def \binstitutionaled#1{#1}\fi
\ifx \bctitle  \undefined \def \bctitle#1{#1}\fi
\ifx \beditor  \undefined \def \beditor#1{#1}\fi
\ifx \bpublisher  \undefined \def \bpublisher#1{#1}\fi
\ifx \bbtitle  \undefined \def \bbtitle#1{#1}\fi
\ifx \bedition  \undefined \def \bedition#1{#1}\fi
\ifx \bseriesno  \undefined \def \bseriesno#1{#1}\fi
\ifx \blocation  \undefined \def \blocation#1{#1}\fi
\ifx \bsertitle  \undefined \def \bsertitle#1{#1}\fi
\ifx \bsnm \undefined \def \bsnm#1{#1}\fi
\ifx \bsuffix \undefined \def \bsuffix#1{#1}\fi
\ifx \bparticle \undefined \def \bparticle#1{#1}\fi
\ifx \barticle \undefined \def \barticle#1{#1}\fi
\bibcommenthead
\ifx \bconfdate \undefined \def \bconfdate #1{#1}\fi
\ifx \botherref \undefined \def \botherref #1{#1}\fi
\ifx \url \undefined \def \url#1{\textsf{#1}}\fi
\ifx \bchapter \undefined \def \bchapter#1{#1}\fi
\ifx \bbook \undefined \def \bbook#1{#1}\fi
\ifx \bcomment \undefined \def \bcomment#1{#1}\fi
\ifx \oauthor \undefined \def \oauthor#1{#1}\fi
\ifx \citeauthoryear \undefined \def \citeauthoryear#1{#1}\fi
\ifx \endbibitem  \undefined \def \endbibitem {}\fi
\ifx \bconflocation  \undefined \def \bconflocation#1{#1}\fi
\ifx \arxivurl  \undefined \def \arxivurl#1{\textsf{#1}}\fi
\csname PreBibitemsHook\endcsname

\bibitem[\protect\citeauthoryear{H{\"o}nig et~al.}{2019}]{honig2019persistent}
\begin{barticle}
\bauthor{\bsnm{H{\"o}nig}, \binits{W.}},
\bauthor{\bsnm{Kiesel}, \binits{S.}},
\bauthor{\bsnm{Tinka}, \binits{A.}},
\bauthor{\bsnm{Durham}, \binits{J.W.}},
\bauthor{\bsnm{Ayanian}, \binits{N.}}:
\batitle{Persistent and robust execution of mapf schedules in warehouses}.
\bjtitle{IEEE Robotics and Automation Letters}
\bvolume{4}(\bissue{2}),
\bfpage{1125}--\blpage{1131}
(\byear{2019})
\end{barticle}
\endbibitem

\bibitem[\protect\citeauthoryear{Ho et~al.}{2020}]{ho2020decentralized}
\begin{barticle}
\bauthor{\bsnm{Ho}, \binits{F.}},
\bauthor{\bsnm{Geraldes}, \binits{R.}},
\bauthor{\bsnm{Gon{\c{c}}alves}, \binits{A.}},
\bauthor{\bsnm{Rigault}, \binits{B.}},
\bauthor{\bsnm{Sportich}, \binits{B.}},
\bauthor{\bsnm{Kubo}, \binits{D.}},
\bauthor{\bsnm{Cavazza}, \binits{M.}},
\bauthor{\bsnm{Prendinger}, \binits{H.}}:
\batitle{Decentralized multi-agent path finding for uav traffic management}.
\bjtitle{IEEE Transactions on Intelligent Transportation Systems}
\bvolume{23}(\bissue{2}),
\bfpage{997}--\blpage{1008}
(\byear{2020})
\end{barticle}
\endbibitem

\bibitem[\protect\citeauthoryear{Li et~al.}{2022}]{li2022mapf}
\begin{bchapter}
\bauthor{\bsnm{Li}, \binits{J.}},
\bauthor{\bsnm{Chen}, \binits{Z.}},
\bauthor{\bsnm{Harabor}, \binits{D.}},
\bauthor{\bsnm{Stuckey}, \binits{P.J.}},
\bauthor{\bsnm{Koenig}, \binits{S.}}:
\bctitle{Mapf-lns2: Fast repairing for multi-agent path finding via large
  neighborhood search}.
In: \bbtitle{Proceedings of the AAAI Conference on Artificial Intelligence},
vol. \bseriesno{36},
pp. \bfpage{10256}--\blpage{10265}
(\byear{2022})
\end{bchapter}
\endbibitem

\bibitem[\protect\citeauthoryear{Ma et~al.}{2019}]{ma2019searching}
\begin{bchapter}
\bauthor{\bsnm{Ma}, \binits{H.}},
\bauthor{\bsnm{Harabor}, \binits{D.}},
\bauthor{\bsnm{Stuckey}, \binits{P.J.}},
\bauthor{\bsnm{Li}, \binits{J.}},
\bauthor{\bsnm{Koenig}, \binits{S.}}:
\bctitle{Searching with consistent prioritization for multi-agent path
  finding}.
In: \bbtitle{Proceedings of the AAAI Conference on Artificial Intelligence},
vol. \bseriesno{33},
pp. \bfpage{7643}--\blpage{7650}
(\byear{2019})
\end{bchapter}
\endbibitem

\bibitem[\protect\citeauthoryear{Sharon et~al.}{2015}]{sharon2015conflict}
\begin{barticle}
\bauthor{\bsnm{Sharon}, \binits{G.}},
\bauthor{\bsnm{Stern}, \binits{R.}},
\bauthor{\bsnm{Felner}, \binits{A.}},
\bauthor{\bsnm{Sturtevant}, \binits{N.R.}}:
\batitle{Conflict-based search for optimal multi-agent pathfinding}.
\bjtitle{Artificial intelligence}
\bvolume{219},
\bfpage{40}--\blpage{66}
(\byear{2015})
\end{barticle}
\endbibitem

\bibitem[\protect\citeauthoryear{Hatem et~al.}{2013}]{hatem2013bounded}
\begin{bchapter}
\bauthor{\bsnm{Hatem}, \binits{M.}},
\bauthor{\bsnm{Stern}, \binits{R.}},
\bauthor{\bsnm{Ruml}, \binits{W.}}:
\bctitle{Bounded suboptimal heuristic search in linear space}.
In: \bbtitle{Proceedings of the International Symposium on Combinatorial
  Search},
vol. \bseriesno{4},
pp. \bfpage{98}--\blpage{104}
(\byear{2013})
\end{bchapter}
\endbibitem

\bibitem[\protect\citeauthoryear{Li et~al.}{2021}]{li2021eecbs}
\begin{bchapter}
\bauthor{\bsnm{Li}, \binits{J.}},
\bauthor{\bsnm{Ruml}, \binits{W.}},
\bauthor{\bsnm{Koenig}, \binits{S.}}:
\bctitle{Eecbs: A bounded-suboptimal search for multi-agent path finding}.
In: \bbtitle{Proceedings of the AAAI Conference on Artificial Intelligence},
vol. \bseriesno{35},
pp. \bfpage{12353}--\blpage{12362}
(\byear{2021})
\end{bchapter}
\endbibitem

\bibitem[\protect\citeauthoryear{Thayer and Ruml}{2011}]{thayer2011bounded}
\begin{bchapter}
\bauthor{\bsnm{Thayer}, \binits{J.T.}},
\bauthor{\bsnm{Ruml}, \binits{W.}}:
\bctitle{Bounded suboptimal search: A direct approach using inadmissible
  estimates}.
In: \bbtitle{IJCAI},
vol. \bseriesno{2011},
pp. \bfpage{674}--\blpage{679}
(\byear{2011})
\end{bchapter}
\endbibitem

\bibitem[\protect\citeauthoryear{Boyarski et~al.}{2015}]{boyarski2015don}
\begin{bchapter}
\bauthor{\bsnm{Boyarski}, \binits{E.}},
\bauthor{\bsnm{Felner}, \binits{A.}},
\bauthor{\bsnm{Sharon}, \binits{G.}},
\bauthor{\bsnm{Stern}, \binits{R.}}:
\bctitle{Don't split, try to work it out: Bypassing conflicts in multi-agent
  pathfinding}.
In: \bbtitle{Proceedings of the International Conference on Automated Planning
  and Scheduling},
vol. \bseriesno{25},
pp. \bfpage{47}--\blpage{51}
(\byear{2015})
\end{bchapter}
\endbibitem

\bibitem[\protect\citeauthoryear{Li et~al.}{2019}]{li2019symmetry}
\begin{bchapter}
\bauthor{\bsnm{Li}, \binits{J.}},
\bauthor{\bsnm{Harabor}, \binits{D.}},
\bauthor{\bsnm{Stuckey}, \binits{P.J.}},
\bauthor{\bsnm{Ma}, \binits{H.}},
\bauthor{\bsnm{Koenig}, \binits{S.}}:
\bctitle{Symmetry-breaking constraints for grid-based multi-agent path
  finding}.
In: \bbtitle{Proceedings of the AAAI Conference on Artificial Intelligence},
vol. \bseriesno{33},
pp. \bfpage{6087}--\blpage{6095}
(\byear{2019})
\end{bchapter}
\endbibitem

\bibitem[\protect\citeauthoryear{Li et~al.}{2020}]{li2020new}
\begin{bchapter}
\bauthor{\bsnm{Li}, \binits{J.}},
\bauthor{\bsnm{Gange}, \binits{G.}},
\bauthor{\bsnm{Harabor}, \binits{D.}},
\bauthor{\bsnm{Stuckey}, \binits{P.J.}},
\bauthor{\bsnm{Ma}, \binits{H.}},
\bauthor{\bsnm{Koenig}, \binits{S.}}:
\bctitle{New techniques for pairwise symmetry breaking in multi-agent path
  finding}.
In: \bbtitle{Proceedings of the International Conference on Automated Planning
  and Scheduling},
vol. \bseriesno{30},
pp. \bfpage{193}--\blpage{201}
(\byear{2020})
\end{bchapter}
\endbibitem

\bibitem[\protect\citeauthoryear{Li et~al.}{2019}]{li2019improved}
\begin{bchapter}
\bauthor{\bsnm{Li}, \binits{J.}},
\bauthor{\bsnm{Felner}, \binits{A.}},
\bauthor{\bsnm{Boyarski}, \binits{E.}},
\bauthor{\bsnm{Ma}, \binits{H.}},
\bauthor{\bsnm{Koenig}, \binits{S.}}:
\bctitle{Improved heuristics for multi-agent path finding with conflict-based
  search.}
In: \bbtitle{IJCAI},
vol. \bseriesno{2019},
pp. \bfpage{442}--\blpage{449}
(\byear{2019})
\end{bchapter}
\endbibitem

\bibitem[\protect\citeauthoryear{Li et~al.}{2021}]{li2021anytime}
\begin{bchapter}
\bauthor{\bsnm{Li}, \binits{J.}},
\bauthor{\bsnm{Chen}, \binits{Z.}},
\bauthor{\bsnm{Harabor}, \binits{D.}},
\bauthor{\bsnm{Stuckey}, \binits{P.J.}},
\bauthor{\bsnm{Koenig}, \binits{S.}}:
\bctitle{Anytime multi-agent path finding via large neighborhood search}.
In: \bbtitle{International Joint Conference on Artificial Intelligence 2021},
pp. \bfpage{4127}--\blpage{4135}
(\byear{2021}).
\bcomment{Association for the Advancement of Artificial Intelligence (AAAI)}
\end{bchapter}
\endbibitem

\bibitem[\protect\citeauthoryear{Chan et~al.}{2023}]{chan2023greedy}
\begin{bchapter}
\bauthor{\bsnm{Chan}, \binits{S.-H.}},
\bauthor{\bsnm{Stern}, \binits{R.}},
\bauthor{\bsnm{Felner}, \binits{A.}},
\bauthor{\bsnm{Koenig}, \binits{S.}}:
\bctitle{Greedy priority-based search for suboptimal multi-agent path finding}.
In: \bbtitle{Proceedings of the International Symposium on Combinatorial
  Search},
vol. \bseriesno{16},
pp. \bfpage{11}--\blpage{19}
(\byear{2023})
\end{bchapter}
\endbibitem

\bibitem[\protect\citeauthoryear{Okumura et~al.}{2019}]{okumura2019priority}
\begin{bchapter}
\bauthor{\bsnm{Okumura}, \binits{K.}},
\bauthor{\bsnm{Machida}, \binits{M.}},
\bauthor{\bsnm{D{\'e}fago}, \binits{X.}},
\bauthor{\bsnm{Tamura}, \binits{Y.}}:
\bctitle{Priority inheritance with backtracking for iterative multi-agent path
  finding}.
In: \bbtitle{Proceedings of the Twenty-Eighth International Joint Conference on
  Artificial Intelligence, {IJCAI-19}},
pp. \bfpage{535}--\blpage{542}.
\bpublisher{International Joint Conferences on Artificial Intelligence
  Organization}, \blocation{???}
(\byear{2019}).
\doiurl{10.24963/ijcai.2019/76} .
\burl{https://doi.org/10.24963/ijcai.2019/76}
\end{bchapter}
\endbibitem

\bibitem[\protect\citeauthoryear{Okumura et~al.}{2022}]{okumura2022priority}
\begin{botherref}
\oauthor{\bsnm{Okumura}, \binits{K.}},
\oauthor{\bsnm{Machida}, \binits{M.}},
\oauthor{\bsnm{Défago}, \binits{X.}},
\oauthor{\bsnm{Tamura}, \binits{Y.}}:
Priority inheritance with backtracking for iterative multi-agent path finding.
Artificial Intelligence,
103752
(2022)
\doiurl{10.1016/j.artint.2022.103752}
\end{botherref}
\endbibitem

\bibitem[\protect\citeauthoryear{Okumura}{2023a}]{okumura2023lacam}
\begin{bchapter}
\bauthor{\bsnm{Okumura}, \binits{K.}}:
\bctitle{Lacam: Search-based algorithm for quick multi-agent pathfinding}.
In: \bbtitle{Proceedings of the AAAI Conference on Artificial Intelligence},
vol. \bseriesno{37},
pp. \bfpage{11655}--\blpage{11662}
(\byear{2023})
\end{bchapter}
\endbibitem

\bibitem[\protect\citeauthoryear{Okumura}{2023b}]{okumura2023improving}
\begin{botherref}
\oauthor{\bsnm{Okumura}, \binits{K.}}:
Improving lacam for scalable eventually optimal multi-agent pathfinding.
arXiv preprint arXiv:2305.03632
(2023)
\end{botherref}
\endbibitem

\bibitem[\protect\citeauthoryear{Standley}{2010}]{Standley2010FindingOS}
\begin{botherref}
\oauthor{\bsnm{Standley}, \binits{T.S.}}:
Finding optimal solutions to cooperative pathfinding problems.
Proceedings of the AAAI Conference on Artificial Intelligence
(2010)
\end{botherref}
\endbibitem

\bibitem[\protect\citeauthoryear{Standley and
  Korf}{2011}]{Standley2011CompleteAF}
\begin{bchapter}
\bauthor{\bsnm{Standley}, \binits{T.S.}},
\bauthor{\bsnm{Korf}, \binits{R.E.}}:
\bctitle{Complete algorithms for cooperative pathfinding problems}.
In: \bbtitle{International Joint Conference on Artificial Intelligence}
(\byear{2011}).
\burl{https://api.semanticscholar.org/CorpusID:11254757}
\end{bchapter}
\endbibitem

\bibitem[\protect\citeauthoryear{Sharon et~al.}{2021}]{Sharon2021MetaAgentCS}
\begin{bchapter}
\bauthor{\bsnm{Sharon}, \binits{G.}},
\bauthor{\bsnm{Stern}, \binits{R.}},
\bauthor{\bsnm{Felner}, \binits{A.}},
\bauthor{\bsnm{Sturtevant}, \binits{N.R.}}:
\bctitle{Meta-agent conflict-based search for optimal multi-agent path
  finding}.
In: \bbtitle{Symposium on Combinatorial Search}
(\byear{2021}).
\burl{https://api.semanticscholar.org/CorpusID:253300147}
\end{bchapter}
\endbibitem

\bibitem[\protect\citeauthoryear{vancara et~al.}{2019}]{Ji2019Online}
\begin{barticle}
\bauthor{\bsnm{vancara}, \binits{J.}},
\bauthor{\bsnm{Vlk}, \binits{M.}},
\bauthor{\bsnm{Stern}, \binits{R.}},
\bauthor{\bsnm{Atzmon}, \binits{D.}},
\bauthor{\bsnm{Barták}, \binits{R.}}:
\batitle{Online multi-agent pathfinding}.
\bjtitle{Proceedings of the AAAI Conference on Artificial Intelligence}
\bvolume{33},
\bfpage{7732}--\blpage{7739}
(\byear{2019})
\end{barticle}
\endbibitem

\bibitem[\protect\citeauthoryear{Ma et~al.}{2017}]{2017Lifelong}
\begin{botherref}
\oauthor{\bsnm{Ma}, \binits{H.}},
\oauthor{\bsnm{Li}, \binits{J.}},
\oauthor{\bsnm{Kumar}, \binits{T.K.S.}},
\oauthor{\bsnm{Koenig}, \binits{S.}}:
Lifelong multi-agent path finding for online pickup and delivery tasks.
International Foundation for Autonomous Agents and Multiagent Systems
(2017)
\end{botherref}
\endbibitem

\bibitem[\protect\citeauthoryear{Ma}{2021}]{ma2021competitive}
\begin{bchapter}
\bauthor{\bsnm{Ma}, \binits{H.}}:
\bctitle{A competitive analysis of online multi-agent path finding}.
In: \bbtitle{Proceedings of the International Conference on Automated Planning
  and Scheduling},
vol. \bseriesno{31},
pp. \bfpage{234}--\blpage{242}
(\byear{2021})
\end{bchapter}
\endbibitem

\bibitem[\protect\citeauthoryear{Atzmon et~al.}{2023}]{DSPAtzmon}
\begin{bchapter}
\bauthor{\bsnm{Atzmon}, \binits{D.}},
\bauthor{\bsnm{Bernardini}, \binits{S.}},
\bauthor{\bsnm{Fagnani}, \binits{F.}},
\bauthor{\bsnm{Fairbairn}, \binits{D.}}:
\bctitle{Exploiting geometric constraints in multi-agent pathfinding}.
In: \bbtitle{Proceedings of the Thirty-Third International Conference on
  Automated Planning and Scheduling}.
\bsertitle{ICAPS '23}.
\bpublisher{AAAI Press}, \blocation{???}
(\byear{2023}).
\doiurl{10.1609/icaps.v33i1.27174} .
\burl{https://doi.org/10.1609/icaps.v33i1.27174}
\end{bchapter}
\endbibitem

\bibitem[\protect\citeauthoryear{Tarjan}{1972}]{tarjan1972depth}
\begin{barticle}
\bauthor{\bsnm{Tarjan}, \binits{R.}}:
\batitle{Depth-first search and linear graph algorithms}.
\bjtitle{SIAM journal on computing}
\bvolume{1}(\bissue{2}),
\bfpage{146}--\blpage{160}
(\byear{1972})
\end{barticle}
\endbibitem

\bibitem[\protect\citeauthoryear{Tarjan}{1983}]{tarjan1983data}
\begin{bbook}
\bauthor{\bsnm{Tarjan}, \binits{R.E.}}:
\bbtitle{Data Structures and Network Algorithms}.
\bpublisher{SIAM}, \blocation{???}
(\byear{1983})
\end{bbook}
\endbibitem

\bibitem[\protect\citeauthoryear{Cormen et~al.}{2022}]{cormen2022introduction}
\begin{bbook}
\bauthor{\bsnm{Cormen}, \binits{T.H.}},
\bauthor{\bsnm{Leiserson}, \binits{C.E.}},
\bauthor{\bsnm{Rivest}, \binits{R.L.}},
\bauthor{\bsnm{Stein}, \binits{C.}}:
\bbtitle{Introduction to Algorithms}.
\bpublisher{MIT press}, \blocation{???}
(\byear{2022})
\end{bbook}
\endbibitem

\bibitem[\protect\citeauthoryear{Stern et~al.}{2019}]{stern2019mapf}
\begin{botherref}
\oauthor{\bsnm{Stern}, \binits{R.}},
\oauthor{\bsnm{Sturtevant}, \binits{N.R.}},
\oauthor{\bsnm{Felner}, \binits{A.}},
\oauthor{\bsnm{Koenig}, \binits{S.}},
\oauthor{\bsnm{Ma}, \binits{H.}},
\oauthor{\bsnm{Walker}, \binits{T.T.}},
\oauthor{\bsnm{Li}, \binits{J.}},
\oauthor{\bsnm{Atzmon}, \binits{D.}},
\oauthor{\bsnm{Cohen}, \binits{L.}},
\oauthor{\bsnm{Kumar}, \binits{T.K.S.}},
\oauthor{\bsnm{Boyarski}, \binits{E.}},
\oauthor{\bsnm{Bartak}, \binits{R.}}:
Multi-agent pathfinding: Definitions, variants, and benchmarks.
Symposium on Combinatorial Search (SoCS),
151--158
(2019)
\end{botherref}
\endbibitem

\bibitem[\protect\citeauthoryear{Surynek et~al.}{2018}]{surynek2018variants}
\begin{bchapter}
\bauthor{\bsnm{Surynek}, \binits{P.}},
\bauthor{\bsnm{{\v{S}}vancara}, \binits{J.}},
\bauthor{\bsnm{Felner}, \binits{A.}},
\bauthor{\bsnm{Boyarski}, \binits{E.}}:
\bctitle{Variants of independence detection in sat-based optimal multi-agent
  path finding}.
In: \bbtitle{Agents and Artificial Intelligence: 9th International Conference,
  ICAART 2017, Porto, Portugal, February 24--26, 2017, Revised Selected Papers
  9},
pp. \bfpage{116}--\blpage{136}
(\byear{2018}).
\bcomment{Springer}
\end{bchapter}
\endbibitem

\bibitem[\protect\citeauthoryear{Luna and Bekris}{2011}]{luna2011push}
\begin{bchapter}
\bauthor{\bsnm{Luna}, \binits{R.}},
\bauthor{\bsnm{Bekris}, \binits{K.E.}}:
\bctitle{Push and swap: Fast cooperative path-finding with completeness
  guarantees}.
In: \bbtitle{IJCAI},
vol. \bseriesno{11},
pp. \bfpage{294}--\blpage{300}
(\byear{2011})
\end{bchapter}
\endbibitem

\bibitem[\protect\citeauthoryear{Silver}{2005}]{silver2005cooperative}
\begin{bchapter}
\bauthor{\bsnm{Silver}, \binits{D.}}:
\bctitle{Cooperative pathfinding}.
In: \bbtitle{Proceedings of the Aaai Conference on Artificial Intelligence and
  Interactive Digital Entertainment},
vol. \bseriesno{1},
pp. \bfpage{117}--\blpage{122}
(\byear{2005})
\end{bchapter}
\endbibitem

\end{thebibliography}



\begin{appendices}

\setcounter{figure}{0}
\renewcommand{\thefigure}{A\arabic{figure}} 

\section{Comparison between Layered MAPF and raw MAPF as the number of agent increasing}

\begin{figure}[h] \tiny

\begin{minipage}{.24\linewidth}
  \centerline{\includegraphics[width=3.2cm]{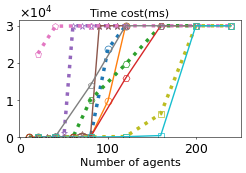}}
\end{minipage}
\hfill
\begin{minipage}{.24\linewidth}
  \centerline{\includegraphics[width=3.2cm]{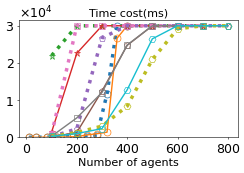}}
\end{minipage}
\hfill
\begin{minipage}{.24\linewidth}
  \centerline{\includegraphics[width=3.2cm]{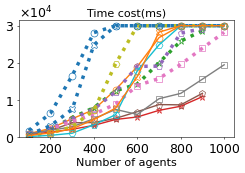}}
\end{minipage}
\hfill
\begin{minipage}{.24\linewidth}
  \centerline{\includegraphics[width=3.2cm]{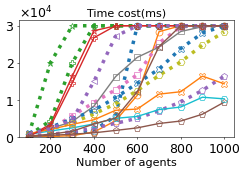}}
\end{minipage}
\vfill

\begin{minipage}{.24\linewidth}
  \centerline{\includegraphics[width=3.2cm]{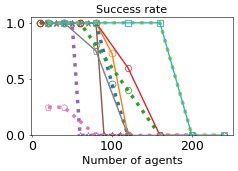}}
\end{minipage}
\hfill
\begin{minipage}{.24\linewidth}
  \centerline{\includegraphics[width=3.2cm]{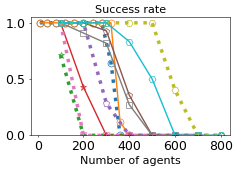}}
\end{minipage}
\hfill
\begin{minipage}{.24\linewidth}
  \centerline{\includegraphics[width=3.2cm]{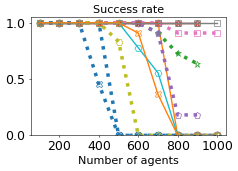}}
\end{minipage}
\hfill
\begin{minipage}{.24\linewidth}
  \centerline{\includegraphics[width=3.2cm]{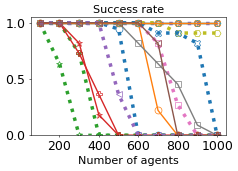}}
\end{minipage}
\vfill

\begin{minipage}{.24\linewidth}
  \centerline{\includegraphics[width=3.2cm]{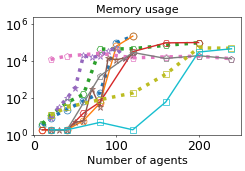}}
\end{minipage}
\hfill
\begin{minipage}{.24\linewidth}
  \centerline{\includegraphics[width=3.2cm]{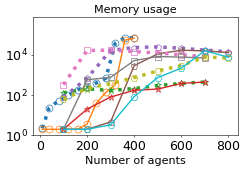}}
\end{minipage}
\hfill
\begin{minipage}{.24\linewidth}
  \centerline{\includegraphics[width=3.2cm]{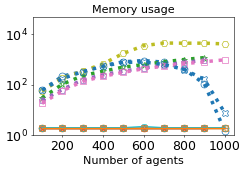}}
\end{minipage}
\hfill
\begin{minipage}{.24\linewidth}
  \centerline{\includegraphics[width=3.2cm]{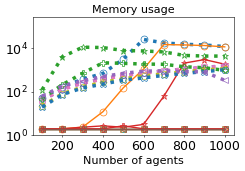}}
\end{minipage}
\vfill

\begin{minipage}{.24\linewidth}
  \centerline{\includegraphics[width=3.2cm]{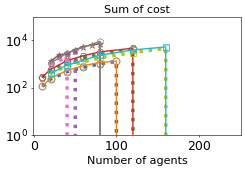}}
\end{minipage}
\hfill
\begin{minipage}{.24\linewidth}
  \centerline{\includegraphics[width=3.2cm]{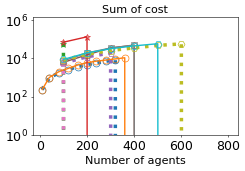}}
\end{minipage}
\hfill
\begin{minipage}{.24\linewidth}
  \centerline{\includegraphics[width=3.2cm]{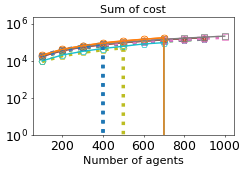}}
\end{minipage}
\hfill
\begin{minipage}{.24\linewidth}
  \centerline{\includegraphics[width=3.2cm]{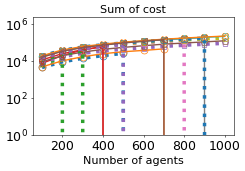}}
\end{minipage}
\vfill

\begin{minipage}{.24\linewidth}
  \centerline{\includegraphics[width=3.2cm]{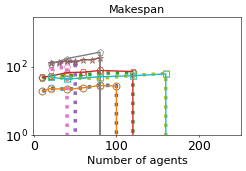}}
\end{minipage}
\hfill
\begin{minipage}{.24\linewidth}
  \centerline{\includegraphics[width=3.2cm]{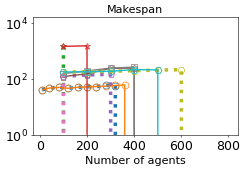}}
\end{minipage}
\hfill
\begin{minipage}{.24\linewidth}
  \centerline{\includegraphics[width=3.2cm]{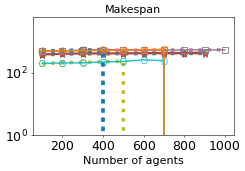}}
\end{minipage}
\hfill
\begin{minipage}{.24\linewidth}
  \centerline{\includegraphics[width=3.2cm]{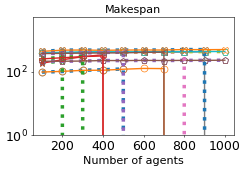}}
\end{minipage}
\vfill

\begin{minipage}{.24\linewidth}
  \centerline{\includegraphics[width=3.2cm]{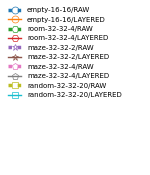}}
\end{minipage}
\hfill
\begin{minipage}{.24\linewidth}
  \centerline{\includegraphics[width=3.2cm]{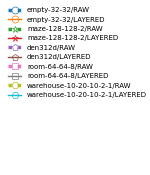}}
\end{minipage}
\hfill
\begin{minipage}{.24\linewidth}
  \centerline{\includegraphics[width=3.2cm]{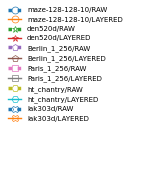}}
\end{minipage}
\hfill
\begin{minipage}{.24\linewidth}
  \centerline{\includegraphics[width=3.2cm]{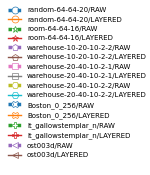}}
\end{minipage}
\vfill

\caption{These figures illustrate how Layered EECBS and raw EECBS work under various maps as the number of agents increases. They are compared in terms of time cost, success rate, memory usage, and solution quality (sum of cost and makespan). The data for Layered EECBS are shown with a solid line, and the data for raw EECBS are shown with a dotted line. More details about the maps can be found in Fig. \ref{decomposition1} and Fig. \ref{decomposition2}.
} 
\label{compare_with_raw_eecbs}
\end{figure}

\begin{figure}[h] \tiny


\begin{minipage}{.24\linewidth}
  \centerline{\includegraphics[width=3.2cm]{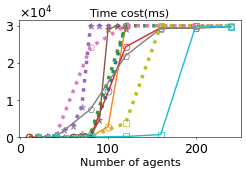}}
\end{minipage}
\hfill
\begin{minipage}{.24\linewidth}
  \centerline{\includegraphics[width=3.2cm]{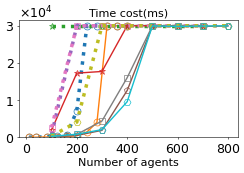}}
\end{minipage}
\hfill
\begin{minipage}{.24\linewidth}
  \centerline{\includegraphics[width=3.2cm]{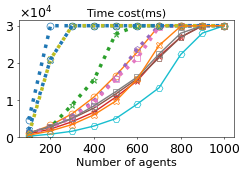}}
\end{minipage}
\hfill
\begin{minipage}{.24\linewidth}
  \centerline{\includegraphics[width=3.2cm]{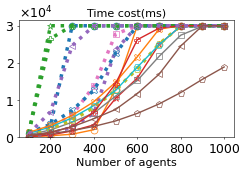}}
\end{minipage}
\vfill

\begin{minipage}{.24\linewidth}
  \centerline{\includegraphics[width=3.2cm]{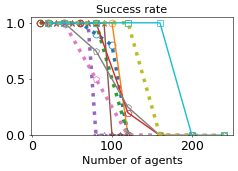}}
\end{minipage}
\hfill
\begin{minipage}{.24\linewidth}
  \centerline{\includegraphics[width=3.2cm]{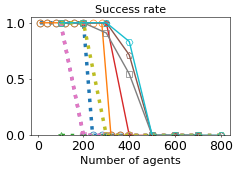}}
\end{minipage}
\hfill
\begin{minipage}{.24\linewidth}
  \centerline{\includegraphics[width=3.2cm]{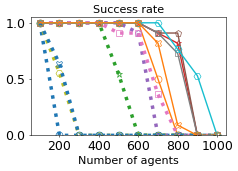}}
\end{minipage}
\hfill
\begin{minipage}{.24\linewidth}
  \centerline{\includegraphics[width=3.2cm]{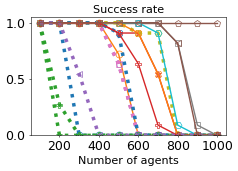}}
\end{minipage}
\vfill

\begin{minipage}{.24\linewidth}
  \centerline{\includegraphics[width=3.2cm]{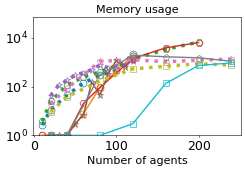}}
\end{minipage}
\hfill
\begin{minipage}{.24\linewidth}
  \centerline{\includegraphics[width=3.2cm]{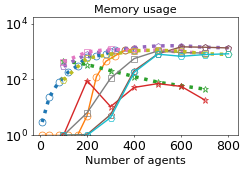}}
\end{minipage}
\hfill
\begin{minipage}{.24\linewidth}
  \centerline{\includegraphics[width=3.2cm]{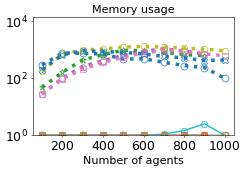}}
\end{minipage}
\hfill
\begin{minipage}{.24\linewidth}
  \centerline{\includegraphics[width=3.2cm]{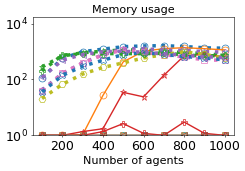}}
\end{minipage}
\vfill

\begin{minipage}{.24\linewidth}
  \centerline{\includegraphics[width=3.2cm]{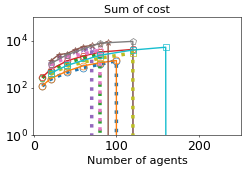}}
\end{minipage}
\hfill
\begin{minipage}{.24\linewidth}
  \centerline{\includegraphics[width=3.2cm]{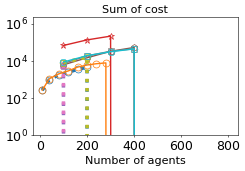}}
\end{minipage}
\hfill
\begin{minipage}{.24\linewidth}
  \centerline{\includegraphics[width=3.2cm]{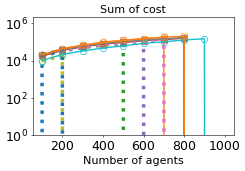}}
\end{minipage}
\hfill
\begin{minipage}{.24\linewidth}
  \centerline{\includegraphics[width=3.2cm]{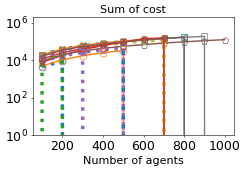}}
\end{minipage}
\vfill

\begin{minipage}{.24\linewidth}
  \centerline{\includegraphics[width=3.2cm]{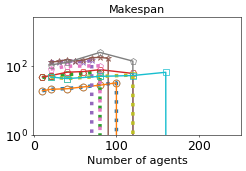}}
\end{minipage}
\hfill
\begin{minipage}{.24\linewidth}
  \centerline{\includegraphics[width=3.2cm]{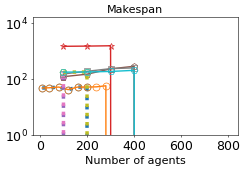}}
\end{minipage}
\hfill
\begin{minipage}{.24\linewidth}
  \centerline{\includegraphics[width=3.2cm]{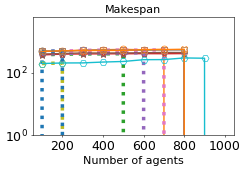}}
\end{minipage}
\hfill
\begin{minipage}{.24\linewidth}
  \centerline{\includegraphics[width=3.2cm]{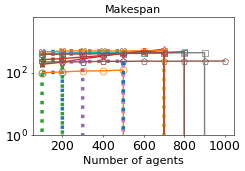}}
\end{minipage}
\vfill

\begin{minipage}{.24\linewidth}
  \centerline{\includegraphics[width=3.2cm]{layered_MAPF/makespan/EECBS/1_legend.png}}
\end{minipage}
\hfill
\begin{minipage}{.24\linewidth}
  \centerline{\includegraphics[width=3.2cm]{layered_MAPF/makespan/EECBS/2_legend.png}}
\end{minipage}
\hfill
\begin{minipage}{.24\linewidth}
  \centerline{\includegraphics[width=3.2cm]{layered_MAPF/makespan/EECBS/3_legend.png}}
\end{minipage}
\hfill
\begin{minipage}{.24\linewidth}
  \centerline{\includegraphics[width=3.2cm]{layered_MAPF/makespan/EECBS/4_legend.png}}
\end{minipage}
\vfill

\caption{These figures illustrate how Layered PBS and raw PBS work under various maps as the number of agents increases. They are compared in terms of time cost, success rate, memory usage, and solution quality (sum of cost and makespan). The data for Layered PBS are shown with a solid line, and the data for raw PBS are shown with a dotted line. More details about the maps can be found in Fig. \ref{decomposition1} and Fig. \ref{decomposition2}.
} 
\label{compare_with_raw_pbs}
\end{figure}

\begin{figure}[h] \tiny


\begin{minipage}{.24\linewidth}
  \centerline{\includegraphics[width=3.2cm]{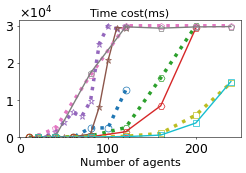}}
\end{minipage}
\hfill
\begin{minipage}{.24\linewidth}
  \centerline{\includegraphics[width=3.2cm]{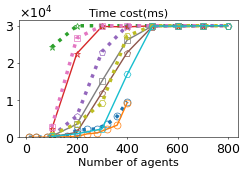}}
\end{minipage}
\hfill
\begin{minipage}{.24\linewidth}
  \centerline{\includegraphics[width=3.2cm]{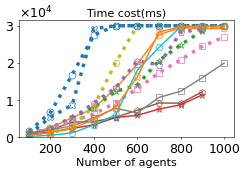}}
\end{minipage}
\hfill
\begin{minipage}{.24\linewidth}
  \centerline{\includegraphics[width=3.2cm]{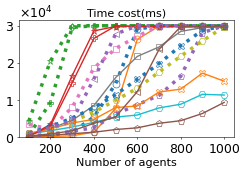}}
\end{minipage}
\vfill

\begin{minipage}{.24\linewidth}
  \centerline{\includegraphics[width=3.2cm]{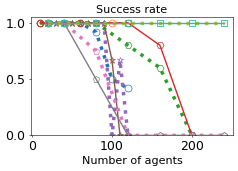}}
\end{minipage}
\hfill
\begin{minipage}{.24\linewidth}
  \centerline{\includegraphics[width=3.2cm]{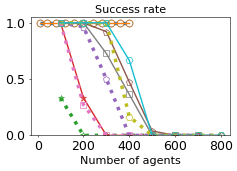}}
\end{minipage}
\hfill
\begin{minipage}{.24\linewidth}
  \centerline{\includegraphics[width=3.2cm]{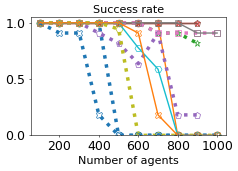}}
\end{minipage}
\hfill
\begin{minipage}{.24\linewidth}
  \centerline{\includegraphics[width=3.2cm]{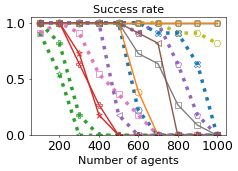}}
\end{minipage}
\vfill

\begin{minipage}{.24\linewidth}
  \centerline{\includegraphics[width=3.2cm]{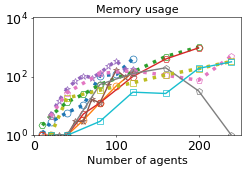}}
\end{minipage}
\hfill
\begin{minipage}{.24\linewidth}
  \centerline{\includegraphics[width=3.2cm]{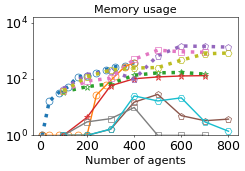}}
\end{minipage}
\hfill
\begin{minipage}{.24\linewidth}
  \centerline{\includegraphics[width=3.2cm]{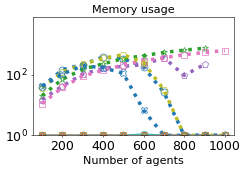}}
\end{minipage}
\hfill
\begin{minipage}{.24\linewidth}
  \centerline{\includegraphics[width=3.2cm]{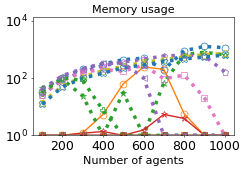}}
\end{minipage}
\vfill

\begin{minipage}{.24\linewidth}
  \centerline{\includegraphics[width=3.2cm]{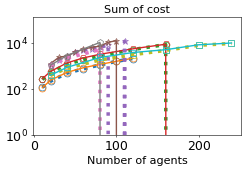}}
\end{minipage}
\hfill
\begin{minipage}{.24\linewidth}
  \centerline{\includegraphics[width=3.2cm]{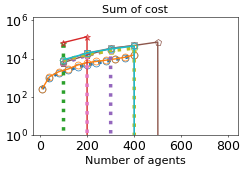}}
\end{minipage}
\hfill
\begin{minipage}{.24\linewidth}
  \centerline{\includegraphics[width=3.2cm]{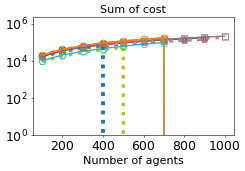}}
\end{minipage}
\hfill
\begin{minipage}{.24\linewidth}
  \centerline{\includegraphics[width=3.2cm]{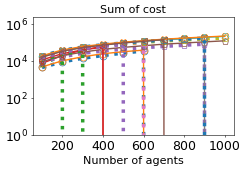}}
\end{minipage}
\vfill

\begin{minipage}{.24\linewidth}
  \centerline{\includegraphics[width=3.2cm]{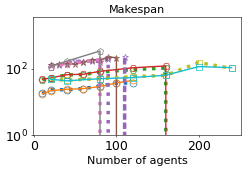}}
\end{minipage}
\hfill
\begin{minipage}{.24\linewidth}
  \centerline{\includegraphics[width=3.2cm]{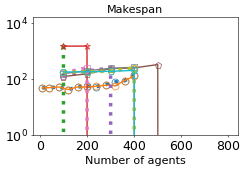}}
\end{minipage}
\hfill
\begin{minipage}{.24\linewidth}
  \centerline{\includegraphics[width=3.2cm]{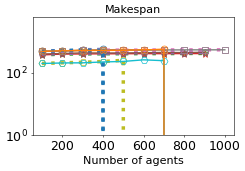}}
\end{minipage}
\hfill
\begin{minipage}{.24\linewidth}
  \centerline{\includegraphics[width=3.2cm]{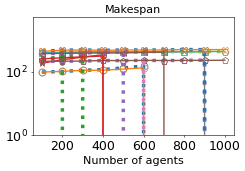}}
\end{minipage}
\vfill

\begin{minipage}{.24\linewidth}
  \centerline{\includegraphics[width=3.2cm]{layered_MAPF/makespan/EECBS/1_legend.png}}
\end{minipage}
\hfill
\begin{minipage}{.24\linewidth}
  \centerline{\includegraphics[width=3.2cm]{layered_MAPF/makespan/EECBS/2_legend.png}}
\end{minipage}
\hfill
\begin{minipage}{.24\linewidth}
  \centerline{\includegraphics[width=3.2cm]{layered_MAPF/makespan/EECBS/3_legend.png}}
\end{minipage}
\hfill
\begin{minipage}{.24\linewidth}
  \centerline{\includegraphics[width=3.2cm]{layered_MAPF/makespan/EECBS/4_legend.png}}
\end{minipage}
\vfill

\caption{These figures illustrate how Layered LNS2 and raw LNS2 works under various of maps as number of agents increasing, they are compared in term of time cost, success rate, memory usage and solution quality (sum of cost and makespan). Layered LNS2's data are shown in solid line and raw LNS2's data are shown in dotted line.  More details about maps can be found in Fig.\ref{decomposition1} and Fig. \ref{decomposition2}. 
} 
\label{compare_with_raw_lns}
\end{figure}

\begin{figure}[h] \tiny


\begin{minipage}{.24\linewidth}
  \centerline{\includegraphics[width=3.2cm]{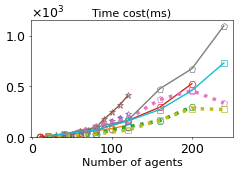}}
\end{minipage}
\hfill
\begin{minipage}{.24\linewidth}
  \centerline{\includegraphics[width=3.2cm]{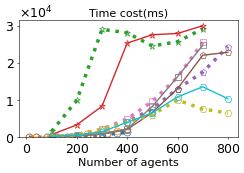}}
\end{minipage}
\hfill
\begin{minipage}{.24\linewidth}
  \centerline{\includegraphics[width=3.2cm]{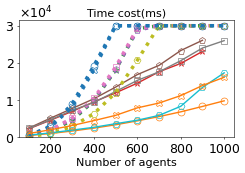}}
\end{minipage}
\hfill
\begin{minipage}{.24\linewidth}
  \centerline{\includegraphics[width=3.2cm]{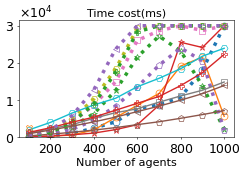}}
\end{minipage}
\vfill

\begin{minipage}{.24\linewidth}
  \centerline{\includegraphics[width=3.2cm]{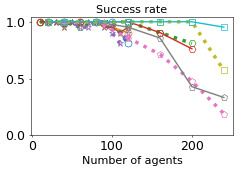}}
\end{minipage}
\hfill
\begin{minipage}{.24\linewidth}
  \centerline{\includegraphics[width=3.2cm]{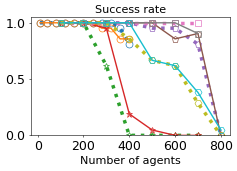}}
\end{minipage}
\hfill
\begin{minipage}{.24\linewidth}
  \centerline{\includegraphics[width=3.2cm]{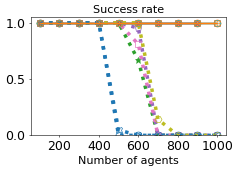}}
\end{minipage}
\hfill
\begin{minipage}{.24\linewidth}
  \centerline{\includegraphics[width=3.2cm]{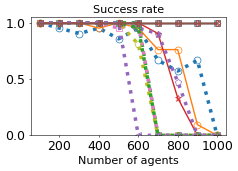}}
\end{minipage}
\vfill

\begin{minipage}{.24\linewidth}
  \centerline{\includegraphics[width=3.2cm]{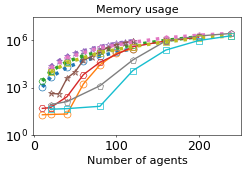}}
\end{minipage}
\hfill
\begin{minipage}{.24\linewidth}
  \centerline{\includegraphics[width=3.2cm]{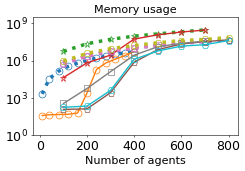}}
\end{minipage}
\hfill
\begin{minipage}{.24\linewidth}
  \centerline{\includegraphics[width=3.2cm]{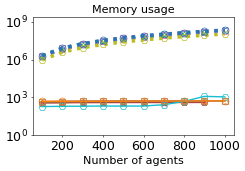}}
\end{minipage}
\hfill
\begin{minipage}{.24\linewidth}
  \centerline{\includegraphics[width=3.2cm]{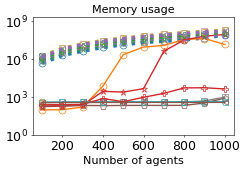}}
\end{minipage}
\vfill

\begin{minipage}{.24\linewidth}
  \centerline{\includegraphics[width=3.2cm]{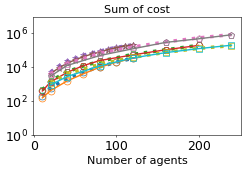}}
\end{minipage}
\hfill
\begin{minipage}{.24\linewidth}
  \centerline{\includegraphics[width=3.2cm]{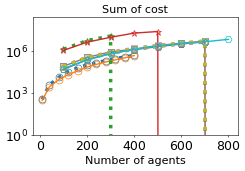}}
\end{minipage}
\hfill
\begin{minipage}{.24\linewidth}
  \centerline{\includegraphics[width=3.2cm]{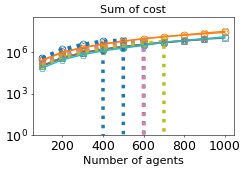}}
\end{minipage}
\hfill
\begin{minipage}{.24\linewidth}
  \centerline{\includegraphics[width=3.2cm]{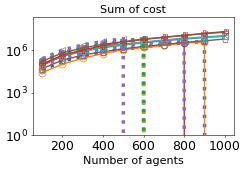}}
\end{minipage}
\vfill

\begin{minipage}{.24\linewidth}
  \centerline{\includegraphics[width=3.2cm]{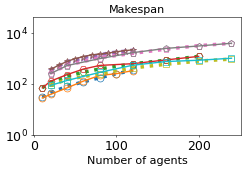}}
\end{minipage}
\hfill
\begin{minipage}{.24\linewidth}
  \centerline{\includegraphics[width=3.2cm]{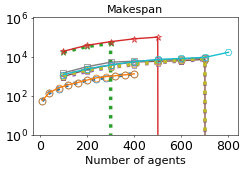}}
\end{minipage}
\hfill
\begin{minipage}{.24\linewidth}
  \centerline{\includegraphics[width=3.2cm]{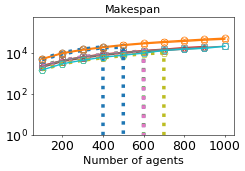}}
\end{minipage}
\hfill
\begin{minipage}{.24\linewidth}
  \centerline{\includegraphics[width=3.2cm]{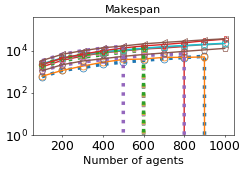}}
\end{minipage}
\vfill

\begin{minipage}{.24\linewidth}
  \centerline{\includegraphics[width=3.2cm]{layered_MAPF/makespan/EECBS/1_legend.png}}
\end{minipage}
\hfill
\begin{minipage}{.24\linewidth}
  \centerline{\includegraphics[width=3.2cm]{layered_MAPF/makespan/EECBS/2_legend.png}}
\end{minipage}
\hfill
\begin{minipage}{.24\linewidth}
  \centerline{\includegraphics[width=3.2cm]{layered_MAPF/makespan/EECBS/3_legend.png}}
\end{minipage}
\hfill
\begin{minipage}{.24\linewidth}
  \centerline{\includegraphics[width=3.2cm]{layered_MAPF/makespan/EECBS/4_legend.png}}
\end{minipage}
\vfill

\caption{These figures illustrate how Layered Push and Swap and raw Push and Swap work under various maps as the number of agents increases. They are compared in terms of time cost, success rate, memory usage, and solution quality (sum of cost and makespan). The data for Layered Push and Swap are shown with a solid line, and the data for raw Push and Swap are shown with a dotted line. More details about the maps can be found in Fig. \ref{decomposition1} and Fig. \ref{decomposition2}.
} 
\label{compare_with_raw_pas}
\end{figure}

\begin{figure}[h] \tiny


\begin{minipage}{.24\linewidth}
  \centerline{\includegraphics[width=3.2cm]{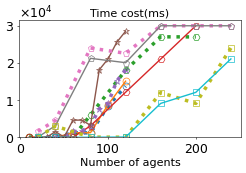}}
\end{minipage}
\hfill
\begin{minipage}{.24\linewidth}
  \centerline{\includegraphics[width=3.2cm]{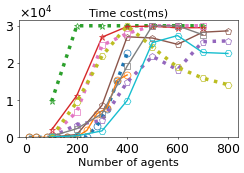}}
\end{minipage}
\hfill
\begin{minipage}{.24\linewidth}
  \centerline{\includegraphics[width=3.2cm]{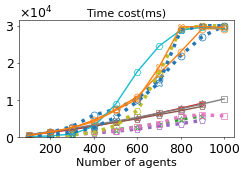}}
\end{minipage}
\hfill
\begin{minipage}{.24\linewidth}
  \centerline{\includegraphics[width=3.2cm]{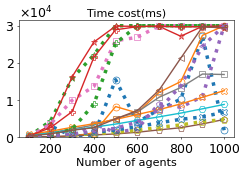}}
\end{minipage}
\vfill

\begin{minipage}{.24\linewidth}
  \centerline{\includegraphics[width=3.2cm]{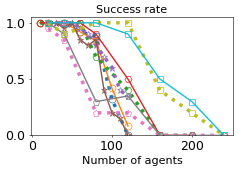}}
\end{minipage}
\hfill
\begin{minipage}{.24\linewidth}
  \centerline{\includegraphics[width=3.2cm]{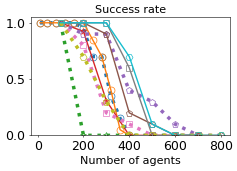}}
\end{minipage}
\hfill
\begin{minipage}{.24\linewidth}
  \centerline{\includegraphics[width=3.2cm]{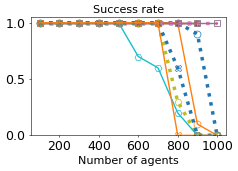}}
\end{minipage}
\hfill
\begin{minipage}{.24\linewidth}
  \centerline{\includegraphics[width=3.2cm]{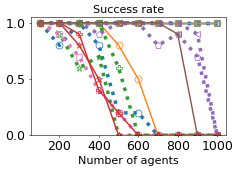}}
\end{minipage}
\vfill

\begin{minipage}{.24\linewidth}
  \centerline{\includegraphics[width=3.2cm]{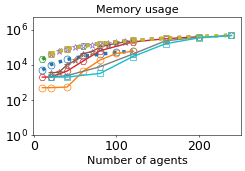}}
\end{minipage}
\hfill
\begin{minipage}{.24\linewidth}
  \centerline{\includegraphics[width=3.2cm]{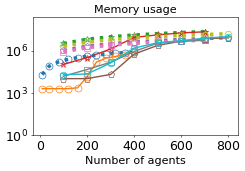}}
\end{minipage}
\hfill
\begin{minipage}{.24\linewidth}
  \centerline{\includegraphics[width=3.2cm]{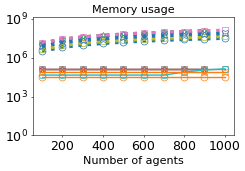}}
\end{minipage}
\hfill
\begin{minipage}{.24\linewidth}
  \centerline{\includegraphics[width=3.2cm]{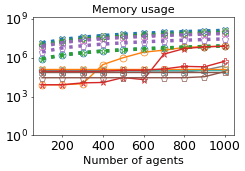}}
\end{minipage}
\vfill

\begin{minipage}{.24\linewidth}
  \centerline{\includegraphics[width=3.2cm]{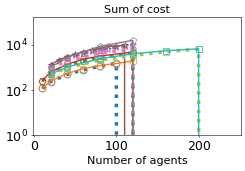}}
\end{minipage}
\hfill
\begin{minipage}{.24\linewidth}
  \centerline{\includegraphics[width=3.2cm]{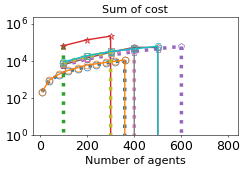}}
\end{minipage}
\hfill
\begin{minipage}{.24\linewidth}
  \centerline{\includegraphics[width=3.2cm]{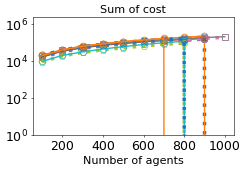}}
\end{minipage}
\hfill
\begin{minipage}{.24\linewidth}
  \centerline{\includegraphics[width=3.2cm]{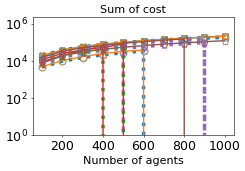}}
\end{minipage}
\vfill

\begin{minipage}{.24\linewidth}
  \centerline{\includegraphics[width=3.2cm]{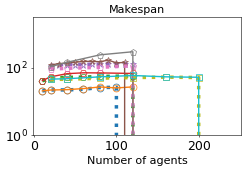}}
\end{minipage}
\hfill
\begin{minipage}{.24\linewidth}
  \centerline{\includegraphics[width=3.2cm]{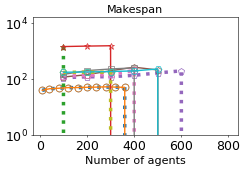}}
\end{minipage}
\hfill
\begin{minipage}{.24\linewidth}
  \centerline{\includegraphics[width=3.2cm]{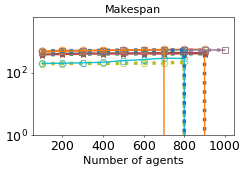}}
\end{minipage}
\hfill
\begin{minipage}{.24\linewidth}
  \centerline{\includegraphics[width=3.2cm]{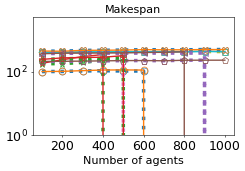}}
\end{minipage}
\vfill

\begin{minipage}{.24\linewidth}
  \centerline{\includegraphics[width=3.2cm]{layered_MAPF/makespan/EECBS/1_legend.png}}
\end{minipage}
\hfill
\begin{minipage}{.24\linewidth}
  \centerline{\includegraphics[width=3.2cm]{layered_MAPF/makespan/EECBS/2_legend.png}}
\end{minipage}
\hfill
\begin{minipage}{.24\linewidth}
  \centerline{\includegraphics[width=3.2cm]{layered_MAPF/makespan/EECBS/3_legend.png}}
\end{minipage}
\hfill
\begin{minipage}{.24\linewidth}
  \centerline{\includegraphics[width=3.2cm]{layered_MAPF/makespan/EECBS/4_legend.png}}
\end{minipage}
\vfill

\caption{These figures illustrate how Layered HCA and raw HCA work under various maps as the number of agents increases. They are compared in terms of time cost, success rate, memory usage, and solution quality (sum of cost and makespan). The data for Layered HCA are shown with a solid line, and the data for raw HCA are shown with a dotted line. More details about the maps can be found in Fig. \ref{decomposition1} and Fig. \ref{decomposition2}.
} 
\label{compare_with_raw_hca}
\end{figure}

\begin{figure}[h] \tiny


\begin{minipage}{.24\linewidth}
  \centerline{\includegraphics[width=3.2cm]{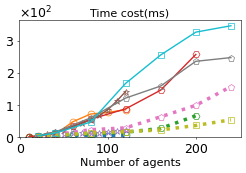}}
\end{minipage}
\hfill
\begin{minipage}{.24\linewidth}
  \centerline{\includegraphics[width=3.2cm]{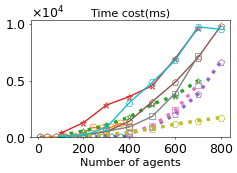}}
\end{minipage}
\hfill
\begin{minipage}{.24\linewidth}
  \centerline{\includegraphics[width=3.2cm]{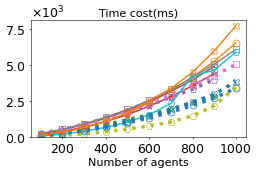}}
\end{minipage}
\hfill
\begin{minipage}{.24\linewidth}
  \centerline{\includegraphics[width=3.2cm]{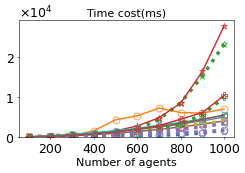}}
\end{minipage}
\vfill

\begin{minipage}{.24\linewidth}
  \centerline{\includegraphics[width=3.2cm]{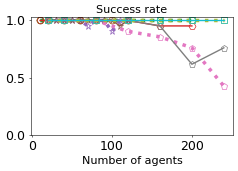}}
\end{minipage}
\hfill
\begin{minipage}{.24\linewidth}
  \centerline{\includegraphics[width=3.2cm]{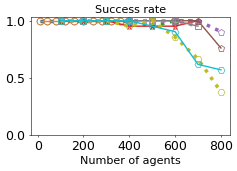}}
\end{minipage}
\hfill
\begin{minipage}{.24\linewidth}
  \centerline{\includegraphics[width=3.2cm]{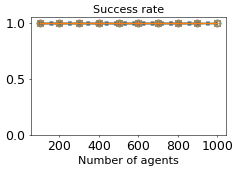}}
\end{minipage}
\hfill
\begin{minipage}{.24\linewidth}
  \centerline{\includegraphics[width=3.2cm]{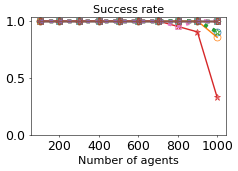}}
\end{minipage}
\vfill

\begin{minipage}{.24\linewidth}
  \centerline{\includegraphics[width=3.2cm]{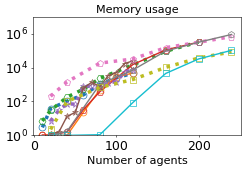}}
\end{minipage}
\hfill
\begin{minipage}{.24\linewidth}
  \centerline{\includegraphics[width=3.2cm]{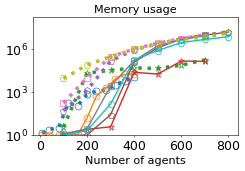}}
\end{minipage}
\hfill
\begin{minipage}{.24\linewidth}
  \centerline{\includegraphics[width=3.2cm]{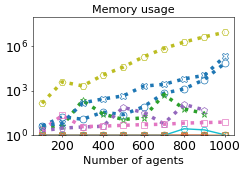}}
\end{minipage}
\hfill
\begin{minipage}{.24\linewidth}
  \centerline{\includegraphics[width=3.2cm]{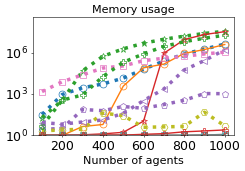}}
\end{minipage}
\vfill

\begin{minipage}{.24\linewidth}
  \centerline{\includegraphics[width=3.2cm]{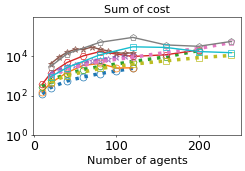}}
\end{minipage}
\hfill
\begin{minipage}{.24\linewidth}
  \centerline{\includegraphics[width=3.2cm]{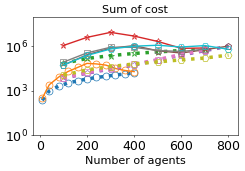}}
\end{minipage}
\hfill
\begin{minipage}{.24\linewidth}
  \centerline{\includegraphics[width=3.2cm]{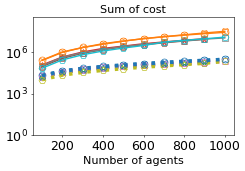}}
\end{minipage}
\hfill
\begin{minipage}{.24\linewidth}
  \centerline{\includegraphics[width=3.2cm]{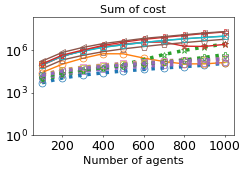}}
\end{minipage}
\vfill

\begin{minipage}{.24\linewidth}
  \centerline{\includegraphics[width=3.2cm]{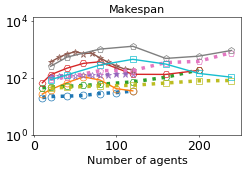}}
\end{minipage}
\hfill
\begin{minipage}{.24\linewidth}
  \centerline{\includegraphics[width=3.2cm]{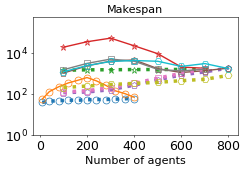}}
\end{minipage}
\hfill
\begin{minipage}{.24\linewidth}
  \centerline{\includegraphics[width=3.2cm]{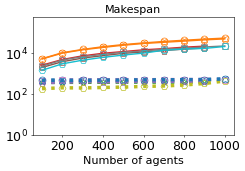}}
\end{minipage}
\hfill
\begin{minipage}{.24\linewidth}
  \centerline{\includegraphics[width=3.2cm]{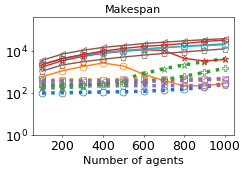}}
\end{minipage}
\vfill

\begin{minipage}{.24\linewidth}
  \centerline{\includegraphics[width=3.2cm]{layered_MAPF/makespan/EECBS/1_legend.png}}
\end{minipage}
\hfill
\begin{minipage}{.24\linewidth}
  \centerline{\includegraphics[width=3.2cm]{layered_MAPF/makespan/EECBS/2_legend.png}}
\end{minipage}
\hfill
\begin{minipage}{.24\linewidth}
  \centerline{\includegraphics[width=3.2cm]{layered_MAPF/makespan/EECBS/3_legend.png}}
\end{minipage}
\hfill
\begin{minipage}{.24\linewidth}
  \centerline{\includegraphics[width=3.2cm]{layered_MAPF/makespan/EECBS/4_legend.png}}
\end{minipage}
\vfill

\caption{These figures illustrate how Layered PIBT+ and raw PIBT+ work under various maps as the number of agents increases. They are compared in terms of time cost, success rate, memory usage, and solution quality (sum of cost and makespan). The data for Layered PIBT+ are shown with a solid line, and the data for raw PIBT+ are shown with a dotted line. More details about the maps can be found in Fig. \ref{decomposition1} and Fig. \ref{decomposition2}.
} 
\label{compare_with_raw_PIBT}
\end{figure}

\begin{figure}[h] \tiny


\begin{minipage}{.24\linewidth}
  \centerline{\includegraphics[width=3.2cm]{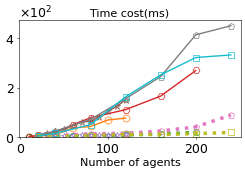}}
\end{minipage}
\hfill
\begin{minipage}{.24\linewidth}
  \centerline{\includegraphics[width=3.2cm]{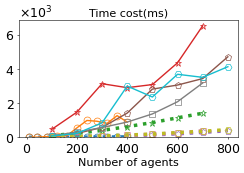}}
\end{minipage}
\hfill
\begin{minipage}{.24\linewidth}
  \centerline{\includegraphics[width=3.2cm]{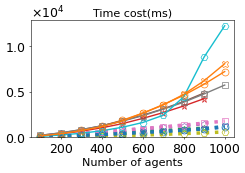}}
\end{minipage}
\hfill
\begin{minipage}{.24\linewidth}
  \centerline{\includegraphics[width=3.2cm]{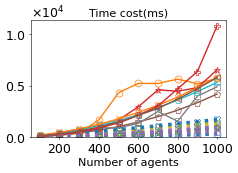}}
\end{minipage}
\vfill

\begin{minipage}{.24\linewidth}
  \centerline{\includegraphics[width=3.2cm]{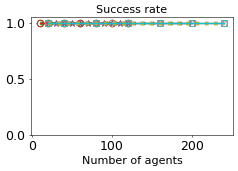}}
\end{minipage}
\hfill
\begin{minipage}{.24\linewidth}
  \centerline{\includegraphics[width=3.2cm]{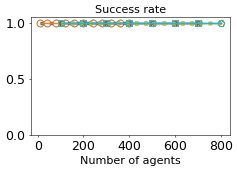}}
\end{minipage}
\hfill
\begin{minipage}{.24\linewidth}
  \centerline{\includegraphics[width=3.2cm]{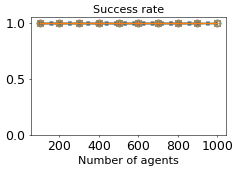}}
\end{minipage}
\hfill
\begin{minipage}{.24\linewidth}
  \centerline{\includegraphics[width=3.2cm]{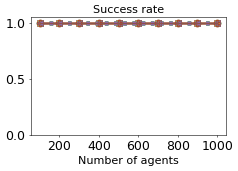}}
\end{minipage}
\vfill

\begin{minipage}{.24\linewidth}
  \centerline{\includegraphics[width=3.2cm]{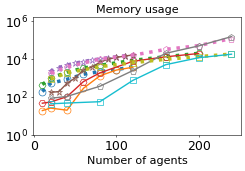}}
\end{minipage}
\hfill
\begin{minipage}{.24\linewidth}
  \centerline{\includegraphics[width=3.2cm]{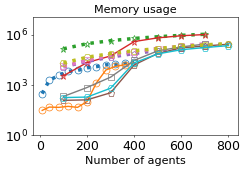}}
\end{minipage}
\hfill
\begin{minipage}{.24\linewidth}
  \centerline{\includegraphics[width=3.2cm]{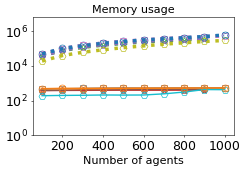}}
\end{minipage}
\hfill
\begin{minipage}{.24\linewidth}
  \centerline{\includegraphics[width=3.2cm]{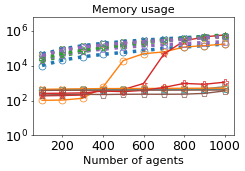}}
\end{minipage}
\vfill

\begin{minipage}{.24\linewidth}
  \centerline{\includegraphics[width=3.2cm]{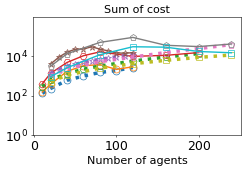}}
\end{minipage}
\hfill
\begin{minipage}{.24\linewidth}
  \centerline{\includegraphics[width=3.2cm]{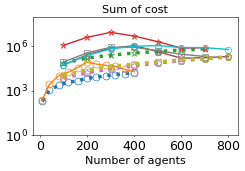}}
\end{minipage}
\hfill
\begin{minipage}{.24\linewidth}
  \centerline{\includegraphics[width=3.2cm]{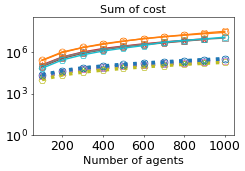}}
\end{minipage}
\hfill
\begin{minipage}{.24\linewidth}
  \centerline{\includegraphics[width=3.2cm]{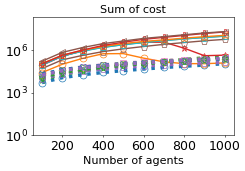}}
\end{minipage}
\vfill

\begin{minipage}{.24\linewidth}
  \centerline{\includegraphics[width=3.2cm]{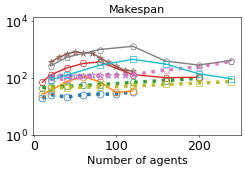}}
\end{minipage}
\hfill
\begin{minipage}{.24\linewidth}
  \centerline{\includegraphics[width=3.2cm]{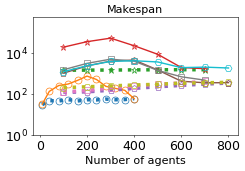}}
\end{minipage}
\hfill
\begin{minipage}{.24\linewidth}
  \centerline{\includegraphics[width=3.2cm]{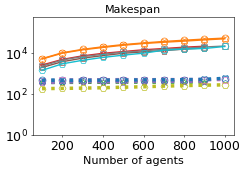}}
\end{minipage}
\hfill
\begin{minipage}{.24\linewidth}
  \centerline{\includegraphics[width=3.2cm]{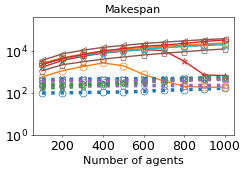}}
\end{minipage}
\vfill

\begin{minipage}{.24\linewidth}
  \centerline{\includegraphics[width=3.2cm]{layered_MAPF/makespan/EECBS/1_legend.png}}
\end{minipage}
\hfill
\begin{minipage}{.24\linewidth}
  \centerline{\includegraphics[width=3.2cm]{layered_MAPF/makespan/EECBS/2_legend.png}}
\end{minipage}
\hfill
\begin{minipage}{.24\linewidth}
  \centerline{\includegraphics[width=3.2cm]{layered_MAPF/makespan/EECBS/3_legend.png}}
\end{minipage}
\hfill
\begin{minipage}{.24\linewidth}
  \centerline{\includegraphics[width=3.2cm]{layered_MAPF/makespan/EECBS/4_legend.png}}
\end{minipage}
\vfill

\caption{These figures illustrate how Layered LaCAM2 and raw LaCAM2 work under various maps as the number of agents increases. They are compared in terms of time cost, success rate, memory usage, and solution quality (sum of cost and makespan). The data for Layered LaCAM2 are shown with a solid line, and the data for raw LaCAM2 are shown with a dotted line. More details about the maps can be found in Fig. \ref{decomposition1} and Fig. \ref{decomposition2}.
} 
\label{compare_with_raw_LaCAM}
\end{figure}

\end{appendices}
\end{document}